\documentclass[a5paper]{book}
\usepackage{geometry}
\usepackage[backend=biber,natbib=true,maxcitenames=2,maxbibnames=99]{biblatex}
\usepackage{footnote}

\newenvironment{abstract}%
{\chapter*{\centering \begin{normalsize}Abstract\end{normalsize}}%
	\begin{quotation}}%
{\end{quotation}}

\DeclareSourcemap{
	\maps[datatype=bibtex]{
		\map[overwrite=true]{
			\step[fieldset=File, null]
			\step[fieldset=Keywords, null]
		}
	}
}

\renewcommand{\thefootnote}{} %

\usepackage[utf8]{inputenc}

\usepackage{multirow}
\usepackage{amsmath,amsfonts,amssymb,amsthm}
\usepackage{subcaption}
\usepackage{hyperref}
\usepackage{color}
\usepackage{adjustbox}
\usepackage[english]{babel}
\usepackage{pifont}%
\usepackage{ctable} %
\usepackage{csquotes}

\newcommand{\figref}[1]{Figure~\ref{#1}}
\newcommand{\secref}[1]{Section~\ref{#1}}
\newcommand{\chpref}[1]{Chapter~\ref{#1}}
\newcommand{\tabref}[1]{Table~\ref{#1}}

\renewcommand{\b}{\ensuremath{\mathbf}}

\newcommand{\T}{^{\raisemath{-1pt}{\mathsf{T}}}}

\makeatletter
\DeclareRobustCommand\onedot{\futurelet\@let@token\@onedot}
\def\@onedot{\ifx\@let@token.\else.\null\fi\xspace}
\def\ia{i.a\onedot} 
 
\def\etc{etc\onedot} 

\def\wrt{with respect to }
\def\etal{\textit{et~al\onedot~}}

\makeatother

\makeatletter
\DeclareRobustCommand\onecomma{\futurelet\@let@token\@onecomma}
\def\@onecomma{\ifx\@let@token,\else,\null\fi\xspace}
\def\eg{e.g.\onecomma} 
\def\ie{i.e.\onecomma} 
\makeatother

\newcommand{\boldparagraph}[1]{\vspace{0.2cm}\noindent{\bf #1:} }

\definecolor{aseem_pink}{RGB}{219,112,147}
\definecolor{fatma_purple}{RGB}{128,0,128}
\newcommand{\cmark}{\ding{52}}%
\newcommand{\courtesy}[1]{Figure courtesy of #1}
\newcommand{\courtesyC}[3]{Figure courtesy of #1 \textcopyright~#2 #3}
\newcommand{\figsource}[1]{Reprinted, with permission, from #1}
\newcommand{\figsourceC}[3]{\textcopyright~#2 #3. Reprinted, with permission, from #1}
\newcommand{\figsourceElsevier}[2]{Reprinted from #1, \textcopyright~#2, with permission from Elsevier}
\newcommand{\figsourceSpringer}[3]{Reprinted by permission from Springer Nature Customer Service Centre GmbH: Springer Nature #3, #1, \textcopyright~#2}
\newcommand{\figsourceSage}[3]{Reprinted by Permission of SAGE Publications, Inc: #3, #1, \textcopyright~#2}

\title{Computer Vision for Autonomous Vehicles:\\ Problems, Datasets and State of the Art}

\usepackage{mwe}

\author{Joel~Janai\textsuperscript{1,$\ast$}\footnote{\textsuperscript{$\ast$}\,The first three authors contributed equally}
	\quad Fatma~G\"{u}ney\textsuperscript{2,$\ast$}
	\quad Aseem~Behl\textsuperscript{1,$\ast$}
	\quad Andreas~Geiger\textsuperscript{1}}

\date{\textsuperscript{1} Max-Planck-Institute~for~Intelligent~Systems~T\"{u}bingen and University~of~T\"{u}bingen, Germany \\
	\textsuperscript{2} College of Engineering, Ko\c{c} University, Turkey\\[4ex]
	\today
}

\begin{document}
	\maketitle
	
	\tableofcontents	
	
	\renewcommand{\thefootnote}{\arabic{footnote}}  %
	
	\begin{abstract}
Recent years have witnessed enormous progress in AI-related fields such as computer vision, machine learning, and autonomous vehicles.
As with any rapidly growing field, it becomes increasingly difficult to stay up-to-date or enter the field as a beginner.
While several survey papers on particular sub-problems have appeared, no comprehensive survey on problems, datasets, and methods in computer vision for autonomous vehicles has been published. 
This book attempts to narrow this gap by providing a survey on the state-of-the-art datasets and techniques. 
Our survey includes both the historically most relevant literature as well as the current state of the art on several specific topics, including recognition, reconstruction, motion estimation, tracking, scene understanding, and end-to-end learning for autonomous driving. Towards this goal, we analyze the performance of the state of the art on several challenging benchmarking datasets, including KITTI, MOT, and Cityscapes. Besides, we discuss open problems and current research challenges.
To ease accessibility and accommodate missing references, we also provide a website that allows navigating topics as well as methods and provides additional information.
\end{abstract}

	\chapter{Introduction}

Since the first successful demonstrations in the 1980s \citep{Dickmanns1992PAMI,Dickmanns1988MVA,Thorpe1988PAMI}, great progress has been made in the field of autonomous vehicles. However, despite these advances and ambitious commercial goals, fully autonomous navigation in general environments has not been realized to date. The reason for this is two-fold: First, autonomous systems which operate in complex dynamic environments require models which generalize to unpredictable situations and reason in a timely manner. Second, informed decisions require accurate perception, yet most of the existing computer vision models are still inferior to human perception and reasoning. 

\begin{figure}[t]
	\centering
	\includegraphics[width=1.00\columnwidth]{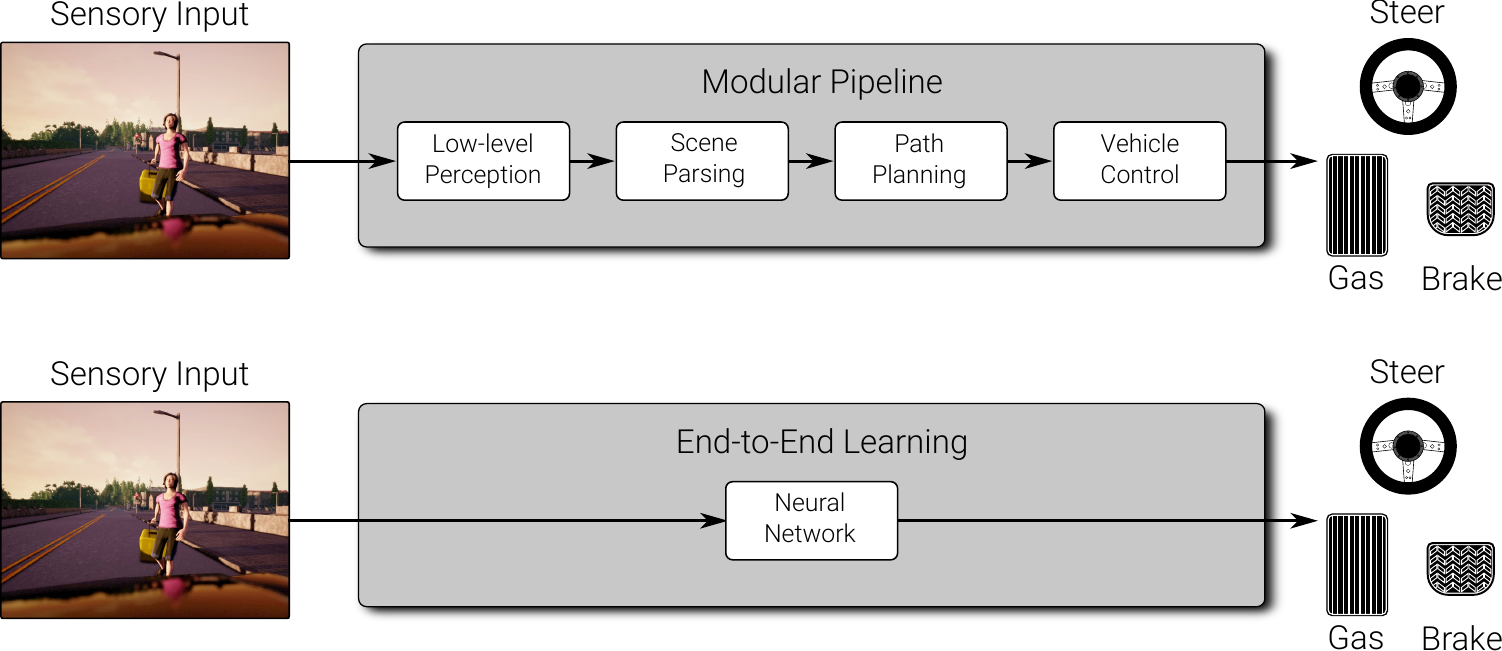}
	\caption[Approaches to Self-Driving]{\textbf{Approaches to Self-Driving.} Classical modular pipeline (top) vs. monolithic end-to-end learning approach (bottom). See text for details.}
	\label{fig:approaches_to_self_driving}
\end{figure} 

Existing approaches to self-driving can be roughly categorized into modular pipelines and monolithic end-to-end learning approaches. Both approaches are contrasted at a conceptual level in \figref{fig:approaches_to_self_driving}. The modular pipeline is the standard approach to autonomous driving, mostly followed in the industry. The key idea is to break down the complex mapping function from high-dimensional inputs to low-dimensional control variables into modules which can be independently developed, trained, and tested. In \figref{fig:approaches_to_self_driving} (top), these modules comprise low-level perception, scene parsing, path planning, and vehicle control. However, this is just one particular example of modularizing a self-driving stack and other or more fine-grained modularizations are also possible. Existing approaches typically leverage machine learning (\eg deep neural networks) to extract low-level features or to parse the scene into individual components. In contrast, path planning and vehicle control are dominated by classical state machines, search algorithms, and control models. 

The major advantage of modular pipelines is that they deploy human interpretable intermediate representations such as detected objects or free space information which allow gaining insights into failure modes of the system. Furthermore, the development of modular pipelines can be easily parallelized within companies where typically different teams work on different aspects of the driving problem simultaneously. Furthermore, it is comparably easy to integrate first principles and prior knowledge about the problem into the system. Examples include traffic laws that can be explicitly enforced in the planner or knowledge about the vehicle dynamics, which lead to improved vehicle control. Other aspects that are more difficult to specify by hand, such as the appearance of pedestrians, are learned from large annotated datasets.

A major drawback of modular approaches is the fact that human-designed intermediate representations are not necessarily optimal for the driving task, which typically includes aspects like safety, comfort, and time for reaching the goal. Moreover, most modules are trained and validated independently from each other, making use of auxiliary loss functions. Consider the problem of object detection as an example. Most objects in the scene are not directly relevant for the driving task, yet the learning algorithm is not informed about the relevance of each object and therefore tasks a neural network to detect all objects with equal importance. Thus, the network is wasting capacity on irrelevant objects while not being able to detect the driving relevant objects with the necessary accuracy. This demonstrates the difficulty of defining appropriate intermediate representations and auxiliary loss functions.

An alternative to modular pipelines is end-to-end learning-based models which try to learn a policy, \ie a function from observations to actions using a generic model such as a deep neural network. This approach is illustrated in \figref{fig:approaches_to_self_driving} (bottom) and discussed in detail in \chpref{sec:end_to_end_learning}. The network parameters can be learned either via imitation learning by replicating the behavior of a teacher or using reinforcement learning by exploring the world and taking actions that are likely to yield a high user-specified reward. However, reinforcement learning approaches suffer from the credit assignment and reward shaping problems, are typically slow and can only be applied in non-safety-critical simulation environments. Imitation learning, on the other hand, suffers from overfitting and does not easily generalize to novel scenarios. Furthermore, holistic neural network-based approaches are often hard to interpret as they present themselves as ``black boxes'' to the user which do not reveal \textit{why} a certain error has occurred.

In this survey, we focus on perception for autonomous vehicles. In particular, we discuss the perception-related modules of the modular pipeline as well as end-to-end learning-based approaches. Other aspects of the self-driving problem are discussed in related surveys: For example, \citet{Winner2015eng} put emphasis on driver assistance systems, considering both their structure and their function. Similarly, \citet{Klette2015} provides an overview of vision-based driver assistance systems. They describe most aspects of the perception problem at a high level but do not provide an in-depth review of the state of the art in each task as we pursue in this survey.
Complementary to our work, \citet{Zhu2017TITS} provide an overview of environment perception for intelligent vehicles, focusing on lane detection, traffic sign/light recognition as well as vehicle tracking.
In contrast, our goal is to bridge the gap between the robotics, intelligent vehicles, and computer vision communities by providing an extensive overview and comparison, including works from all three fields.

This survey is structured as follows: first, we provide a brief history of autonomous driving, followed by an introduction to camera models and calibration techniques. We then provide an overview of autonomous driving-related datasets with a particular focus on perception before surveying the relevant perception tasks and the state-of-the-art algorithms for solving them. More specifically, we review object detection, tracking, semantic (instance) segmentation, reconstruction, motion estimation, and scene understanding techniques. Each chapter starts with the problem definition, an overview over the most important methods and main design choices, a qualitative and quantitative analysis of the top-performing techniques on the most popular datasets, as well as a discussion of the state of the art in this area. Finally, we provide an overview of state-of-the-art end-to-end models for autonomous driving before concluding this survey.
To ease navigation, we also provide an interactive online tool\footnote{\url{http://www.cvlibs.net/projects/autonomous_vision_survey}} which visualizes the surveyed papers with an interactive graph and additional information in an easily accessible manner.
We hope that our survey will become a useful tool for researchers in the field of autonomous vision and lowers the entry barrier for beginners by providing a thorough overview of the field.
	\chapter{History of Autonomous Driving}

Similar to the invention of the automobile by Carl Benz in 1886, self-driving technology promises to profoundly impact our mobility. In this chapter, we briefly review the history of driverless and self-driving vehicles from 1925 to 2019.

The first demonstration of a driverless vehicle was reported in 1925 when Houdina Radio Control demonstrated the ``American Wonder'', a remote-controlled vehicle that traveled along Broadway in New York City trailed by an operator in another vehicle\citep{TIME1925}. Several years later, General Motors approached Norman Bel Geddes to sketch his vision about mobility 20 years into the future, culminating in Futurama, the most successful exhibition at the New York World Fair in 1939. Besides multi-lane highways, this vision sketched radio-controlled electric cars that navigated via electromagnetic circuits installed in the roadway. This vision led to several prototypes such as the GM Firebird II \citep{GM2017} in 1956, and RCA Labs' wire controlled car in 1960 as well as a demonstration of Citroen with its DS 19 and the Cabinentaxi\footnote{\url{https://www.youtube.com/watch?v=ERdF0FK-2io}} of Demag/MBB in 1970. However, the idea of infrastructure-based autonomous navigation is largely restricted to specific use cases such as ground transportation at airports, park shuttles, or automated facilities due to its limited scalability and high cost. 

\begin{figure}[t]
	\centering
	\includegraphics[width=1.00\columnwidth]{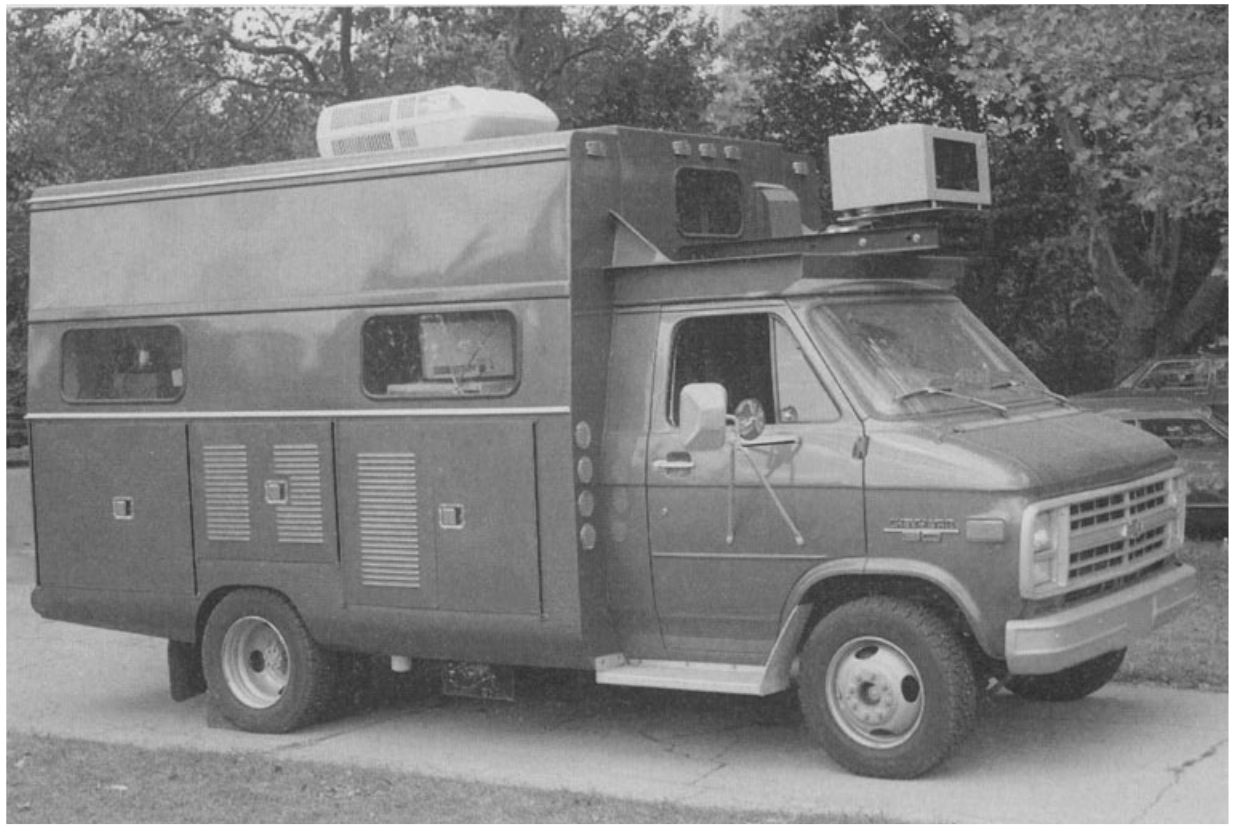}
	\caption[The Navlab]{\textbf{The Navlab.} The self-contained laboratory from CMU for navigational vision system research. \figsourceSpringer{\citet{Thorpe1988PAMI}}{1988}{High Precision Navigation}.}
	\label{fig:Thorpe1988PAMI}
\end{figure} 
In 1986, the first self-driving car prototypes which did not rely on dedicated infrastructure hit the road. This pioneering effort was led by the Navlab team at CMU in the US as well as Ernst Dickmanns's team at the Bundeswehr University Munich in Germany.
Carnegie Mellon University's Navlab team \citep{Thorpe1988PAMI} (\figref{fig:Thorpe1988PAMI}) achieved another major milestone in 1995, by driving from Washington, D.C., to San Diego, CA, 98\% autonomously with manual longitudinal control in the ’No hands across America’ tour \citep{Pomerleau2015CMU, Pomerleau1996EXPERT}. With ALVINN \citep{Pomerleau1988NIPS}, the Navlab team at CMU demonstrated an imitation learning approach where a relatively small neural network was optimized in an end-to-end fashion to keep the vehicle on the road based on user demonstrations. On the contrary, Dickmanns presented a modular approach in which a vehicle and road model was used for continuously estimating the state and controlling the vehicle \citep{Dickmann1995}. The project was conducted in the context of the European PROMETHEUS project, which involved more than 13 vehicle manufacturers and several research units from governments and universities of 19 European countries. In 1995, the PROMETHEUS team demonstrated the first autonomous long-distance drive from Munich, Germany, to Odense, Denmark, at velocities up to 175 km/h with about 95\% autonomous driving \citep{Dickmanns1990SMC,Franke1994IV,Dickmanns1994IV}.

Motivated by the success of the PROMETHEUS projects to drive autonomously on highways, \citet{Franke1998IS} describe a real-time vision system for autonomous driving in complex urban traffic situations. While highway scenarios have been studied intensively, urban scenes have not been addressed before. Their system included depth-based obstacle detection and tracking from stereo as well as a framework for monocular detection and recognition of relevant objects such as traffic signs. 
Many approaches to the challenging task of autonomous driving developed during these projects are presented and discussed in \citep{Bertozzi2000RAS}. They concluded that sufficient computing power is becoming increasingly available, but difficulties like reflections, wet roads, direct sunshine, tunnels, and shadows still make data interpretation challenging. Thus, they suggested the enhancement of sensor capabilities. They also pointed out that the legal aspects related to the responsibility and impact of automatic driving on human passengers need to be considered carefully. In summary, the automation will likely be restricted to special infrastructures and will be extended gradually.

While full self-driving has remained unsolved to date, driver assistance systems have reached commercial success, enriching driving comfort and safety. In 1995, Mitsubishi presented the first LiDAR-based distance control \citep{Mitsubishi1998}, and in 1999 Mercedes-Benz implemented the radar-assisted adaptive cruise control. 
In 2000, navigation systems and digital road maps became available. Today, differential GPS in combination with inertial measurement units (IMU) allows for localization at an accuracy of 5cm in good conditions, enabling the use of detailed lane-level road maps (HD maps) and providing redundancy for noisy vision-based localization. 

\begin{figure}[t]
	\centering
	\includegraphics[width=1.00\columnwidth]{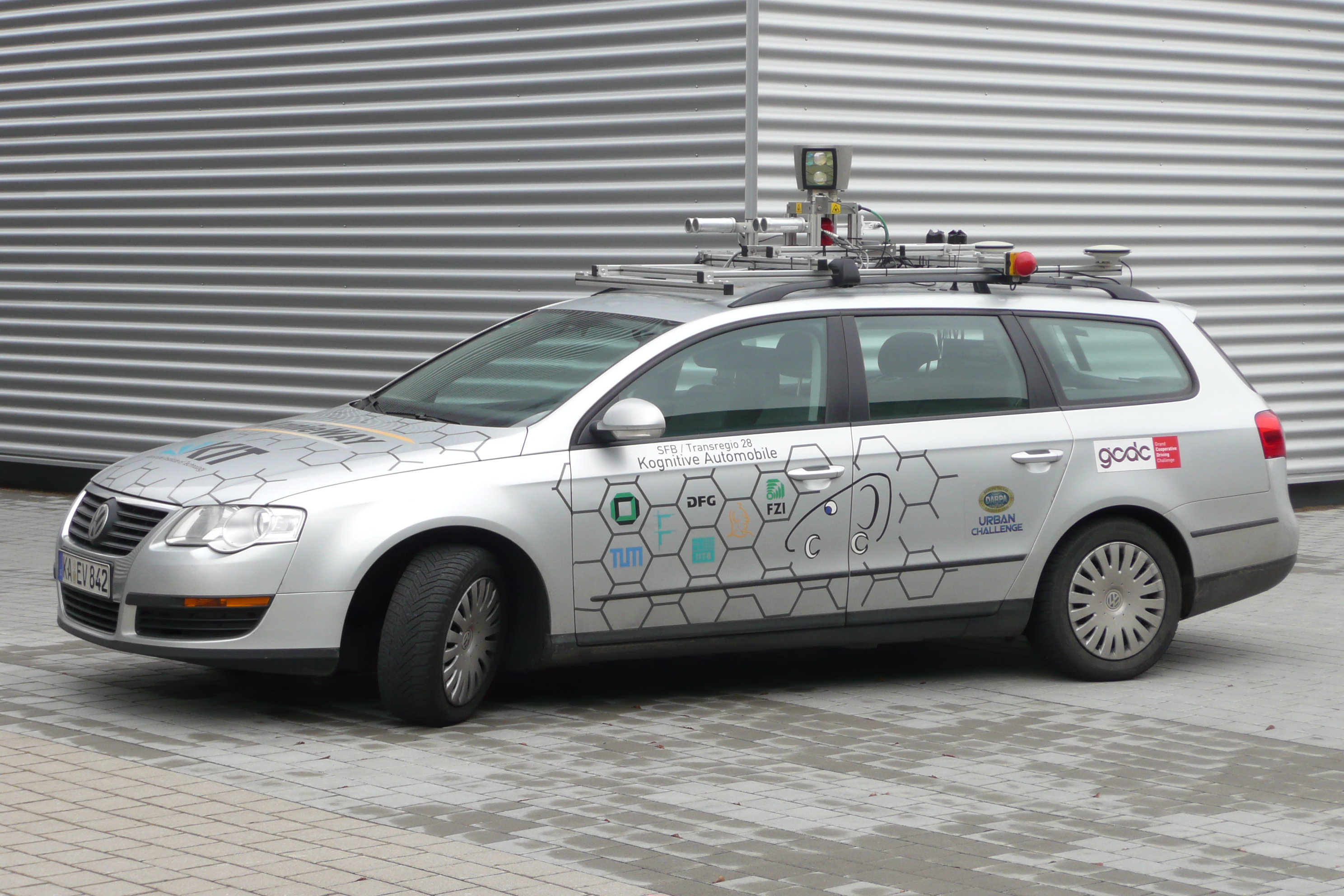}
	\caption[Participant in the DARPA Urban Challenge]{\textbf{AnnieWAY.} Participant in the DARPA Urban Challenge. \figsource{Andreas Geiger and Karlsruher Institut für Technologie - MRT}.}
	\label{fig:AnnieWay}
\end{figure} 
In 2004, the Defense Advanced Research Projects Agency (DARPA) of the US Department of Defense started to organize and sponsor a series of 3 races to foster the development of self-driving technology \citep{DARPA2014}. The first race, the Darpa Grand Challenge 2004, was limited to US participants. DARPA offered a prize money of \$1 million for the first team  autonomously completing a 240km long dirt route from California to Nevada through the Mojave desert, guided by GPS waypoints. However, none of the robot vehicles completed the route. One year later, in 2005, DARPA announced a second edition of its challenge with 5 vehicles successfully completing the route \citep{Buehler2007} and Stanford taking the lead, arriving 10 minutes before the CMU team which ranked second. In 2007, DARPA organized the last race of this series, the Darpa Urban Challenge \citep{Buehler2009DARPAChallenge}, where also international participants were allowed (\figref{fig:AnnieWay}). In contrast to the previous challenges, this competition required vehicles to drive a 96 km route through a mock-up town at George Air Force Base while obeying traffic laws, avoiding obstacles, negotiating with other vehicles, and merging into traffic. This time, the CMU team finished first, followed by the Stanford team, which ranked second. Notably, most of the successful teams relied heavily on the emerging multi-beam LiDAR technology developed in a pioneering effort by Velodyne\footnote{\url{https://www.velodynelidar.com/}}. This spinning multi-beam LiDAR scanner allowed for obtaining precise depth measurements with a 360-degree field-of-view around the vehicle, which turned out crucial for navigating urban environments.

\begin{figure}[t]
	\centering
	\includegraphics[width=0.50\columnwidth]{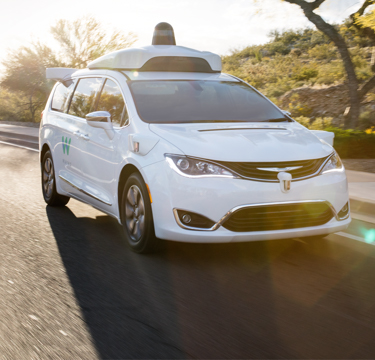}
	\caption[Waymo Autonomous Vehicle]{\textbf{Waymo Autonomous Vehicle.} Source: Waymo \textcopyright~2019 Waymo.}
	\label{fig:Waymo}
\end{figure} 
In 2009, Google took the lead and hired a range of star scientists who had participated in the Darpa Challenges (including Sebastian Thrun, Chris Urmson, and Mike Montemerlo). They started their own self-driving car program which included the development of a new driving platform and a custom, affordable multi-beam LiDAR scanner. According to accident reports \citep{DMV2019}, Google's self-driving cars were involved in 14 collisions, while 13 were caused by others until 2016.

In 2010, the VisLab team led by Alberto Broggi at the University of Parma in Italy conducted the VisLab Intercontinental Autonomous Challenge (VIAC)\citep{Broggi2010ITSC}. Based on the experience with various prototype vehicles \citep{Broggi1999,Braid2006JFR,Grisleri2010IFAC}, VIAC \citep{Bertozzi2011IV} was an effort to drive semi-autonomously from Parma in Italy to Shanghai in China. In this demonstration, a second vehicle automatically followed a route defined by a manually driven lead vehicle either visually or based on GPS waypoints sent by the lead vehicle. The onboard system allowed for detecting obstacles, lane marking, ditches, berms, and to identify the presence and position of the preceding vehicle.

In the same year, Audi demonstrated a self-driven car ride to the summit of Pikes Peak at 4300 meters above sea level and the Technical University of Braunschweig showcased their Stadtpilot\citep{TUB2010} which was able to navigate in a small geofenced innercity area based on LiDAR, cameras, and HD maps. In 2015, the VisLab team conducted the PROUD project \citep{Broggi2015TITS}, a demonstration of inner-city and freeway driving in Parma.

In 2011, TNO organized the Grand Cooperative Driving Challenge \citep{Lauer2011ITSM}, a competition focusing on autonomous cooperative driving behavior. It was held in Helmond, Netherlands in 2011 for the first time and in 2016 for the second edition. During the competition, the semi-autonomous vehicles had to negotiate convoys, join convoys, and lead convoys. While longitudinal control was autonomous, lateral control was provided by a human safety driver. The winner (team KIT in 2011 \citep{Geiger2012TITS} and team Halmstad in 2016) was selected based on a system that assigned points to randomly mixed teams.

In 2012, the KITTI Vision Benchmark\footnote{\url{http://www.cvlibs.net/datasets/kitti/}} \citep{Geiger2012CVPR, Geiger2013IJRR} was released. For the first time, researchers around the globe were able to evaluate their progress on various self-driving perception tasks (including reconstruction, motion estimation, and object recognition) in a fair and objective manner. At the same time, deep learning started to revolutionize many fields, including computer vision and robotics, which laid the foundations for significant improvements in particular in terms of accuracy, robustness, and run-time of the perception components of self-driving vehicles.

In 2013, Mercedes Benz demonstrated the S500 Intelligent Drive, a 103 km autonomous ride on the historic Bertha Benz route from Mannheim to Pforzheim in Germany. The system was developed by Daimler research in collaboration with the Karlsruhe Institute of Technology (KIT) \citep{Ziegler2014ITSM}. The Mercedes S500 vehicle was equipped with close-to-production sensor hardware. Object detection and free-space analysis were performed using radar and stereo vision. Monocular vision was used for traffic light detection and object classification. Two complementary vision algorithms, point-feature-based and lane-marking-based, allowed for centimeter-accurate localization relative to manually annotated HD maps. While focusing on a single route, the effort demonstrated that autonomous driving in complex inner-city environments based on close-to-production hardware and HD maps is feasible.

The EU funded collaborative project V-Charge \citep{Furgale2013IV} conducted by Volkswagen, Bosch, and several academic partners (ETHZ, Oxford, Parma, Braunschweig)
aimed for fully autonomous charging and parking of electric vehicles.
In the context of this project, a fully operational system has been demonstrated which included vision-only localization, mapping, navigation and control. The project supported many publications on different problems such as calibration \citep{Heng2013IROS,Heng2015JFR}, stereo \citep{Haene2014THREEDV}, reconstruction \citep{Haene2012THREEDIMPVT,Haene2013CVPR,Haene2014CVPR}, SLAM \citep{Grimmett2015ICRA} and free space detection \citep{Haene2015IROS}.

\begin{figure}[t]
	\centering
	\includegraphics[width=1.00\columnwidth]{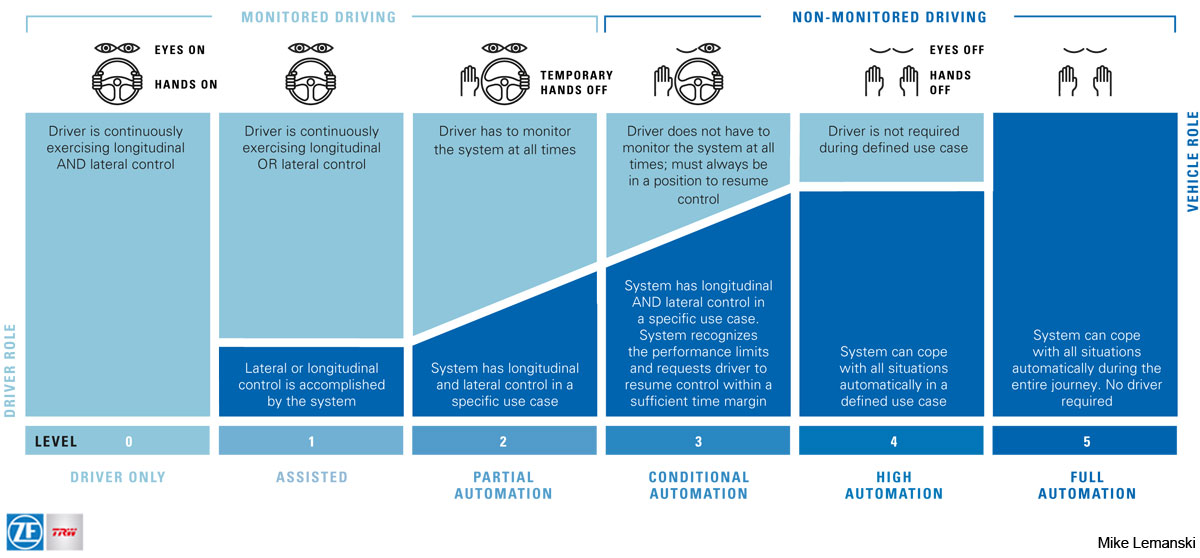}
	\caption[SAE Levels of Autonomy]{\textbf{SAE Levels of Autonomy.} \figsource{Mike Lemanski}.}
	\label{fig:sae_levels}
\end{figure} 

In 2014, the society of automotive engineers released their classification of autonomous driving systems into 6 SAE levels of autonomy, ranging from level 0 (no autonomy) to level 5 (full autonomy), illustrated in \figref{fig:sae_levels}. In the same year, Mercedes released its S Class and Tesla its Autopilot \citep{Tesla2014} with level 2 autonomy (the driver has to monitor the system at all times), providing autonomous steering, lane keeping, acceleration, and braking on the highway. One year later, ride-hailing company Uber launched its own self-driving effort \citep{Uber2015}, hiring a large number of robotics researchers from CMU. From October 2016, all vehicles produced by Tesla are equipped with eight cameras, twelve ultrasonic sensors, and a forward-facing radar with the goal of enabling full self-driving in the future. However, both Uber and Tesla witnessed fatal accidents in which neither the driver was attentive, nor the self-driving system was functioning properly.

In 2016, after completing over 1,5 million miles, Google's self-driving efforts became Waymo, a stand-alone subsidiary of Alphabet Inc. Today, Waymo offers 400 citizens of Phoenix access to its early rider program \citep{Waymo2019} which features full self-driving in several geo-fenced districts of Phoenix (\figref{fig:Waymo}) with a safety driver on the back seat.

In the same year, NVIDIA \citep{Bojarski2016ARXIV} demonstrated a $98\%$ autonomous ride from Holmdel to Atlantic Highlands in Monmouth County NJ using a single convolutional neural network. The network was trained via imitation learning to predict vehicle control directly from input images. In 2018, several last-mile delivery projects were launched, including Nuro \citep{Cade2018}, a project founded by two former Google self-driving car engineers and Scout \citep{Sean2019}, a fully-electric delivery system designed to safely get packages to Amazon customers using autonomous delivery devices.
In 2019, Bosch and Daimler announced a fleet of autonomous cars, providing customers a shuttle service with automated vehicles on selected routes in California \citep{Daimler2019}.

	\chapter{Sensors}
The navigation of autonomous systems is usually addressed with a sensor suite which comprises various different types of sensors, including cameras, wheel odometry, and range sensors (SONAR, RADAR, and LiDAR). 
As an example, Tesla uses several cameras, RADAR, and ultrasonics for their advanced driver-assistance system Autopilot.
Fusing information from several sensors allows exploiting their complementary characteristics and addressing the limitations of individual sensors, \eg the loss of structure information in cameras or missing color information in range data.

Wheel odometry measures the rotation of a wheel and can be used to estimate the distance covered by the autonomous vehicle. 
However, wheel odometry does not provide the full vehicle pose (\ie all six degrees of freedom) and is thus typically combined with visual odometry or SLAM techniques discussed in \chpref{chap:EgoMotionEstimation}.
Range sensors, \ie SONAR, RADAR, LiDAR, provide additional information about the geometry and structure of the scene.
Ultrasonic sensors (SONAR) emit high-frequency sound waves and measure the time for sound waves to travel to nearby objects.
The distance to objects is computed from the travel time since the speed of sound waves is known.
RADAR and LiDAR work with the same principle but use electromagnetic waves and laser light pulses instead of sound waves.
Because of the larger wavelength, RADAR sensors benefit from a larger working distance than LiDAR and SONAR but at the price of lower accuracy.

As cameras are cheap, passive, and easy to deploy, they are an attractive sensor choice for self-driving cars, and several existing driver assistance systems rely on cameras for lane keeping or pedestrian detection.
We now briefly discuss the most dominant camera types and give a short overview of popular calibration pipelines for estimating intrinsic and extrinsic sensor parameters.

\begin{figure}[t]
	\centering
	\includegraphics[width=\linewidth]{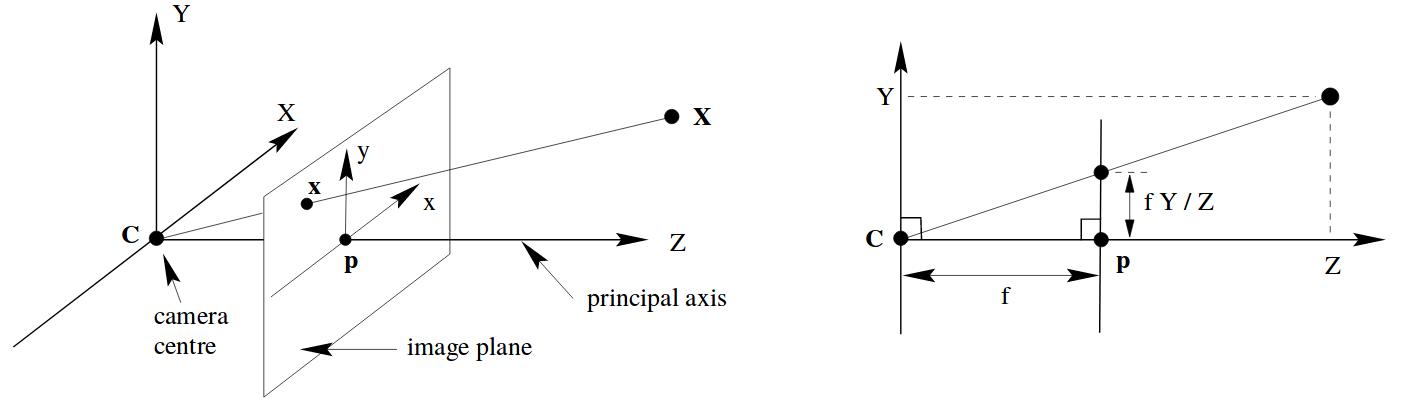}
	\caption[Pinhole Camera Model]{\textbf{Pinhole Camera Model.} In the pinhole model the three-dimensional world coordinates ($X,Y,Z$) are mapped to a two-dimensional image plane ($x,y$) using a perspective projection defined by the principle point ($p$) and focal length ($f$).  
	\figsourceC{\protect\citet{Hartley2004}}{2004}{Cambridge University Press}.}
	\label{fig:pinhole_model}
\end{figure}
\section{Camera Models}

Most conventional cameras comprise an aperture and one or multiple lenses and can be well approximated by the pinhole camera model (\figref{fig:pinhole_model}). Omnidirectional cameras allow to significantly increase the field of view by exploiting mirrors or special lenses.
Event cameras enable the acquisition of intensity changes at very high temporal resolutions. In the following, we provide a brief overview of omnidirectional and event cameras. We refer the reader to \cite{Szeliski2011,Hartley2004} for an in-depth discussion of the pinhole camera model and projective geometry.

\subsection{Omnidirectional Cameras}
\label{sec:calibration_omnidirectional_cam}
\begin{figure}[t]
	\centering
	\begin{subfigure}[t]{0.66\columnwidth}
		\centering
		\includegraphics[width=\linewidth]{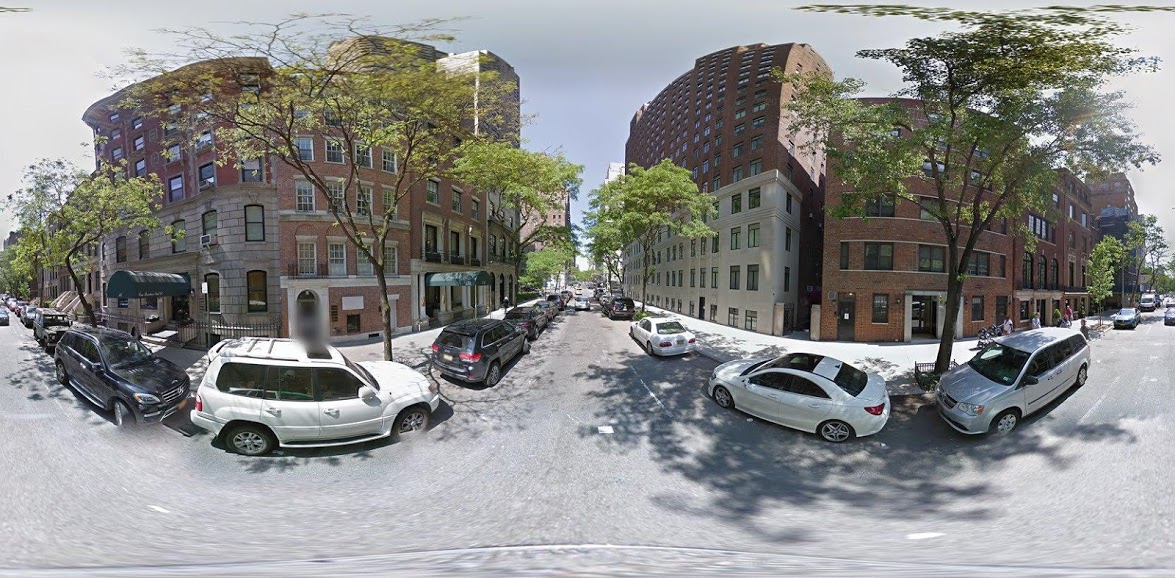}
		\caption{Equirectangular Projection}
		\label{fig:equi_view}
	\end{subfigure}%
	~ 
	\begin{subfigure}[t]{0.34\columnwidth}
		\centering
		\includegraphics[width=\linewidth]{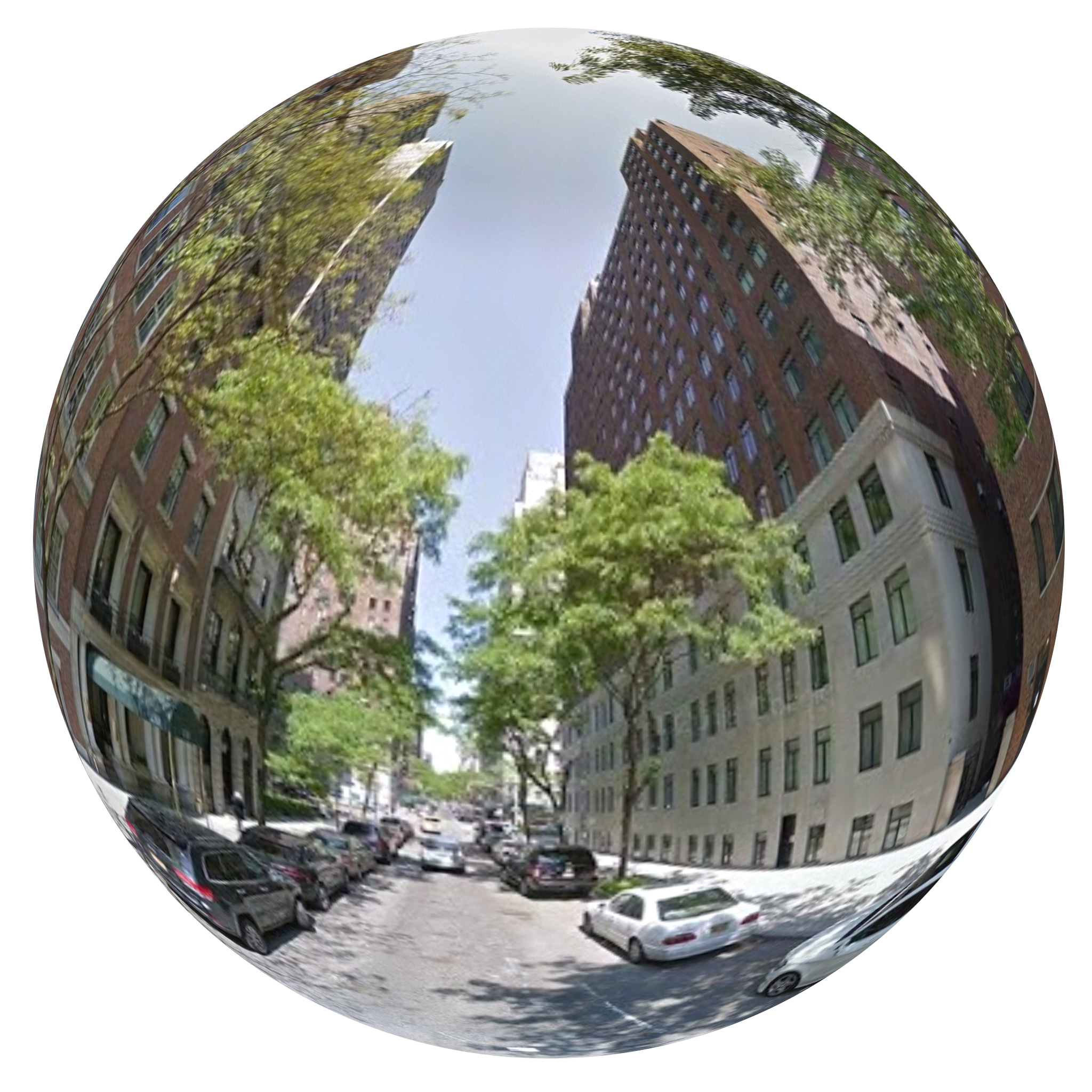}
		\caption{Spherical Projection} 
		\label{fig:spheric_view}
	\end{subfigure}
	\caption[Omnidirectional Cameras]{\textbf{Omnidirectional Cameras.} Equirectangular (\subref{fig:equi_view}) and spherical (\subref{fig:spheric_view}) view of a panorama from DeepMind’s StreetLearn research project \protect\citep{Mirowski2019ARXIV}. \figsourceC{DeepMind’s StreetLearn research project}{2020}{Google LCC} www.streetlearn.cc}
	\label{fig:omnidirectional_cams}
\end{figure}

A panoramic field of view is desirable in autonomous driving to gain maximum information about the surrounding area for safe navigation. Omnidirectional cameras with a 360-degree field of view (see \figref{fig:omnidirectional_cams}) provides enhanced coverage by eliminating the need for more cameras or mechanically turnable cameras. There are different types of omnidirectional cameras. Catadioptric cameras combine a standard camera with a shaped mirror, such as a parabolic, hyperbolic, or elliptical mirror, while dioptric cameras use purely dioptric fisheye lenses. Polydioptric cameras use multiple cameras with overlapping field of view to provide a full spherical field of view.

\citet{Geyer2000ECCV} provide a unifying theory for all central catadioptric systems which is known as unified projection model in the literature and widely used by different calibration toolboxes \citep{Mei2007ICRA, Heng2013IROS, Heng2015JFR}. \citet{Scaramuzza2006IROS} propose to model the imaging function using the Taylor series expansion. \citet{Mei2007ICRA} improve upon the unified projection model of \citep{Geyer2000ECCV} to account for real-world errors by modeling distortions. \citet{Schoenbein2014ICRA} propose a fast approximation to computationally expensive non-central camera models.

Omnidirectional cameras are gaining popularity in autonomous driving research. For feature-based applications such as navigation, motion estimation, and mapping, a large field of view enables the extraction and matching of interest points from all around the car. Thus, omnidirectional cameras have been successfully used to improve ego-motion estimation of vehicles \citep{Scaramuzza2008TR} and 3D reconstruction of static scenes \citep{Schoenbein2014IROS, Haene2014THREEDV}.

\subsection{Event Cameras}
\begin{figure}[t]
    \centering
    \begin{subfigure}[t]{0.4\columnwidth}
        \centering
        \includegraphics[width=\linewidth]{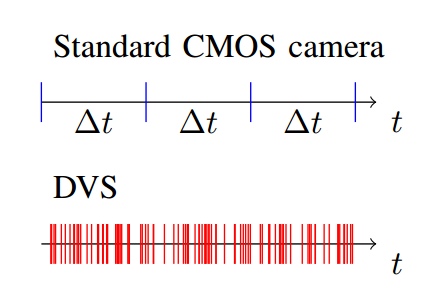}
        \caption{CMOS vs. DVS}
        \label{fig:event_cam_frame_rate}
    \end{subfigure}%
    ~ 
    \begin{subfigure}[t]{0.6\columnwidth}
        \centering
        \includegraphics[width=\linewidth]{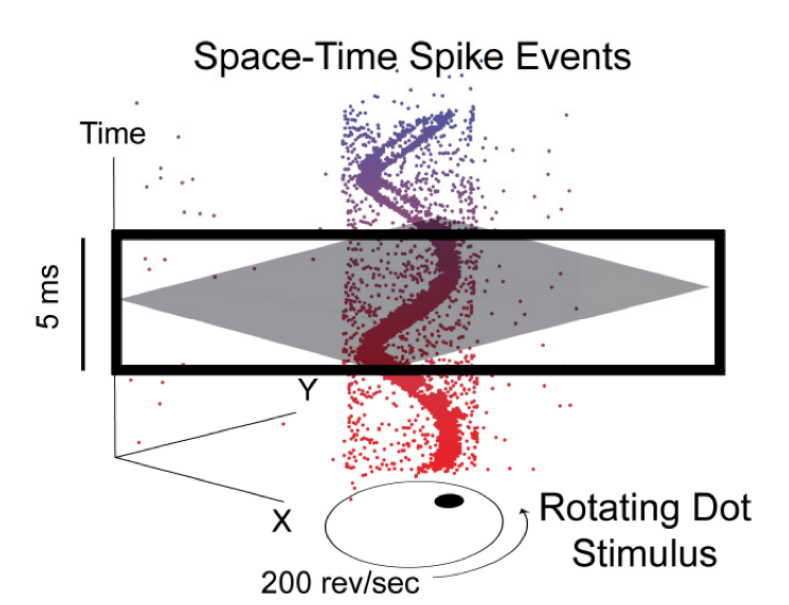}
        \caption{DVS with Stimulus} 
        \label{fig:event_cam_output}
    \end{subfigure}
    \caption[Event Cameras]{\textbf{Event Cameras.} \textbf{(\subref{fig:event_cam_frame_rate})} A standard CMOS camera sends images at a fixed frame rate (blue) while a Dynamic Vision Sensor (DVS) sends spike events at the time they occur (red). Each event corresponds to a local, pixel-level change of brightness. \figsourceC{\protect\citet{Mueggler2015RSS}}{2015}{RSS}.
    \textbf{(\subref{fig:event_cam_output})} Visualization of the output of a DVS looking at a rotating dot. Colored dots mark individual events. Events that are not part of the spiral are caused by sensor noise. \figsourceElsevier{\protect\citet{Liu2010CON}}{2010}.}
    \label{fig:event_cam}
\end{figure}
Contrary to conventional frame-based cameras,
event cameras produce a stream of asynchronous events of brightness changes surpassing a pre-defined threshold at microsecond resolution, as illustrated in \figref{fig:event_cam}. An event comprises the location, sign, and timestamp of the change. As events are sparse in both space and time, this representation has the potential to reduce transmission and processing demands. The high temporal resolution enables the development of highly reactive systems.

Dynamic and Active-Pixel Vision Sensors (DAVIS) output both CMOS images at fixed frame rates as well as asynchronous events, hence combining the benefits of both sensors. \citet{Mueggler2017IJRR} provide a collection of real and synthetic datasets captured with DAVIS to push research on event-based methods. \citet{Binas2017ICMLWORK} present the DAVIS Driving Dataset and demonstrate end-to-end learning of steering angles. Recent work exploits DAVIS for feature tracking \citep{Gehrig2018ECCV} and SLAM \citep{Vidal2018RAL}, improving accuracy and robustness over using only a single modality.

Several methods have been developed which exploit the high temporal resolution and the asynchronous nature of event sensor for various problems. The majority of these methods focus on the application in unmanned aerial vehicles (UAVs) since very efficient methods are necessary to navigate these systems. In this context, event-based cameras have been used for ego-motion estimation \citep{Mueggler2015RSS}, simultaneous localization and mapping (SLAM) \citep{Rebecq2016RAL} as well as for finding feature correspondences \citep{Gallego2018CVPR}. More recently, the benefits of event-based sensors have been exploited for autonomous vehicles by learning steering angles end-to-end \citep{Maqueda2018CVPR}.

\section{Calibration}
\label{sec:calibration}
Geometric calibration is the problem of estimating intrinsic and extrinsic parameters of one or multiple sensors in order to accurately relate 3D world points to 2D measurements. Fiducial markers and checkerboards are often used to facilitate parameter estimation \citep{Zhang2004IROS,Bouguet2010,Kassir2010ACRA,Andreasson2010RAS,Geiger2012ICRA}.

Various methods for camera calibration can be found since the beginning of the 1970s. \citet{Heikkila1997CVPR} were the first to consider the entire calibration pipeline, including control point extraction, model fitting, and image correction. They proposed a four-step procedure to obtain the parameters of a physical camera model and address the problem of compensating image distortions.

Modern vehicles are typically equipped with multiple different sensors with the goal of increasing robustness and coverage. Several calibration procedures have been proposed to address the needs of such big sensor suites. While early approaches \citep{Zhang2004IROS, Bouguet2010} rely on manual extraction of interest points in laser scans, \citet{Kassir2010ACRA} and \citet{Andreasson2010RAS} propose the first complete automatic camera-to-range calibration systems. \citet{Geiger2012ICRA} demonstrate how to automatically calibrate a setup involving two cameras and a single range sensor such as Kinect or Velodyne laser scanner. \citet{Heng2013IROS} tackle the problem of estimating the intrinsic and extrinsic parameters of a multi-camera rig without overlapping field of view. \citet{Heng2015JFR} extend this work by removing the requirement to modify the environment by using a map and natural features instead of fiducial markings.
	\chapter{Datasets \& Benchmarks}
\label{chap:Datasets}
Datasets have played a key role in the progress of many research fields by providing problem-specific examples with ground truth. Quantitative evaluations of different approaches provide key insights about their capacities and limitations. Landmark examples in the field of computer vision include the Middlebury benchmarks for stereo and optical flow \cite{Scharstein2002IJCV} and the PASCAL VOC object recognition challenges \cite{Everingham2010IJCV}.
In particular, many of these datasets \citep{Scharstein2002IJCV,Everingham2010IJCV,Baker2011IJCV,Geiger2012CVPR,Butler2012ECCV,Leal-Taixe2015ARXIV,Cordts2016CVPR,Kondermann2016CVPRWORK,Schoeps2017CVPR} also provide online evaluation servers that allow for a fair comparison on held-out test sets and provide researchers in the field an up-to-date overview over the state of the art. 
This way, current progress and remaining challenges can be easily identified by the research community.

In the context of autonomous vehicles, \citep{Geiger2012CVPR, Cordts2016CVPR, Kondermann2016CVPRWORK, Neuhold2017ICCV, Aly2008IV, Dollar2009CVPR, Leal-Taixe2015ARXIV} have introduced challenging benchmarks for reconstruction, motion estimation, recognition tasks, and tracking, and contributed to closing the gap between laboratory settings and challenging real-world situations.
\citet{Kang2019TIV} provide a detailed overview of different datasets and testing environments in the context of autonomous driving. 

\begin{table*}[p]
	\begin{center}
		\begin{adjustbox}{width=1\textwidth}{
\setlength{\arrayrulewidth}{.1em}
\begin{tabular}{l|c|c|c|c|c|c|c|c|c|c|c|c|c}
	{\textbf{Dataset}} & \rotatebox{90}{{\textbf{Realism}}} & \rotatebox{90}{{\textbf{Diversity}}} & \rotatebox{90}{{\textbf{Autonomous Driving}}} & \rotatebox{90}{{\textbf{Evaluation Server}}} & \rotatebox{90}{{\textbf{Stereo}}} & \rotatebox{90}{{\textbf{Reconstruction}}} & \rotatebox{90}{{\textbf{Optical Flow}}} & \rotatebox{90}{{\textbf{Object Detection}}} & \rotatebox{90}{{\textbf{Traffic Sign Detection}}} & \rotatebox{90}{{\textbf{Semantic Segmentation}}} & \rotatebox{90}{{\textbf{Road Detection}}} & \rotatebox{90}{{\textbf{Lane Detection}}} & \rotatebox{90}{{\textbf{Tracking}}} \\ 
	\hline
	Middlebury \citep{Scharstein2002IJCV} &  +  &  -{}-   &  &  \cmark  &  XS  &  XS  &  XS  &    &  &    &    &    &  \\ 
	\specialrule{.05em}{0em}{0em} 
	EPFL Multi-View \citep{Strecha2008CVPR}  &  ++  &  +  &  &  \cmark  &    &  XS  &    &    &  &    &    &    &  \\ 
	\specialrule{.05em}{0em}{0em} 
	DTU MVS \citep{Jensen2014CVPR}  &  +  &  -  &  &    &     &  S  &    &    &  &    &    &    &  \\ 
	\specialrule{.05em}{0em}{0em} 
	ETH3D \citep{Schoeps2017CVPR}  &  ++  &  +  &  &  \cmark  &  S  &  S  &    &    &  &    &    &    &  \\ 
	\specialrule{.05em}{0em}{0em} 
	Tanks and Temples \citep{Knapitsch2017SIGGRAPH}  &  ++  &  +  &  &  \cmark  &    &  S  &    &    &  &    &    &    &  \\ 
	\hline
	SlowFlow \citep{Janai2017CVPR}  &  ++  &  ++   &  &     &    &    &  S  &    &  &    &    &    &  \\ 
	\specialrule{.05em}{0em}{0em} 
	HCI Benchmark \citep{Kondermann2016CVPRWORK}  &  ++  &  +   &  \cmark  &  \cmark  &    &    &  M  &    &  &    &    &     &  \\ 
	\specialrule{.05em}{0em}{0em} 
	MPI Sintel \citep{Butler2012ECCV}  &  O  &  +   &  &  \cmark  &  M  &    &  M  &    &  &    &    &    &  \\ 
	\specialrule{.05em}{0em}{0em} 
	Flying Chairs \citep{Dosovitskiy2015ICCV}  &  -{}-  &  -{}-  &  &    &    &    &  L  &    &  &    &    &    &  \\ 
	\specialrule{.05em}{0em}{0em} 
	Flying Things \citep{Mayer2016CVPR}  &  -  &  O  & ( \cmark ) &    &  L  &    &  L  &    &  &    &    &    &  \\ 
	\hline
	ImageNet \citep{Deng2009CVPR}  &  ++  &  ++  &  &     &     &    &    &  XL  &  &  XL  &    &    &  \\ 
	\specialrule{.05em}{0em}{0em} 
	PASCAL VOC \citep{Everingham2010IJCV}  &  ++  &  ++  &  &     &     &    &    &  XL  &  &  XL  &    &    &  \\ 
	\specialrule{.05em}{0em}{0em} 
	Microsoft Coco \citep{Lin2014ECCV}  &  ++  &  ++  &  &     &    &    &    &  XL  &  &  XL  &    &    &  \\ 
	\specialrule{.05em}{0em}{0em} 
	Cityscapes \citep{Cordts2016CVPR}  &  ++  &  +   &  \cmark  &  \cmark  &    &    &    &  L  &  &  L  &    &    &  \\ 
	\specialrule{.05em}{0em}{0em} 
	EuroCity Persons Dataset \citep{Braun2019PAMI}  &  ++  &  ++  &  \cmark  &  \cmark  &    &    &    &  L  &  &   &    &    &  \\ 
	\specialrule{.05em}{0em}{0em} 
	Mapillary \citep{Neuhold2017ICCV}  &  ++  &  ++  &  \cmark  &     &    &    &    &    &  &  L  &    &    &  \\ 
	\specialrule{.05em}{0em}{0em} 
	ApolloScape \citep{Huang2018CVPR}  &  ++  &  +  &  \cmark  &    &     &    &    &  L  &  &  XL  &    &  XL  &  XL  \\ 
	\specialrule{.05em}{0em}{0em} 
	NuScenes \citep{Caesar2019ARXIV}  &  ++  &  +  &  \cmark  &    &     &    &    &  XL  &  &  XL  &    &    &  \\ 
	\specialrule{.05em}{0em}{0em} 
	Berkeley DeepDrive \citep{Yu2018ARXIV}  &  ++  &  +  &  \cmark  &    &     &    &    &  XL  &  &  XL  &  XL  &  XL  &  \\ 
	\specialrule{.05em}{0em}{0em} 
	German Traffic Sign Recognition Benchmark \citep{Stallkamp2011IJCNN}  &  ++  &  +  &  \cmark  &  \cmark  &     &    &    &  XL  & L &  XL  &  XL  &  XL  &  \\ 
	\specialrule{.05em}{0em}{0em} 
	German Traffic Sign Detection Benchmark \citep{Houben2013IJCNN}  &  ++  &  +  &  \cmark  &  \cmark  &     &    &    &  XL  & M &  XL  &  XL  &  XL  &  \\ 
	\specialrule{.05em}{0em}{0em} 
	Tsinghua-Tencent 100K \citep{Zhu2016CVPR}  &  ++  &  +  &  \cmark  &    &     &    &    &  XL  &  XL  &  XL  &  XL  &  XL  &  \\ 
	\specialrule{.05em}{0em}{0em} 
	SYNTHIA \citep{Ros2016CVPR}  &  O  &  +  &  \cmark  &   &    &    &    &    &  &  XL  &    &    &  \\ 
	\specialrule{.05em}{0em}{0em} 
	Playing for Data \citep{Richter2016ECCV}   &  +  &  +  &  \cmark  &    &    &    &    &    &  &  L  &    &    &  \\ 
	\specialrule{.05em}{0em}{0em} 
	Playing for Benchmarks \citep{Richter2017ICCV}   &  +  &  +  &  \cmark  &  \cmark &    &    &  XL  &  XL  &  &  XL  &    &    & XL \\ 
	\hline
	Caltech Lanes Dataset \citep{Aly2008IV}   &  ++  &  +  &  \cmark  &    &    &    &    &    &  &    &    &  M  &  \\ 
	\specialrule{.05em}{0em}{0em} 
	VPGNet Dataset \citep{Lee2017ICCV}   &  ++  &  +  &  \cmark  &    &    &    &    &    &  &    &    &  L  &  \\ 
	\hline
	MOTChallenge \citep{Leal-Taixe2015ARXIV}  &  ++  &  +  &  &  \cmark   &    &    &    &    &  &    &    &    &  M  \\ 
	\specialrule{.05em}{0em}{0em} 
	Caltech Pedestrian Detection \citep{Dollar2009CVPR}  &  ++  &  +  &  \cmark  &    &    &    &    &    &  &    &    &    &  XL  \\ 
	\specialrule{.05em}{0em}{0em} 
	Argoverse \citep{Chang2019CVPR}  &  ++  &  +  &  \cmark  &  \cmark  &    &    &    &    &  &    &    &    &  L  \\ 
	\specialrule{.05em}{0em}{0em} 
	Waymo Open Dataset \citep{Sun2019ARXIVa}  &  ++  &  +  &  \cmark  &  \cmark  &    &    &    &    XL &  & XL  &    &    &  XL  \\ 
	\specialrule{.05em}{0em}{0em} 
	\specialrule{.05em}{0em}{0em} 
	\hline
	KITTI \citep{Geiger2012CVPR}  &  ++  &  +  &  \cmark  &  \cmark  &  S  &  S  &  S  &  M  &  &  S  &  S  &  S  &  M \\ 
	\specialrule{.05em}{0em}{0em} 
	VirtualKITTI \citep{Gaidon2016CVPR}  &  O  &  +  &  \cmark  &  \cmark  &  S  &  S  &  L  &  L  &  &  L  &    &    &  L \\ 
\end{tabular}
}

\end{adjustbox}
	\end{center}
	\vspace{-0.4cm}
	\caption[Popular Datasets for Computer Vision and Self-Driving]{{\bf Popular Datasets in Computer Vision and Self-Driving.} Overview of popular datasets for Stereo, Reconstruction, Optical Flow, Object Detection, Traffic Sign Detection, Semantic Segmentation, Road Detection, Lane Detection, Tracking. Datasets specific to the autonomous driving scenario are marked with a checkmark in the corresponding column. The size of extra small datasets (XS) are in the order of tens examples/scenes for training, small sized (S) in the order of hundreds, medium sized (M) in the order of thousands, large (L) and extra large (XL) sized datasets in the order of 10 and \textgreater100 thousands, respectively. We (subjectively) rate realism and diversity with \{-{}-,-,O,+,++\} from low to high.
	}
	\label{tab:datasets}
\end{table*}

Only a few years ago, datasets with a few hundred annotated examples were considered sufficient for many problems. The introduction of datasets with many hundred to thousands of labeled examples has led to spectacular breakthroughs in many computer vision disciplines by allowing to train high-capacity deep models in a supervised fashion. However, collecting a large amount of annotated data is not an easy endeavor, in particular for tasks such as optical flow or semantic segmentation where pixel-level annotations are required. For optical flow, \citet{Scharstein2002IJCV,Baker2011IJCV} acquire dense pixel-level annotations in a controlled lab environment using a time-consuming procedure whereas \citet{Geiger2012CVPR,Kondermann2016CVPRWORK} are only able to provide sparse pixel-level annotations of real street scenes using a LiDAR laser scanner. \citet{Janai2017CVPR} pursued a different approach to obtain dense pixel-level annotations in arbitrary real scenes by using a high-speed camera to solve the optical flow problem in a simpler setting. Recently, crowdsourcing with Amazon’s Mechanical Turk platform\footnote{\url{https://www.mturk.com/mturk/welcome}} has been popularized for annotating large scale datasets, \eg \citep{Deng2009CVPR,Lin2014ECCV,Leal-Taixe2015ARXIV,Milan2016ARXIV,Dollar2009CVPR}. However, the annotation quality obtained via Mechanical Turk is often not sufficient and significant efforts in post-processing and clean-up are typically required.

An alternative to manual annotation is offered by modern computer graphic techniques which allow generating large-scale synthetic datasets with pixel-level ground truth. However, the creation of photorealistic virtual worlds is time-consuming and expensive. Nevertheless, the popularity of movies and video games has led to an industry creating very realistic 3D content which nourishes the hope to replace real data completely using synthetic datasets. Consequently, several synthetic datasets \citep{Butler2012ECCV,Dosovitskiy2015ICCV,Mayer2016CVPR,Gaidon2016CVPR, Ros2016CVPR} have been proposed and are being used by AI researchers. It remains an open question, however, whether the realism and variety attained will be sufficient to replace real-world datasets and if models trained on synthetic data will be able to generalize to real-world inputs. Challenges include complex object shape and appearances as well as adversarial environmental conditions such as direct lighting, reflections from specular surfaces, fog, or rain.

Studying the performance of a system over time, \eg in case of environmental changes or rare situations, is another important aspect for autonomous vehicles. In \secref{sec:long_term_autonomy} we discuss several recent datasets for long-term autonomy. While most of these datasets focus on environmental changes, it is more difficult to capture rare situations which can only be captured with a large fleet of vehicles that log these situations in real-world driving. A notable exception is the Tesla Shadow Mode \citep{TeslaSoftware9} of the Autopilot system which is a dormant logging-only mode that allows validating the Autopilot system running in the background in real and particularly rare situations.

In the following, we will first introduce the most popular computer vision datasets and benchmarks addressing tasks relevant to autonomous vehicles. Thereafter, in \secref{sec:datasets_autonomous_driving}, we will focus on datasets particularly dedicated to autonomous vehicles. We also provide a detailed overview of the most popular datasets in computer vision in \tabref{tab:datasets} and discuss them in the following.

\begin{figure}[t]
	\centering
	\includegraphics[width=1.00\columnwidth]{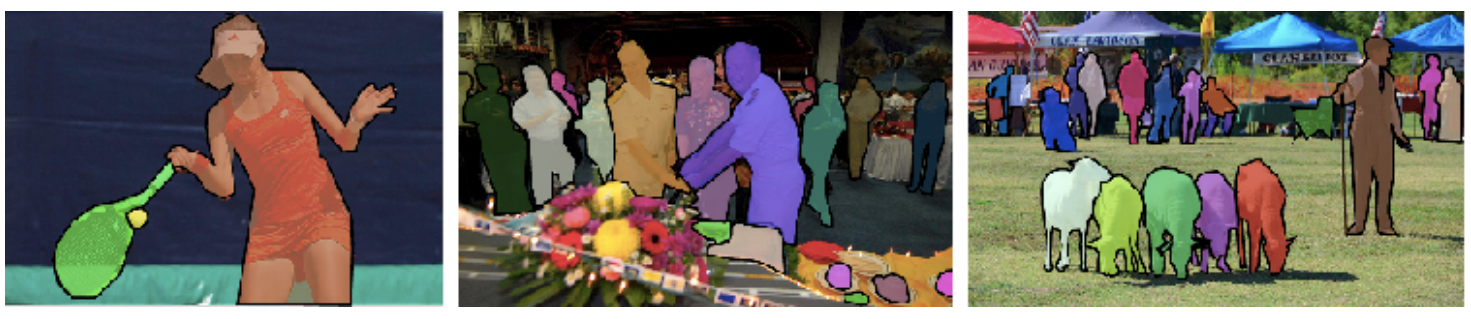}
	\caption[MS COCO Object Recognition Dataset]{\textbf{MS COCO Object Recognition Dataset.} Examples from the MS COCO \protect\citep{Lin2014ECCV} object detection task. \courtesyC{\url{www.cocodataset.org}}{2015}{COCO Consortium}.}
	\label{fig:Lin2014ECCV}
\end{figure}

\section{Computer Vision Datasets}
In this section, we introduce the most popular computer vision datasets and benchmarks relevant to autonomous driving tasks. In particular, we discuss datasets for object recognition and tracking, stereo and 3D reconstruction, and optical flow estimation.

\subsection{Object Recognition}
\label{sec:datasets_recognition}
The availability of large-scale, publicly available datasets such as ImageNet \citep{Deng2009CVPR}, PASCAL VOC \citep{Everingham2010IJCV} and Microsoft COCO \citep{Lin2014ECCV} propelled the development of novel computer vision algorithms, in particular, deep learning techniques, for recognition tasks such as object classification, detection, and semantic segmentation. 

The EU funded PASCAL Visual Object Classes (VOC) challenge\footnote{\url{http://host.robots.ox.ac.uk/pascal/VOC/}} by \citet{Everingham2010IJCV} is a benchmark for object classification, object detection, object segmentation, and action recognition. It consists of challenging consumer photographs collected from Flickr with high-quality annotations and contains a large variability in pose, illumination, and occlusion. Since its introduction, the VOC challenge has become one of the most popular testbeds for benchmarking recognition algorithms. It has been regularly adapted to the needs of the community until the end of the PASCAL program in 2012. Over the years, the benchmark grew in size, reaching a total of 11,530 images with 27,450 annotated objects in 2012.

In 2014, \citet{Lin2014ECCV} introduced the Microsoft COCO dataset\footnote{\url{http://mscoco.org/}} (\figref{fig:Lin2014ECCV}) for object detection, instance segmentation, and contextual reasoning. They provide images of complex everyday scenes containing common objects in their natural context. The dataset comprises 91 object classes, 2.5 million annotated instances, and 328k images in total.
Microsoft COCO is significantly larger in the number of instances per class than the PASCAL VOC object segmentation benchmark. All objects have been annotated with per-instance segmentations. 

ImageNet \citep{Deng2009CVPR}, PASCAL VOC \citep{Everingham2010IJCV} and Microsoft COCO \citep{Lin2014ECCV} are to date the largest and most diverse datasets for object classification, detection, and segmentation (\tabref{tab:datasets}). 

\subsection{Object Tracking}
For tracking multiple objects, the first centralized benchmark, MOTChallenge\footnote{\url{https://motchallenge.net/}}, was introduced by  \citet{Leal-Taixe2015ARXIV, Milan2016ARXIV}. The benchmark contains 14 challenging video sequences in unconstrained environments filmed with static and moving cameras. MOTChallenge combines several existing multi-object tracking benchmarks such as PETS \citep{Ferryman2009PETS} and KITTI \citep{Geiger2012CVPR}. Public detections provided by the benchmark allow analyzing the performance of tracking systems independent of the detector.

\begin{figure}[t]
	\centering
	\includegraphics[width=1.00\columnwidth]{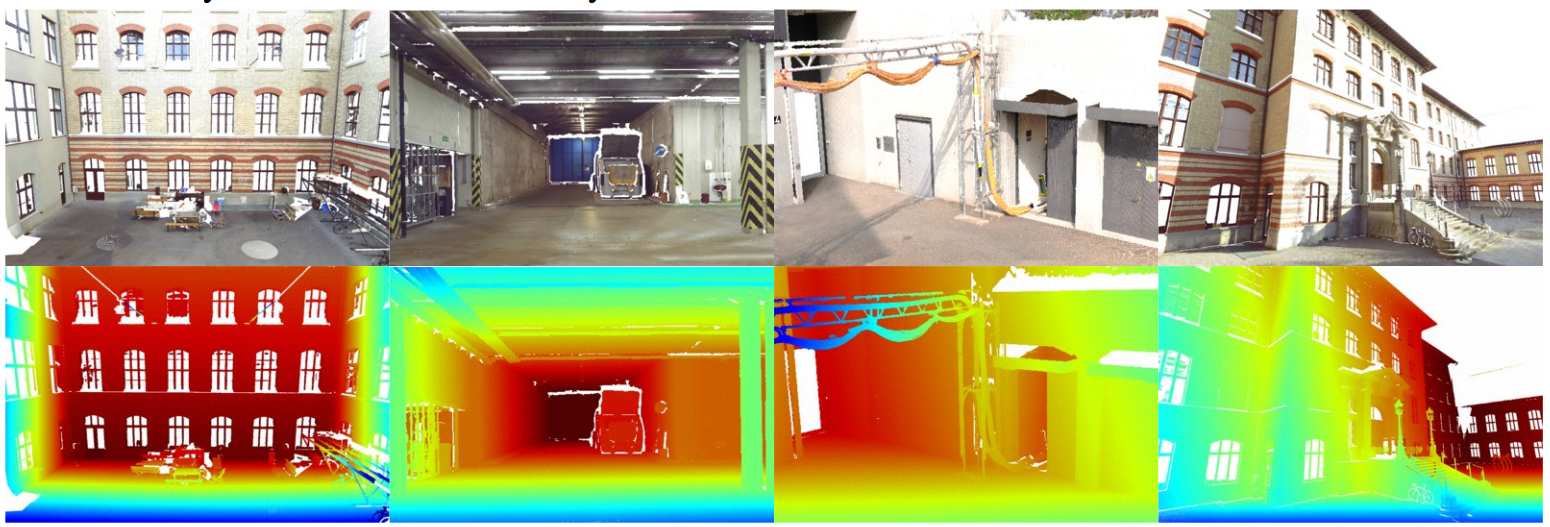}
	\caption[ETH3D Reconstruction Dataset]{\textbf{ETH3D Reconstruction Dataset.} Examples from the ETH3D \protect\citep{Schoeps2017CVPR} dataset. Colored 3D point cloud renderings in the upper row and depth in the lower row. \courtesy{\url{www.eth3d.net}}.}
	\label{fig:Schoeps2017CVPR}
\end{figure}

\subsection{Stereo and 3D Reconstruction}
\label{sec:stereo_datasets}
For stereo vision and multi-view reconstruction, there are several publicly available datasets. The Middlebury stereo benchmark\footnote{\url{http://vision.middlebury.edu/stereo/}} introduced by \citep{Scharstein2002IJCV,Scharstein2003CVPR,Scharstein2014GCPR} was proposed with the goal of providing a unified testbed for a fair comparison of stereo matching algorithms. An evaluation server was created, allowing for a direct comparison of the latest approaches. The success of the Middlebury stereo benchmark in fostering research in binocular vision motivated \citet{Seitz2006CVPR} to create the Middlebury multi-view stereo (MVS) benchmark\footnote{\url{http://vision.middlebury.edu/mview/}}. The dataset consists of calibrated high-resolution multi-view images with registered 3D ground truth models and played a key role in advancing research in MVS.

However, the Middlebury datasets lack in size and diversity in comparison to other datasets for stereo and reconstruction (\tabref{tab:datasets}). The DTU MVS dataset\footnote{\url{http://roboimagedata.compute.dtu.dk/?page_id=36}} by \citet{Jensen2014CVPR} provides 124 different scenes which were recorded in a controlled laboratory environment.  Reference data is obtained by combining structured light scans from different camera positions. 
While the DTU MVS dataset is more diverse than Middlebury in terms of the number of objects used as well as their complexity, neither of these two datasets exhibits the full spectrum of complexities of real-world scenes.

With the goal of moving multi-view stereo out of the laboratory, \citet{Strecha2008CVPR} presented the EPFL Multi-View dataset  \footnote{\url{https://www.epfl.ch/labs/cvlab/data/data-strechamvs/}}, which comprises images and LiDAR scans of 5 different buildings as well as a fountain.

Recently, \citet{Schoeps2017CVPR} published the ETH3D \footnote{\url{https://www.eth3d.net}} dataset (\figref{fig:Schoeps2017CVPR}) providing high-resolution DSLR imagery as well as synchronized low-resolution stereo videos for a variety of indoor and outdoor scenes. They used a high-precision laser scanner as \citep{Strecha2008CVPR} and registered all images using a robust optimization technique.

Similarly, Tanks and Temples\footnote{\url{https://www.tanksandtemples.org}} presented by \citet{Knapitsch2017SIGGRAPH} used a high-precision laser scanner and two high-resolution cameras (one with global and the other with rolling shutter) to create a novel dataset of outdoor and indoor scenes. The dataset consists of 14 scenes comprising sculptures, large vehicles, house-scale buildings as well as large indoor and outdoor scenes.

For large-scale reconstruction, multiple Internet photo collections have been proposed over time. The most popular collections are combined in the BigSFM dataset \footnote{\url{http://www.cs.cornell.edu/projects/bigsfm/}} and comprise Vienna \citep{Irschara2009CVPR}, Dubrovnik \citep{Li2010ECCV}, and Rome \citep{Crandall2011CVPR}. While Dubrovnik and Rome were retrieved from Flickr, Vienna was recorded with a calibrated camera. Besides large-scale reconstruction, these datasets are also frequently used for evaluating loop-closure detection (\secref{sec:LoopClosure}) and localization methods (\secref{sec:Localization}).

\subsection{Optical Flow}
\label{sec:dataset_optical_flow}
Similar to stereo vision, the Middlebury flow benchmark\footnote{\url{http://vision.middlebury.edu/flow/}} by \citet{Baker2011IJCV} provided the first unified test environment and evaluation server for optical flow approaches. The benchmark comprises sequences with non-rigid motion, synthetic sequences, and a subset of the Middlebury stereo benchmark (static scenes). For all non-rigid sequences, ground truth flow is obtained by tracking hidden fluorescent textures sprayed onto the objects. In comparison to other optical flow datasets (\tabref{tab:datasets}), the Middlebury flow dataset is limited in size and missing real-world challenges like complex structures, lighting variation, and shadows due to the laboratory conditions in which it has been recorded. In addition, Middlebury only contains small motions of up to twelve pixels which do not allow the investigation of challenges related to fast motions.

The acquisition of optical flow ground truth is very difficult since no sensor exists that can capture optical flow ground-truth in general natural scenes. While \citep{Geiger2012CVPR,Kondermann2016CVPRWORK} use a LiDAR laser scanner for this purpose, they only obtain sparse pixel-level annotations and are restricted to static scenes (only camera motion). 
\citet{Janai2017CVPR} present a novel approach to obtain accurate reference data from High-Speed video cameras by tracking pixels through densely sampled space-time volumes. This method allows the acquisition of optical flow ground truth in challenging everyday scenes and the data augmentation with realistic effects such as motion blur to compare methods in varying conditions. \citet{Janai2017CVPR} provide 160 diverse real-world sequences of dynamic scenes with a significantly larger resolution ($1280 \times 1024$ pixels) than previous optical datasets. 

The problem of acquiring optical flow ground truth can also be resolved by creating synthetic datasets. Towards this goal, \citet{Butler2012ECCV} take advantage of the open-source movie Sintel, a short animated film. They create the MPI Sintel optical flow benchmark\footnote{\url{http://sintel.is.tue.mpg.de/}} by rendering scenes with optical flow ground truth. 
Sintel consists of 1,628 frames and provides three different datasets with varying complexity that are obtained using different passes of the rendering pipeline. Similar to Middlebury, they provide an evaluation server for comparison.

The limited size of optical flow datasets hampered the training of deep high-capacity models. Thus, \citet{Dosovitskiy2015ICCV} introduced a simple synthetic 2D dataset of flying 3D chairs rendered on top of random background images from Flickr to train a convolutional neural network. As the limited realism of this dataset proved insufficient to learn highly accurate models, \citet{Mayer2016CVPR} presented another large-scale dataset consisting of three synthetic stereo video datasets with optical flow ground truth: FlyingThings3D, Monkaa, and Driving. FlyingThings3D provides everyday 3D objects flying along randomized 3D trajectories in a randomly created scene. Inspired by the KITTI dataset, a driving dataset has been created which uses car models from the same pool as FlyingThings3D and additionally highly detailed tree and building models from 3D Warehouse. Monkaa is an animated short movie similar to Sintel used in the MPI Sintel benchmark.

While synthetic optical flow datasets provide numerous examples for training deep neural networks, they lack realism and are limited in diversity, as indicated in \tabref{tab:datasets}. Therefore, large-scale synthetic datasets are typically used for pre-training, and, afterwards, the pre-trained models are fine-tuned on small, more realistic datasets.

\begin{figure}[t]
	\centering
	\includegraphics[width=1.00\columnwidth]{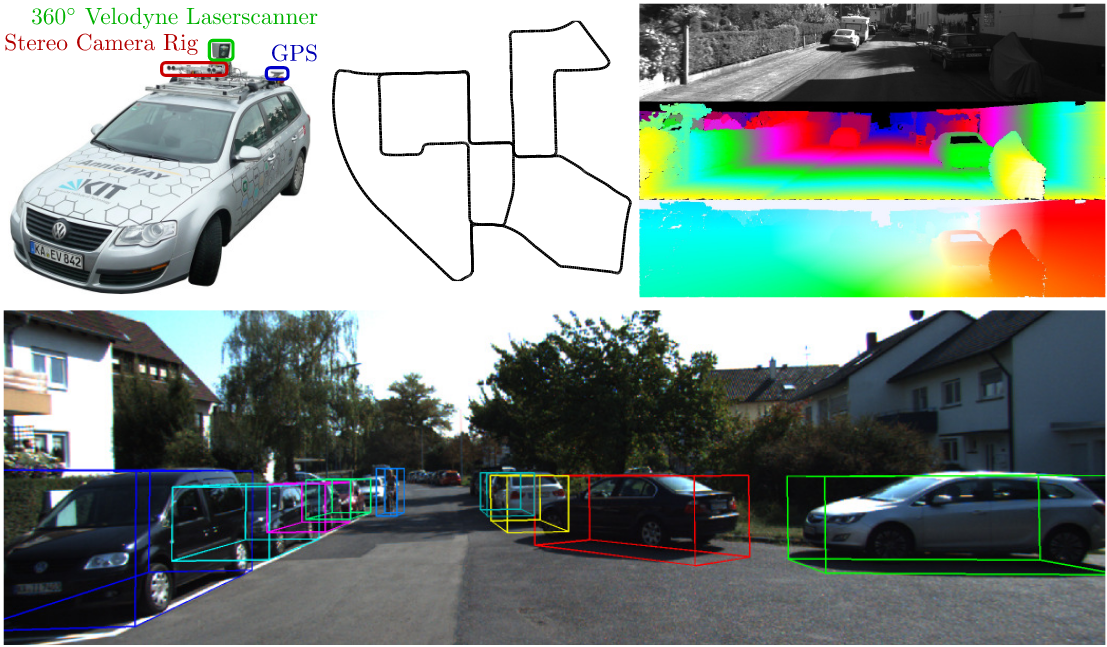}
	\caption[KITTI Dataset]{\textbf{KITTI Dataset.} The recording platform with sensors (top left), trajectory (top center), disparity and optical flow (top right) and 3D object labels (bottom) from the KITTI benchmark proposed by \protect\citet{Geiger2012CVPR}. \figsourceC{\protect\citet{Geiger2012CVPR}}{2012}{IEEE}.}
	\label{fig:Geiger2012CVPR}
\end{figure}

\section{Autonomous Driving Datasets}
\label{sec:datasets_autonomous_driving}
Several datasets have been proposed to specifically address the problem of autonomous driving. The KITTI Vision Benchmark\footnote{\url{http://www.cvlibs.net/datasets/kitti/}} introduced by \citet{Geiger2012CVPR, Geiger2013IJRR} was the first publicly available benchmark for stereo, optical flow, visual odometry/SLAM, and 3D object detection (\figref{fig:Geiger2012CVPR}) in the autonomous driving context. The dataset has been captured from an autonomous driving platform equipped with high-resolution color and grayscale stereo cameras, a Velodyne 3D laser scanner, and high-precision GPS/IMU inertial navigation system. 

Due to the limitations of the rotating laser scanner used as reference sensor, the stereo and optical flow benchmark were restricted to static scenes with camera motion. In the 2015 version of the optical flow and stereo Benchmark, \citet{Menze2015CVPR} provide ground truth for dynamic scenes by fitting 3D CAD models to all vehicles in motion. This new version of KITTI also combined the stereo and flow ground truth to form a novel 3D scene flow benchmark.
For the KITTI object detection challenge, a special 3D labeling tool has been developed to annotate all 3D objects with 3D bounding boxes in 7481 training and 7518 test images. The benchmark for object detection was separated into a vehicle, pedestrian and cyclist detection tasks, allowing to focus the analysis on the most important problems in the context of autonomous vehicles. The visual odometry / SLAM challenge consists of 22 stereo sequences, with a total length of 39.2 km. The ground truth pose is obtained by using GPS/IMU localization unit which was fed with RTK correction signals. 

The KITTI dataset has established itself as one of the standard benchmarks in all of the aforementioned tasks, in particular in the context of autonomous driving applications.
While KITTI provides annotated data and an evaluation server for all problems considered in this work (\tabref{tab:datasets}), it is still comparably limited in size. Therefore, the KITTI dataset is usually used mostly for evaluation and fine-tuning.

Very recently, major companies working on autonomous driving solutions also started making their annotated data publicly available. The autonomous driving project Apollo from Baidu created the Data Open Platform\footnote{\url{http://data.apollo.auto}} consisting of simulation, annotation, and demonstration data for autonomous driving. The ApolloScape dataset\citep{Huang2018CVPR} provides  annotated street view images for semantic (144K images) and instance segmentation (90K images), lane detection (160K images), car detection (70K) and tracking of traffic participants (100K images). The dataset allows evaluating the performance of methods in various weather conditions and at different day times.

The company Nutonomy released the NuScenes dataset\footnote{\url{https://www.nuscenes.org}} \citep{Caesar2019ARXIV}, which provides data for semantic segmentation and object detection. The dataset consists of over 1 million camera images. However, both ApolloScape and NuScenes have been recorded only in one or two cities, respectively, and are therefore still limited in diversity.

ARGO AI \citep{Chang2019CVPR} presented Argoverse\footnote{\url{https://www.argoverse.org}}, a novel 3D object tracking dataset. A fleet of autonomous vehicles collected different sensor data, such as $360^{\circ}$ images, forward-facing stereo imagery, LiDAR, and 6-DOF pose. They also provide 290km of lane markings and 10k human-annotated tracked objects.

Finally, Waymo Open Dataset\footnote{\url{https://waymo.com/open/}} \citep{Sun2019ARXIVa} and Lyft Level 5 AV Dataset\footnote{\url{https://level5.lyft.com/dataset/}} \citep{LyftDataset} were presented providing semantic annotations and object detections for many driving scenarios. Both companies started challenges to push the research in 2D and 3D object detection.

\begin{figure}[t]
	\centering
	\includegraphics[width=1.00\columnwidth]{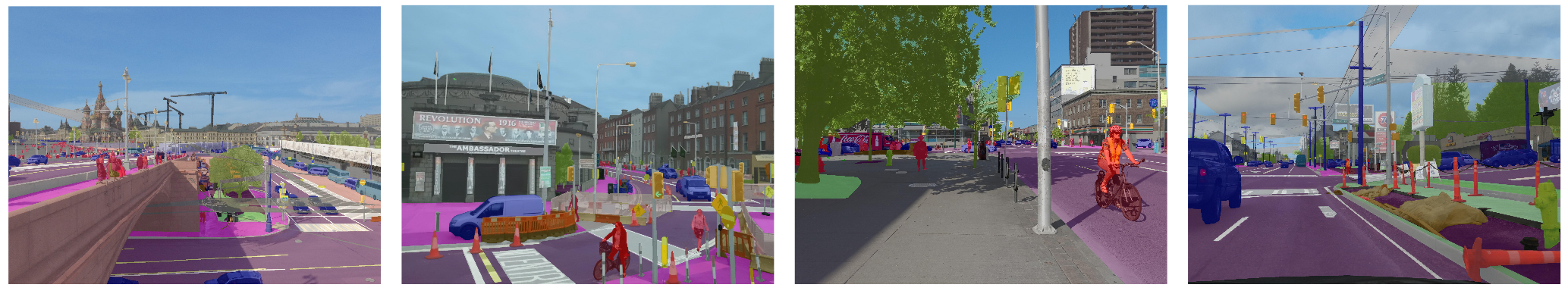}
	\caption[Mapillary Vistas Dataset]{\textbf{Mapillary Vistas Dataset.} Examples colorized according to the class definition of the Mapillary Vistas Dataset proposed by \protect\citet{Neuhold2017ICCV}. \figsourceC{\protect\citet{Neuhold2017ICCV}}{2017}{IEEE}.}
	\label{fig:Neuhold2017ICCV}
\end{figure}

\subsection{Object Detection and Semantic Segmentation}

The Cityscapes Dataset\footnote{\url{https://www.cityscapes-dataset.com/}} by \citet{Cordts2016CVPR} provides a benchmark and large-scale dataset for pixel-level and instance-level semantic labeling that captures the complexity of real-world urban scenes. High-quality pixel-level annotations are provided for 5,000 images, while 20,000 additional images have been annotated with coarse labels obtained using crowdsourcing. While Cityscapes provides an evaluation server for a fair comparison of methods, the dataset is limited in size and diversity.  

For object detection, \citet{Braun2019PAMI} presented a large-scale dataset recorded in 31 cities of 12 European countries. Similar to Cityscapes, an evaluation server allows a fair comparison of methods. However, they only provide bounding box, occlusion, and orientation annotations for pedestrians, cyclists, and other riders in urban traffic.

The crowdsourcing company Mapillary\footnote{\url{https://www.mapillary.com/app}} has collected 282 million street-level images covering 4.5 million road kilometers around the world. Based on this data, the Mapillary Vistas Dataset \footnote{\url{https://www.mapillary.com/dataset/vistas}} \citep{Neuhold2017ICCV} has been created and shared with the community, providing 25,000 high-resolution images with dense annotation for 66 object categories and instance-specific labels for 37 classes (\figref{fig:Neuhold2017ICCV}).

The Berkeley DeepDrive dataset\footnote{\url{https://bdd-data.berkeley.edu}} \citep{Yu2018ARXIV} for object detection, instance segmentation, road, and lane detection provides 100K partially annotated driving videos from New York, Berkeley, San Francisco, and the Bay Area. The dataset is more diverse in scenes and weather conditions than Cityscapes, but it is still limited in the number of cities used for the recording. In this context, the Mapillary Vistas Dataset is the most diverse autonomous driving-related dataset for semantic segmentation and object recognition (\tabref{tab:datasets}). However, datasets like Mapillary Vistas Dataset, ImageNet, PASCAL VOC, and Microsoft Coco are less suited for training and testing temporal coherence of methods since they provide only single images in contrast to KITTI, Cityscapes, and Berkeley DeepDrive which provide image sequences.

So far, datasets for 3D semantic segmentation have been limited in size \cite{Munoz2009CVPR, Behley2012ICRA, Hackel2017APRS, Zhang2015ICRAa} and the number of classes \citep{Geiger2013IJRR} due to the large labeling effort required for annotating detailed object boundaries. Recently, \citet{Behley2019ARXIV} present a large dataset for 3D semantic segmentation based on the KITTI Visual Odometry Benchmark \citep{Geiger2013IJRR}. In contrast to previous annotations for KITTI, they provide dense point-wise annotations for the complete 360-degree field-of-view of the LiDAR. The dataset comprises over 20,000 scans with 25 different classes.

\subsection{Tracking}

The Caltech Pedestrian Detection Benchmark\footnote{\url{http://www.vision.caltech.edu/Image_Datasets/CaltechPedestrians/}} proposed by \citet{Dollar2009CVPR} provides 250,000 frames of sequences recorded while driving through regular traffic in an urban environment. 350,000 bounding boxes and 2,300 unique pedestrians were annotated, including temporal correspondence between bounding boxes and detailed occlusion labels. 

\subsection{Traffic Sign Detection}
While all previously discussed detection datasets focus on the detection of generic objects or traffic participants, only a few datasets exist for the recognition and detection of traffic signs. 
The most popular datasets for this task are the German Traffic Sign Recognition Benchmark (GTSRB\footnote{\url{http://benchmark.ini.rub.de/?section=gtsrb}}) \citep{Stallkamp2011IJCNN} and the German Traffic Sign Detection Benchmark (GTSDB\footnote{\url{http://benchmark.ini.rub.de/?section=gtsdb}}) \citep{Houben2013IJCNN}. GTSRB considers the task of classifying traffic signs into their corresponding category and consists of 50,000 images. In contrast, GTSDB provides 600 training and 300 test images for the task of detecting traffic signs. Reliable ground truth annotations for 40 different classes were created using a semi-automatic annotation tool. 
Recently, the limits of both datasets have been reached by state-of-the-art detection systems and \citet{Zhu2016CVPR} presented Tsinghua-Tencent 100K\footnote{\url{https://cg.cs.tsinghua.edu.cn/traffic-sign/}}, a new traffic sign detection benchmark. In contrast to GTSDB, their benchmark consists of 100,000 images with 30,000 signs. They provide high resolution images with pixel mask annotations and bounding boxes for each traffic sign. 

\subsection{Road and Lane Detection}

The KITTI benchmark was extended by \citet{Fritsch2013ITSC} to the task of road/lane detection. In total, 600 diverse training and test images have been selected for manual annotation of road and lane areas. \citet{Mattyus2016CVPR} used aerial images to enhance the KITTI dataset with fine-grained segmentation categories such as parking spots and sidewalk as well as the number and location of road lanes. 

A larger dataset for lane detection, the Caltech Lane Detection dataset\footnote{\url{http://www.mohamedaly.info/datasets/caltech-lanes}}, has been proposed by \citet{Aly2008IV}. The dataset was recorded in Pasadena in California at different day times and consists of over 1200 frames.
The first large-scale lane detection dataset was presented by \citet{Lee2017ICCV} and provides over 20,000 images. In contrast to previous datasets, they also consider different weather conditions. The Berkeley DeepDrive dataset\footnote{\url{https://bdd-data.berkeley.edu}} \citep{Yu2018ARXIV} with 100,000 images is so far the largest and most diverse lane/road detection dataset.

\subsection{Flow and Stereo}

Complementary to the datasets presented in \secref{sec:dataset_optical_flow} and KITTI, the HCI benchmark\footnote{\url{http://hci-benchmark.org}} proposed by \citet{Kondermann2016CVPRWORK} includes realistic, systematically varied radiometric and geometric challenges for autonomous driving. Overall, a total of 28,504 stereo pairs with stereo and flow ground truth is provided. The major limitation of the HCI Benchmark is that all sequences were recorded in a single street section, and thus the dataset lacks diversity. However, the controlled environment allows for more easily simulating rare events such as accidents which are of great interest for validating autonomous driving systems.

\begin{figure}[t]
	\centering
	\includegraphics[width=1.00\columnwidth]{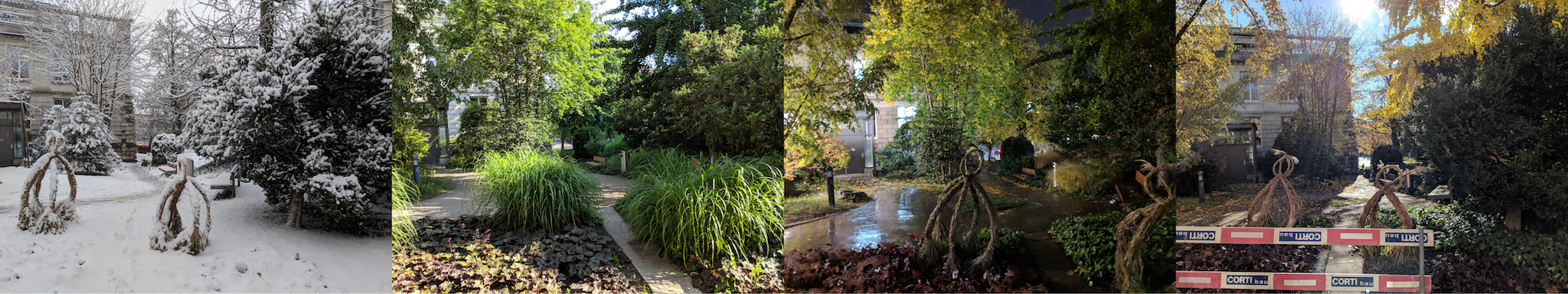}
	\caption[Long-Term Autonomy]{\textbf{Long-Term Autonomy.} Examples for different weather conditions, seasons and day times for a scene from the Workshop organized by \protect\citet{Hammarstrand2019CVPRWORK}. \figsource{\protect\citet{Hammarstrand2019CVPRWORK}}.}
	\label{fig:Hammarstrand2019CVPRWORK}
\end{figure}

\subsection{Long-Term Autonomy}
\label{sec:long_term_autonomy}
Several datasets such as KITTI or Cityscapes focus on the development of algorithmic competences for autonomous driving but do not address challenges of long-term autonomy, as for example environmental changes over time. In order to address this problem, \citet{Carlevaris-Bianco2016IJRR} presented a new long-term vision and LiDAR dataset comprising 27 sessions. However, the dataset was not recorded from a vehicle but instead using a Segway robot on the campus of the University of Michigan. A novel dataset for long-term autonomous driving has been presented by \citet{Maddern2016IJRR}. They collected images, LiDAR, and GPS data while traversing 1,000 km in central Oxford, UK during an entire year. This allowed them to capture large variations in scene appearance due to illumination, weather and seasonal changes, dynamic objects, and constructions. Such long-term datasets allow for an in-depth investigation of problems that detain the realization of autonomous vehicles such as localization at different times of the year, as illustrated in \figref{fig:Hammarstrand2019CVPRWORK}.

Several datasets have been proposed which address environmental changes for multi-view reconstruction. The structure-from-motion dataset BigSFM discussed in \secref{sec:stereo_datasets}, for instance, consists of Internet photos taken with different cameras at different times. Recently, \citet{Sattler2018CVPR} presented three datasets for visual localization (Aachen Day-Night, RobotCar Seasons and CMU Seasons) recorded under different weather conditions, seasons and during night and day. While the Aachen Day-Night dataset consists of images recorded using consumer cameras, RobotCar Seasons, and CMU Seasons were obtained using a car-mounted camera. More recently, Scape Technologies\footnote{\url{https://scape.io/}} presented a long-term dataset captured around the Imperial College London campus using a low-end, consumer spherical camera \citep{Balntas2019MEDIUM}. The dataset was recorded over a period of one year and incorporates different weather conditions, day times, and seasons.

\begin{figure}[t]
	\centering
	\includegraphics[width=1.00\columnwidth]{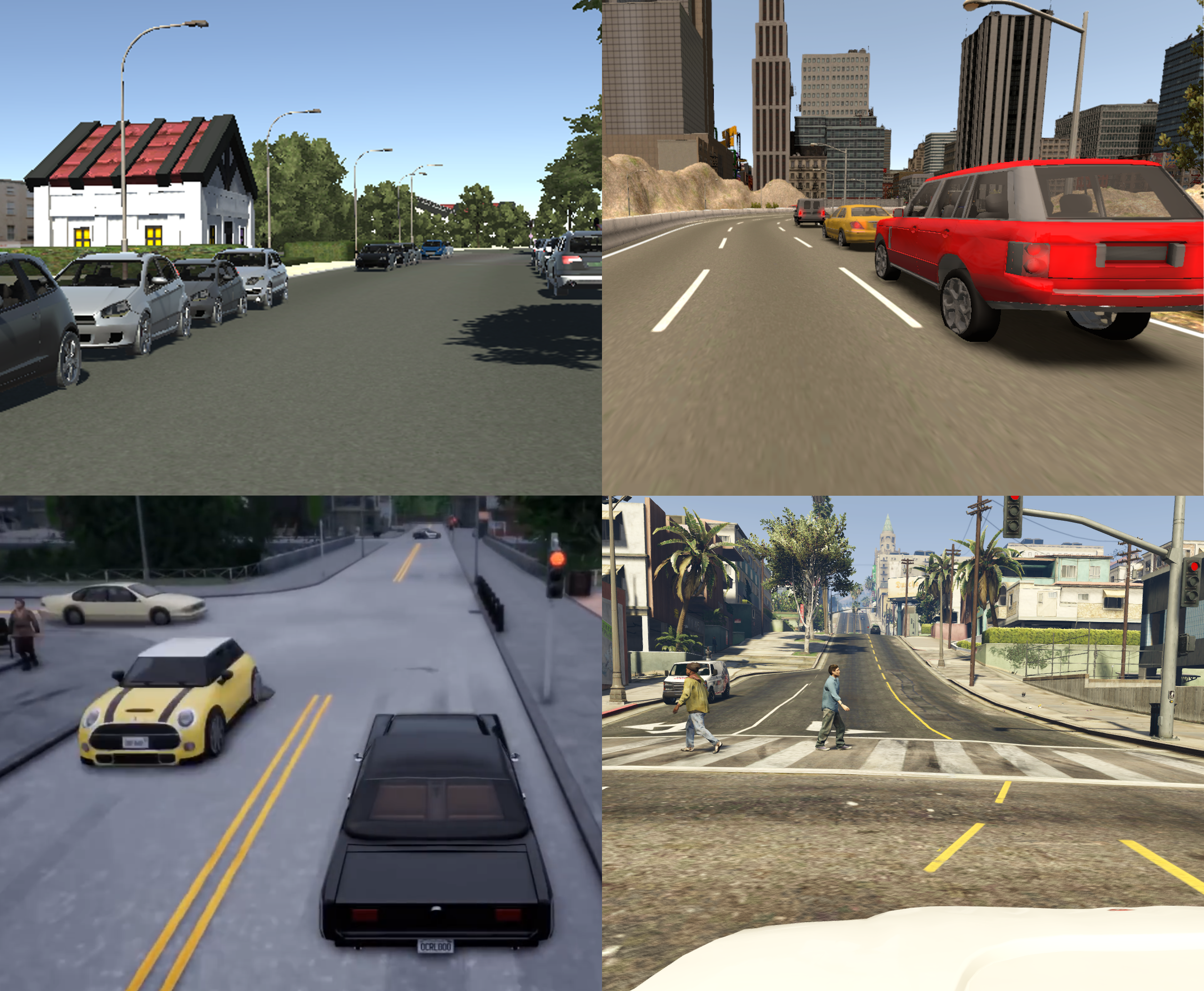}
	\caption[Synthetic Datasets]{\textbf{Synthetic Datasets.} Examples from Virtual KITTI \protect\citep{Gaidon2016CVPR}, SYNTHIA \protect\citep{Ros2016CVPR}, Carla \protect\citep{Dosovitskiy2017CORL} and Playing for Data \protect\citep{Richter2016ECCV}. \courtesy{\protect\citet{VirtualKITTI2016, Synthia2016, Carla2017, PlayingData2016}}}
	\label{fig:synthetic_data}
\end{figure}

\section{Synthetic Data Generation using Game Engines}
Data from animated movies as used in \citep{Butler2012ECCV,Mayer2016CVPR} is very limited since the content is hard to change, and such movies are rarely open-source. Moreover, rendering 3D models into random scenes as in \citep{Dosovitskiy2015ICCV,Mayer2016CVPR} lacks realism and diversity. In contrast, game engines allow for creating an infinite amount of more realistic and diverse data. 

One of the first datasets exploring game engines is the Virtual KITTI dataset\footnote{\url{https://europe.naverlabs.com/research/computer-vision/proxy-virtual-worlds}} presented by \citet{Gaidon2016CVPR}. They propose a real-to-virtual world cloning method to create realistic proxy worlds that resemble real scenarios. A cloned virtual world allows varying conditions such as weather or illumination and using different camera settings. This way, the proxy world can be used for virtual data augmentation to train deep networks. Virtual KITTI contains 35 photo-realistic synthetic videos with a total of 17,000 high resolution frames. They provide ground truth for object detection, tracking, scene and instance segmentation, depth, and optical flow. 

In concurrent work, \citet{Ros2016CVPR} created SYNTHIA\footnote{\url{http://synthia-dataset.net/}}, a synthetic collection of imagery and annotations of urban scenarios for semantic segmentation. They rendered a virtual city using the Unity Engine. The dataset consists of 13,400 randomly taken virtual images from the city and four video sequences with 200,000 frames in total. Pixel-level semantic annotations are provided for 13 classes.

In the Playing for Data project\footnote{\url{https://download.visinf.tu-darmstadt.de/data/from_games/}}, \citet{Richter2016ECCV} extracted pixel-accurate semantic label maps for images from the commercial video game Grand Theft Auto V. Towards this goal, they developed a tool that operates between the game and the graphics hardware to obtain pixel-accurate object signatures across time. Their algorithm allows them to produce dense semantic annotations for 25,000 images synthesized by the photorealistic open-world computer game with minimal human supervision. 
This work was extended in Playing for Benchmarks\footnote{\url{https://playing-for-benchmarks.org/}} \citep{Richter2017ICCV} to obtain dense correspondences and semantic instances from the game engine. The benchmark consists of about 250,000 images with dense annotations for semantic segmentation, instance segmentation, object detection, tracking, 3D scene layout, visual odometry, and optical flow. They provide an online evaluation server for semantic segmentation, instance segmentation, visual odometry, and optical flow.
Similarly, \citet{Qiu2017ACM} provide an open-source tool to create virtual worlds by accessing and modifying the internal data structure of Unreal Engine 4. They show how virtual worlds can be used to test deep learning algorithms by linking them with the deep learning framework Caffe \citep{Jia2014ICM}.

Recently, Carla\footnote{\url{http://carla.org/}}, an open-source simulator for autonomous driving, was introduced by \citet{Dosovitskiy2017CORL}. Carla allows generating synthetic data for control and perception of an autonomous driving system in urban environments. Complete access to the engine and digital assets are provided for non-commercial usage. Based on the Unreal Engine 4, extensions for Carla can be easily integrated by the community. 

Modern game engines as used in Carla \citep{Dosovitskiy2017CORL} and Playing for Data \citep{Richter2016ECCV} allow creating impressively realistic data for training large models, as shown in \figref{fig:synthetic_data}. While there is still a large gap between real and synthetic data and the creation of 3D content is costly and time-consuming, game engines enable the generation of large datasets and the investigation of dangerous situations that can only be rarely observed in real data.

	\chapter{Object Detection}
\label{chap:detection}

\begin{figure}[t]
	\centering
	\includegraphics[width=1.00\columnwidth]{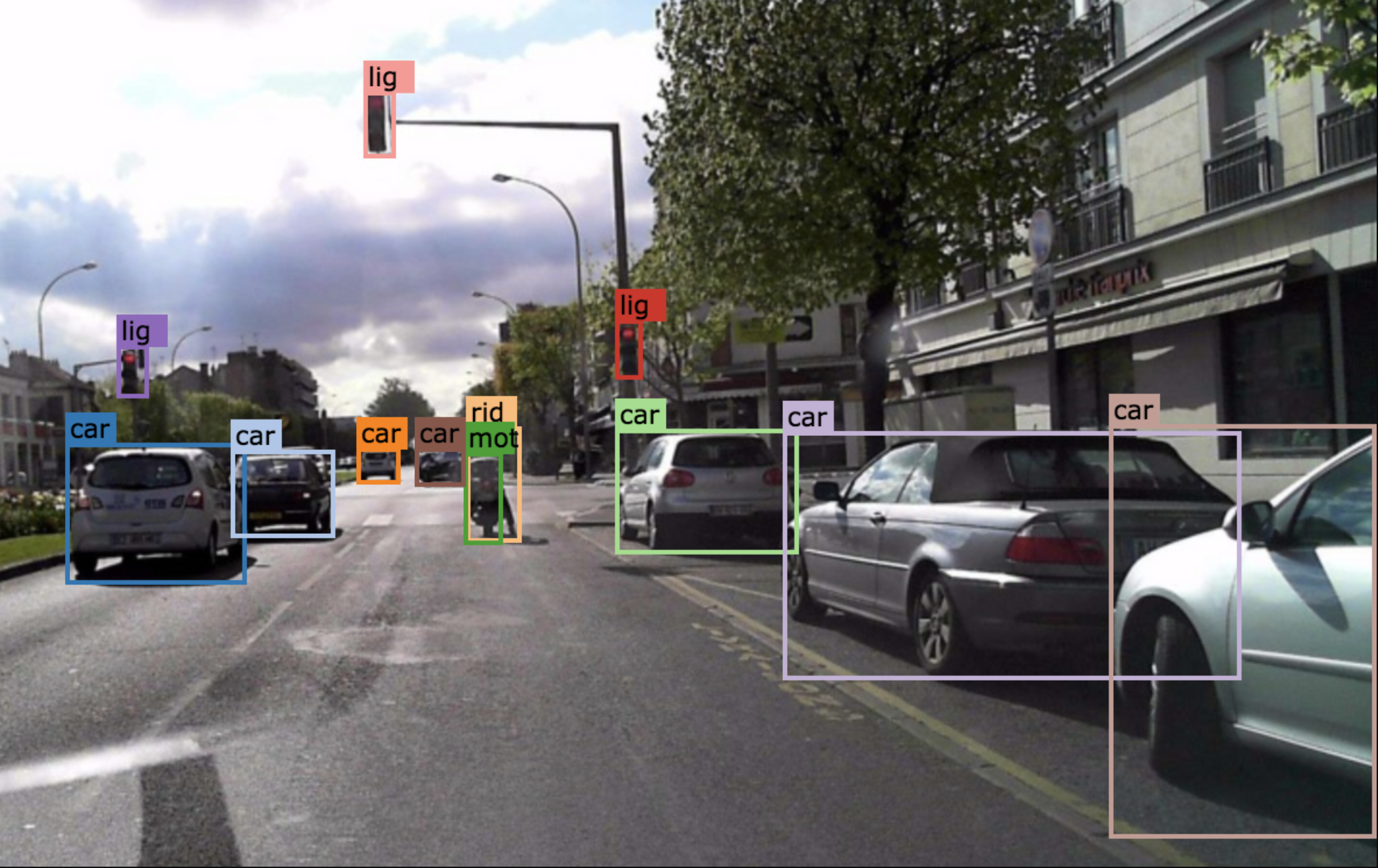}
	\caption[Object Detection]{\textbf{Object Detection.} In object detection, we are interested in finding all objects of certain classes in an image. These detections are usually represented with bounding boxes. \courtesy{Berkeley DeepDrive \protect\citep{BerkeleyDeepDrive}}.}
	\label{fig:ObjectDetection}
\end{figure}
\section{Problem Definition}
Reliable detection of objects, as shown in \figref{fig:ObjectDetection}, is a crucial requirement to realize autonomous driving. As the vehicle shares the road with many other traffic participants, particularly in urban areas, the awareness of other traffic participants or obstacles is necessary to avoid accidents that might be life-threatening. The detection in urban areas is hard because of the wide variety of object appearances and occlusions caused by other objects or the object of interest itself. In addition, the resemblance of objects to each other or to the background and physical effects like cast shadows or reflections can make the detection of objects difficult.

Reliable pedestrian detection is particularly difficult because of their complex, highly varying motion and the large variety of appearances due to different clothing and articulated poses. Furthermore, the interaction of pedestrians with each other and the world often cause partial occlusions. This problem has been deeply investigated as for example in advanced driver assistance systems to increase road safety. 
Pedestrian protection systems (PPS) detect the presence of stationary and moving people around a moving vehicle in order to warn the driver against dangerous situations. \citet{Geronimo2010PAMI} survey pedestrian detection for Advanced Driver Assistance Systems. While the driver can still handle missed detections of a PPS, an autonomous car needs a flawless pedestrian detection system which is robust against all weather conditions and efficient for real-time detection.

The object detection problem has been approached using a variety of input modalities. Video cameras are the cheapest and most commonly used type of sensors for the detection of objects. The visible spectrum (VS) is typically used for daytime detections, whereas the infrared spectrum offers more visibility for night-time detection\citep{Suard2006IV}. Thermal infrared (TIR) cameras capture relative temperature, which allows distinguishing warm objects like pedestrians from cold objects like vegetation or the road. Active sensors that emit signals and observe their reflection, like laser scanners can provide range information which is helpful for detecting an object and localizing it in 3D. However, laser scanners often have a smaller resolution compared with video cameras. Depending on the weather conditions, time of day, or material properties, it can be problematic to rely on a single type of sensor alone. VS cameras and laser scanners are affected by reflective or transparent surfaces, while hot objects (like engines) or warm temperatures can influence TIR cameras. The combination of information from different sensors via sensor fusion \citep{Enzweiler2011TIP,Chen2018PAMIb,Gonzalez2016TCYB} allows for the robust integration of this complementary information.

\section{Methods}
Classical object detection systems usually consist of multiple steps that are applied consecutively to solve the object detection task. With the success of deep neural networks, most of these steps \citep{Sermanet2013CVPR,Girshick2014CVPR,He2014ECCV,Girshick2015ICCV} and even the complete pipeline have been replaced by learned models \citep{Sermanet2014ICLR,Ren2015NIPS,Redmon2016CVPR,Liu2016ECCV,Redmon2017CVPR,Lin2017ICCV}. 
We start our discussion with classical pipelines, followed by more modern approaches.

\subsection{Classical Pipeline}
A classical detection pipeline usually comprises the following steps: preprocessing, region of interest extraction (ROI), object classification, and verification or refinement. In the preprocessing step, tasks such as exposure and gain adjustment, as well as camera calibration and image rectification, are usually performed. Some approaches leverage temporal information with a joint detection and tracking system. We discuss tracking approaches in-depth in \chpref{chap:tracking}. 

Regions of interest can be extracted using a sliding window approach, which shifts a window over the image at different scales. As exhaustive search is very expensive, several heuristics have been proposed for reducing the search space. Typically, the number of evaluations is reduced by assuming a certain ratio, size, and position of candidate bounding boxes. Apart from that, image features, stereo, or optical flow can be leveraged for focusing the search on relevant regions. \citet{Broggi2000IV}, for instance, leverage morphological characteristics (size, ratio, and shape), vertical symmetry of human shape, and distance information obtained from stereo for the extraction of relevant ROIs. Selective Search \citep{Uijlings2013IJCV} is an alternative approach to generate regions of interest. Instead of an exhaustive search over the full image domain, selective search exploits a segmentation of the image to extract approximate locations efficiently. For a more detailed discussion, we refer the reader to \citet{Dollar2011PAMI}, presenting an extensive evaluation of pedestrian detection systems from monocular images with a focus on sliding window approaches. 

The next step is the processing of candidate image regions from sliding window to verify them and classify objects.
The classification of all candidates in an image can be quite costly due to the vast amount of image regions that need to be processed. Therefore, a fast decision is necessary which quickly discards candidates in the background region of the image. \citet{Viola2005IJCV} combine simple and efficient classifiers, learned using AdaBoost, in a cascade that allows them to quickly discard false candidates while spending more time on promising regions. With the work of \citet{Dalal2005CVPR}, linear Support Vector Machines (SVMs) in combination with Histogram of Orientation (HOG) features have become popular tools for classification. \citet{Enzweiler2009PAMI} provide an overview of classical approaches for monocular pedestrian detection. They make the observation that SVM with HOG features work well at higher resolutions while having a higher processing time than cascaded approaches that are superior at lower resolutions and achieve near real-time performance. In their survey, \citet{Benenson2014ECCV} found no clear evidence that a certain type of classifier (\eg SVM or decision forests) is better suited than any other. In particular, \citet{Wojek2008DAGM} show that AdaBoost and linear SVM perform roughly the same if enough features are given. \citet{Benenson2014ECCV} conclude that the number and diversity of features is clearly an important factor for the performance of classifiers since the classification problem becomes easier with higher dimensional representations. Consequently, today, all state-of-the-art object detection systems use convolutional neural networks to learn expressive features in an end-to-end fashion from large datasets \citep{Cai2016ECCV,Xiang2017WACV,Zhu2016ACCV,Yang2016CVPR,Chen2015NIPS,Ren2015NIPS,Girshick2015ICCV}.

\begin{figure}[t]
	\centering
	\includegraphics[width=1.00\columnwidth]{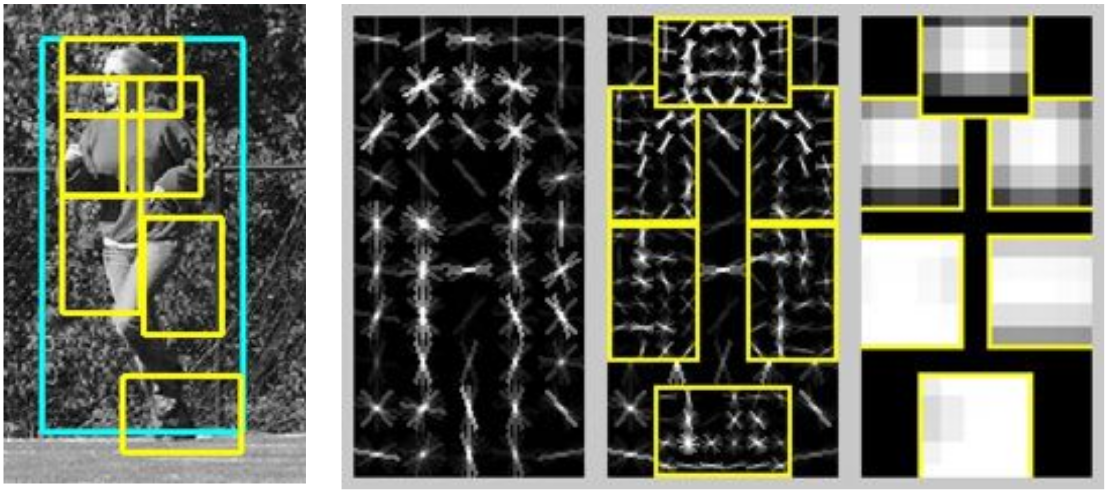}
	\caption[Part-based Approaches]{\textbf{Part-based Approaches.} Illustration of the Deformable Part Model (DPM) proposed by \protect\citet{Felzenszwalb2008CVPR}. The model consists of a coarse global template (middle-left), several high resolution part templates (middle-right) and the location (right). \figsourceC{\protect\citet{Felzenszwalb2008CVPR}}{2008}{IEEE}.}
	\label{fig:Felzenszwalb2008CVPR}
\end{figure}
\boldparagraph{Multi-Cue Object Detection}
While most object detection systems rely on single images as input, there are several approaches which show that using multiple cues such as temporal and structure information can boost performance. Temporal information from video sequences can provide important additional constraints to solve the detection task better. \citet{Shashua2004IV} integrate additional cues measured over time (dynamic gait, motion parallax) and situation-specific features (such as leg positions at certain poses) into a detection system to obtain more reliable detections. \citet{Wojek2009CVPR} show significant improvement in detection performance by incorporating motion cues and combining different complementary feature types. The dense correspondences between two frames (optical flow) \citep{Walk2010CVPR} or joint tracking as discussed in \chpref{chap:tracking} also lead to significant performance gain since more information about the same object can be aggregated over time.
Structure information can be beneficial to generate region of interests and provide additional information about the shape of objects to improve classification. Towards this goal, \citet{Keller2011TITS} jointly detect objects and estimate dense depth maps from stereo images. 

\boldparagraph{Generative Models for Augmenting Training Data}
As object detection is typically formulated as a supervised learning task, large amounts of annotated training data are required to obtain good performance. Unfortunately, generating examples belonging to the target class is usually time-consuming because of manual labeling, while negative examples can be more easily obtained. \citet{Enzweiler2008CVPR} address this bottleneck by creating synthesized virtual samples with a learned generative model. The generative model consists of probabilistic shape and texture models for a set of generic poses. As the discriminative model, they consider a neural network \citep{Wohler1999IVC} and SVMs with Haar features \citep{Papageorgiou2000IJCV} to demonstrate the generality of their approach. The generative model captures prior knowledge about the pedestrian class and allows significant improvement in classification performance. 

\subsection{Part-based Approaches}
Learning the appearance of articulated objects is difficult because all possible articulations need to be considered. The idea of part-based approaches is to split the complex appearance of non-rigidly moving objects like humans into simpler parts and to represent articulation using these parts, as illustrated in \figref{fig:Felzenszwalb2008CVPR}. This provides greater flexibility and reduces the number of training examples required for learning the appearance of each part. 

The Deformable Part Model (DPM), by \citet{Felzenszwalb2008CVPR}, attempts to break down the complex appearance of objects into easier parts. As a classifier, they train a SVM with latent structure variables which represent the model configuration (part positions) and need to be inferred at training time. They use a coarse global template covering the entire object and higher resolution part templates to model the appearance of each part. An alternative to this representation is the Implicit Shape Model proposed by \citet{Leibe2008IJCV}, which learns a highly flexible representation of object shape. They extract local features around interest points and perform clustering to construct a codebook of local appearances that are characteristic for the particular object class under consideration. Finally, they learn the occurrences of codebook entries for each object.
However, \citet{Benenson2014ECCV} observe in their survey on detection approaches that part-based models like \citep{Felzenszwalb2008CVPR, Leibe2008IJCV} improve results only slightly compared to the much simpler approach of \citet{Dalal2005CVPR}.

The discussed part-based models can not represent relationships between different objects, their parts, and the scene, which, for instance, is necessary to reason about occlusions. 
Usually, a separate context model \citep{Hoiem2008IJCV, Tu2010PAMI, Desai2011IJCV, Yang2012CVPRa} is learned which puts the detected objects in context to the 3D scene. In contrast, \citet{Wu2016PAMI} propose to learn an And-Or model that embeds a grammar to represent large structural and appearance variations in a reconfigurable hierarchy.
The learned model takes into account structural and appearance variations at multi-car, single-car, and part-levels jointly to represent both context and occlusions.  

\subsection{Deep Learning for Detection}
\label{sec:deep_learning_detection}
\begin{figure}[t]
	\centering
	\includegraphics[width=1.00\columnwidth]{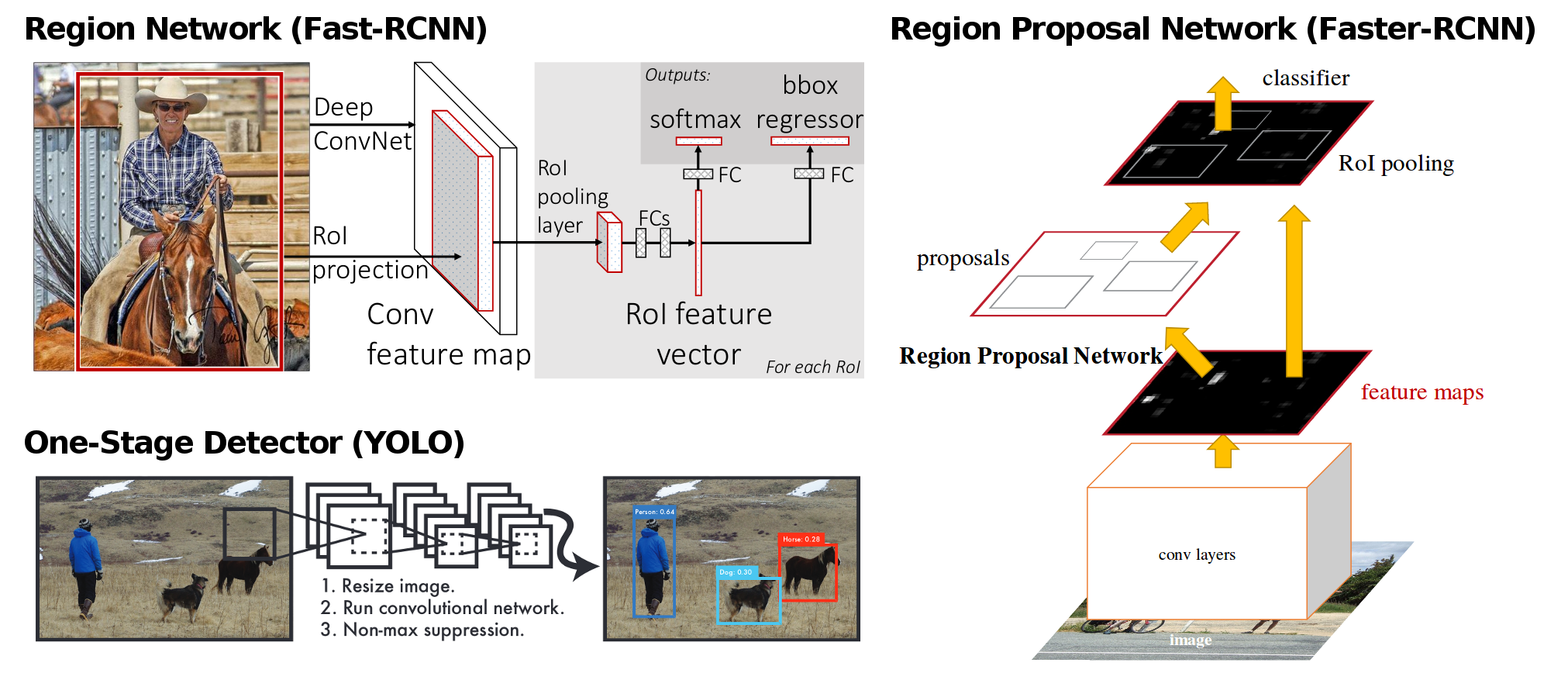}
	\caption[Object Detection Networks]{\textbf{Object Detection Networks.} Illustration of the three popular object detection networks. Upper left: Region-based network Fast-RCNN \protect\citep{Girshick2015ICCV} that works on regions. Right: Region proposal network Faster-RCNN \protect\citep{Ren2015NIPS} that learn to extract regions. Lower left: One-stage detector YOLO \protect\citep{Redmon2016CVPR} that formulates the detection task as regression problem. 
	\textcopyright~2015,~2016 IEEE. Reprinted, with permission, from \protect\citet{Girshick2015ICCV, Redmon2016CVPR, Ren2017PAMI}.}
	\label{fig:ObjectDetectionNetworks}
\end{figure}

All previous methods rely on hand-crafted features that are difficult to design and limited in their representation capabilities. With the renaissance of deep learning \citep{Krizhevsky2012NIPS}, convolutional neural networks have been applied to the object detection problem, resulting in significantly increased performance. Examples of the three most popular architectures are illustrated in \figref{fig:ObjectDetectionNetworks}.

\citet{Sermanet2013CVPR} introduced CNNs to the pedestrian detection problem by learning the extraction of expressive features in an unsupervised fashion using convolutional sparse auto-encoders. 
Eventually, they train a classifier in an end-to-end supervised fashion while extracting the features with a sliding window scheme and jointly fine-tuning the auto-encoders. However, they use a shallow network with a small receptive field, which allows precise localization of the objects using a sliding window approach. In contrast, deeper networks with larger receptive fields complicate the precise localization because local information is extracted in earlier layers, while high-level information is represented in deeper layers. Therefore, \citet{Girshick2014CVPR} propose R-CNNs to solve the CNN localization problem via a ``recognition using regions'' paradigm. They generate many region proposals using selective search \citep{Uijlings2013IJCV}, extract a fixed-length feature vector for each proposal using a CNN and classify each region with a linear SVM. Region-based CNNs are computationally expensive but several improvements have been proposed to reduce the computational burden \citep{He2014ECCV, Girshick2015ICCV}. \citet{He2014ECCV} use spatial pyramid pooling which allows computing a convolutional feature map for the entire image with only one run of the CNN in contrast to R-CNN that needs to be applied on many image regions. \citet{Girshick2015ICCV} (Fast-RCNN) further improve upon these results by proposing a single-stage training algorithm using a multi-task loss that jointly learns to classify object proposals and refine their spatial locations. 

In region-based CNNs, the classical region proposal algorithm remained the primary computational bottleneck and the main factor limiting performance. Therefore, \citet{Ren2015NIPS} (Faster-RCNN) introduced Region Proposal Networks (RPN), which share full-image convolutional features with the detection network and thus do not incur additional computational costs. RPNs are trained end-to-end to generate high-quality region proposals, which are classified using the Fast R-CNN detector \citep{Girshick2015ICCV}. 

Eventually, one-stage detectors \citep{Sermanet2014ICLR, Redmon2016CVPR, Liu2016ECCV, Redmon2017CVPR, Lin2017ICCV} completely removed the region proposal step by formulating the object detection task as a regression problem. The first one-stage detector by \citet{Sermanet2014ICLR} was a deep convolutional version of the sliding window approach. They extract features with a CNN and apply a classifier network based on AlexNet \citep{Krizhevsky2012NIPS} on the extracted feature maps in a sliding window fashion. \citet{Redmon2016CVPR} (YOLO) instead suggest to jointly learn spatially separated bounding boxes and class probabilities from the topmost feature maps of a network based on GoogLeNet \citep{Szegedy2015CVPR}. This allows them to achieve real-time performance and eventually YOLO9000 \citep{Redmon2017CVPR} to outperform the Region Proposal Networks. \citet{Liu2016ECCV} further improve in accuracy and efficiency by incorporating feature maps from different scales and considering a fixed set of bounding boxes. 
However, one-stage detectors \citep{Sermanet2014ICLR, Redmon2016CVPR, Liu2016ECCV, Lin2017ICCV} could not compete with region proposal algorithms. One reason for the performance gap is the foreground-background class imbalance \citep{Lin2017ICCV}. To alleviate this problem and improve training, \citet{Lin2017ICCV} propose a dynamically scaled cross-entropy loss allowing them to reduce the contribution of easy examples.

All previous one-stage detectors use anchor bounding boxes that are densely placed over the image, verified, and refined using regression. In contrast, \citet{Law2018ECCV} propose to directly predict heatmaps for the top-left and bottom-right corners of all bounding boxes. Finally, they need to identify corners belonging to the same bounding box. Towards this goal, they train a network to predict similar embedding vectors for corners of the same bounding box, which allows them to group the corners according to the distance between the embeddings.

Part-based models have also been introduced to CNN-based approaches. \citet{Zhang2014ECCVb} propose to extract deep convolutional features from bottom-up proposals obtained from a selective search algorithm and learn part appearance models. This allows them to enforce geometric constraints between parts and to outperform previous methods.

\subsection{Real-time Pedestrian Detection}
In case of a potential collision with pedestrians, a fast detection system allows the early intervention of the autonomous system. In classical literature, \citet{Benenson2012CVPR} provide fast pedestrian detections based on better handling of scales and exploiting depth extracted from stereo. Instead of resizing the images, they scale HOG features similar to \citet{Viola2004IJCV}. However, CNN-based approaches recently also reached real-time efficiency due to strong parallelization on the GPU. 
While Fast R-CNN \citep{Girshick2015ICCV} could only be applied at 0.5 Hz, the faster version with the Region Proposal Network Faster-RCNN \citep{Ren2015NIPS} already achieves 17 Hz. Finally, YOLO9000 \citep{Redmon2017CVPR} can be applied at up to 90 Hz at $288\times288$ pixels resolution and achieves 40 Hz at $544\times544$ pixels resolution.

\begin{figure}[t]
	\centering
	\includegraphics[width=0.70\columnwidth]{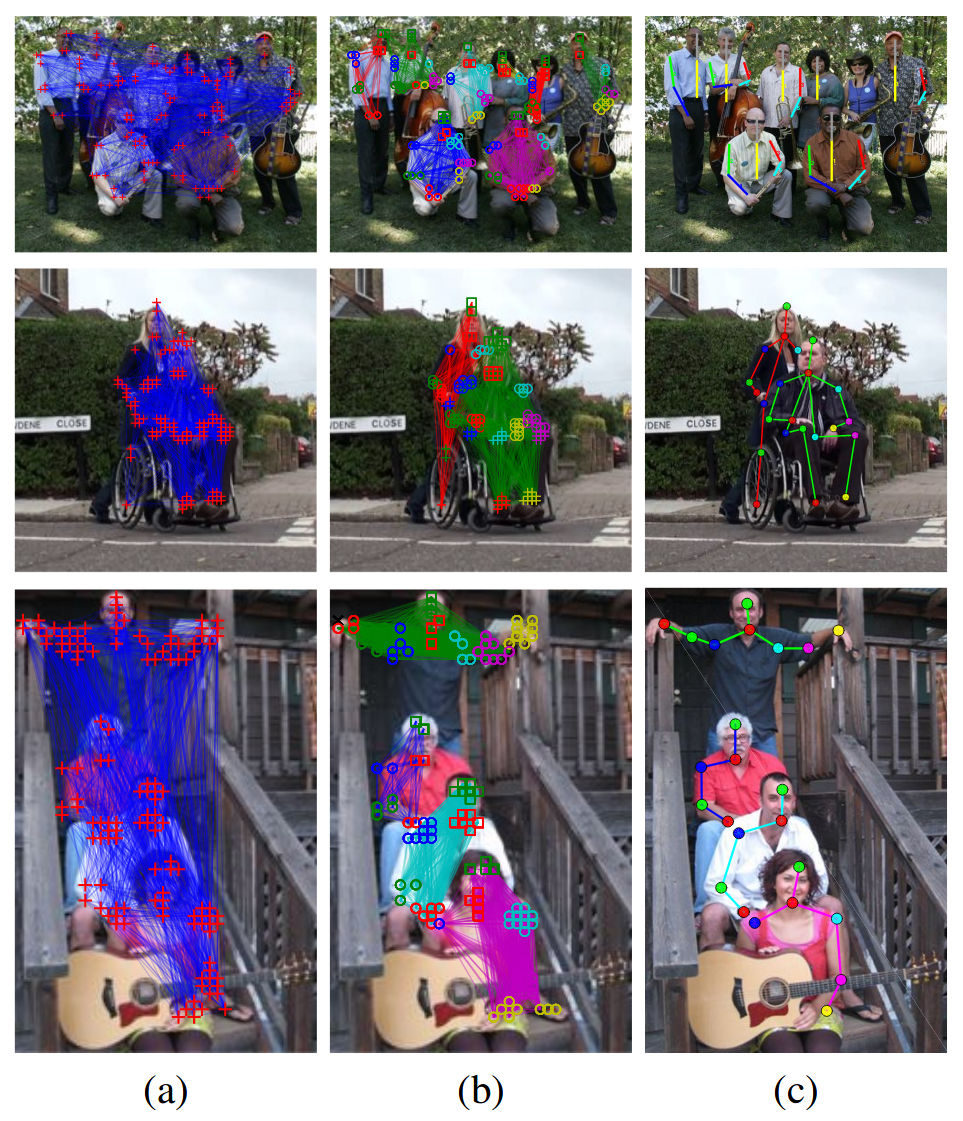}
	\caption[Human Pose Estimation]{\textbf{Human Pose Estimation.} Illustration of the DeepCut \protect\citep{Pishchulin2016CVPR} showing the initial detections and pairwise terms (a), joint clustering of nodes belonging to the same person visualized by colors (b) and the predicted poses (c). \figsourceC{\protect\citet{Pishchulin2016CVPR}}{2016}{IEEE}.}
	\label{fig:HumanPoseEstimation}
\end{figure}
\subsection{Human Pose Estimation}
The pose of a person provides important information to the autonomous vehicle about the behavior and intention of the person. However, the pose estimation problem is challenging since the pose space is very large, and typically, people can only be observed at low resolutions because of their size and distance to the vehicle. 
Several approaches have been proposed to jointly estimate the pose and body parts of a person. Traditionally, a two-staged approach was used by first detecting body parts and then estimating the pose as in \citep{Pishchulin2012CVPR, Gkioxari2014CVPR, Sun2011ICCV}. This is problematic in cases when people are in proximity to each other because body-parts can be wrongly assigned to different instances. 

\citet{Pishchulin2016CVPR} present DeepCut, visualized in \figref{fig:HumanPoseEstimation}, a model that jointly estimates the poses of all people in an image. The formulation is based on partitioning and labeling a set of body-part hypotheses obtained from a CNN-based part detector. The model jointly infers the number of people, their poses, spatial proximity, and part-level occlusions. \citet{Bogo2016ECCV} use DeepCut to estimate the 3D pose and 3D shape of a human body from a single unconstrained image. Towards this goal, SMPL, a 3D body shape model proposed by \citet{Loper2015SIGGRAPH}, is fit to predictions of the 2D body joint locations from DeepCut. SMPL captures correlations in human shape across the population, which allows fitting human poses robustly even in the presence of weak observations.

\subsection{Traffic Sign Detection}
Reliable detection and recognition of traffic signs are essential for autonomous vehicles. The introduction of the German Traffic Sign Recognition Benchmark (GTSRB) by \citet{Stallkamp2011IJCNN} and the German Traffic Sign Detection Benchmark (GTSDB) by \citet{Houben2013IJCNN} are the most popular datasets for traffic sign detection. However, recent CNNs already reach the limits of GTSRB and GTSDB with a recall and precision of 100\%. Therefore, \citet{Zhu2016CVPR} recently presented Tsinghua-Tencent 100K, a new traffic sign detection benchmark, introducing new challenges to the community.

Several object detectors have been considered for traffic sign detection, \ie SVMs \citep{Maldonado2007TITS}, pattern matching techniques \citep{Broggi2007IV}, voting schemes such as radial symmetric detectors \citep{Barnes2008TITS} and integral channel features \citep{Dollar2009BMVC, Mathias2013IJCNN}. However, the recent progress in deep learning also led to better traffic sign classifiers \citep{Ciresan2011IJCNN, Sermanet2011IJCNN, Ciresan2012NN, Jin2014TITS}. \citet{Ciresan2011IJCNN} propose a committee consisting of a CNN trained on images and an MLP trained on HOG feature descriptors to classify traffic signs. In contrast, \citet{Sermanet2011IJCNN} propose a multi-scale CNN to learn meaningful features instead of using handcrafted features such as HOG. For faster training, \citet{Jin2014TITS} present a stochastic gradient descent method with a cost function similar to the objective function of the SVM. Similar to \citep{Sermanet2013CVPR}, \citet{Aghdam2016RAS} propose a sliding window detector that extracts features using a CNN. However, they apply the CNN using dilated convolutions on several resolutions to learn the detection of traffic signs at different scales.
Finally, they train a convolutional network with fully connected layers to classify the extracted features.

\citet{Garcia2018NEURO} compare generic object detectors on the popular GTSDB dataset. Region-based networks \citep{Girshick2014CVPR,He2014ECCV,Girshick2015ICCV} and one-stage generic detectors \citep{Redmon2016CVPR,Redmon2017CVPR,Lin2017ICCV} have difficulties with traffic signs at small scales. Traffic signs can appear very small in the image depending on the size, distance, and occlusions.
Region-Proposal networks \citep{Ren2015NIPS} give the best performance of generic detectors and in combination with Inception V2 \citep{Ioffe2015ICML} for feature extraction achieve comparable results with \citep{Aghdam2016RAS} on GTSDB.
\citet{Yang2018CN} adapt Faster-RCNN \citep{Ren2015NIPS} to the traffic sign detection task by extracting region proposals in a coarse-to-fine fashion. 
A novel attention network is proposed to roughly locate and classify RoIs before using the finer Region Proposal Network. 
This allowed them to improve upon Faster-RCNN on both GTSDB and Tsinghua-Tencent 100K datasets.

\subsection{3D Object Detection from 2D Images}
Geometric 3D representations of object classes can recover far more details than just 2D or 3D bounding boxes, however, most of today's object detectors are focused on robust 2D matching. In contrast, \citet{Zia2013PAMI} exploit the fact that high-quality 3D CAD models are available for many important classes. From these models, they obtain coarse 3D wireframe models using principal components analysis and train detectors for the vertices of the wireframe. At test time, they generate evidence for vertices by densely applying the detectors. \citet{Zia2015IJCV} extend this work by directly using detailed 3D CAD models in their formulation, combining them with explicit representations of likely occlusion patterns. Further, a ground plane is jointly estimated to stabilize the pose estimation process. This extension outperforms the pseudo-3D model \citep{Zia2013PAMI} and shows the benefits of reasoning in true metric 3D space. 

While these 3D representations provide more expressive descriptions of objects, they can not yet compete with state-of-the-art detectors using 2D bounding boxes. To overcome this problem, \citet{Pepik2015PAMI} propose a 3D extension of the deformable parts model \citep{Felzenszwalb2008CVPR} that combines the 3D geometric representation with robust matching to real-world images. They further add 3D CAD information of the object class of interest as geometry cue to enrich the appearance model. 

\citet{Kundu2018CVPR} train a CNN to map 2D object proposals to full 3D shape and pose. They add region-wise subnetworks for 3D shape and 3D pose prediction to a Faster-RCNN/Network-on-Convolution \citep{Ren2015NIPS, Ren2017PAMIa} architecture. To facilitate the problem, they learn a low dimensional shape-space from CAD models and use it as shape prior. The 3D shape estimation is then formulated as a prediction problem of a set of low dimensional shape parameters. With a differentiable Render-and-Compare loss, they are able to learn 3D shape and pose from 2D supervision (instance segmentation or depth).
In contrast, \citet{Ku2019CVPR} suggest a more flexible approach using LiDAR point clouds as supervision to avoid the dependency on annotated datasets of CAD models. They use 2D detections of MS-CNN \citep{Cai2016ECCV} and learn a model based on Faster-RCNN \citep{Ren2015NIPS} to regress amodal, oriented 3D bounding boxes.
\citet{Manhardt2019CVPR} also first extract 2D detection using an architecture based on \citep{Ren2015NIPS}. They propose a fully-differentiable mapping to lift the 2D detections, orientation, and scale estimation to the 3D space while using monocular depth predictions \citep{Pillai2019ICRA} to guide the distance reasoning.

\subsection{3D Object Detection from 3D Point Clouds} 
In contrast to cameras, laser range sensors directly provide accurate 3D information, which simplifies the extraction of object candidates and can be helpful for the classification task as it provides 3D shape information. 

\citet{Li2016RSS} exploit a fully convolutional neural network for detecting vehicles from range data. They use a 2D representation of the 3D range data analogous to cylindrical images with the channels encoding the 3D location of the points. Given this representation, they simultaneously predict an objectness confidence and bounding box using a single 2D CNN. In contrast, \citet{Wang2015RSS} propose an efficient scheme to apply the common 2D sliding window detection approach to 3D data. More specifically, they discretize the space into a 3D voxel grid and exploit the sparse nature of the problem with a voting scheme on top of a linear classifier, which is shown to be equivalent to convolutions on the full 3D point cloud. \citet{Engelcke2017ICRA} extend this feature-centric voting scheme by implementing a novel convolutional layer to apply sparse convolutions across the 3D point cloud. Additionally, they encourage sparsity in the intermediate representation using ReLU non-linearities and $L_1$ penalty. While \citep{Wang2015RSS, Engelcke2017ICRA} extract hand-crafted features from the voxels, VoxelNet from \citet{Zhou2018CVPR} learns the features in an end-to-end trainable deep network. They propose a voxel feature encoding layer that learns a unified feature representation for the points of the voxels. Eventually, a region proposal network generates detections from these feature representations.

Relying only on laser range data makes the detection task challenging due to the limited density of the laser scans and lack of appearance information. 
Thus, existing LiDAR-based approaches perform weaker compared to their image-based counterparts on the 2D detection problem of KITTI. However, recently, it has been shown that the fusion of LiDAR and camera information allows reducing the gap and eventually even outperforming state-of-the-art 2D detectors \citep{Chen2017CVPR, Ku2018IROS, Costea2017CVPR, Qi2017CVPRa, Du2018ICRA}. We will discuss these methods in detail in \secref{sec:Object_Detection_Results}.

\section{Datasets}
The most popular datasets for object detection are ImageNet \citep{Deng2009CVPR}, PASCAL VOC \citep{Everingham2010IJCV}, Microsoft COCO \citep{Lin2014ECCV}, KITTI \citep{Geiger2012CVPR} and Caltech Pedestrian Detection \citep{Dollar2012PAMI}. While ImageNet, PASCAL VOC, and Microsoft COCO consider the general detection problem, KITTI and Caltech Pedestrian Detection benchmark focus on classes that are relevant for the autonomous driving context. KITTI provides separate benchmarks for 2D and 3D detection of cars, pedestrians, and cyclists with 2D and 3D input modalities for both benchmarks. In contrast, the Caltech Detection benchmark focuses on the pedestrian detection problem only.

Recently, EuroCity Persons \citep{Braun2019PAMI}, a new large-scale benchmark for pedestrian detection, was presented. Also, several companies, \ie ApolloScape \citep{Huang2018CVPR}, NuScenes \citep{Caesar2019ARXIV}, and Berkeley DeepDrive \citep{Yu2018ARXIV}, presented new publicly available datasets for object detection in street scenes. Similarly to KITTI, ApolloScape provides annotation for 3D car detection but is not considering other classes than cars. The Berkeley DeepDrive dataset even provides additional classes (traffic light, traffic sign, train) for the road object detection problem. However, these datasets and benchmarks \citep{Braun2019PAMI,Huang2018CVPR,Caesar2019ARXIV,Yu2018ARXIV} are not yet established in the field.

In this work, we focus our attention on the KITTI benchmark since it allows us to compare generic object and specific pedestrian detection systems on the same data. 
We refer the interested reader to the survey papers \citep{Benenson2014ECCV, Zhang2016CVPRa} for an in-depth comparison of pedestrian detection systems on Caltech-USA. 

\section{Metrics}
The most popular measures for the performance of object detection systems are the average precision (AP) and average recall (AR) \citep{Deng2009CVPR, Everingham2010IJCV, Lin2014ECCV, Cordts2016CVPR, Geiger2012CVPR}. In addition, the precision-recall curve is usually used to evaluate methods \cite{Everingham2010IJCV, Geiger2012CVPR}. For calculating precision and recall, the detections are categorized into true positives, false positives, and false negatives.
Towards this goal, the intersection-over-union (IOU) between the detected bounding boxes and the ground truth bounding boxes is considered. 
A popular threshold for true positives is an IOU of at least 50\%. The AP with an IOU of 50\% is known as the PASCAL VOC \citep{Everingham2010IJCV} metric and used in many different benchmarks \cite{Geiger2012CVPR, Dollar2012PAMI, Deng2009CVPR}. In addition to the standard  PASCAL VOC metric, MS COCO \citep{Deng2009CVPR} considers several additional metrics: the AP with an IOU of at least 75\%, for small, medium and large objects, and several AR metrics.

In our discussion here, we consider the metrics reported on the KITTI benchmark \cite{Geiger2012CVPR}. The performance is assessed for three levels of difficulty (easy, moderate, hard) using PASCAL VOC intersection-over-union (IOU) \citep{Everingham2010IJCV}.
While the PASCAL VOC metric (IOU of 50\%) is used for pedestrians and cyclists on KITTI, the metric used for cars is more strict and requires an overlap of 70\%. Easy examples have a minimum bounding box height of 40 px and are fully visible, whereas moderate examples have a minimum height of 25 px and include partial occlusions. Hard examples have the same minimum height but include large levels of occlusion. 
In \tabref{tab:kitti_object_det2d_ori}, the estimation of the object's orientation is evaluated using the average orientation similarity (AOS) proposed in \citep{Geiger2012CVPR}.

\section{State of the Art on KITTI}
\label{sec:Object_Detection_Results}
\begin{table*}[p!]
	\begin{subtable}{\linewidth}
		\begin{adjustbox}{width=1\textwidth}\begin{tabular}{l l | c | c | c | c}
   & {\bf Method} & {\bf Moderate} & {\bf Easy} & {\bf Hard} & {\bf Runtime}\\ \hline
1. & RRC \citep{Ren2017CVPR} & 90.23 \% & 90.61 \% & 87.44 \% & 3.6 s / GPU \\
2. & SJTU-HW \citep{Zhang2018ICIP} & 90.08 \% & 90.81 \% & 79.98 \% & 0.85s / GPU \\
3. & Deep MANTA \citep{Chabot2017CVPR} & 90.03 \% & 97.25 \% & 80.62 \% & 0.7 s / GPU \\
4. & sensekitti \citep{Yang2016CVPRb} & 90.00 \% & 90.76 \% & 81.83 \% & 4.5 s / GPU \\
5. & SINet+ \citep{Hu2018TITS} & 89.73 \% & 90.51 \% & 77.82 \% & 0.3 s / GPU \\
\hline
11. & SubCNN \citep{Xiang2015CVPR} & 88.86 \% & 90.75 \% & 79.24 \% & 2 s / GPU \\
12. & Deep3DBox \citep{Mousavian2017CVPR} & 88.86 \% & 90.47 \% & 77.60 \% & 1.5 s / GPU \\
13. & MS-CNN \citep{Cai2016ECCV} & 88.83 \% & 90.46 \% & 74.76 \% & 0.4 s / GPU \\
23. & Faster R-CNN \cite{Ren2015NIPS} & 79.11 \% & 87.90 \% & 70.19 \% & 2 s / GPU \\
41. & YOLOv2 \cite{Redmon2017CVPR} & 19.31 \% & 28.37 \% & 15.94 \% & 0.02 s / GPU \\
\end{tabular}\end{adjustbox}
		\caption{KITTI Car Detection Leaderboard}
		\label{tab:kitti_object_det2d_car_image}
		\vspace{0.2cm}
	\end{subtable}
	\begin{subtable}{\linewidth}
		\begin{adjustbox}{width=1\textwidth}\begin{tabular}{l l | c | c | c | c}
   & {\bf Method} & {\bf Moderate} & {\bf Easy} & {\bf Hard} & {\bf Runtime}\\ \hline
1. & RRC \citep{Ren2017CVPR} & 75.33 \% & 84.16 \% & 70.39 \% & 3.6 s / GPU \\
2. & SJTU-HW \citep{Zhang2018ICIP} & 74.24 \% & 85.42 \% & 69.34 \% & 0.85s / GPU \\
3. & MS-CNN \citep{Cai2016ECCV} & 73.62 \% & 83.70 \% & 68.28 \% & 0.4 s / GPU \\
4. & GN \citep{Jung2017PRL} & 71.55 \% & 80.73 \% & 64.82 \% & 1 s / GPU \\
5. & SubCNN \citep{Xiang2017WACV} & 71.34 \% & 83.17 \% & 66.36 \% & 2 s / GPU \\
\hline
10. & sensekitti \cite{Yang2016CVPRb} & 67.28 \% & 80.12 \% & 62.25 \% & 4.5 s / GPU \\
11. & Mono3D \citep{Chen2016CVPR} & 66.66 \% & 77.30 \% & 63.44 \% & 4.2 s / GPU \\
12. & Faster R-CNN \citep{Ren2015NIPS} & 65.91 \% & 78.35 \% & 61.19 \% & 2 s / GPU \\
43. & YOLOv2 \cite{Redmon2017CVPR} & 16.19 \% & 20.80 \% & 15.43 \% & 0.02 s / GPU \\
\end{tabular}
\end{adjustbox}
		\caption{KITTI Pedestrian Detection Leaderboard}
		\label{tab:kitti_object_det2d_pedestrian_image}
		\vspace{0.2cm}
	\end{subtable}
	\begin{subtable}{\linewidth}
		\begin{adjustbox}{width=1\textwidth}\begin{tabular}{l l | c | c | c | c}
   & {\bf Method} & {\bf Moderate} & {\bf Easy} & {\bf Hard} & {\bf Runtime}\\ \hline
1. & RRC \citep{Ren2017CVPR} & 76.49 \% & 84.96 \% & 65.46 \% & 3.6 s / GPU \\
2. & MS-CNN \citep{Cai2016ECCV} & 74.45 \% & 82.34 \% & 64.91 \% & 0.4 s / GPU \\
3. & Deep3DBox \citep{Mousavian2017CVPR} & 73.48 \% & 82.65 \% & 64.11 \% & 1.5 s / GPU \\
4. & SDP+RPN \citep{Yang2016CVPR} & 73.08 \% & 81.05 \% & 64.88 \% & 0.4 s / GPU \\
5. & sensekitti \citep{Yang2016CVPRb} & 72.50 \% & 81.76 \% & 64.00 \% & 4.5 s / GPU \\
6. & SubCNN \citep{Xiang2017WACV} & 70.77 \% & 77.82 \% & 62.71 \% & 2 s / GPU \\
\hline
12. & Mono3D \citep{Chen2016CVPR} & 63.85 \% & 75.22 \% & 58.96 \% & 4.2 s / GPU \\
13. & Faster R-CNN \citep{Ren2015NIPS} & 62.81 \% & 71.41 \% & 55.44 \% & 2 s / GPU \\
26. & YOLOv2 \citep{Redmon2017CVPR} & 4.55 \% & 4.55 \% & 4.55 \% & 0.02 s / GPU 
\end{tabular}
\end{adjustbox}
		\caption{KITTI Cyclist Detection Leaderboard}
		\label{tab:kitti_object_det2d_cyclist_image}
	\end{subtable}
	\caption{{\bf KITTI Object Detection Leaderboard.} Only image-based methods are shown in these tables, \ie no laser scan data is used. The numbers represent average precision at different levels of difficulty based on the object size and the level of occlusion/truncation. Higher numbers indicate better performance. Methods below the horizontal line show older entries for reference. Accessed on: June 2019.}
	\label{tab:kitti_object_det2d_img}
\end{table*}
In Tables \ref{tab:kitti_object_det2d_img} and \ref{tab:kitti_object_det2d_lidar}, we show the current state of the art on the KITTI benchmark for object, pedestrian, and cyclist detection from images. Note that for all result tables in this book, we list only public methods that have a technical paper associated with them that describes the details of the method. 

Region-based networks \citep{Girshick2014CVPR,He2014ECCV,Girshick2015ICCV} have proven to be very successful on the PASCAL VOC benchmark. However, they could not achieve similar performance on KITTI benchmark. The main reason is that objects occur at many different scales, and objects are often partially occluded. These objects are hard to detect using generic region-based networks. 

In contrast, Region Proposal Networks \citep{Ren2015NIPS,Yang2016CVPR,Cai2016ECCV,Xiang2017WACV} have been more successful on the KITTI dataset. In the case of small objects, strong activations of convolutional neurons are more likely to occur in earlier layers. Therefore, \citet{Yang2016CVPR} (SDP+PRN) propose cascaded rejection classifiers that gradually reject negative proposals using stronger features. Combined with a scale-dependent pooling approach that provides convolutional features from the corresponding scale for each proposal, they achieve competitive results on KITTI cyclist \ref{tab:kitti_object_det2d_cyclist_image}. \citet{Xiang2017WACV} (SubCNN) improve on the orientation estimation task by guiding the proposal generating and detection network using subcategory information obtained from 3DVP \citep{Xiang2015CVPR}. Object subcategories are defined for objects with similar properties or attributes such as appearance, pose, or shape. This formulation allows them to achieve the best performance in pedestrian orientation estimation (\tabref{tab:kitti_object_detori2d_pedestrian_image}).
\begin{figure}[t]
	\centering
	\includegraphics[width=1.00\columnwidth]{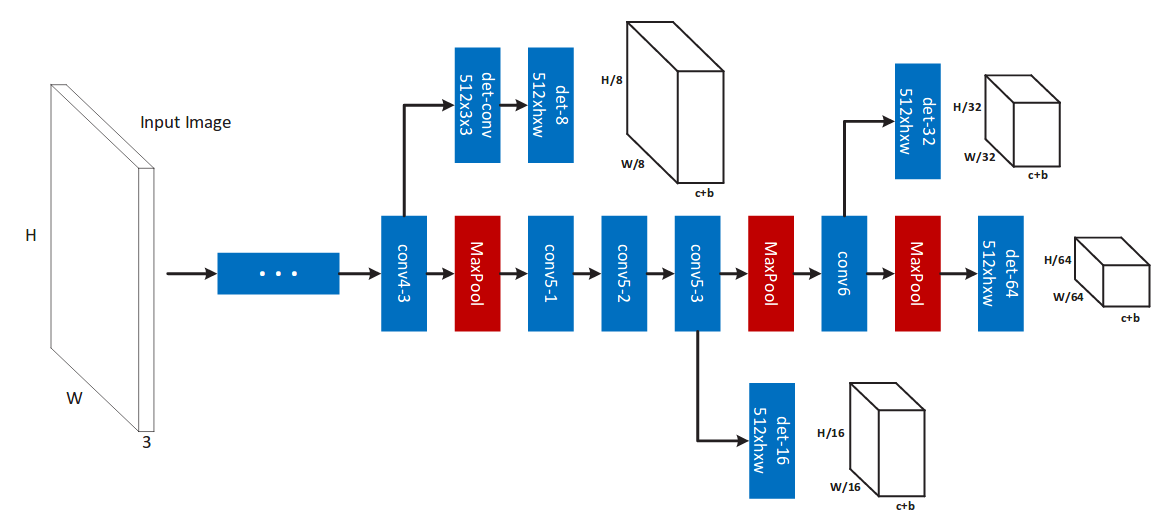}
	\caption[Multi-scale Deep CNN for Object Detection]{\textbf{Multi-scale Deep CNN for Object Detection.} The proposal sub-network presented by \protect\citet{Cai2016ECCV} performs detection at multiple output layers to match objects at different scales. Scale-specific detectors are combined to produce a strong multi-scale object detector. \figsourceSpringer{\protect\citet{Cai2016ECCV}}{2016}{ECCV}.
	}
	\label{fig:Cai2016ECCV}
\end{figure}
The best performing Region-Proposal networks are presented by \citet{Cai2016ECCV, Chabot2017CVPR}. MS-CNN \citep{Cai2016ECCV} consists of two subnetworks, \ie a multi-scale proposal network and a detection network.
The proposal network, illustrated in \figref{fig:Cai2016ECCV}, has several output layers corresponding to different scales. On each output layer, the detection network is applied that allows detecting objects at different scales. Their multi-scale CNN performs well on pedestrians and cyclists (Tables \ref{tab:kitti_object_det2d_pedestrian_image},\ref{tab:kitti_object_det2d_cyclist_image}). Deep MANTA \citep{Chabot2017CVPR} leverages a 3D vehicle model dataset for 3D vehicle detection from images. They first extract region proposals from the input image with an iterative refinement of region proposals using a coarse-to-fine CNN (Deep MANTA network). Afterwards, they use a network to choose the closest 3D model from the 3D dataset and perform matching between the 2D regions from the image and 3D models to recover the vehicle orientation and 3D location. This allows them even to detect parts of cars that are occluded and estimate the orientation of cars. 
They achieve competitive results on car detection (\tabref{tab:kitti_object_det2d_car_image}) and the best performance on the car orientation estimation task (\tabref{tab:kitti_object_detori2d_car_image}).

One-stage detectors \citep{Redmon2016CVPR,Redmon2017CVPR,Lin2017ICCV} have similar difficulties with objects at different scales and occlusions as region-based networks on the KITTI dataset. However, by leveraging feature pyramids as in \citep{Liu2016ECCV}, one-stage detectors \citep{Ren2017CVPR, Zhang2018ICIP} can reach state-of-the-art performance. \citet{Zhang2018ICIP} (SJTU-HW) propose to improve the localization by embedding a localization-quality estimation into the detector. They fuse features from classification and box regression subnetworks to estimate the localization quality. During inference, they combine the localization quality with the classification confidence to obtain more accurate detections. This approach outperforms all Region Proposal Networks on the pedestrian and car detection tasks. However, the best performance on all detection tasks (Tables \ref{tab:kitti_object_det2d_img}) is achieved by \citet{Ren2017CVPR}. Inspired by feature pyramids used in \citep{Liu2016ECCV}, they propose a Recurrent Rolling Convolution architecture that aggregates contextual information from multiple scales. By providing this rich contextual information to the classifier and box regressors, they achieve state-of-the-art performance on all KITTI detection tasks.

\begin{table*}[p!]
	\begin{subtable}{\linewidth}
		\begin{adjustbox}{width=1\textwidth}\begin{tabular}{l l | c | c | c | c}
   & {\bf Method} & {\bf Moderate} & {\bf Easy} & {\bf Hard} & {\bf Runtime}\\ \hline
1. & Deep MANTA \citep{Chabot2017CVPR} & 89.86 \% & 97.19 \% & 80.39 \% & 0.7 s / GPU \\
2. & Deep3DBox \citep{Mousavian2017CVPR} & 88.56 \% & 90.39 \% & 77.17 \% & 1.5 s / GPU \\
3. & SubCNN \citep{Xiang2017WACV} & 88.43 \% & 90.61 \% & 78.63 \% & 2 s / GPU \\
4. & AVOD (LiDAR) \citep{Ku2018IROS} & 87.46 \% & 89.59 \% & 79.54 \% & 0.08 s / GPU \\
5. & AVOD-FPN (LiDAR) \citep{Ku2018IROS} & 87.13 \% & 89.95 \% & 79.74 \% & 0.1 s / GPU \\
\hline
13. & Pose-RCNN \citep{Braun2016ITSC} & 75.35 \% & 88.78 \% & 61.47 \% & 2 s / >8 cores \\
25. & sensekitti \citep{Yang2016CVPRb} & 44.56 \% & 47.06 \% & 41.50 \% & 4.5 s / GPU \\
\end{tabular}\end{adjustbox}
		\caption{KITTI Car Detection and Orientation Estimation Leaderboard}
		\label{tab:kitti_object_detori2d_car_image}
		\vspace{0.2cm}
	\end{subtable}
	\begin{subtable}{\linewidth}
		\begin{adjustbox}{width=1\textwidth}\begin{tabular}{l l | c | c | c | c}
   & {\bf Method} & {\bf Moderate} & {\bf Easy} & {\bf Hard} & {\bf Runtime}\\ \hline
1. & SubCNN \citep{Xiang2017WACV} & 63.41 \% & 71.39 \% & 56.34 \% & 2 s / GPU \\
2. & Pose-RCNN \citep{Braun2016ITSC} & 62.25 \% & 74.85 \% & 55.09 \% & 2 s / >8 cores \\
3. & Deep3DBox \citep{Mousavian2017CVPR} & 59.37 \% & 68.58 \% & 51.97 \% & 1.5 s / GPU \\
4. & 3DOP (Stereo) \citep{Chen2015NIPS} & 58.59 \% & 71.95 \% & 52.35 \% & 3s / GPU \\
5. & AVOD-FPN (LiDAR) \citep{Ku2018IROS} & 57.53 \% & 67.61 \% & 54.16 \% & 0.1 s / GPU \\
\hline
8. & AVOD (LiDAR)\citep{Ku2018IROS} & 54.43 \% & 64.36 \% & 47.67 \% & 0.08 s / GPU \\
9. & Mono3D \citep{Chen2016CVPR} & 53.11 \% & 65.74 \% & 48.87 \% & 4.2 s / GPU \\
10. & FRCNN+Or \citep{Guindel2018ITSM} & 50.91 \% & 63.41 \% & 45.46 \% & 0.09 s / GPU \\
11. & sensekitti \citep{Yang2016CVPRb} & 42.12 \% & 46.65 \% & 36.66 \% & 4.5 s / GPU \\
\end{tabular}

\end{adjustbox}
		\caption{KITTI Pedestrian Detection and Orientation Estimation Leaderboard}
		\label{tab:kitti_object_detori2d_pedestrian_image}
		\vspace{0.2cm}
	\end{subtable}
	\begin{subtable}{\linewidth}
		\begin{adjustbox}{width=1\textwidth}\begin{tabular}{l l | c | c | c | c}
   & {\bf Method} & {\bf Moderate} & {\bf Easy} & {\bf Hard} & {\bf Runtime}\\ \hline
1. & SubCNN \citep{Xiang2017WACV} & 66.28 \% & 78.33 \% & 61.37 \% & 2 s / GPU \\
2. & Pose-RCNN \citep{Braun2016ITSC} & 59.89 \% & 74.10 \% & 54.21 \% & 2 s / >8 cores \\
3. & 3DOP (Stereo) \citep{Chen2015NIPS} & 59.79 \% & 73.46 \% & 57.04 \% & 3s / GPU \\
4. & DeepStereoOP \citep{Pham2017SPIC} & 59.28 \% & 73.37 \% & 56.87 \% & 3.4 s / GPU \\
5. & Mono3D \citep{Chen2016CVPR} & 58.12 \% & 68.58 \% & 54.94 \% & 4.2 s / GPU \\
\hline
6. & AVOD-FPN (LiDAR) \citep{Ku2018IROS} & 44.92 \% & 53.36 \% & 43.77 \% & 0.1 s / GPU \\
7. & SECOND \citep{Yan2018SEN} & 43.51 \% & 51.56 \% & 38.78 \% & 0.04 s / GPU \\
8. & DPM-VOC+VP \citep{Pepik2015PAMI} & 39.83 \% & 53.66 \% & 35.73 \% & 8 s / 1 core \\
9. & sensekitti \citep{Yang2016CVPRb} & 37.50 \% & 43.55 \% & 35.08 \% & 4.5 s / GPU \\
10. & AVOD (LiDAR) \citep{Ku2018IROS} & 36.38 \% & 44.12 \% & 31.81 \% & 0.08 s / GPU \\
\end{tabular}\end{adjustbox}
		\caption{KITTI Cyclist Detection and Orientation Estimation Leaderboard}
		\label{tab:kitti_object_detori2d_cyclist_image}
	\end{subtable}
	\caption{{\bf KITTI Detection and Orientation Estimation Leaderboard.} Only image-based methods are shown in these tables, \ie no laser scan data is used. The numbers represent average orientation similarity as described in \cite{Geiger2012CVPR}. Higher numbers indicate better detection and orientation estimation. Methods below the horizontal line show older entries for reference. Accessed on: June 2019.}
	\label{tab:kitti_object_det2d_ori}
\end{table*}

\begin{table*}[p!]
	\begin{subtable}{\linewidth}
		\begin{adjustbox}{width=1\textwidth}\begin{tabular}{l l | c | c | c | c}
   & {\bf Method} & {\bf Moderate} & {\bf Easy} & {\bf Hard} & {\bf Runtime}\\ \hline
1. & PC-CNN-V2 \citep{Du2018ICRA} & 90.15 \% & 90.79 \% & 87.58 \% & 0.5 s / GPU \\
2. & F-PointNet \citep{Qi2017CVPRa} & 90.00 \% & 90.78 \% & 80.80 \% & 0.17 s / GPU \\
3. & MV3D \citep{Chen2017CVPR} & 89.17 \% & 90.53 \% & 80.16 \% & 0.36 s / GPU \\
4. & MM-MRFC \citep{Costea2017CVPR} & 88.20 \% & 90.93 \% & 78.02 \% & 0.05 s / GPU \\
5. & AVOD \citep{Ku2018IROS} & 88.08 \% & 89.73 \% & 80.14 \% & 0.08 s / GPU \\
\hline
18. & CSoR \cite{Plotkin2015} & 26.13 \% & 35.24 \% & 22.69 \% & 3.5 s / 4 cores \\
19. & mBoW \cite{Behley2013IROS} & 23.76 \% & 37.63 \% & 18.44 \% & 10 s / 1 core \\
\end{tabular}\end{adjustbox}
		\caption{KITTI Car Detection Leaderboard}
		\label{tab:kitti_object_det2d_car_lidar}
		\vspace{0.2cm}
	\end{subtable}
	\begin{subtable}{\linewidth}
		\begin{adjustbox}{width=1\textwidth}\begin{tabular}{l l | c | c | c | c}
   & {\bf Method} & {\bf Moderate} & {\bf Easy} & {\bf Hard} & {\bf Runtime}\\ \hline
1. & F-PointNet \citep{Qi2017CVPRa} & 77.25 \% & 87.81 \% & 74.46 \% & 0.17 s / GPU \\
2. & MM-MRFC \citep{Costea2017CVPR} & 69.96 \% & 82.37 \% & 64.76 \% & 0.05 s / GPU \\
3. & AVOD-FPN \citep{Ku2018IROS} & 58.42 \% & 67.32 \% & 57.44 \% & 0.1 s / GPU \\
4. & MV-RGBD-RF \citep{Gonzalez2016TCYB} & 56.59 \% & 73.05 \% & 49.63 \% & 4 s / 4 cores \\
5. & Vote3Deep \citep{Engelcke2017ICRA} & 55.38 \% & 67.94 \% & 52.62 \% & 1.5 s / 4 cores \\
8. & AVOD \cite{Ku2018IROS} & 43.49 \% & 51.64 \% & 37.79 \% & 0.08 s / GPU \\
9. & Vote3D \cite{Wang2015RSS} & 35.74 \% & 44.47 \% & 33.72 \% & 0.5 s / 4 cores \\
10. & mBoW \cite{Behley2013IROS} & 31.37 \% & 44.36 \% & 30.62 \% & 10 s / 1 core \\
\end{tabular}\end{adjustbox}
		\caption{KITTI Pedestrian Detection Leaderboard}
		\label{tab:kitti_object_det2d_pedestrian_lidar}
		\vspace{0.2cm}
	\end{subtable}
	\begin{subtable}{\linewidth}
		\begin{adjustbox}{width=1\textwidth}\begin{tabular}{l l | c | c | c | c}
   & {\bf Method} & {\bf Moderate} & {\bf Easy} & {\bf Hard} & {\bf Runtime}\\ \hline
1. & F-PointNet \cite{Qi2017CVPRa} & 72.25 \% & 84.90 \% & 65.14 \% & 0.17 s / GPU \\
2. & Vote3Deep \cite{Engelcke2017ICRA} & 67.96 \% & 76.49 \% & 62.88 \% & 1.5 s / 4 cores \\
3. & AVOD-FPN \cite{Ku2018IROS} & 59.32 \% & 68.65 \% & 55.82 \% & 0.1 s / GPU \\
4. & AVOD \cite{Ku2018IROS} & 56.01 \% & 65.72 \% & 48.89 \% & 0.08 s / GPU \\
5. & BirdNet \cite{Beltran2018ITSC} & 49.04 \% & 64.88 \% & 46.61 \% & 0.11 s / GPU \\
6. & MV-RGBD-RF \cite{Gonzalez2016TCYB} & 42.61 \% & 51.46 \% & 37.42 \% & 4 s / 4 cores \\
7. & Vote3D \cite{Wang2015RSS} & 31.24 \% & 41.45 \% & 28.60 \% & 0.5 s / 4 cores \\
8. & mBoW \cite{Behley2013IROS} & 21.62 \% & 28.19 \% & 20.93 \% & 10 s / 1 core \\
\end{tabular}\end{adjustbox}
		\caption{KITTI Cyclist Detection Leaderboard}
		\label{tab:kitti_object_det2d_cyclist_lidar}
	\end{subtable}
	\caption{{\bf KITTI LiDAR Detection Leaderboard.} Methods that focus on LiDAR scans and methods combining LiDAR with RGB images are presented. The numbers represent average precision at different levels of difficulty. Higher numbers indicate better performance. Methods below the horizontal line show older entries for reference. Accessed on: June 2019.}
	\label{tab:kitti_object_det2d_lidar}
\end{table*}

\boldparagraph{3D Object Detection from 3D Point Clouds} 
In \tabref{tab:kitti_object_det2d_lidar}, we show the LiDAR-based state of the art on the KITTI benchmark for object, pedestrian, and cyclist detection. The performance is assessed similarly to the image-based approaches using the intersection-over-union by projecting the 3D bounding boxes into the image plane. 

\begin{figure}[t]
	\centering
	\includegraphics[width=1.00\columnwidth]{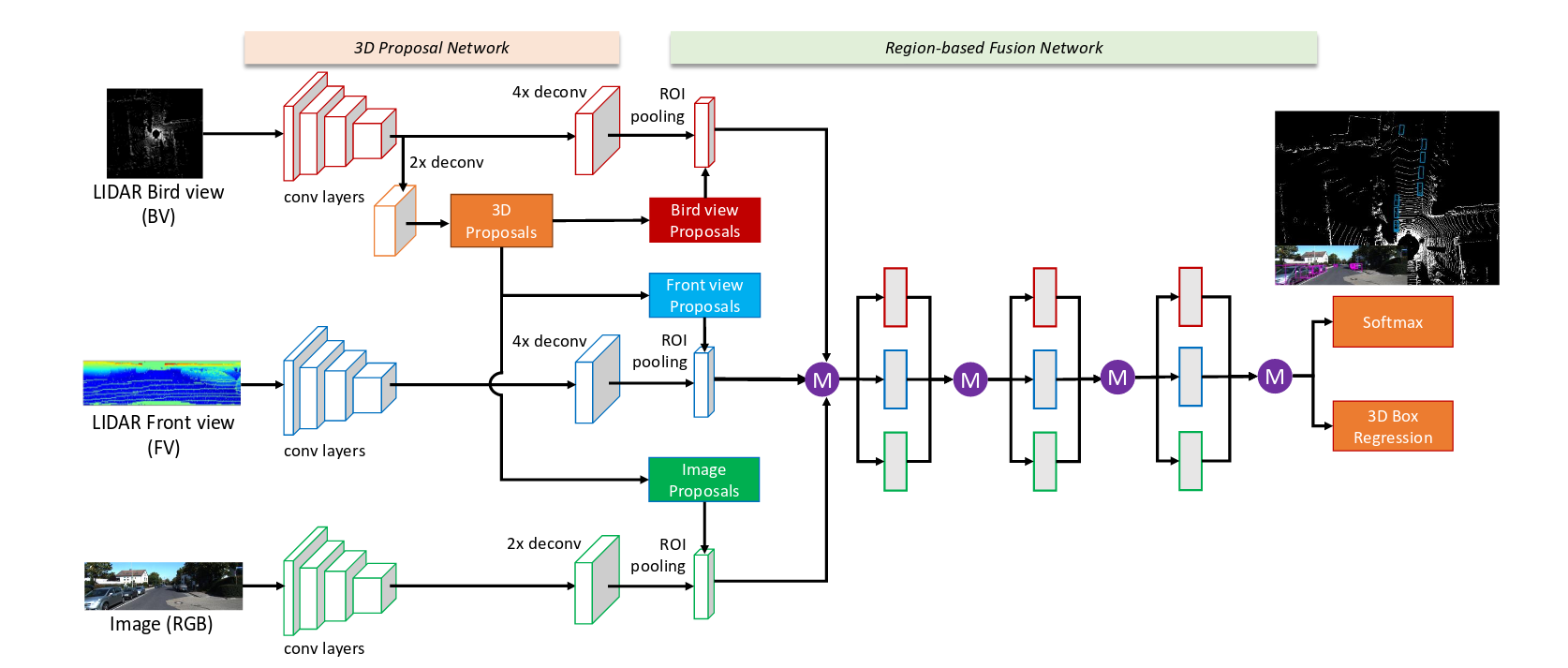}
	\caption[Multi-View 3D Object Detection Network]{\textbf{Multi-View 3D Object Detection.} The network proposed by \protect\citet{Chen2017CVPR} combines region-wise features from the bird's eye view, the front view of the LiDAR point cloud as well as the RGB  image as input for a deep fusion network. \figsourceC{\protect\citet{Chen2017CVPR}}{2017}{IEEE}.}
	\label{fig:Chen2017CVPR}
\end{figure}
\citet{Chen2017CVPR} encode sparse point clouds using a compact multi-view representation. While the proposal generation network utilizes the bird's eye view to generate 3D candidates, they eventually combine region-wise features from multiple views via deep fusion for the final detection and box regression scheme, as illustrated in \figref{fig:Chen2017CVPR}. Instead of using an intermediate representation, \citet{Ku2018IROS} propose to directly share features extracted from LiDAR point clouds and RGB images with a Region Proposal Network (RPN) and detector network.  \citet{Costea2017CVPR} improve on the pedestrian and car detection task by considering dense optical flow as additional input. They use multi-modal, multi-resolution filtering of intensities, gradient magnitudes and orientations to obtain discriminative features for detection. In contrast to \citep{Chen2017CVPR, Ku2018IROS, Qi2017CVPRa, Du2018ICRA}, they follow a boosting-based sliding window approach and achieve competitive results while being faster than the deep learning-based approaches. 

\citet{Qi2017CVPRa} propose to directly work on the 3D point clouds by reducing the search space using 2D detections in image space. This allows them to use two variants of PointNet \citep{GarciaGarcia2016IJCNN}; one for 3D object instance segmentation and the other for 3D box regression. With this approach, they outperform all other 3D-based  detectors on the categories pedestrian and cyclist (Tables \ref{tab:kitti_object_det2d_pedestrian_lidar}, \ref{tab:kitti_object_det2d_cyclist_lidar}) and even all image-based detectors on pedestrians (\tabref{tab:kitti_object_det2d_pedestrian_image}). Similar to \citep{Qi2017CVPRa}, \citet{Du2018ICRA} leverage 2D detections to obtain accurate 3D detections. Instead of using PointNets, they propose to fit a generalized 3D car model to the points corresponding to 2D detections. Finally, they use the points matching the model in a two-stage refinement CNN to predict the final 3D box and an objectiveness score. The combination of 2D and 3D detection allows them to outperform all 3D-based detectors on cars (\tabref{tab:kitti_object_det2d_car_lidar}) while achieving a performance on par with the best performing 2D-based detector \citep{Ren2017CVPR} on the pedestrian category (\tabref{tab:kitti_object_det2d_car_image}).

\section{Discussion}
\label{sec:detection_discussion}
\begin{figure*}[p]
\begin{subfigure}{\linewidth}
\includegraphics[width=0.33\linewidth]{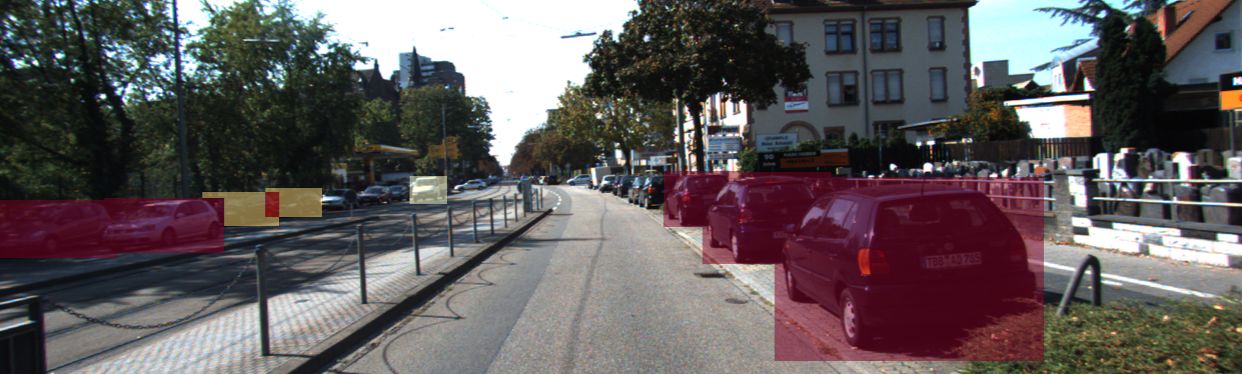}%
\includegraphics[width=0.33\linewidth]{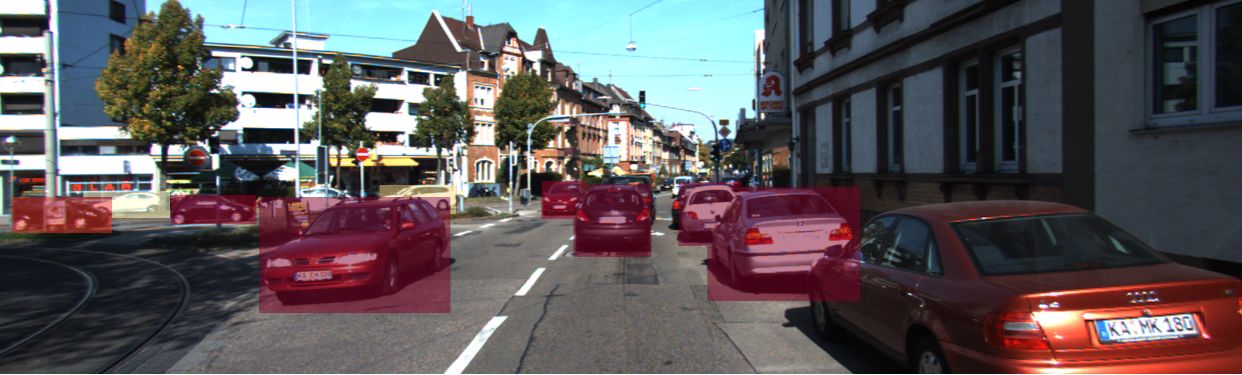}%
\includegraphics[width=0.33\linewidth]{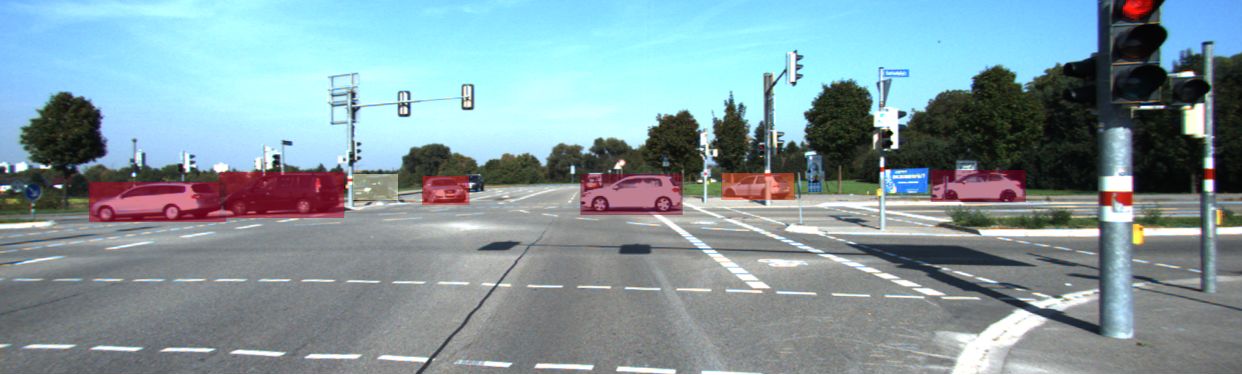}\\%
\includegraphics[width=0.33\linewidth]{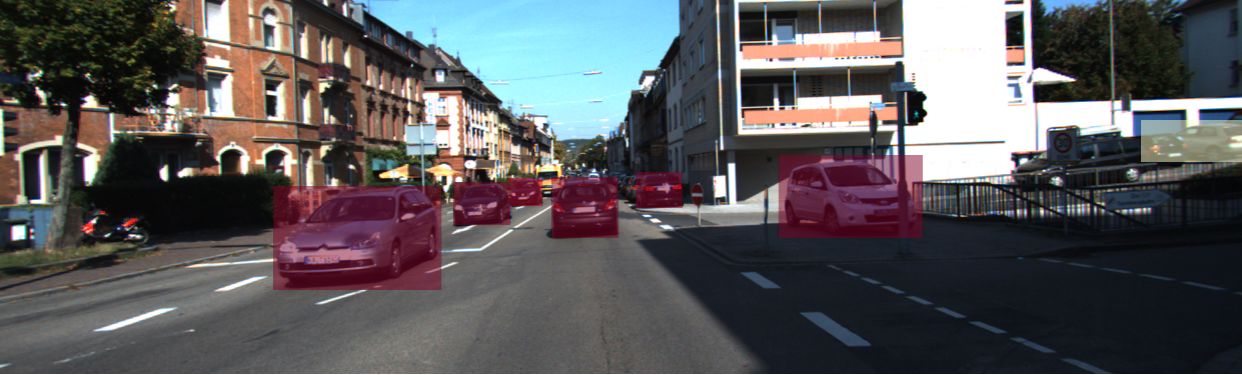}%
\includegraphics[width=0.33\linewidth]{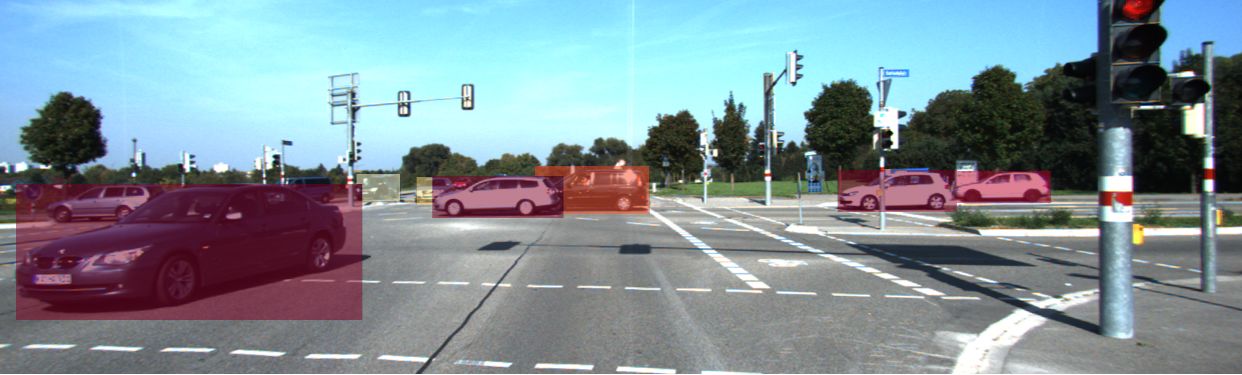}%
\includegraphics[width=0.33\linewidth]{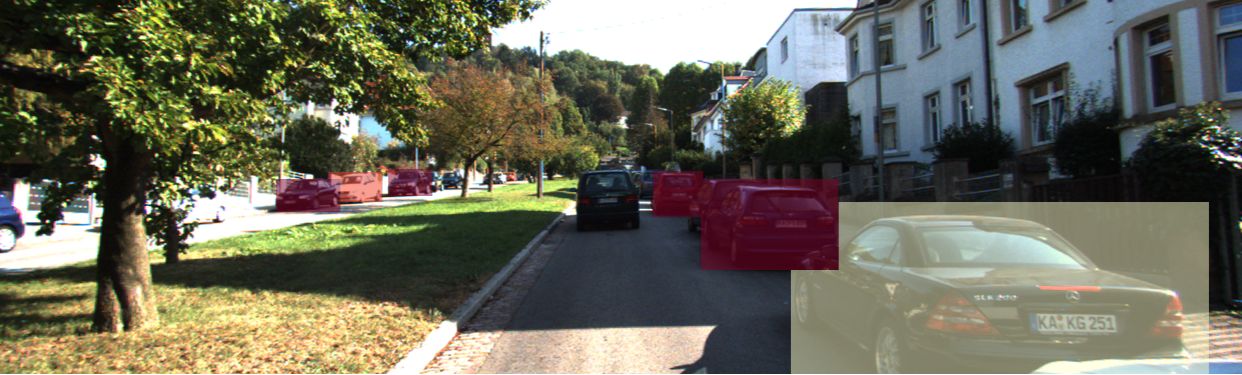}\\%
\includegraphics[width=0.33\linewidth]{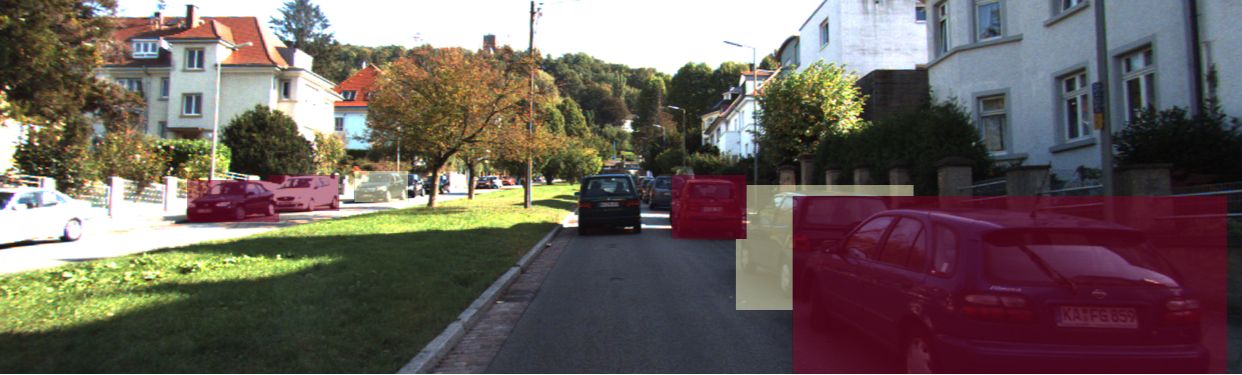}%
\includegraphics[width=0.33\linewidth]{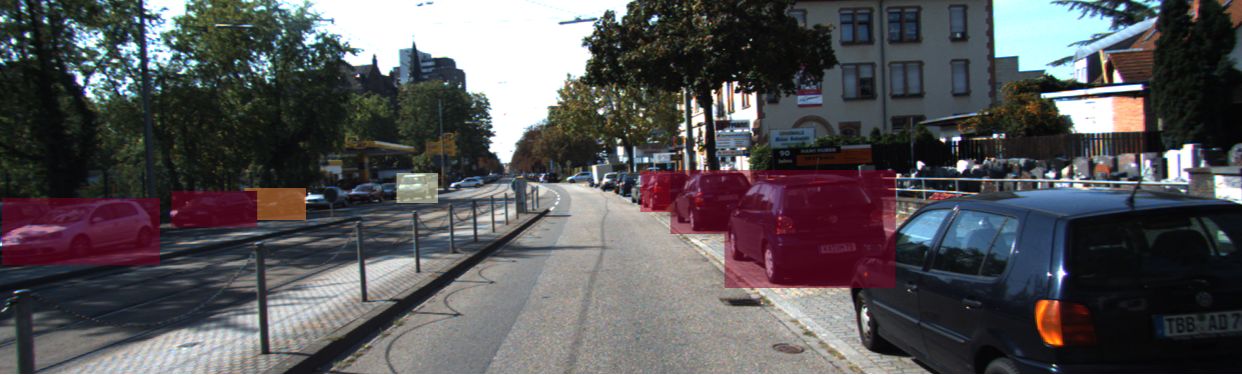}%
\includegraphics[width=0.33\linewidth]{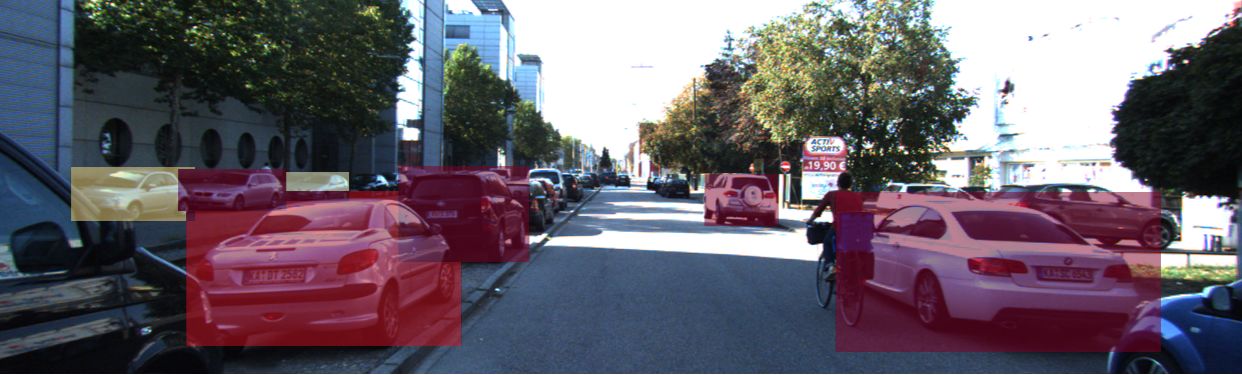}%
\caption{Images with Largest Number of True Positive Detections}
\end{subfigure}
\begin{subfigure}{\linewidth}
\includegraphics[width=0.33\linewidth]{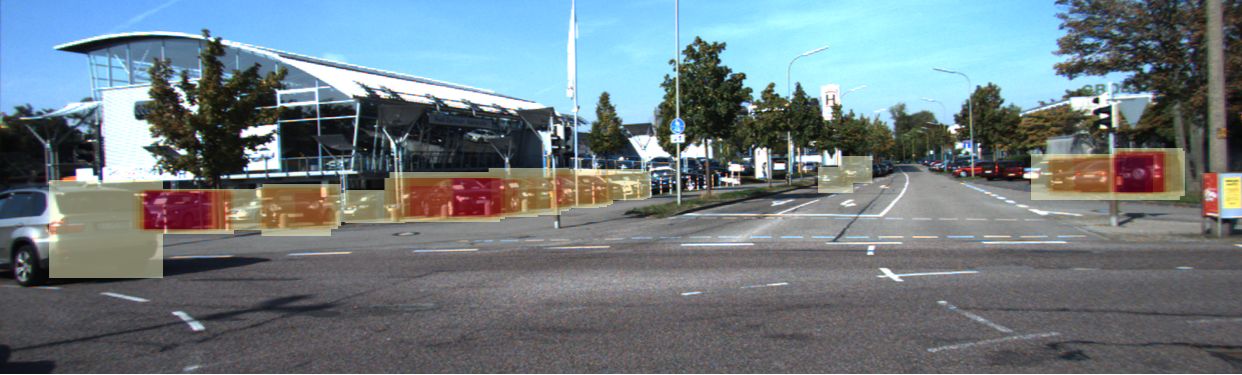}%
\includegraphics[width=0.33\linewidth]{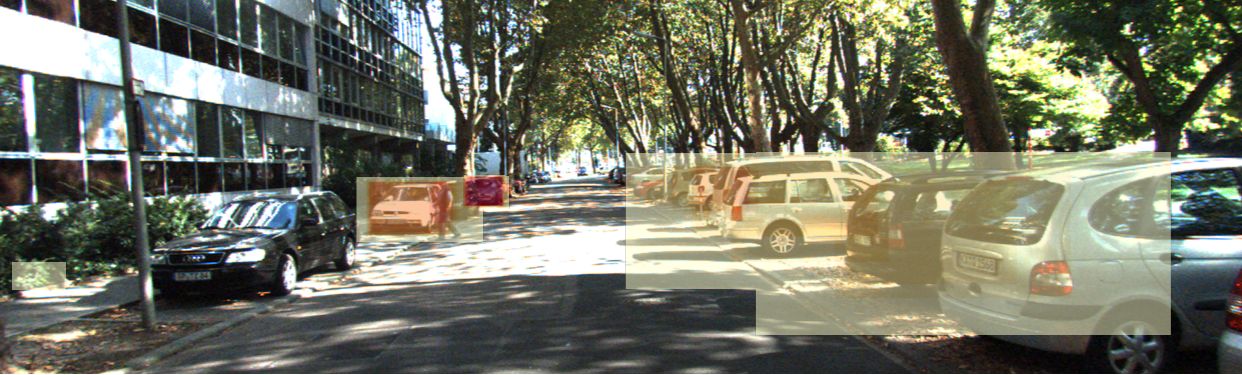}%
\includegraphics[width=0.33\linewidth]{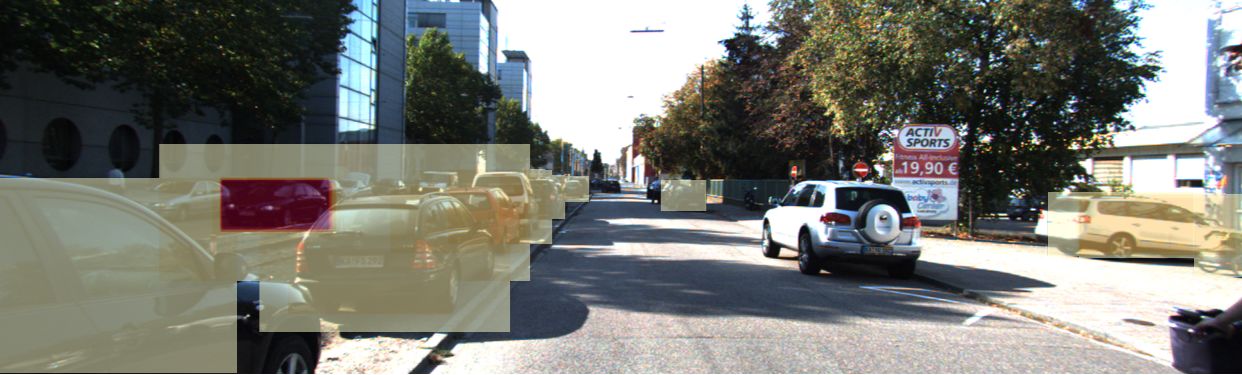}\\%
\includegraphics[width=0.33\linewidth]{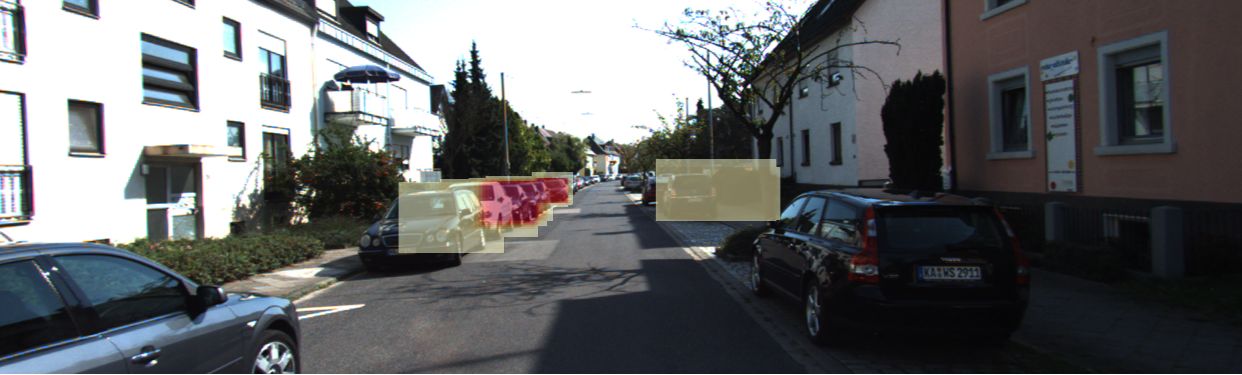}%
\includegraphics[width=0.33\linewidth]{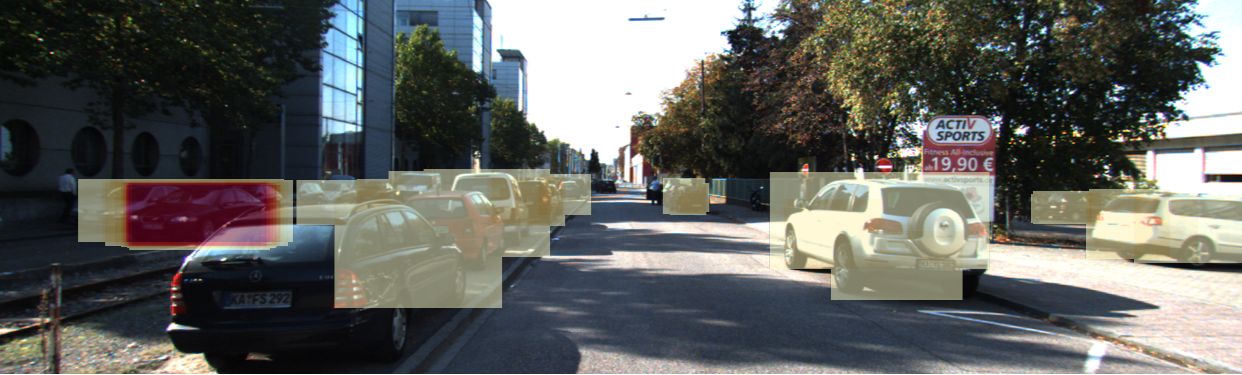}%
\includegraphics[width=0.33\linewidth]{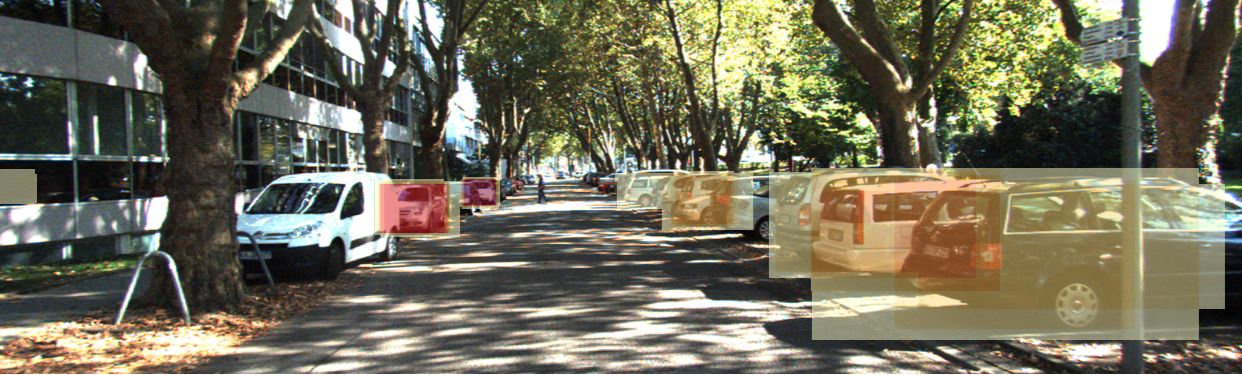}\\%
\includegraphics[width=0.33\linewidth]{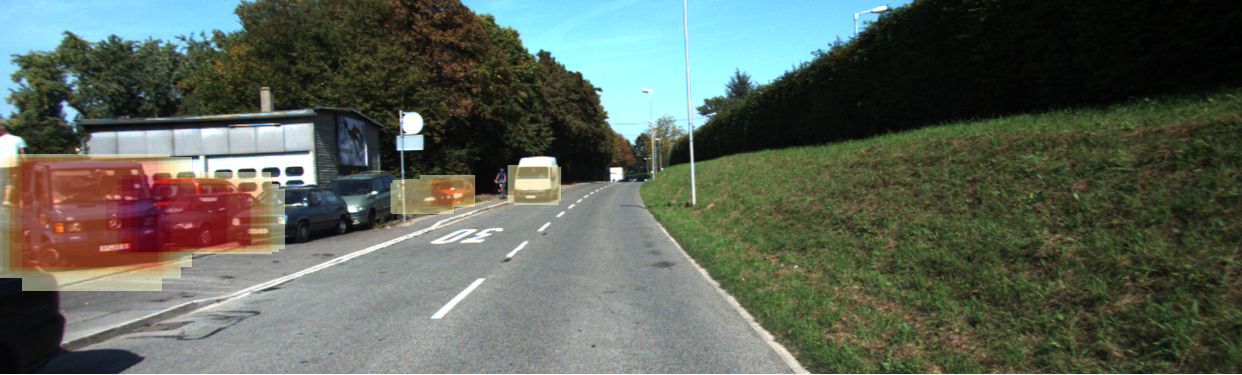}%
\includegraphics[width=0.33\linewidth]{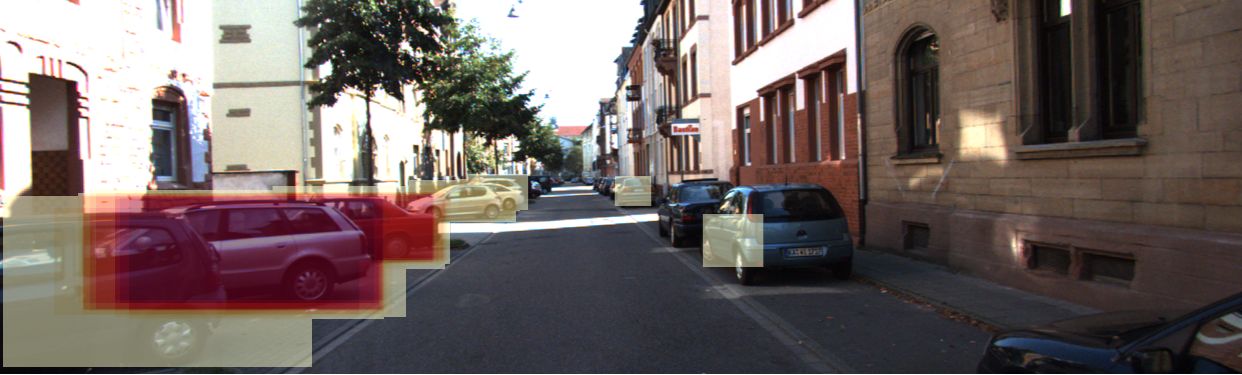}%
\includegraphics[width=0.33\linewidth]{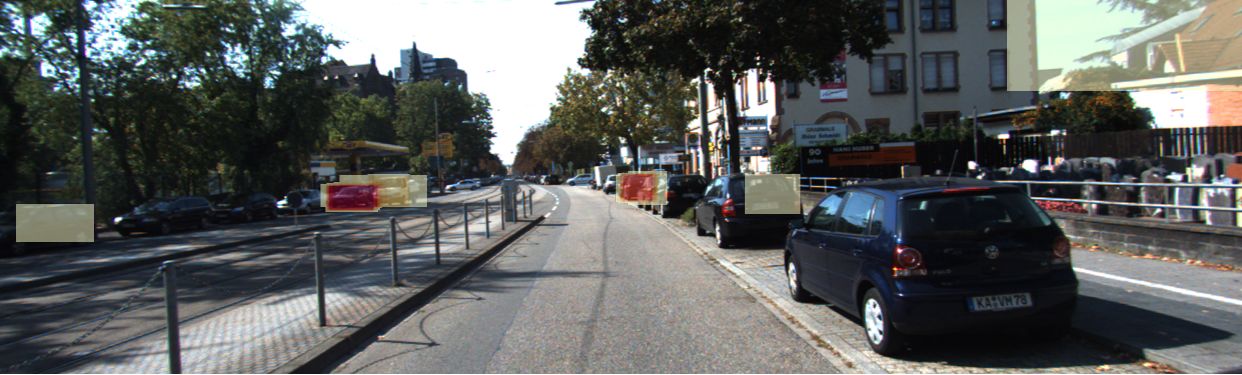}%
\caption{Images with Largest Number of False Positive Detections}
\end{subfigure}
\begin{subfigure}{\linewidth}
\includegraphics[width=0.33\linewidth]{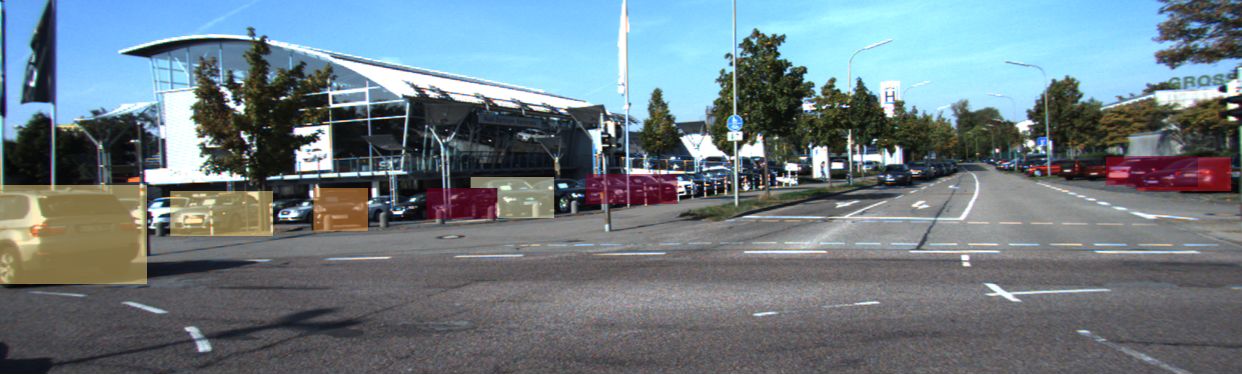}%
\includegraphics[width=0.33\linewidth]{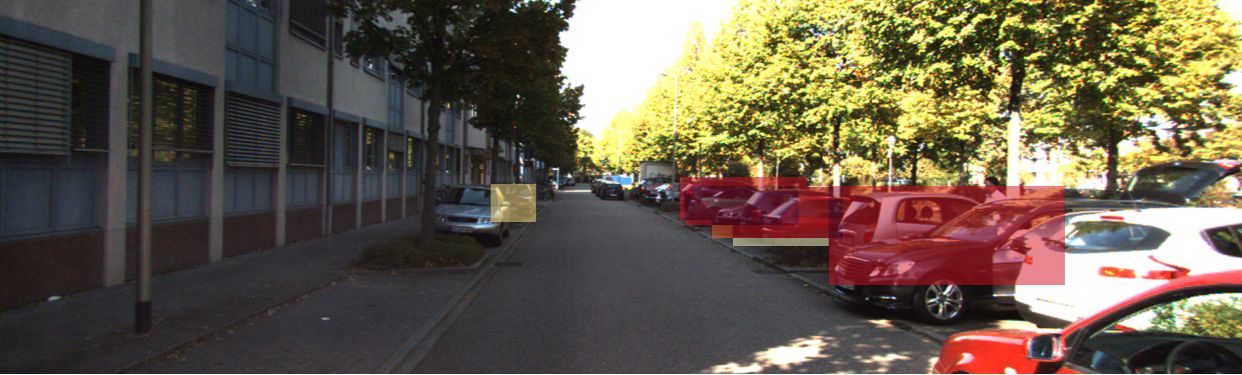}%
\includegraphics[width=0.33\linewidth]{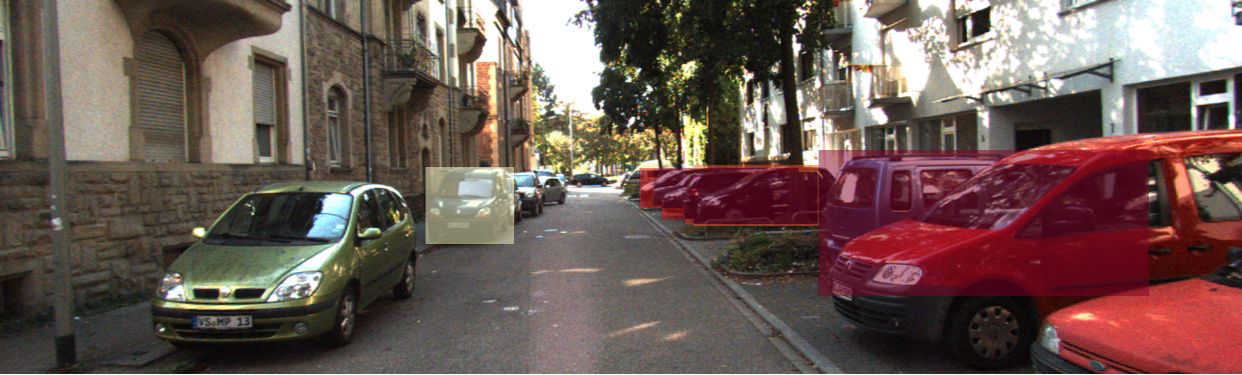}\\%
\includegraphics[width=0.33\linewidth]{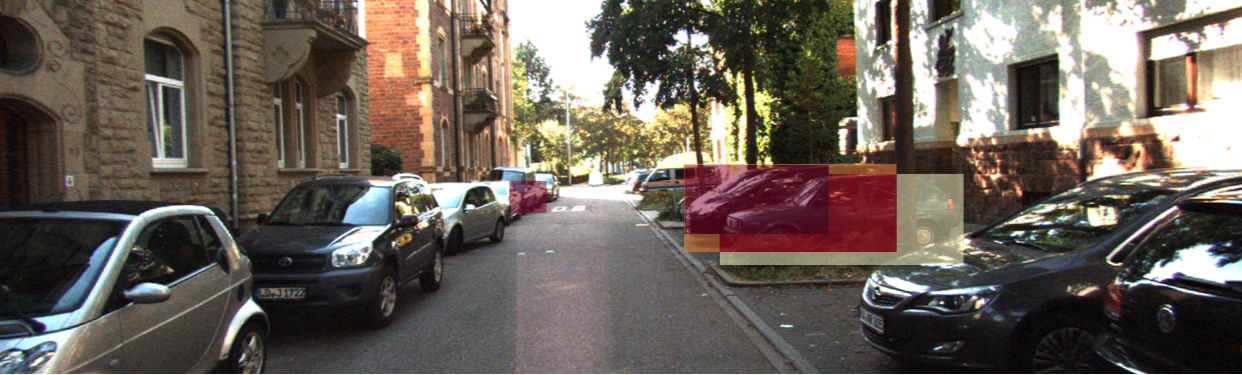}%
\includegraphics[width=0.33\linewidth]{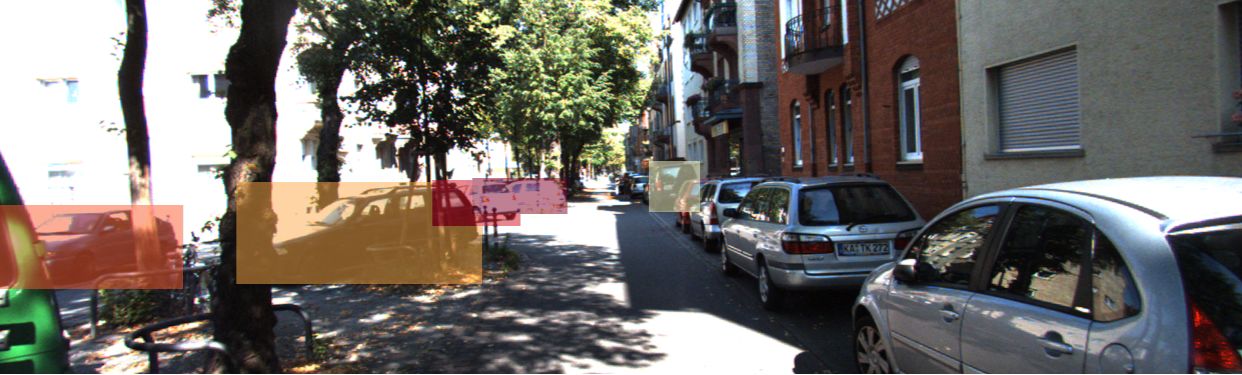}%
\includegraphics[width=0.33\linewidth]{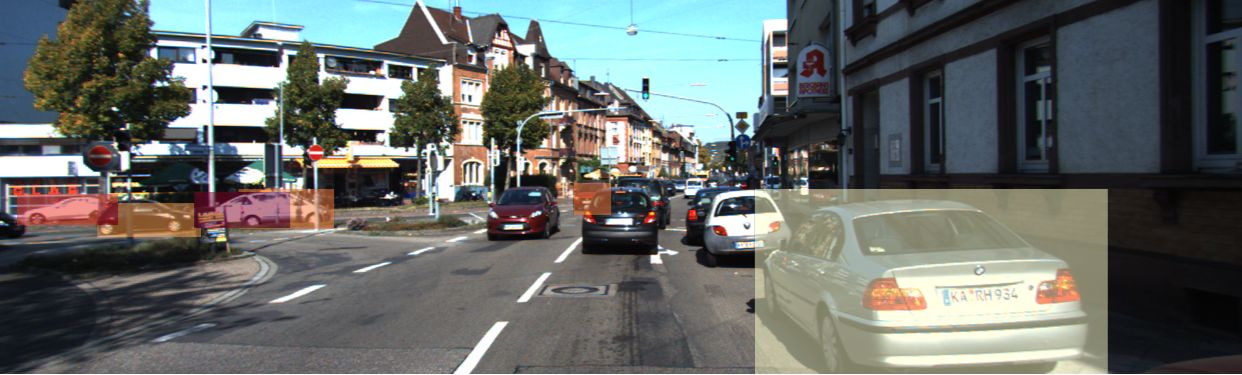}\\%
\includegraphics[width=0.33\linewidth]{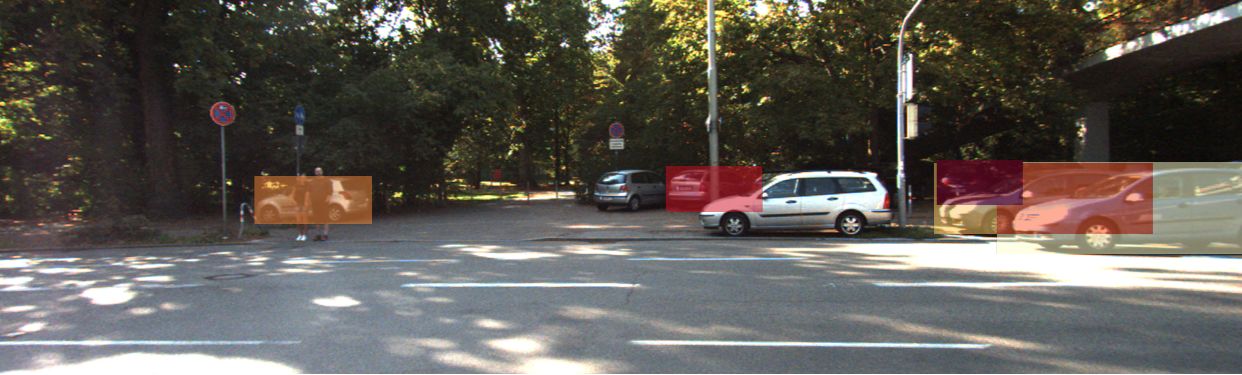}%
\includegraphics[width=0.33\linewidth]{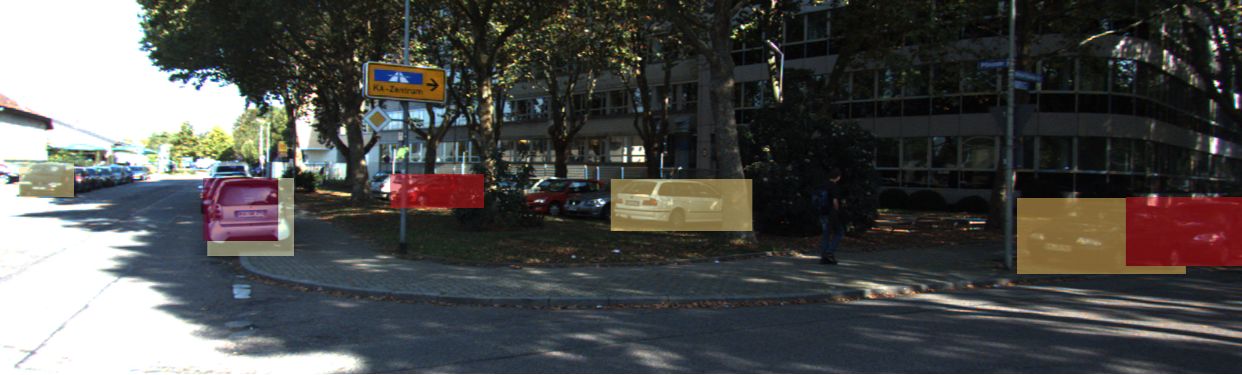}%
\includegraphics[width=0.33\linewidth]{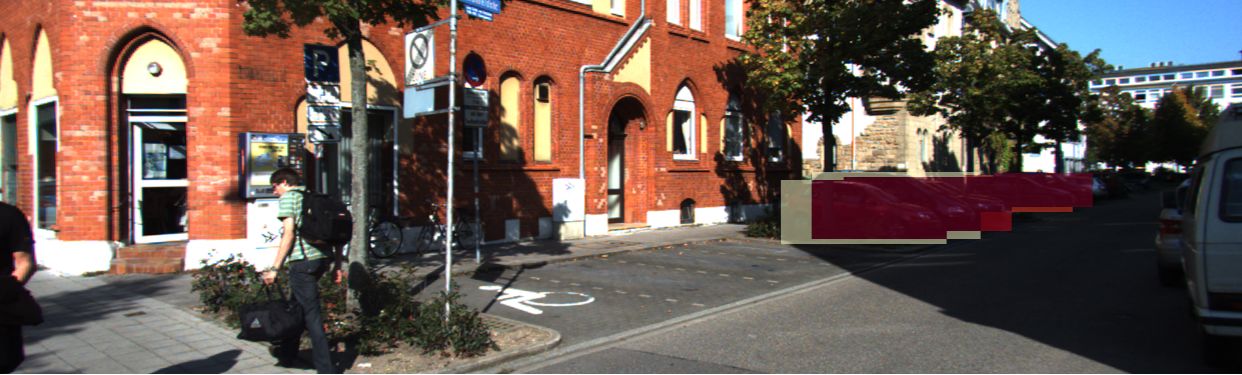}%
\caption{Images with Largest Number of False Negative Detections}
\end{subfigure}
\caption{{\bf KITTI Vehicle Detection Analysis.} Each figure shows images with a large number of true positive (TP) detections, false positive (FP) detections and false negative (FN) detections, respectively. If all detectors agree on TP, FP or FN, the object is marked in red. If only some of the detectors agree, the object is marked in yellow. The ranking has been established by considering the 15 leading methods published on the KITTI evaluation server at time of submission.}
\label{fig:car_detection_qualitative_results}
\end{figure*}

\begin{figure*}[p]
\begin{subfigure}{\linewidth}
\includegraphics[width=0.33\linewidth]{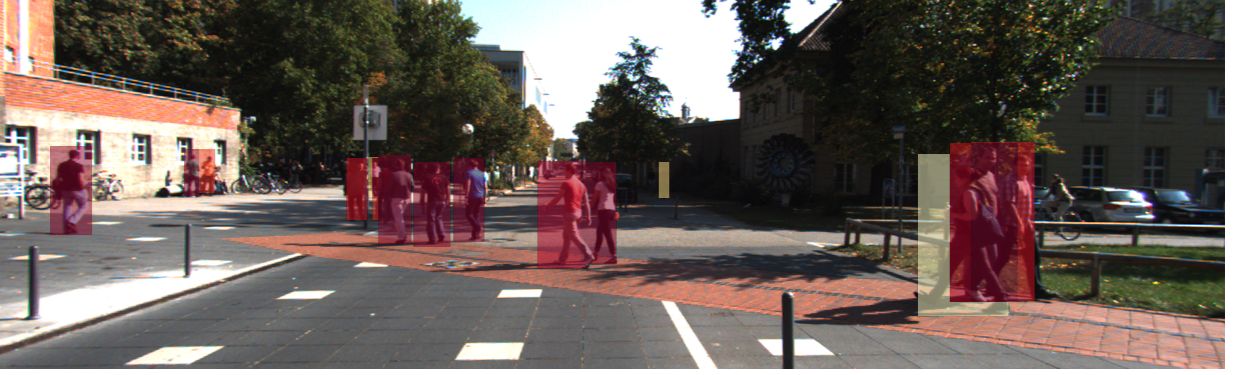}%
\includegraphics[width=0.33\linewidth]{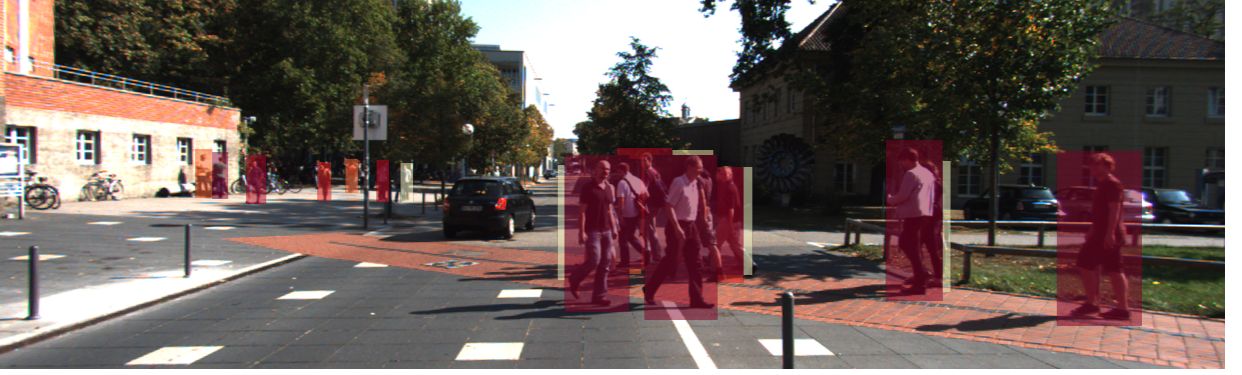}%
\includegraphics[width=0.33\linewidth]{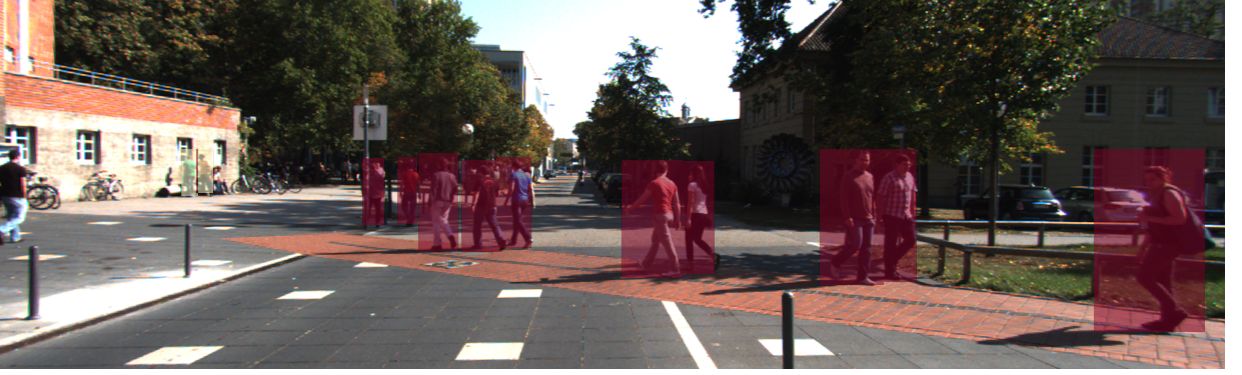}\\%
\includegraphics[width=0.33\linewidth]{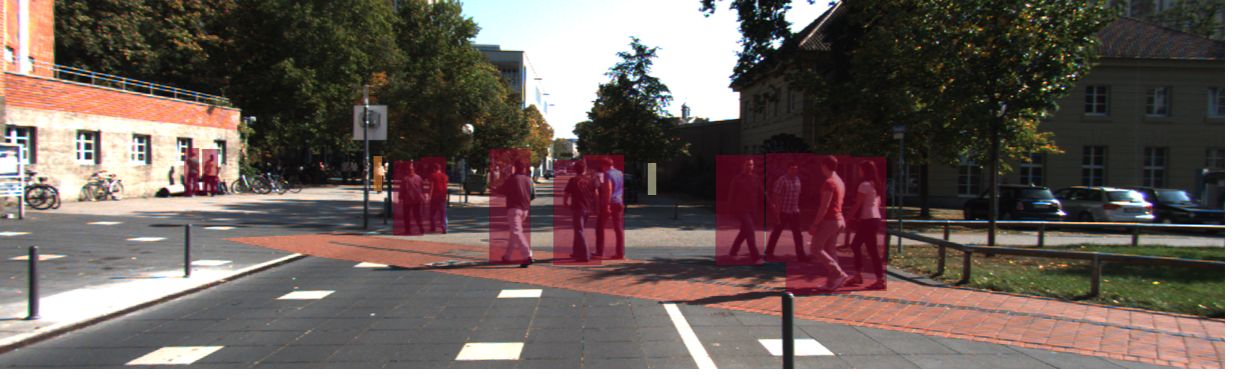}%
\includegraphics[width=0.33\linewidth]{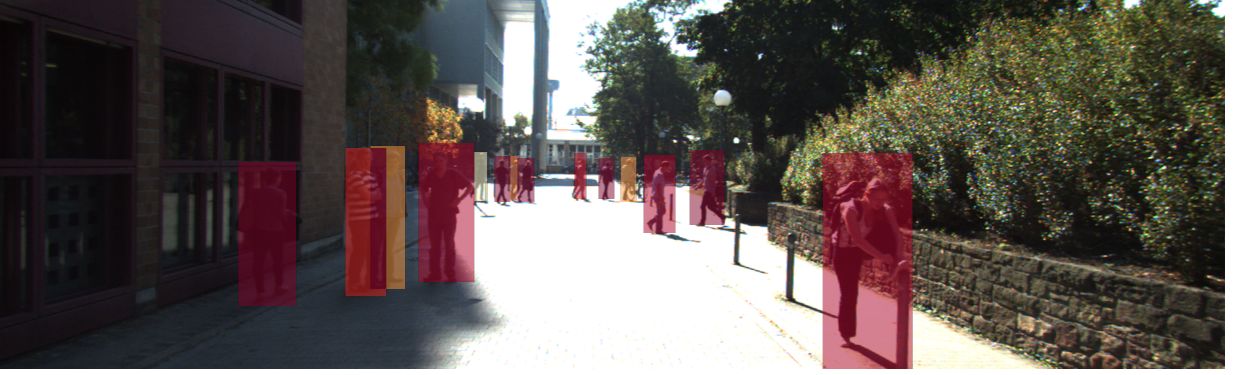}%
\includegraphics[width=0.33\linewidth]{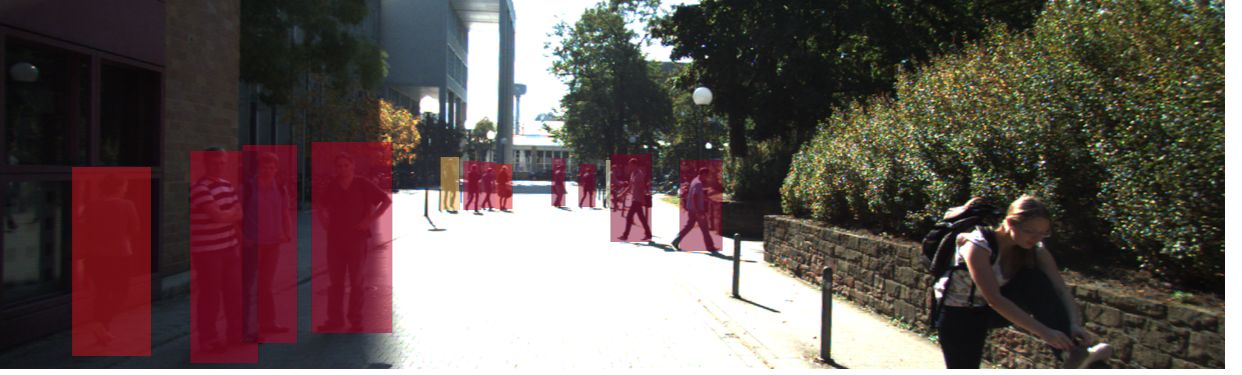}\\%
\includegraphics[width=0.33\linewidth]{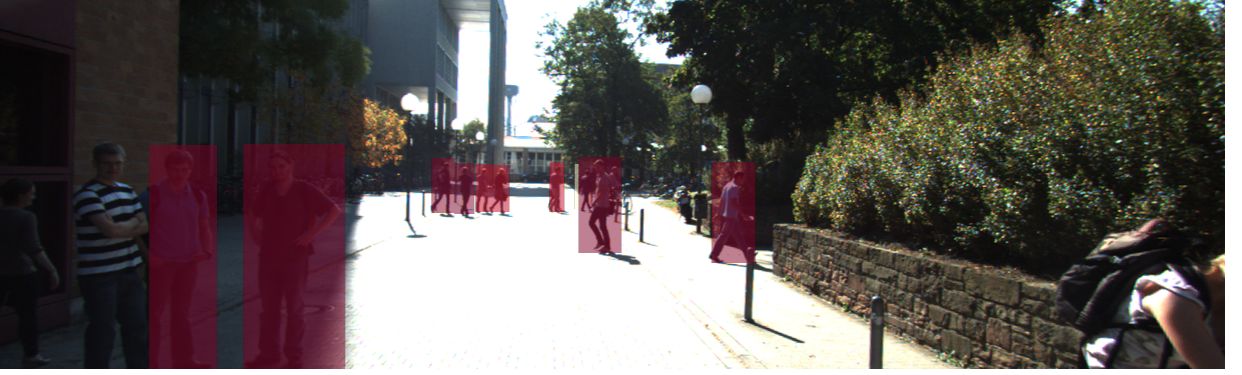}%
\includegraphics[width=0.33\linewidth]{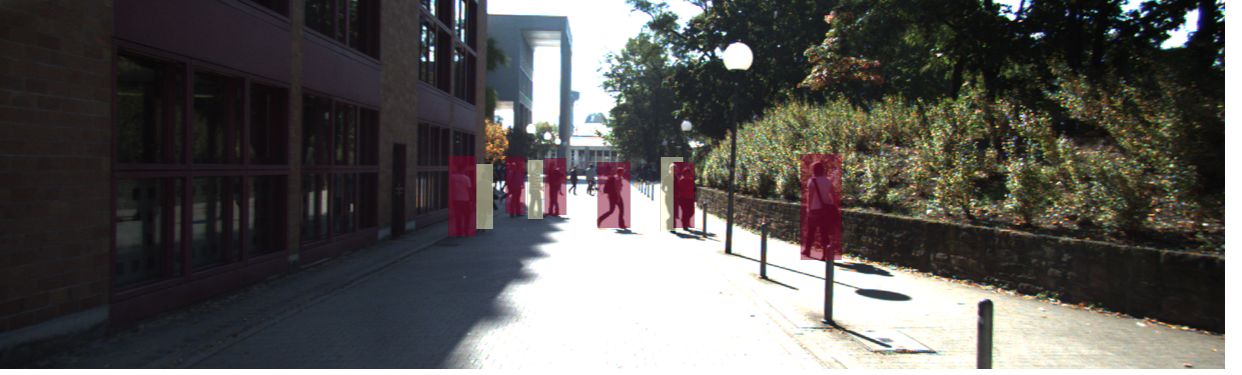}%
\includegraphics[width=0.33\linewidth]{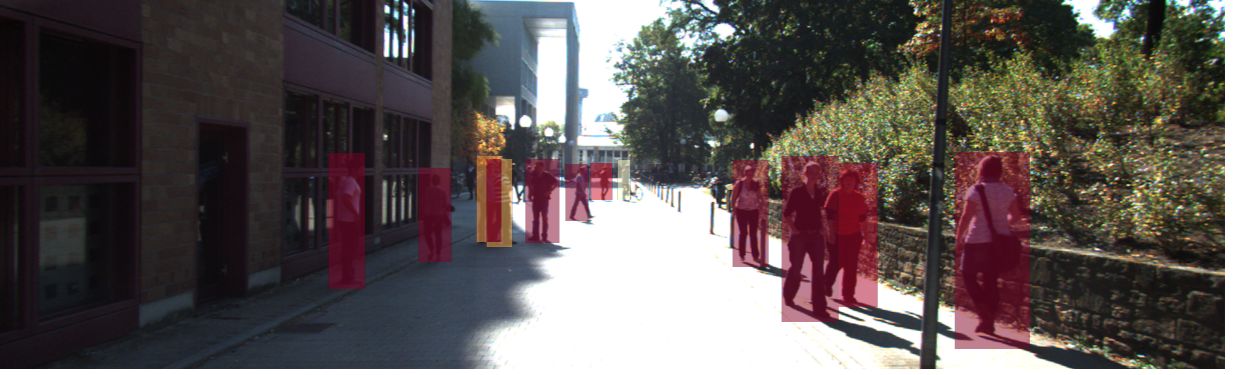}%
\caption{Images with Largest Number of True Positive Detections}
\end{subfigure}
\begin{subfigure}{\linewidth}
\includegraphics[width=0.33\linewidth]{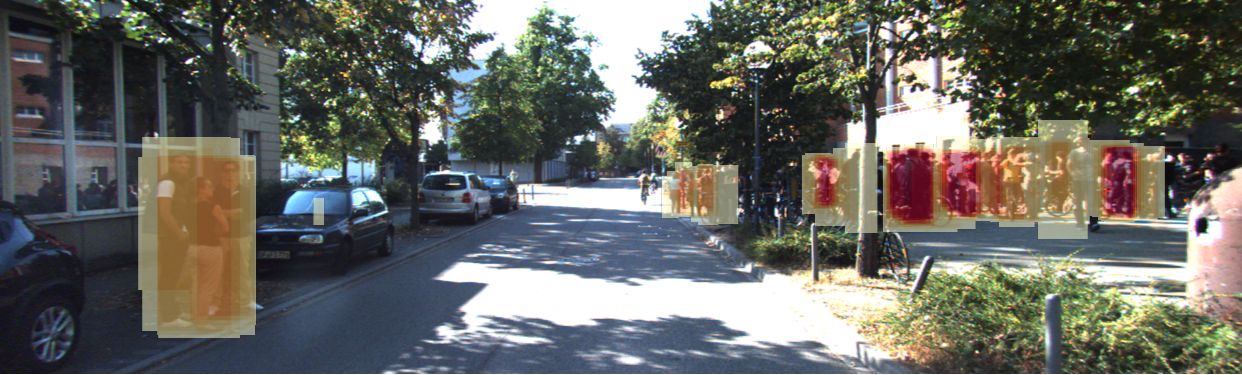}%
\includegraphics[width=0.33\linewidth]{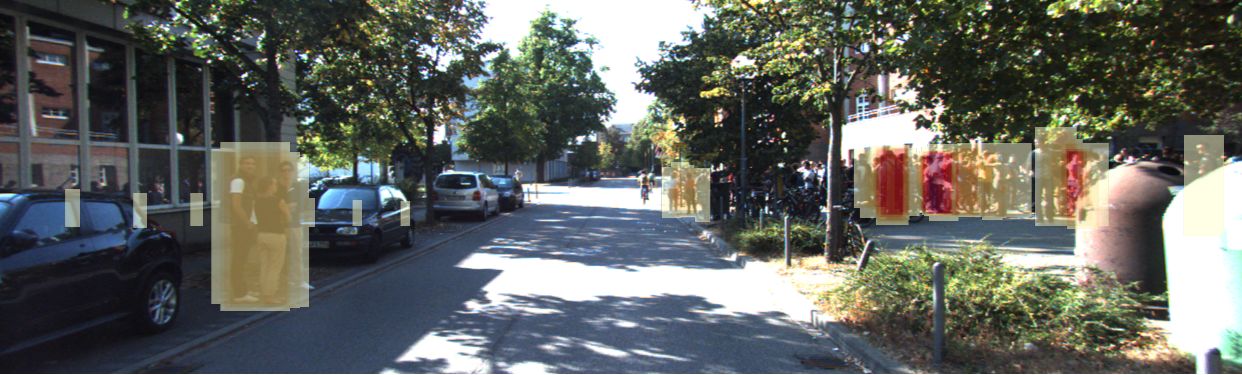}%
\includegraphics[width=0.33\linewidth]{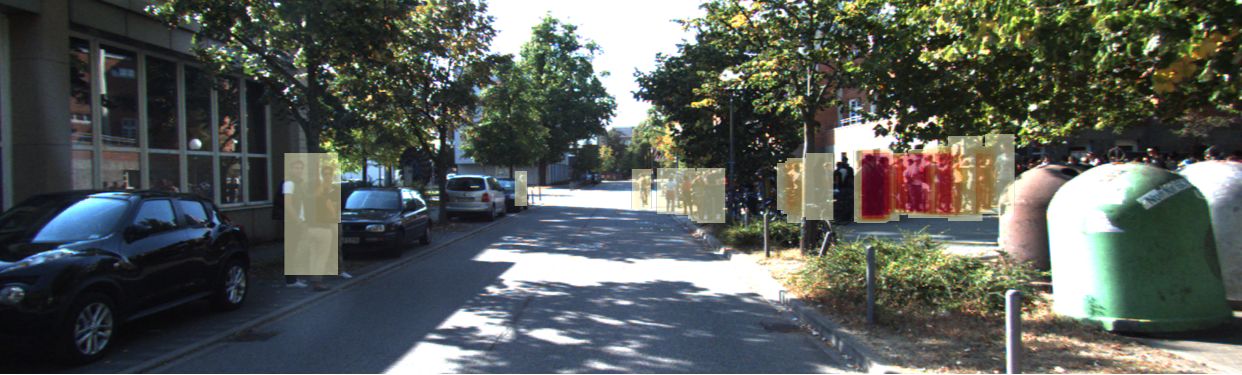}\\%
\includegraphics[width=0.33\linewidth]{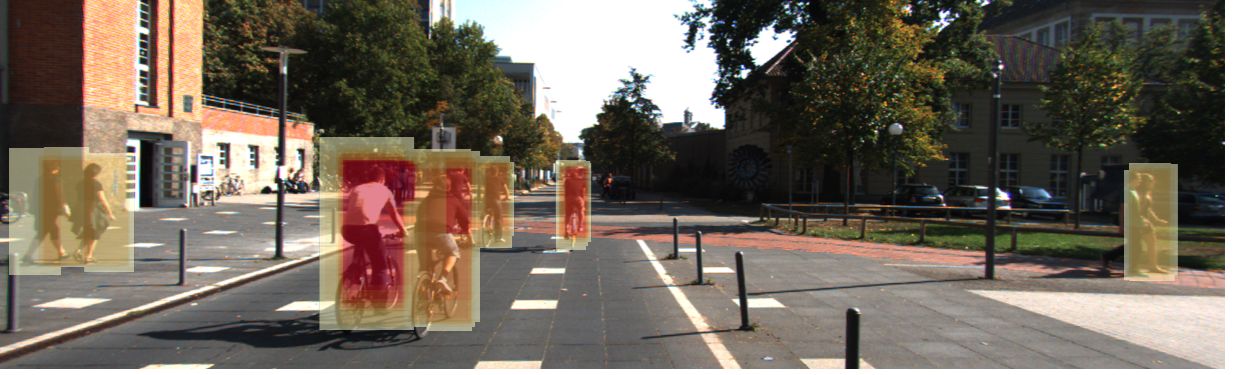}%
\includegraphics[width=0.33\linewidth]{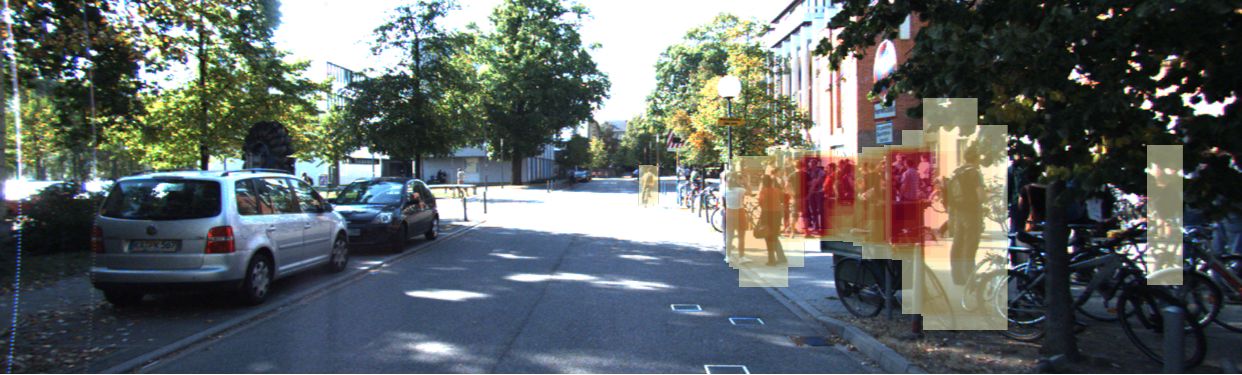}%
\includegraphics[width=0.33\linewidth]{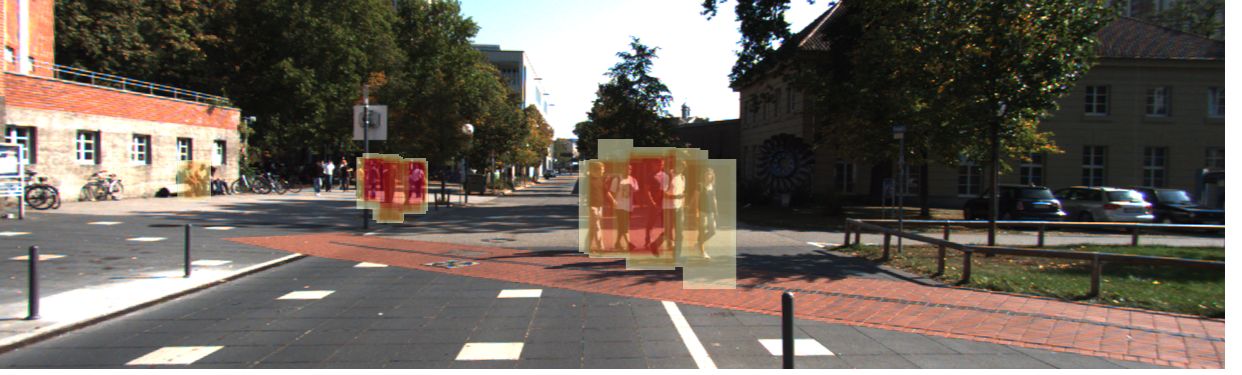}\\%
\includegraphics[width=0.33\linewidth]{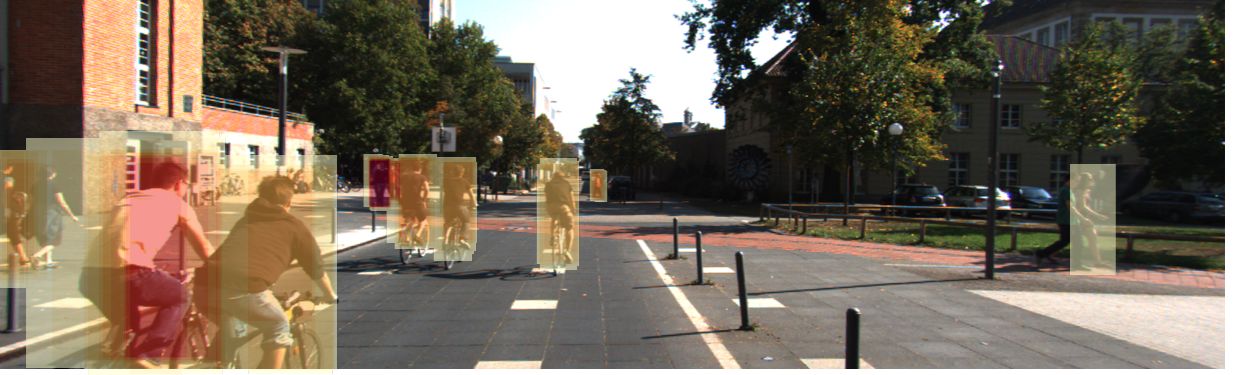}%
\includegraphics[width=0.33\linewidth]{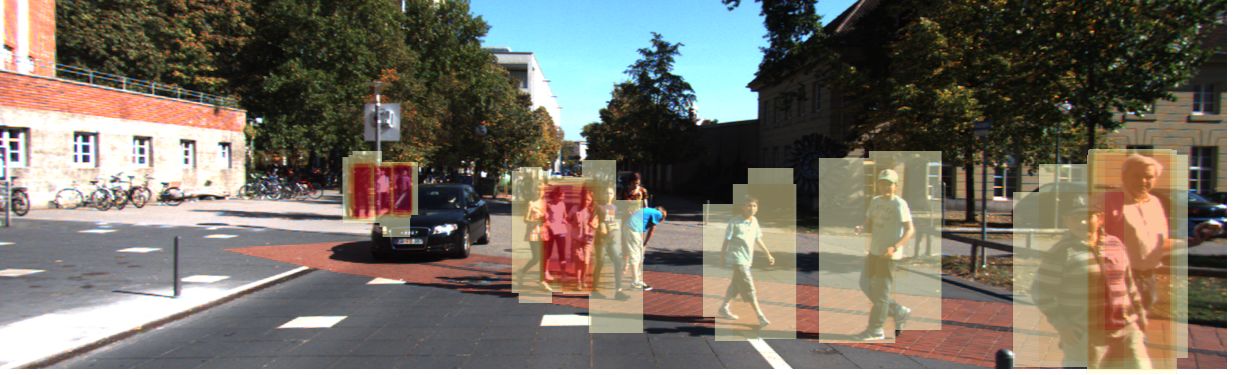}%
\includegraphics[width=0.33\linewidth]{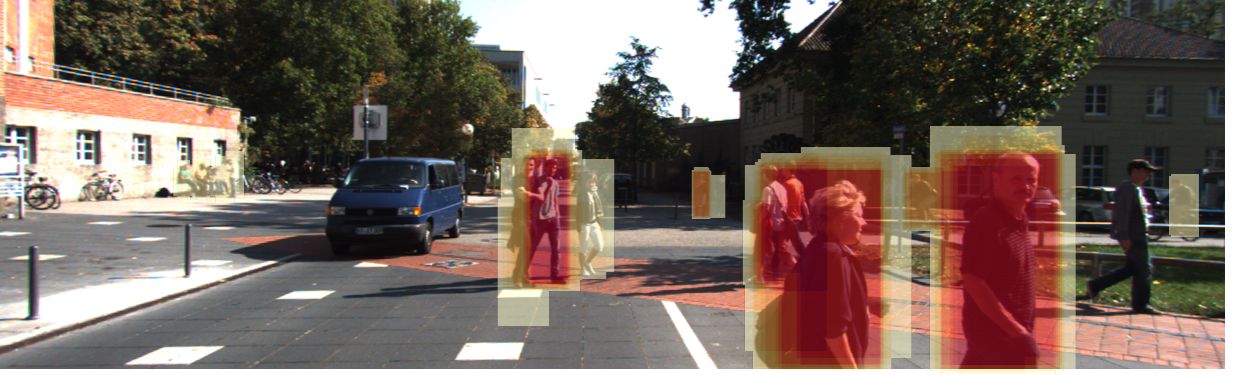}%
\caption{Images with Largest Number of False Positive Detections}
\end{subfigure}
\begin{subfigure}{\linewidth}
\includegraphics[width=0.33\linewidth]{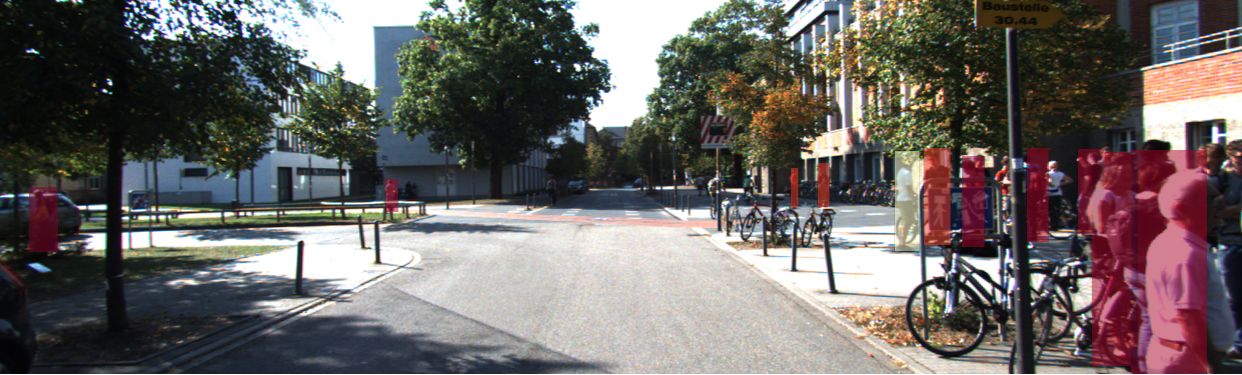}%
\includegraphics[width=0.33\linewidth]{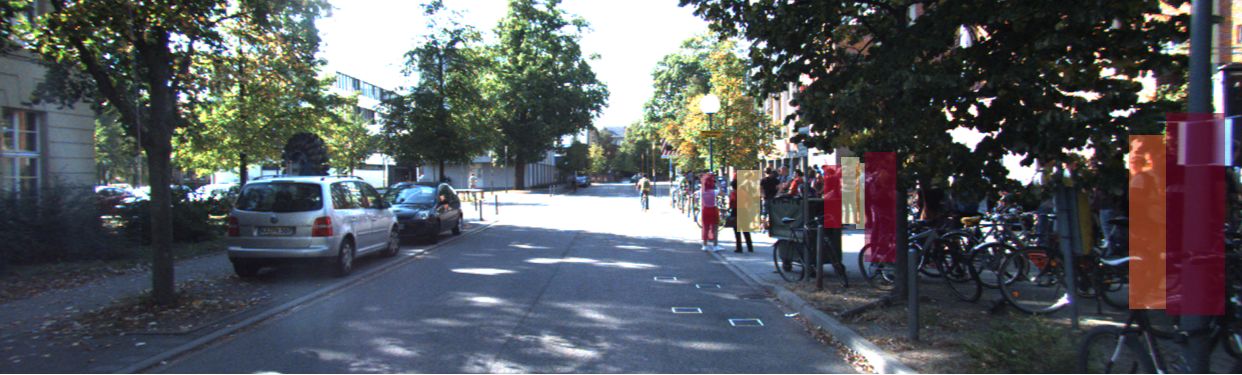}%
\includegraphics[width=0.33\linewidth]{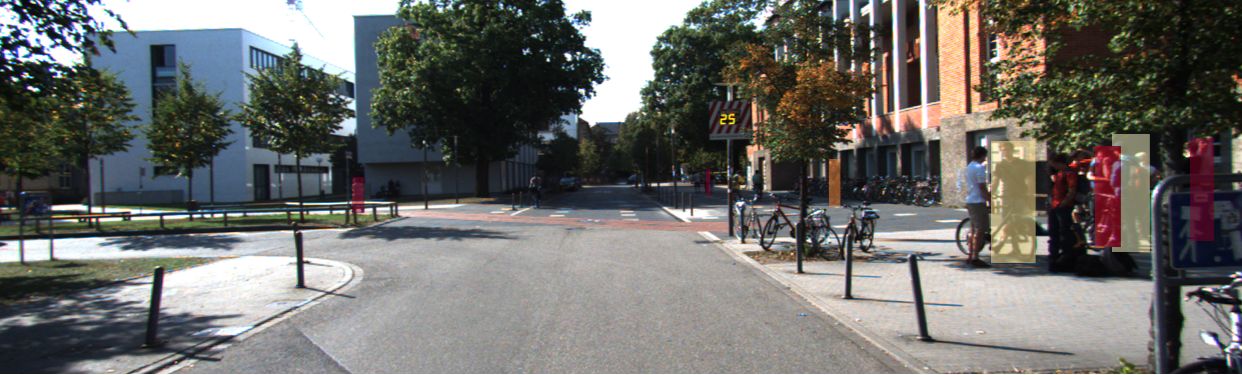}\\%
\includegraphics[width=0.33\linewidth]{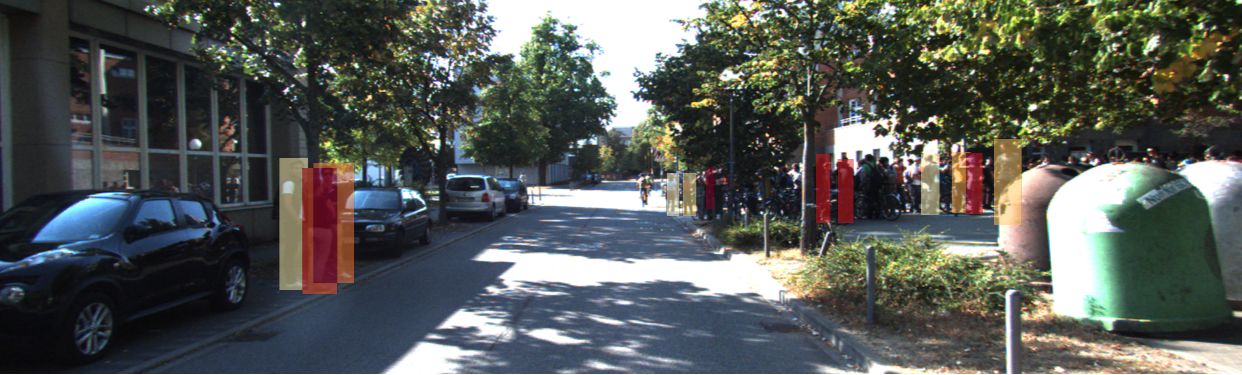}%
\includegraphics[width=0.33\linewidth]{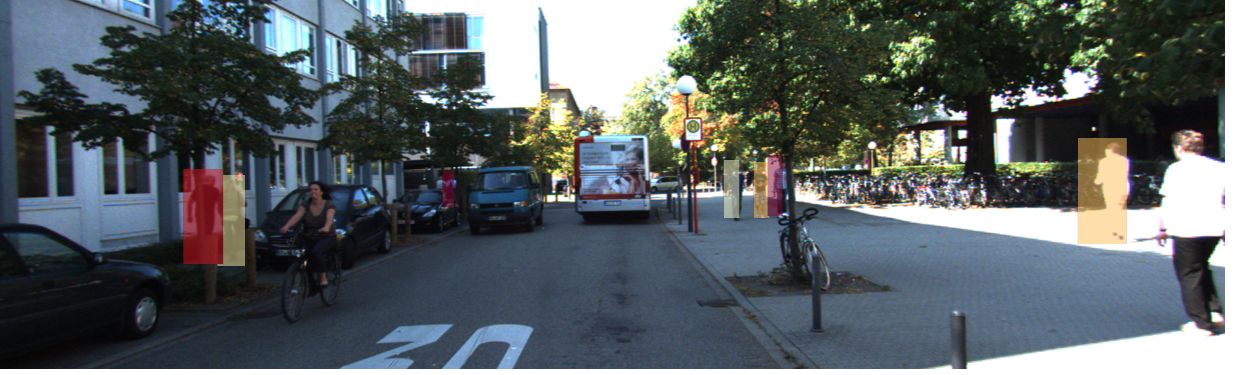}%
\includegraphics[width=0.33\linewidth]{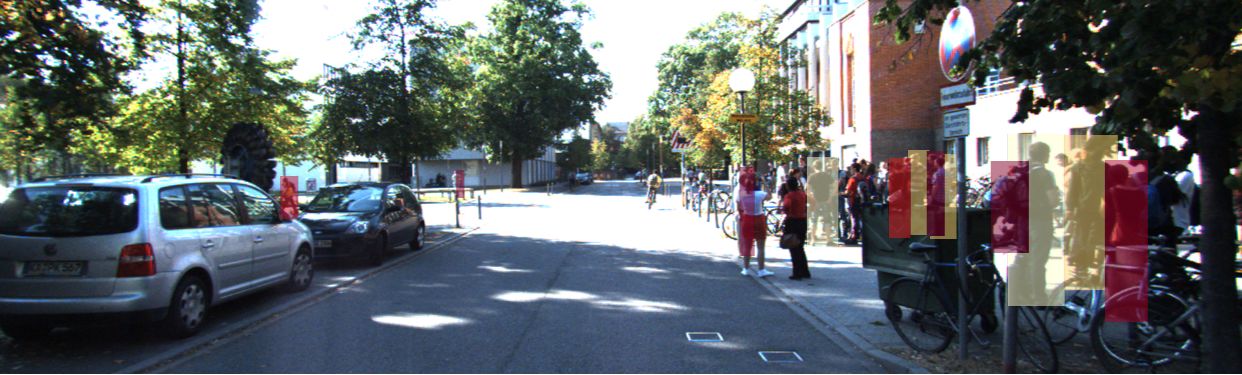}\\%
\includegraphics[width=0.33\linewidth]{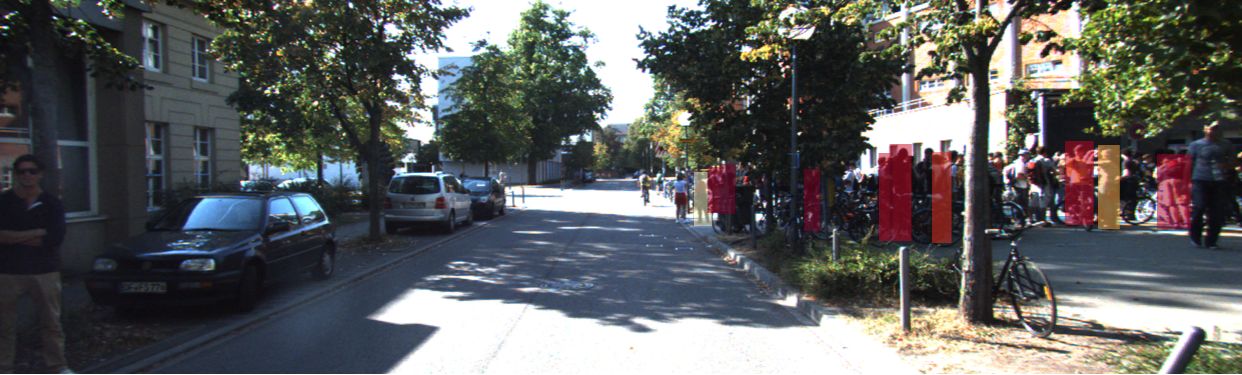}%
\includegraphics[width=0.33\linewidth]{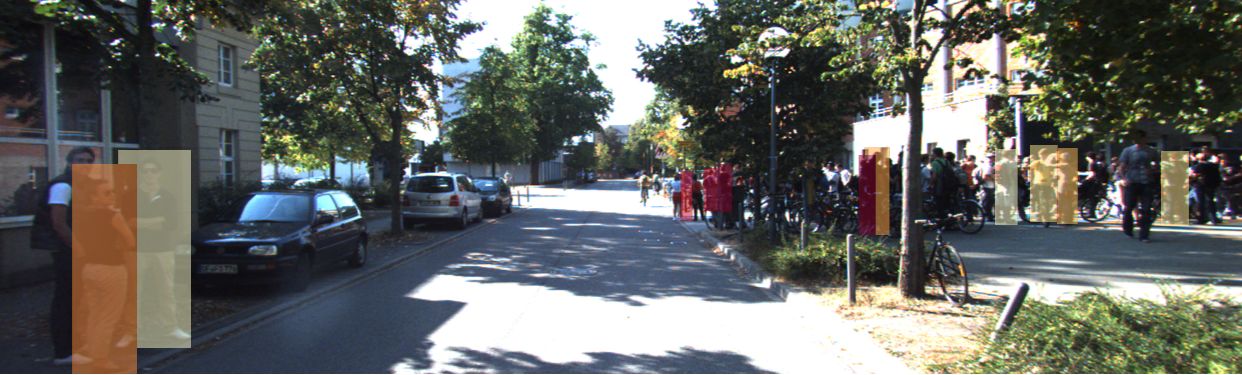}%
\includegraphics[width=0.33\linewidth]{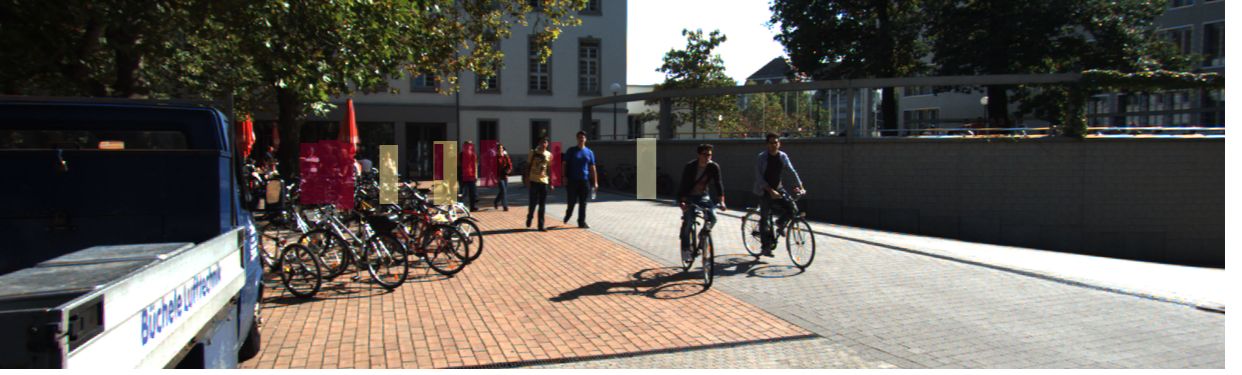}%
\caption{Images with Largest Number of False Negative Detections}
\end{subfigure}
\caption{{\bf KITTI Pedestrian Detection Analysis.} Each figure shows images with a large number of true positive (TP) detections, false positive (FP) detections and false negative (FN) detections, respectively. If all detectors agree on TP, FP or FN, the object is marked in red. If only some of the detectors agree, the object is marked in yellow. The ranking has been established by considering the 15 leading methods published on the KITTI evaluation server at time of submission.}
\label{fig:pedestrian_detection_qualitative_results}
\end{figure*}

\begin{figure*}[p]
\begin{subfigure}{\linewidth}
\includegraphics[width=0.33\linewidth]{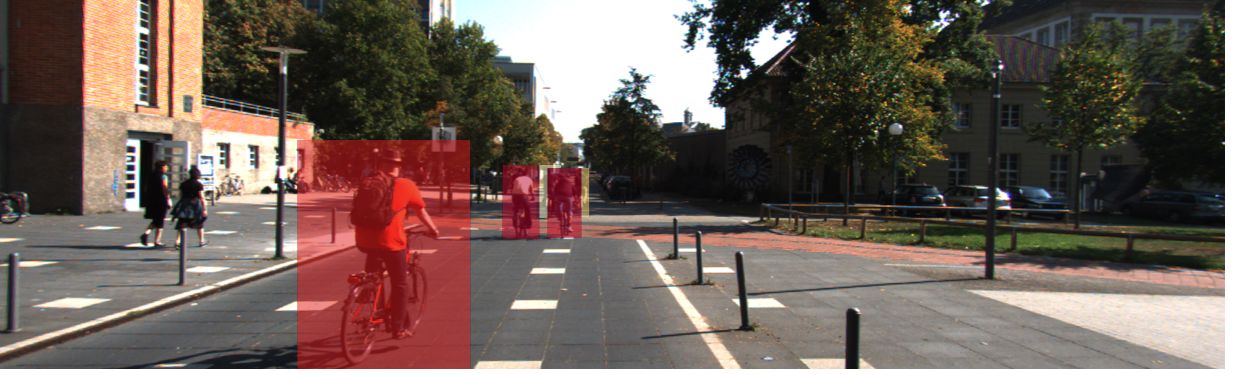}%
\includegraphics[width=0.33\linewidth]{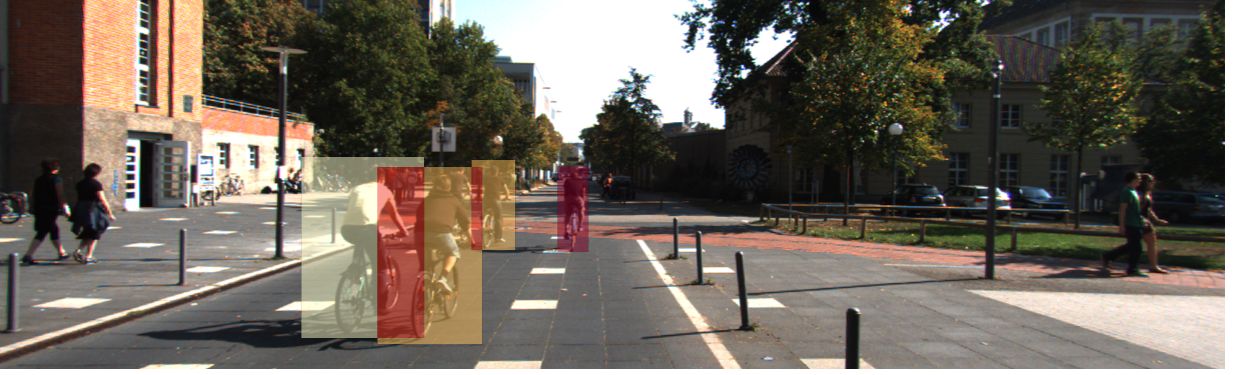}%
\includegraphics[width=0.33\linewidth]{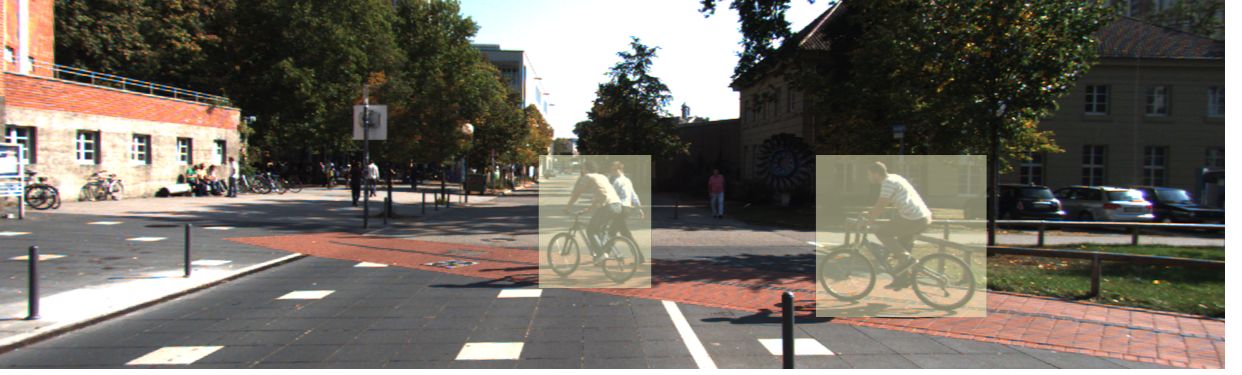}\\%
\includegraphics[width=0.33\linewidth]{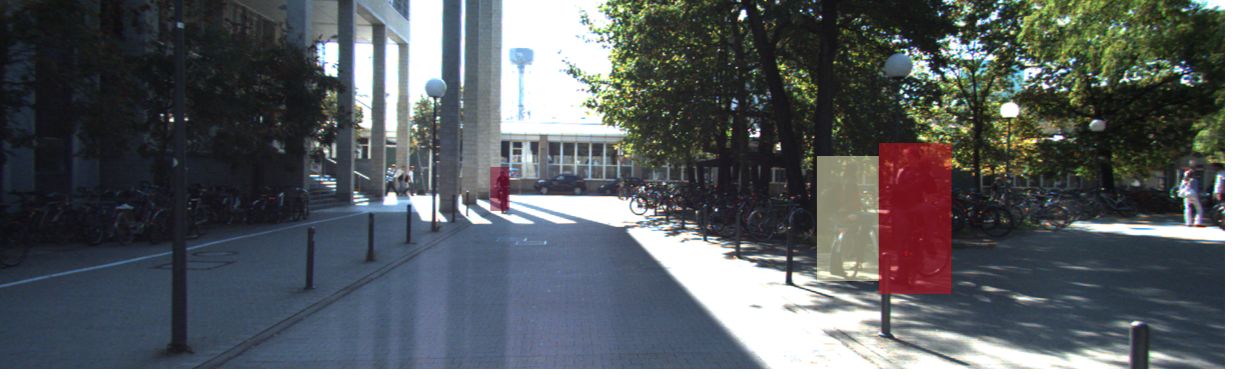}%
\includegraphics[width=0.33\linewidth]{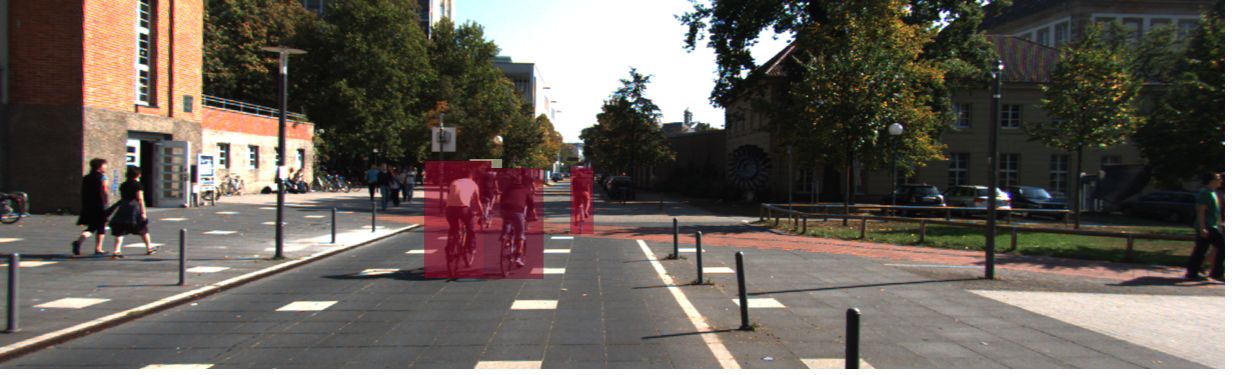}%
\includegraphics[width=0.33\linewidth]{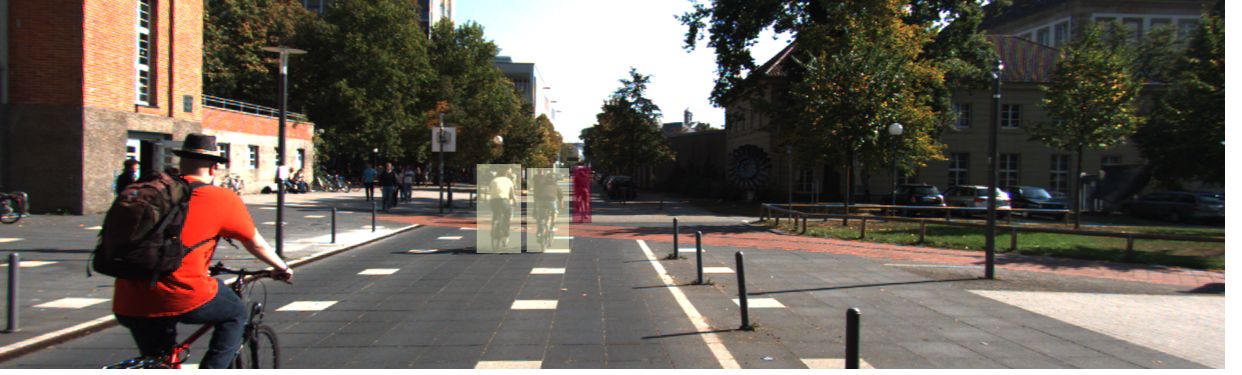}\\%
\includegraphics[width=0.33\linewidth]{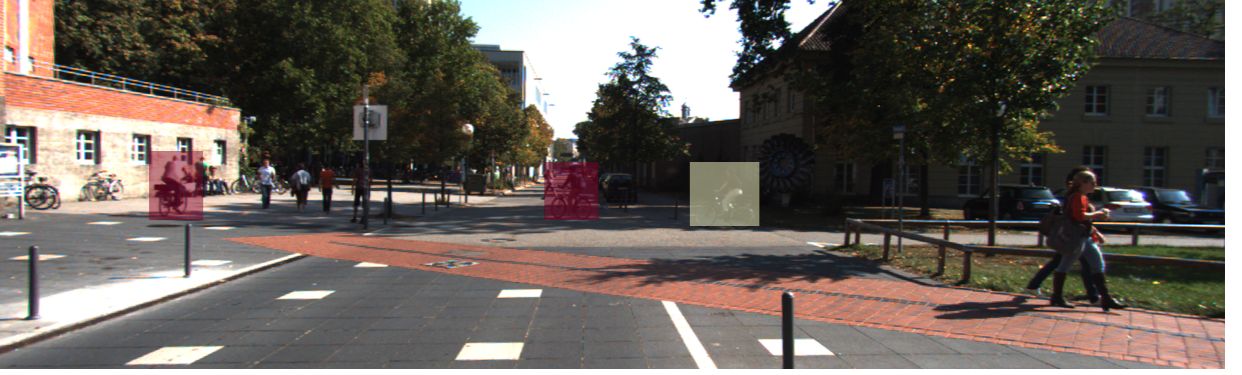}%
\includegraphics[width=0.33\linewidth]{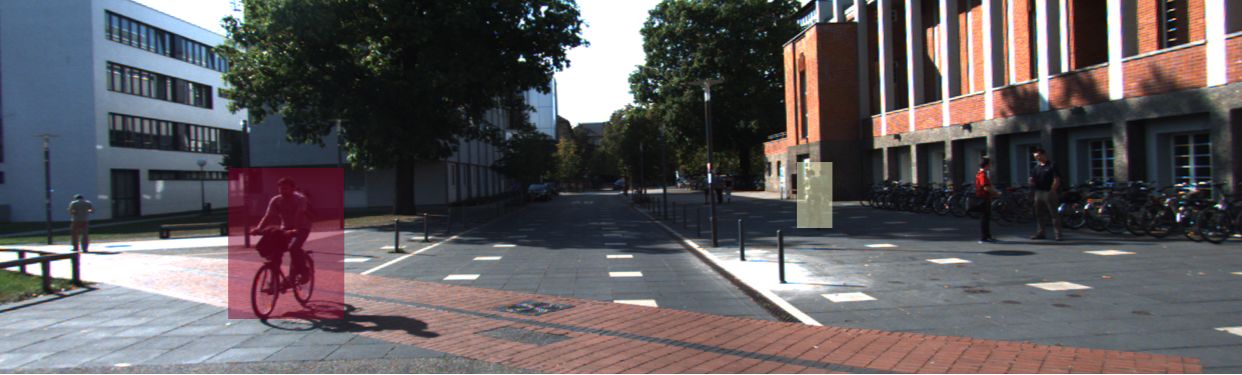}%
\includegraphics[width=0.33\linewidth]{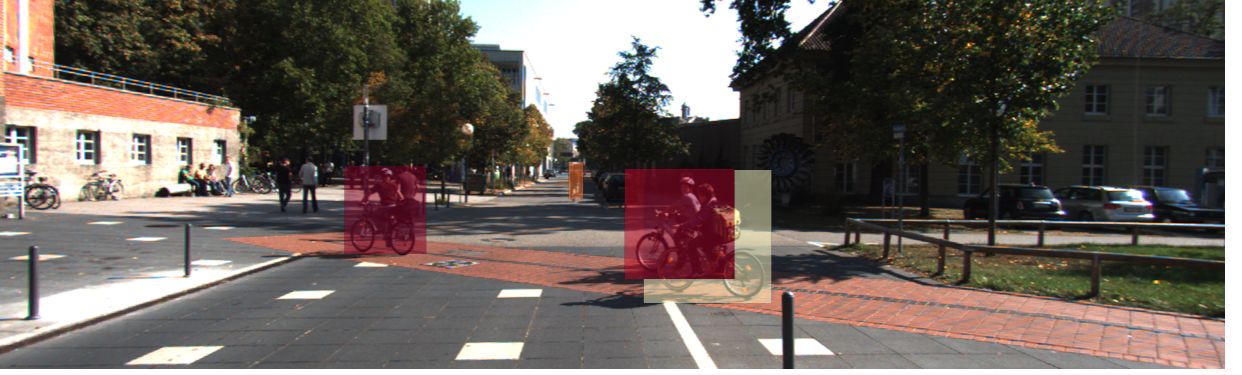}%
\caption{Images with Largest Number of True Positive Detections}
\end{subfigure}
\begin{subfigure}{\linewidth}
\includegraphics[width=0.33\linewidth]{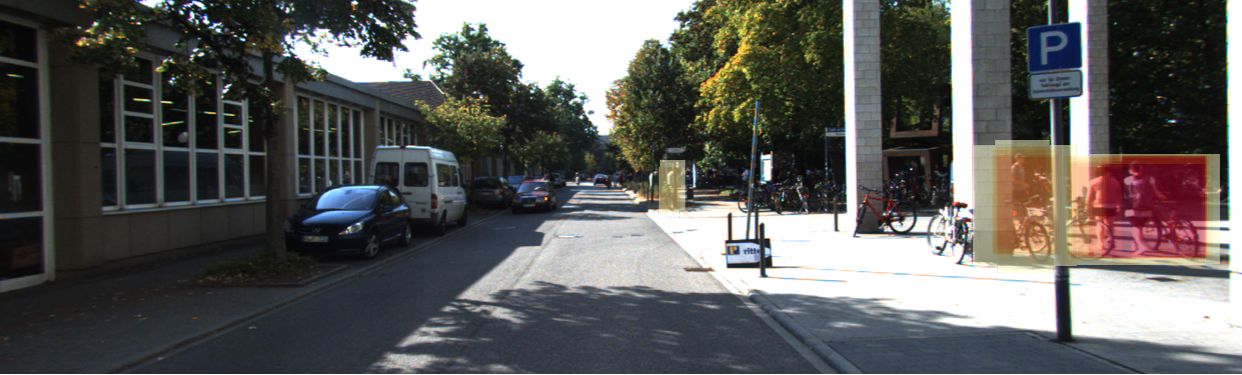}%
\includegraphics[width=0.33\linewidth]{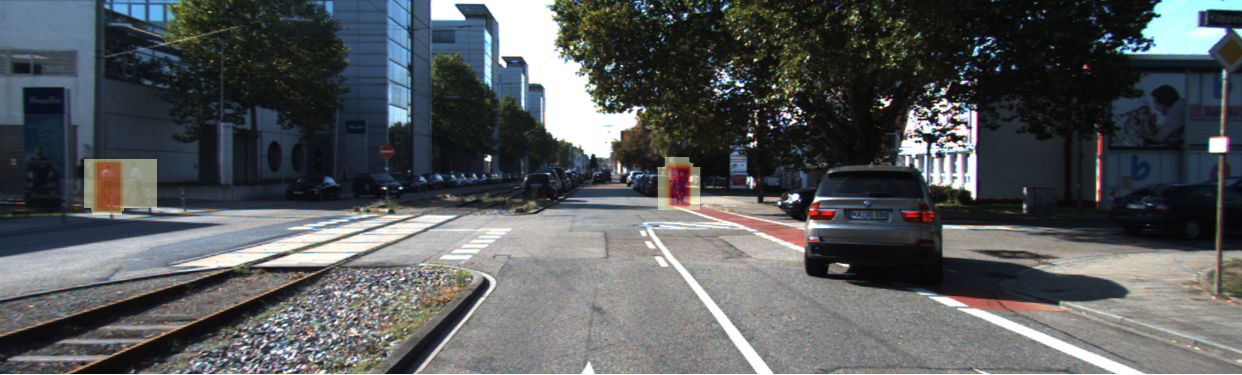}%
\includegraphics[width=0.33\linewidth]{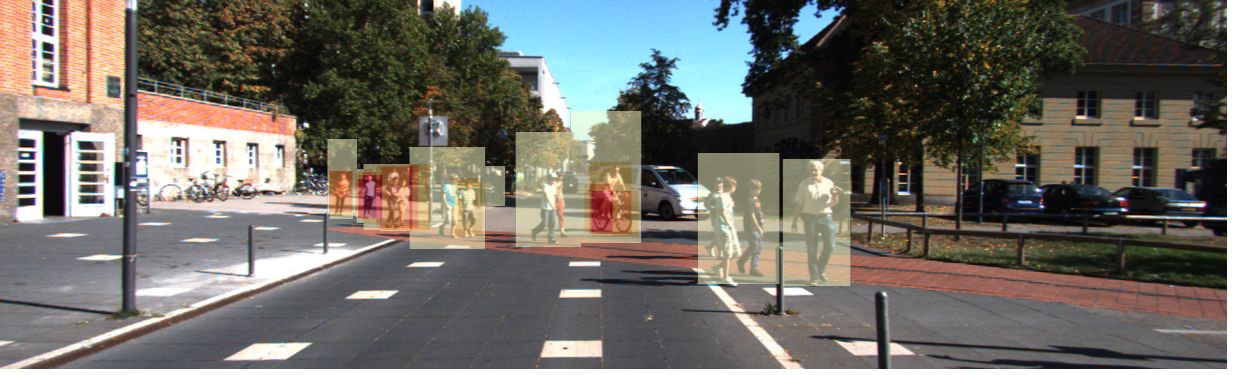}\\%
\includegraphics[width=0.33\linewidth]{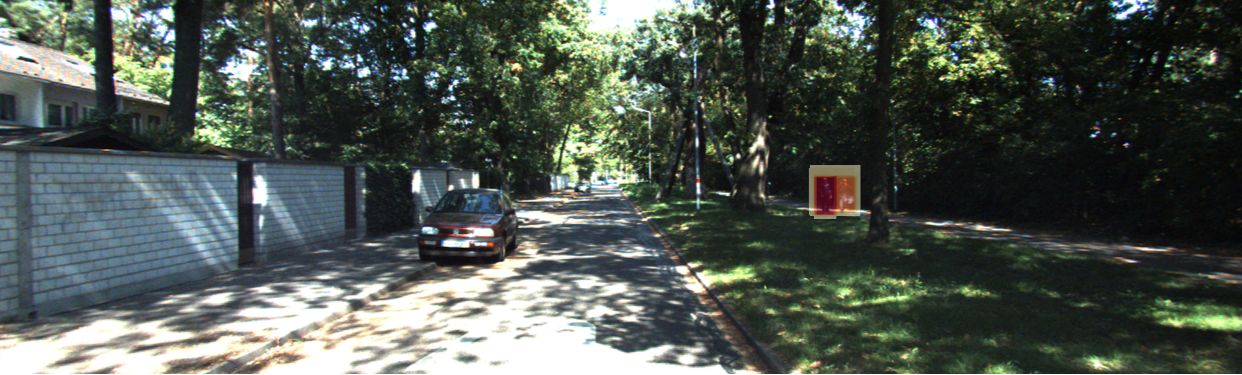}%
\includegraphics[width=0.33\linewidth]{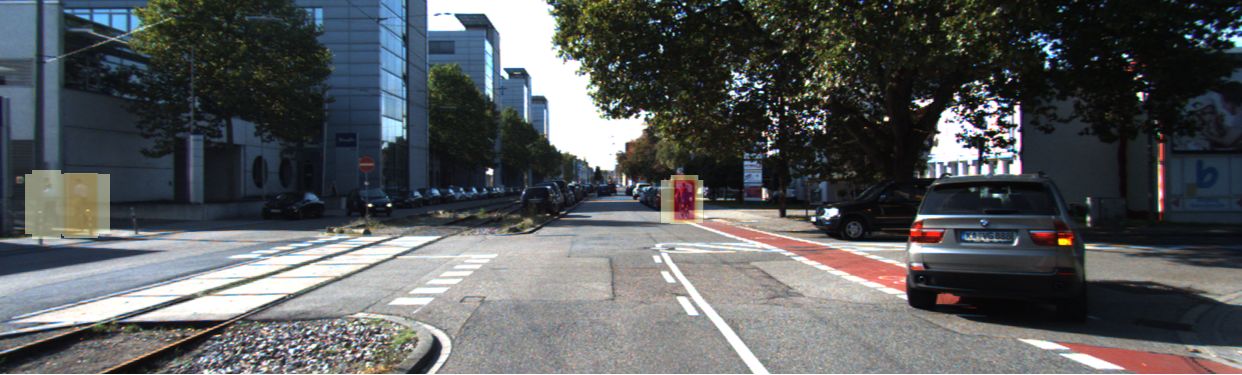}%
\includegraphics[width=0.33\linewidth]{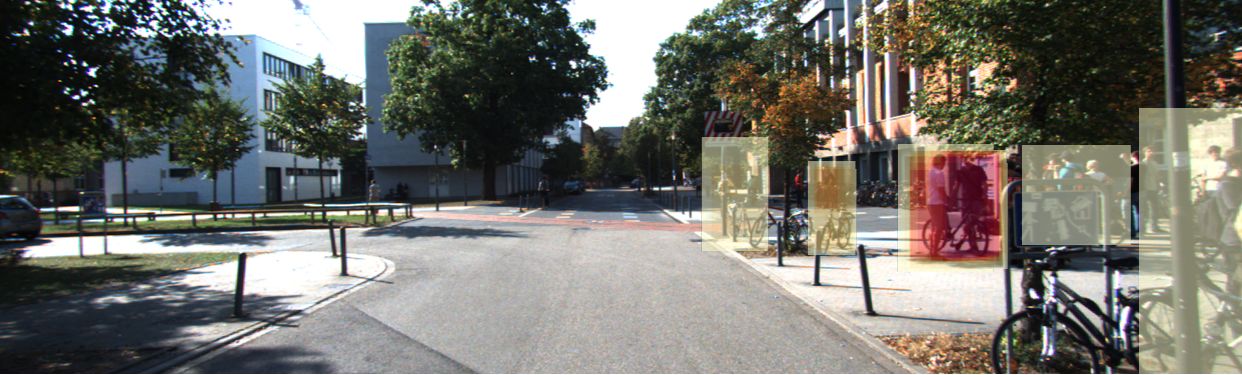}\\%
\includegraphics[width=0.33\linewidth]{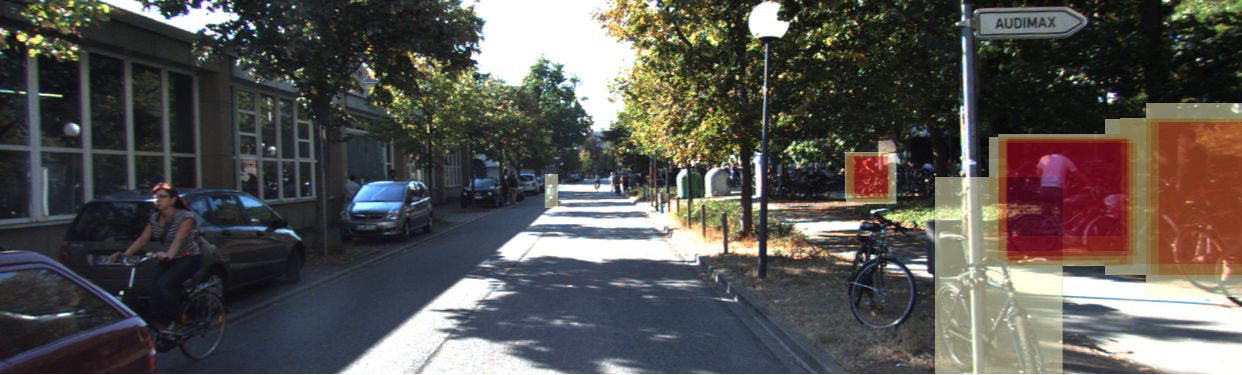}%
\includegraphics[width=0.33\linewidth]{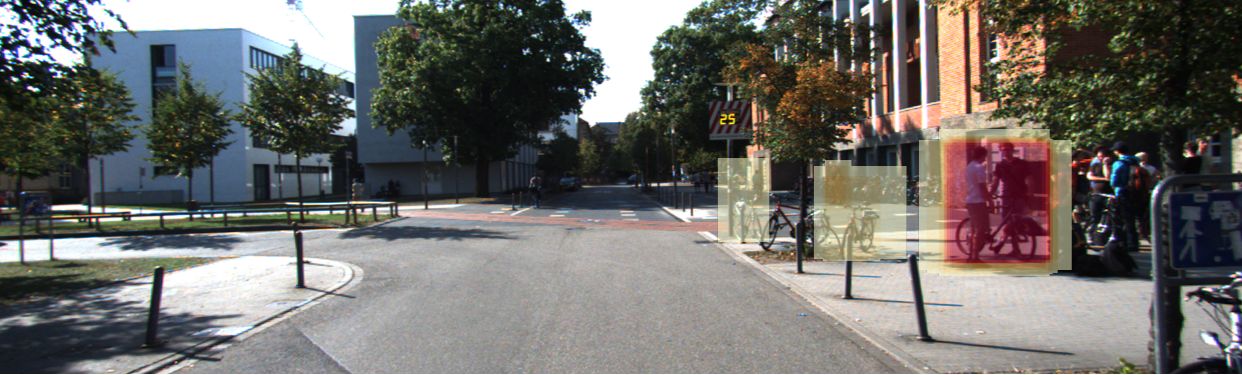}%
\includegraphics[width=0.33\linewidth]{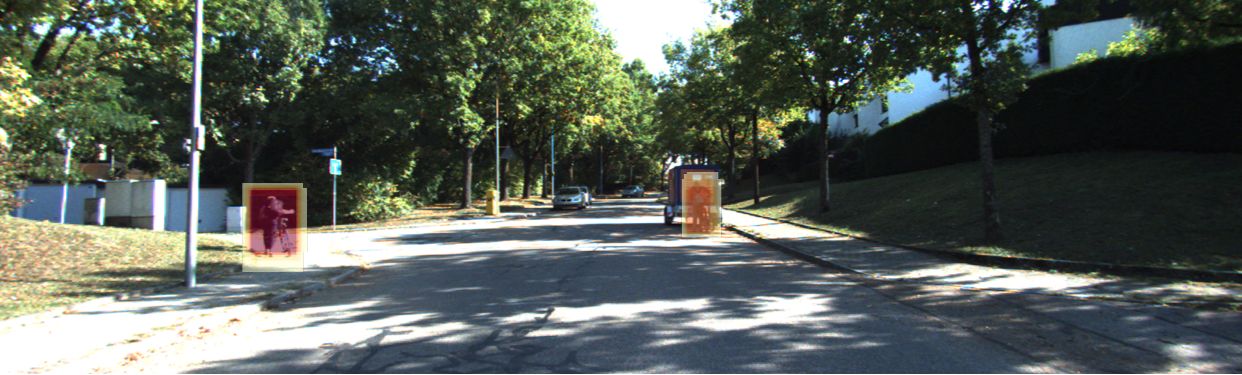}%
\caption{Images with Largest Number of False Positive Detections}
\end{subfigure}
\begin{subfigure}{\linewidth}
\includegraphics[width=0.33\linewidth]{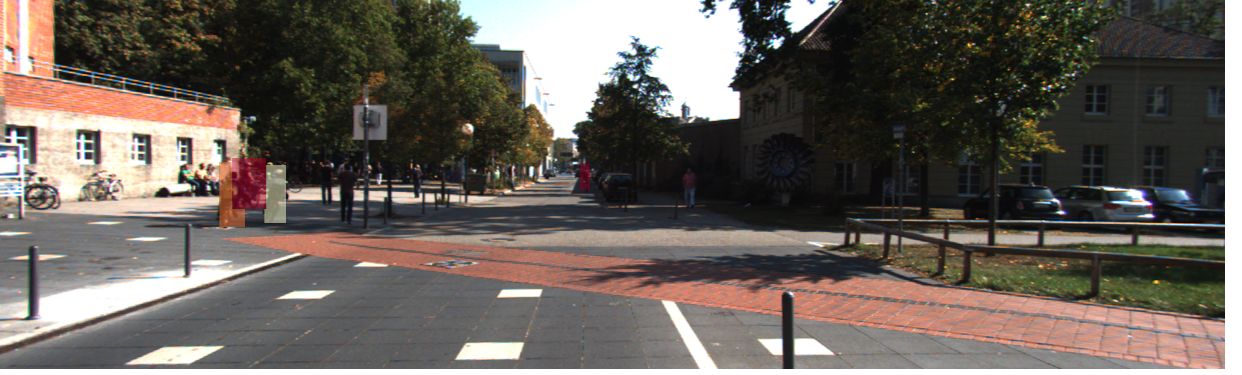}%
\includegraphics[width=0.33\linewidth]{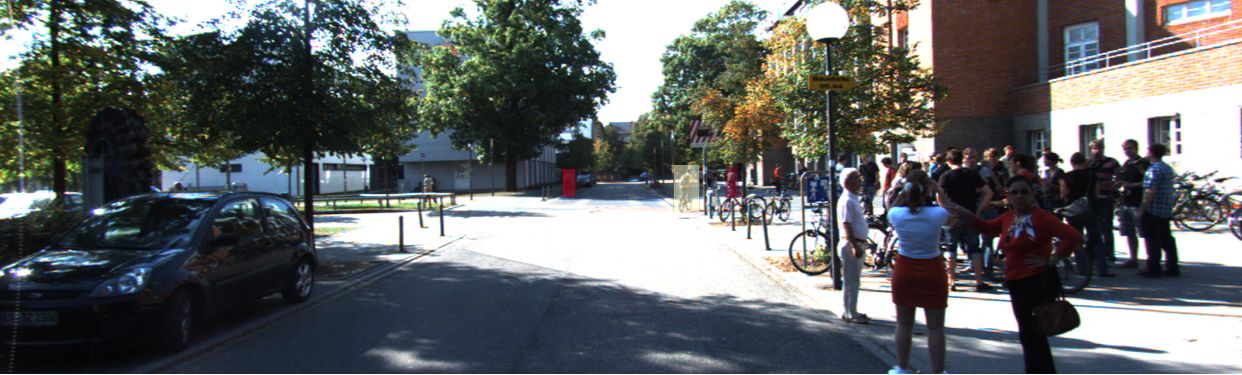}%
\includegraphics[width=0.33\linewidth]{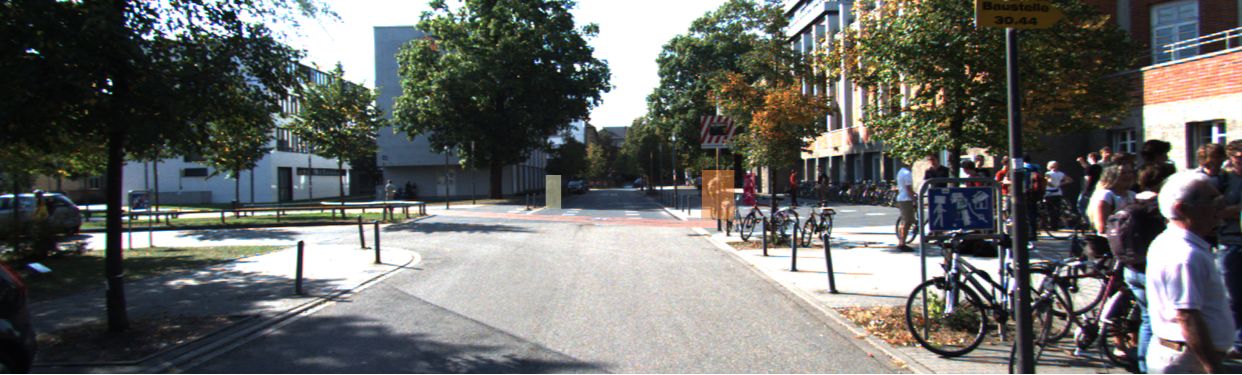}\\%
\includegraphics[width=0.33\linewidth]{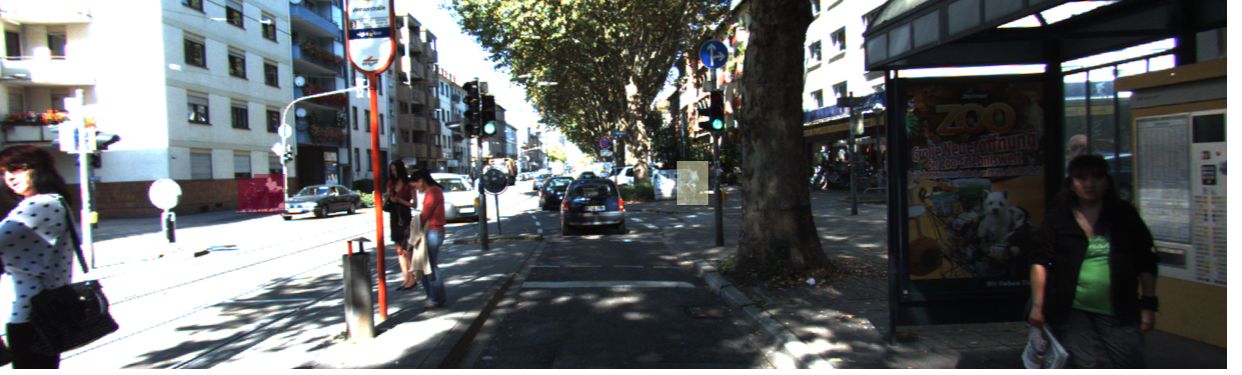}%
\includegraphics[width=0.33\linewidth]{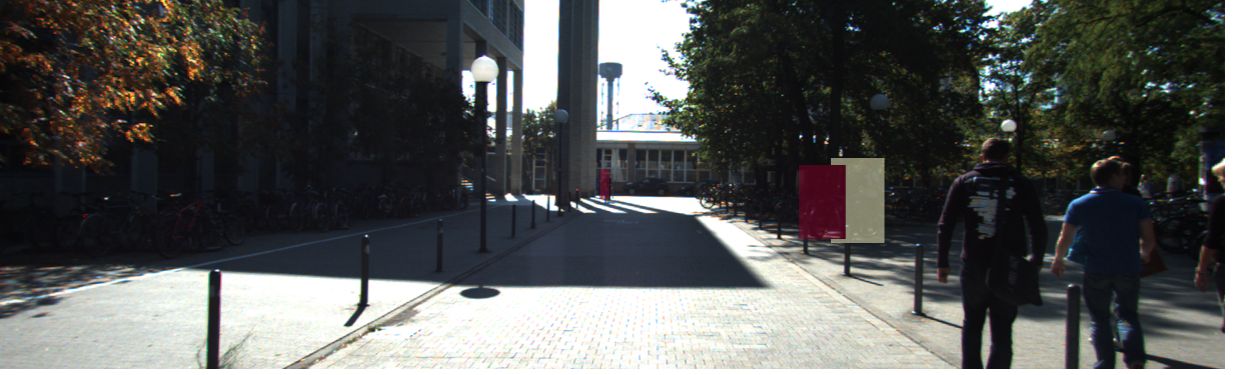}%
\includegraphics[width=0.33\linewidth]{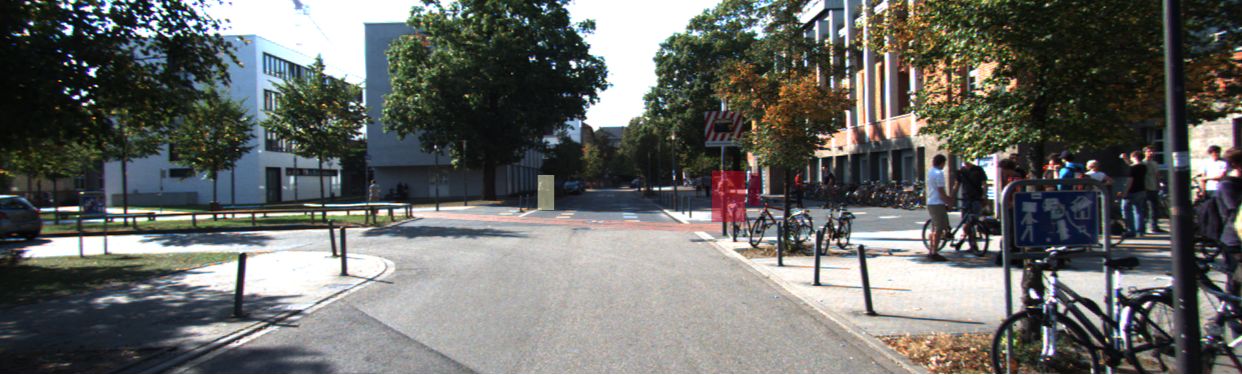}\\%
\includegraphics[width=0.33\linewidth]{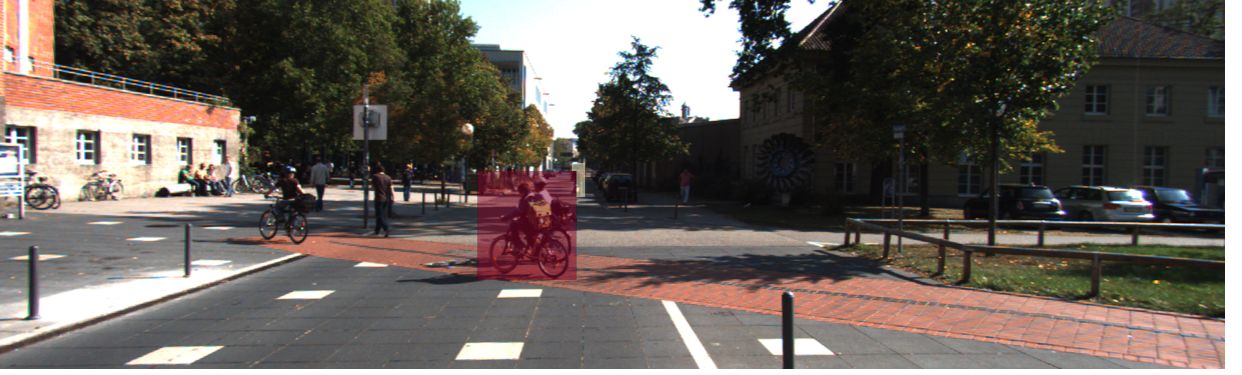}%
\includegraphics[width=0.33\linewidth]{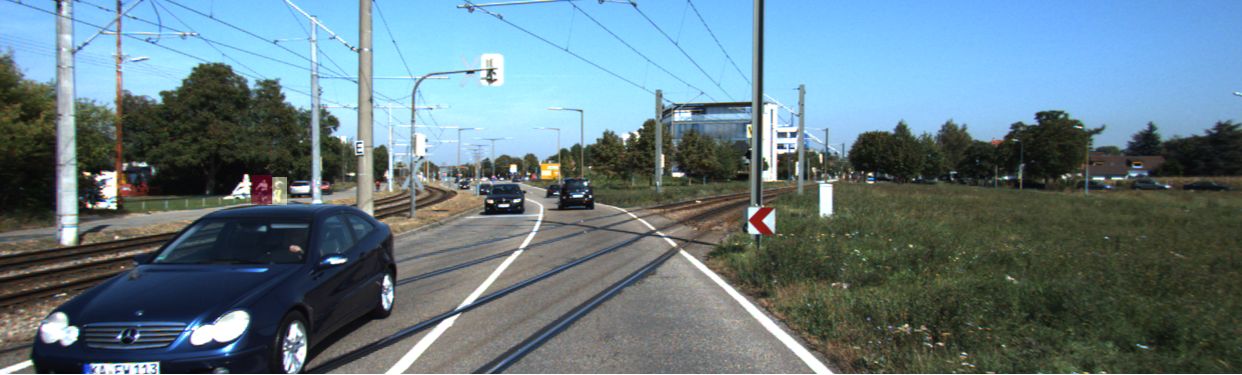}%
\includegraphics[width=0.33\linewidth]{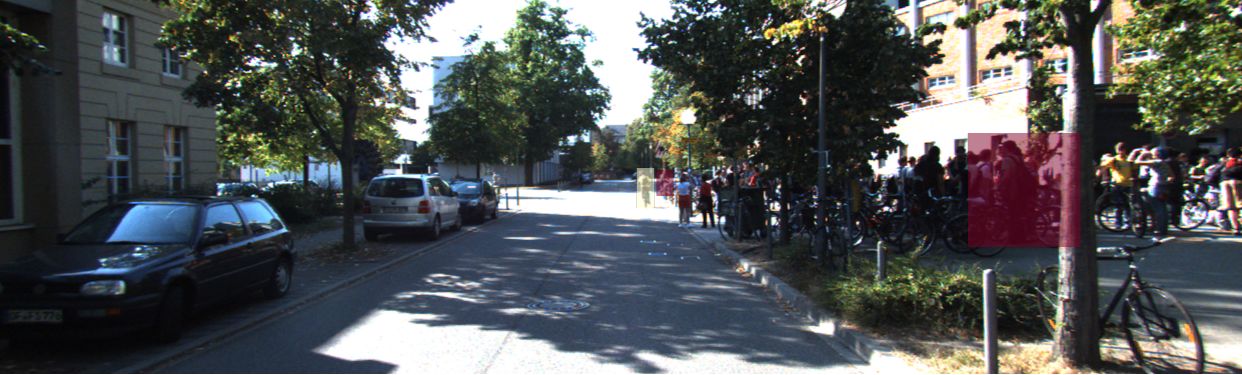}%
\caption{Images with Largest Number of False Negative Detections}
\end{subfigure}
\caption{{\bf KITTI Cyclist Detection Analysis.} Each figure shows images with a large number of true positive (TP) detections, false positive (FP) detections and false negative (FN) detections, respectively. If all detectors agree on TP, FP or FN, the object is marked in red. If only some of the detectors agree, the object is marked in yellow. The ranking has been established by considering the 15 leading methods published on the KITTI evaluation server at time of submission.}
\label{fig:cyclist_detection_qualitative_results}
\end{figure*}

Object detection has demonstrated impressive performance in case of high resolution images with little occlusions. For the easy and moderate cases of the car detection task (\tabref{tab:kitti_object_det2d_car_image}), many methods provide accurate detections. The pedestrian and cyclist detection task (Tables \ref{tab:kitti_object_det2d_pedestrian_image}, \ref{tab:kitti_object_det2d_cyclist_image}) is more challenging, as demonstrated by the overall weaker performance of all methods.
One reason for this is the limited number of training examples and the possibility of confusing cyclists and pedestrians which differ only via their context and semantics. Remaining major problems across tasks are the detection of small objects and highly occluded objects. In the leaderboards, this manifests in a significant drop in performance when comparing easy, moderate, and hard examples. Qualitatively, this can be observed in Figures \ref{fig:car_detection_qualitative_results}, \ref{fig:pedestrian_detection_qualitative_results},\ref{fig:cyclist_detection_qualitative_results} where we show typical estimation errors of the best-performing methods on the KITTI dataset. Major sources of errors are crowds of pedestrians, groups of cyclists, and parked cars that cause occlusions and lead to missing detections for all methods. Furthermore, distant objects still prove to be challenging for modern methods due to the low amount of image evidence provided for these objects.
	\chapter{Object Tracking}
\label{chap:tracking}

\section{Problem Definition}
In tracking, the goal is to estimate the state of one or multiple objects over time given measurements of a sensor. This is in contrast to object detection where each frame is typically processed independently and no associations over time are established. Typically, the state of an object is represented by its location, velocity and acceleration at a certain time. Tracking of other traffic participants is a very important task for autonomous driving.
Consider for instance, the braking distance of a vehicle which increases quadratically with its speed. Because of the braking distance it is necessary to detect possible collisions with other traffic participants early on.
This is only possible with good predictions of future trajectories. In the case of pedestrians and bicyclists, it is particularly difficult to predict the future behavior because they can abruptly change the direction of their movements. Therefore, humans tend to drive more carefully around pedestrians and bicyclists. Similarly, tracking in combination with the classification of traffic participants allows adapting the speed of the vehicle accordingly. In addition, tracking of other cars can be used for automatic distance control and to anticipate possible driving maneuvers of other traffic participants (such as takeovers) early on.

Tracking systems must cope with a variety of challenges such as cluttered backgrounds, the variety and complexity of motion, and occlusions.
The problem of associating instances of the same object over time becomes particularly challenging due to the resemblance of different objects, especially of the same class. 
In addition to the lack of discriminative information due to similarities with other objects, instances of the same object might not look similar enough for association in different time steps. Often objects are partially or fully occluded by other objects or themselves. 
The interaction of objects, especially in the case of pedestrians, further increases the amount of occlusions and makes it difficult to track each individual object. Difficult lighting conditions and reflections in mirrors or windows pose additional challenges. 

\section{Methods}
Historically, tracking has been formulated as a Bayesian inference problem \citep{Thrun2005} where the goal is to estimate the posterior probability density function of a state given the current observation and the previous state(s). The posterior is usually updated in a recursive manner with a prediction step using a motion model and a correction step using an observation model. In each iteration, the data association problem is solved to assign new observations to the tracked objects. Extended Kalman and particle filtering algorithms \citep{Giebel2004ECCV, Breitenstein2011PAMI, Choi2013PAMI} are widely used models in this context. Unfortunately, the recursive approach makes it hard to recover from detection errors and to track through occlusions because of missing observations. Therefore, non-recursive approaches \citep{Andriyenko2011CVPR,Andriyenko2012CVPR} that optimize a global energy function with respect to all trajectories in a temporal window, have gained popularity. However, the large number of possible target trajectories per object and the large number of potential objects in a scene lead to a very large search space.

\subsection{Tracking by Detection}
\begin{figure}[t]
	\centering
	\includegraphics[width=1.00\columnwidth]{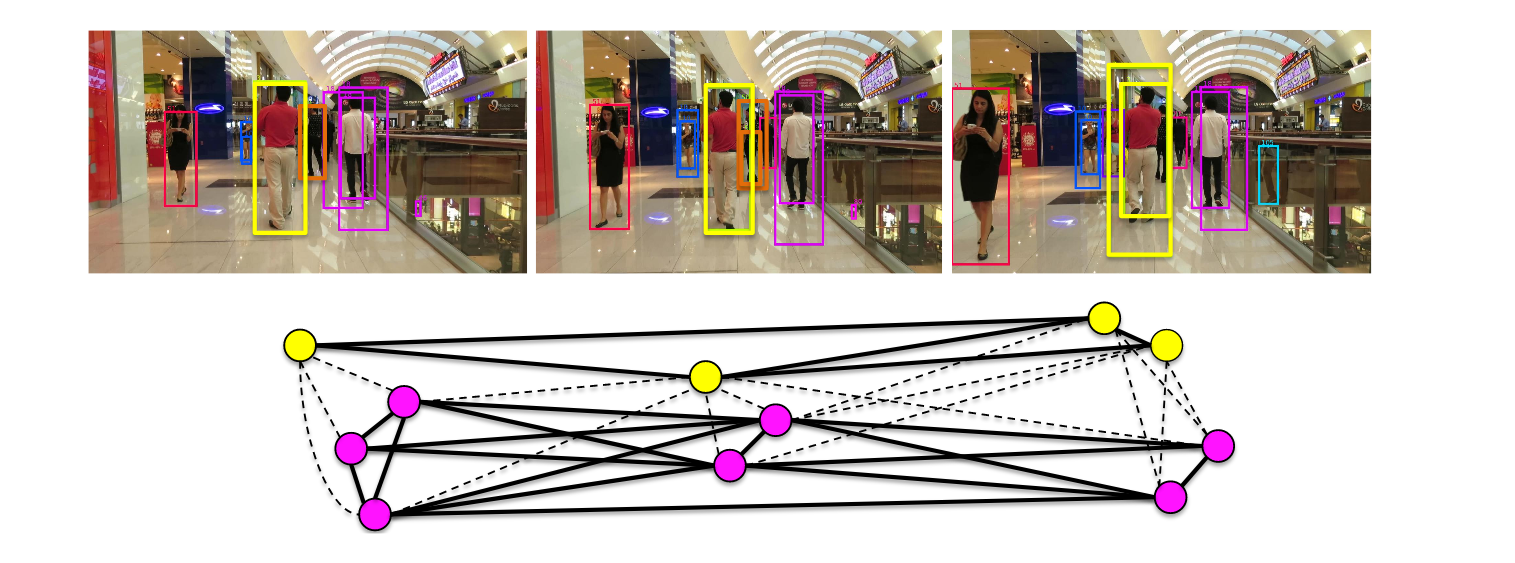}
	\caption[Graph-based Data Association]{\textbf{Graph-based Data Association.} Graph-based representation solved with a multi-cut formulation presented by \protect\citet{Tang2016ECCVWORK}. The graph is created from detections in the upper images and the colorization as well as connections in the graph are obtained by solving the multi-cut problem. \figsourceSpringer{\protect\citet{Tang2016ECCVWORK}}{2016}{ECCV Workshop}.}
	\label{fig:tracking_graph_representation}
\end{figure}
Given the success of static object detectors, a common paradigm often used in tracking is tracking-by-detection. This approach splits the task into two steps: first detect the people and second associate detections of the same person across time. Tracking-by-detection has become very popular since the tracking problem is reduced to a data association problem. However, the tracking system still needs to handle and recover from errors of the detection system, such as false and missing detections.

\boldparagraph{Tracking on Graphs}
Graph representations illustrated in \figref{fig:tracking_graph_representation} are widely adopted for inferring associations in tracking. In the simplest case, bipartite matching between the trajectories and the detections can be considered a graph-based approach with two disjoint sets of nodes. The assignment between the two sets can be performed either greedily \citep{Wu2007IJCV, Breitenstein2009ICCV, Shu2012CVPR} or by applying the optimal Hungarian algorithm \citep{Perera2006CVPR, Huang2008ECCV, Xing2009CVPR, Reilly2010ECCV, Qin2012CVPR} running in polynomial time. 

In network flow approaches \citep{Jiang2007CVPR,Zhang2008CVPR, Berclaz2009PETS, Wu2011CVPRb, Berclaz2011PAMI, Pirsiavash2011CVPR, Choi2012ECCV, Wu2012CVPR}, a graph is first constructed with nodes as detections and edges representing spatial and temporal links between detections. Then, a simple set of constraints is defined to ensure that produced tracks are valid and continuous between the start and the end nodes. Typically, these constraints are formulated as an integer program which is then relaxed to a linear program in order to avoid the NP-Hardness of the integer program. Various dynamic programming approaches have been proposed to solve the network flow using linear programming \citep{Jiang2007CVPR, Berclaz2009PETS}, k-shortest paths \citep{Berclaz2011PAMI, Pirsiavash2011CVPR, Choi2012ECCV} or set cover \citep{Wu2011CVPRb} for optimization.

Another line of work on graphs phrases tracking as clustering problem.
Minimum Clique \citep{Zamir2012ECCV, Dehghan2015CVPRb} and Minimum Cost Multicut approaches \citep{Tang2015CVPR, Tang2016ECCVWORK, Tang2017CVPR, Babaee2018ARXIV} find a decomposition of the graph that has the minimal sum of costs. Maximum-weight independent set formulations \citep{Shafique2008CVPR, Brendel2011CVPR} first solve the pairwise (two-frame) association problem independently and link the pairwise solutions using a learned distance measure. 
Graphical models \citep{Yang2011CVPR, Yang2012CVPR, Milan2013CVPR, Le2016ECCVWORK} minimize a global energy function defined on the nodes with pairwise and higher-order potentials.

\boldparagraph{Continuous Optimization}
\begin{figure}[t]
	\centering
	\includegraphics[width=1.00\columnwidth]{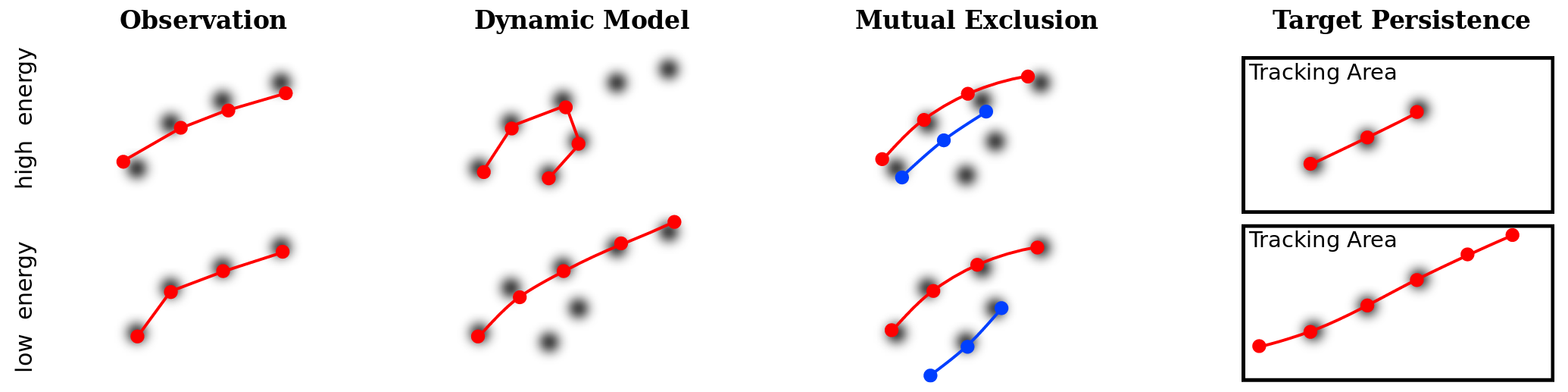}
	\caption[Continuous Energy Function for Multi-Target Tracking]{\textbf{Continuous Energy Formulation.} Components of the energy function proposed by \protect\citet{Andriyenko2011CVPR}. The upper and lower row show a configuration with a higher and smaller energy. The darker grey-values correspond to higher target likelihoods. \figsourceC{\protect\citet{Andriyenko2011CVPR}}{2011}{IEEE}.}
	\label{fig:tracking_continuous_energy}
\end{figure}
As an alternative to discretization, continuous energy minimization approaches have been proposed. For this highly non-convex problem, \citet{Andriyenko2011CVPR} use a heuristic energy minimization scheme with repeated jump moves to prevent poor local minima and better explore the variable-dimensional search space. The effects of different components of their energy function are illustrated in \figref{fig:tracking_continuous_energy}. \citet{Milan2014PAMI} extend the continuous energy function of \citep{Andriyenko2011CVPR} to take into account physical constraints such as target dynamics, mutual exclusion, and track persistence. Assigning each observation to a certain target in data association is intrinsically a discrete optimization problem. Therefore, \citet{Andriyenko2012CVPR} argue that a joint discrete and continuous formulation describes the tracking problem more naturally. Their method alternates between solving the data association problem using discrete optimization with label costs and analytically fitting continuous trajectories while disregarding the label costs. \citet{Milan2013CVPR} propose a mixed discrete-continuous conditional random field model that specifically addresses mutual exclusion in the data association and the trajectory estimation. During data association, each observation should be assigned to at most one target while in the trajectory estimation, two trajectories should always remain spatially separated.

\boldparagraph{Multiple Cues}
For data association, various complementary cues can be used in combination in order to improve the robustness of tracking systems. 
\citet{Giebel2004ECCV} learn a spatio-temporal shape representation based on distinct linear subspace models. They handle appearance changes by combining shape, texture, and depth from stereo in the observation model of a particle filter. 
\citet{Gavrila2007IJCV} employ the same set of cues with a cascade of modules in a detection and tracking system, namely region of interest generation, shape-based detection, texture-based classification, and stereo-based verification. Their system can focus on relevant image regions inferred by a stereo-based region of interest approach. They propose a novel mixture-of-experts architecture by weighting texture-based component classifiers according to the outcome of the shape matching. 
In their appearance-based approach, \citet{Choi2013PAMI} use a combination of detection systems, each specialized in a different task such as pedestrian and upper body, face, skin color, depth-based shape, and motion. The response of all detection systems is combined in the observation likelihood to improve matching.

\subsection{Pedestrian Tracking}
Tracking of pedestrians is of particular importance for autonomous driving as mentioned before. However, the identification of pedestrians remains difficult, especially because of false positives of detection systems.   
\citet{Andriluka2008CVPR} address this problem with a joint detection and articulated human pose tracking formulation. They extend an existing person detector to a limb-based structure model and model the dynamics of the detected limbs with a hierarchical Gaussian process latent variable model (hGPLVM). This allows them to detect people more reliably than approaches considering only one frame. \citet{Andriluka2010CVPR} extend this idea towards 3D pose estimation from monocular images. In the first stage, they estimate 2D articulation and the viewpoint of people and associate them across a small number of frames. This accumulated 2D image evidence is then used to estimate the 3D pose with a hGPLVM. This approach allows them to accurately estimate the 3D poses of multiple people from monocular images. In combination with a Hidden Markov Model (HMM), these approaches can track people over very long sequences. 

\subsection{Joint Detection and Tracking}
While the typical tracking-by-detection approach assumes detections to be available, \citet{Dehghan2015CVPRa, Tian2018PAMI} propose to solve detection and association jointly with a network flow approach by learning a model for each target and modifying the graph to encode the assignment probabilities between the targets and nodes.

\citet{Kang2016CVPR, Kang2017CVPR} introduce a tubelet proposal module that combines object detection and tracking for video object detection. A tubelet represents detections of the same object over consecutive frames. The performance is improved by first generating static object proposals as spatial anchors (\eg from a Region Proposal Network) and then predicting the relative movements to adjust the anchors. 
Instead of propagating bounding boxes, \citet{Tang2019PAMI} link objects in the same frame and propagate box scores across frames. In addition, per-frame proposals in \citep{Kang2016CVPR, Kang2017CVPR} are replaced by spatio-temporal proposals that are directly generated for video segments.

Another line of work uses optical flow for feature aggregation in videos \citep{Zhu2017CVPR, Zhu2017ICCV, Zhu2018CVPR}. The feature maps of nearby frames are warped according to the estimated optical flow and aggregated by learning an adaptive weighting. The motivation is to improve the detection of fast-moving objects which are hard to detect on some frames due to motion blur. A more efficient version is proposed by \citet{Zhu2018CVPR} with key-frame selection, \ie selecting frames or parts of the frames to aggregate. 
\citet{Wang2018ECCVb} also use flow at different levels, namely at the pixel-level by per-pixel warping and at the instance-level by predicting instance movements. Then, the two levels are combined according to the motion pattern observed, \eg by relying on pixel-level more in the case of non-rigid motion.
To avoid expensive optical flow computation, \citet{Bertasius2018ECCV} propose a spatio-temporal sampling mechanism based on deformable convolutional layers.

\subsection{Deep Learning for Multi-Object Tracking}
\label{sec:tracking_DL}
\begin{figure}[t]
	\centering
	\includegraphics[width=1.00\columnwidth]{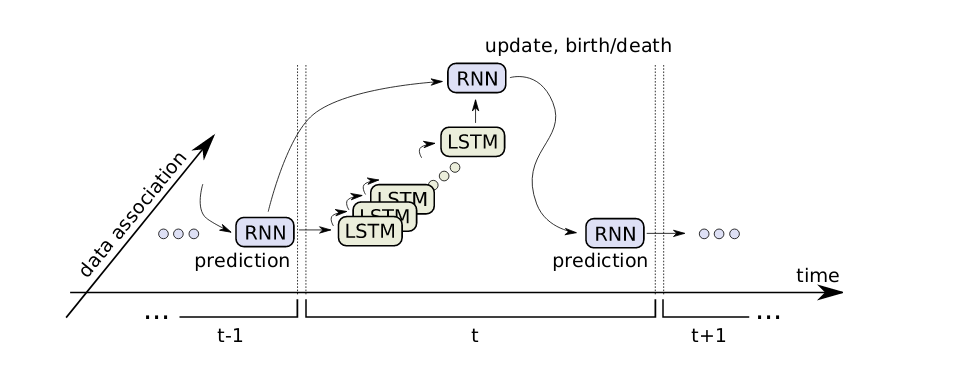}
	\caption[Deep Learning for Tracking]{\textbf{Deep Learning for Tracking.} The end-to-end learning method by \protect\citet{Milan2017AAAI} uses RNNs \citep{Rumelhart1986NATURE} for state estimation and LSTMs \citep{Hochreiter1997NC} for data association. \figsourceC{\protect\citet{Milan2017AAAI}}{2017}{AAAI}.}
	\label{fig:deep_tracking}
\end{figure}
Tracking has strongly benefited from the success of deep learning in object detection discussed in \chpref{sec:deep_learning_detection}. Moreover, deep learning has been used for representation learning to verify detections belonging to the same person \citep{Leal-Taixe2016CVPRWORK, Tang2016ECCVWORK, Tang2017CVPR} or more recently, for learning track representations using sequential models \citep{Sadeghian2017ICCV, Kim2018ECCV, Babaee2018ARXIV}.
Learned sequential models combined with a traditional model for the association have shown to improve performance in comparison to their predecessors. Examples include the combination of appearance, motion, and interaction LSTM networks \citep{Sadeghian2017ICCV} in contrast to Markov decision process tracking with hand-crafted features \citep{Xiang2015ICCV}, a modified bi-linear LSTM \citep{Kim2018ECCV} in contrast to the multiple hypothesis tracking model with CNN features \citep{Kim2015ICCV}, and a hierarchical clustering method based on tracklet similarity using an RNN \citep{Babaee2018ARXIV} in contrast to the lifted multi-cut approach with Siamese networks \citep{Tang2017CVPR}. In these examples, the common approach is to learn a good track representation and then use an established method for the association.

Recently, several approaches \citep{Schulter2017CVPR, Milan2017AAAI, Frossard2018ICRA} proposed end-to-end learning of multi-object tracking. 
The challenges are mainly the scarcity of labeled data, the structured nature of the problem both in the input and the output space, and the combinatorial search space. 
\citet{Schulter2017CVPR} propose a network layer to learn the network flow cost functions based on hand-designed representations of bounding boxes. The first end-to-end learning method for tracking presented by \citet{Milan2017AAAI} illustrated in \figref{fig:deep_tracking} uses a RNN to estimate the states of targets and LSTMs for the association. The model, however, is trained on synthetic data and lacks an appearance model, which makes it unable to match the performance of previous approaches.
\citet{Frossard2018ICRA} propose an end-to-end learning method for detection and tracking of vehicles in 3D using a deep structured loss to backpropagate through the linear program which solves the association problem.
In contrast, \citet{Feichtenhofer2017ICCV} present a more general approach for end-to-end learning of detection and tracking by extending the convolutional object detector proposed in \citep{Dai2016NIPS} with a tracking loss that regresses object coordinates across frames. However, they only evaluate on ImageNet VID challenge \citep{Russakovsky2015IJCV} which mostly consists of sequences with one or a few objects at the center of the video.
\begin{figure}[t]
	\centering
	\includegraphics[width=1.00\columnwidth]{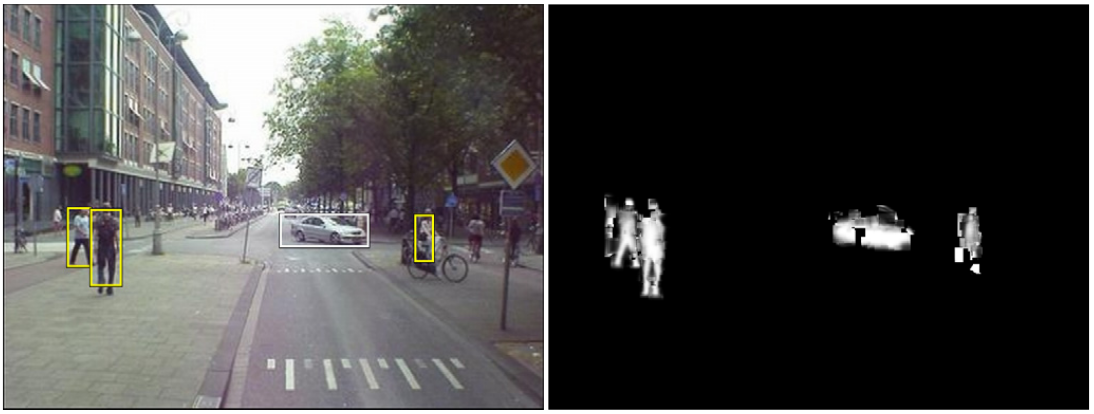}
	\caption[Object Detections and Segmentations for Tracking]{\textbf{Object Detections and Segmentations for Tracking.} The detections (left) and corresponding top-down segmentations (right) used by \protect\citet{Leibe2008PAMI} to learn an object-specific color model for tracking. \figsourceC{\protect\citet{Leibe2008PAMI}}{2008}{IEEE}.}
	\label{fig:object_detection_for_tracking}
\end{figure}

\subsection{3D Object Tracking}
Some works have investigated a joint formulation for object tracking and depth estimation to obtain the structure of the scene while estimating the trajectories of objects in the scene. The structure of the scene allows the tracking system to focus on more plausible solutions. Different input modalities were considered to estimate or obtain depth information, \ie monocular \citep{Hu2018ICCV} and stereo imagery \citep{Giebel2004ECCV, Leibe2007CVPR, Leibe2008PAMI, Ess2009PAMI, Mitzel2012ECCV, Luiten2019ARXIV}. A few approaches \citep{Moosmann2013ICRA, Dewan2016ICRA, Chang2019CVPR} address the tracking problem solely using LiDAR data. However, the missing appearance information and decreasing density of the laser range information with increasing distance complicate the tracking problem. 

\citet{Leibe2007CVPR,Leibe2008PAMI} propose an approach integrating scene geometry estimation, 2D object detection, 3D localization, trajectory estimation, and tracking. They learn object-specific color models using the detection and top-down segmentation of objects, as illustrated in \figref{fig:object_detection_for_tracking}. The structure of the scene guides the extraction of physically plausible space-time trajectories, and a final global optimization criterion takes object-object interactions into account to refine the 3D localization and trajectory estimation results. \citet{Ess2009PAMI} jointly estimate the camera position, stereo depth, object detection, and the pose of all objects over time using a graphical model. Thereby, the graphical model represents the interplay between the different components and incorporates object-to-object interactions. \citet{Luiten2019ARXIV} propose a two-stage approach that first estimates short-term tracks from images and afterwards fuses these tracks by reconstructing the 3D scene from depth. The short-term tracking is performed with a segmentation of the images and a temporal association using optical flow estimates. 

In contrast to previous approaches, \citet{Hu2018ICCV} train a network for 3D box estimation from monocular images. The network learns to regress the 3D location, orientation, and dimension of vehicles. The 3D boxes are used as observation in a tracking formulation and assigned to tracks using a weighted bipartite matching algorithm, taking into consideration the depth ordering. In addition, they use LSTMs to learn two motion models, one for predicting new locations and the other one to update the locations. 

\boldparagraph{Tracking-Before-Detection}
In addition to facilitating the tracking problem, 3D information also allows the segmentation of the scene into different objects, independently of their class. In tracking-before-detection, these segmented class agnostic objects are directly considered as observations in the tracking formulation. This way, the tracking system is independent of a classifier and, thus, is able to track unknown objects which have not been seen before or for which only a little amount of training data exists. Furthermore, motion information from the object's estimated trajectory can be used as another cue to detect a certain class of objects. \citet{Mitzel2012ECCV} extract observations of objects by segmenting the scene using depth from stereo. With a compact 3D representation, they can robustly track known and unknown object categories. This representation also allows them to detect anomalous shapes such as carried items.

\begin{table*}[t!]
	\begin{subtable}{\linewidth}
		\begin{adjustbox}{width=1\textwidth}\begin{tabular}{l|c|c|c|c|c|c|c|c}
& {\textbf{Method}} & {\textbf{MOTA}} & {\textbf{IDF1}} & {\textbf{MT}} & {\textbf{ML}} & {\textbf{IDS}} & {\textbf{FRAG}} & {\textbf{Hz}} \\ \hline
1. & HCC \citep{Ma2018ACCV} & 49.3\% & 50.7\% & 17.8 \% & 39.9\% & 391 & 535 & 0.8 \\
2. & eTC  \citep{Wang2018ARXIV} & 49.2\% & 56.1\% & 17.3 \% & 40.3 \% & 606 & 882 & 0.7 \\
3. & AFN \citep{Shen2018ARXIV} & 49.0\% & 48.2\% & 19.1 \% & 35.7 \% & 899 & 1,383 & 0.6 \\
4. & KCF \citep{Chu2019WACV} & 48.8\% & 47.2\% & 15.8 \% & 38.1 \% & 906 & 1,116 & 0.1 \\
5. & LMP \citep{Tang2017CVPR} & 48.8\% & 51.3\% & 18.2 \% & 40.1 \% & 481 & 595 & 0.5 \\
\hline
16. & NOMT \citep{Choi2015ICCV} & 46.4\% &  53.3\% & 18.3\%  & 41.4\%  & 359 & 504 & 2.6 \\ 
17. & JMC \citep{Tang2016ECCVWORK} & 46.3\% & 46.3\% & 15.5\%  & 39.7\%  & 657 & 1,114 & 0.8 \\ 
18. & STAM \citep{Chu2017ICCV} & 46.0\%	 & 50.0\%	& 14.6\% & 43.6\% & 473 & 1,422 &	0.2 \\
22. & MHT\_DAM  \citep{Kim2015ICCV} & 42.9\% &  46.1\% & 13.6\%  & 46.9\%  & 499 & 659 & 0.8 \\ 
38. & GMMCP \citep{Dehghan2015CVPRa} & 38.1\% & 35.5\% & 8.6 \% & 50.9 \% & 937 & 1,669 & 0.5 \\
44. & CEM  \citep{Milan2014PAMI} & 33.2\% & 0.0\% & 7.8\%  & 54.4\%  & 642 & 731 & 0.3 \\ 
49. & DP\_NMS  \citep{Pirsiavash2011CVPR} & 32.2\% &  31.2\% & 5.4\%  & 62.1\%  & 972 & 944 & 5.9 \\ 
\end{tabular}\end{adjustbox}
		\caption{MOT16 Leaderboard using Public DPM detections}
		\label{tab:mot_challenge_public_16}
		\vspace{0.2cm}
	\end{subtable}
	\begin{subtable}{\linewidth}
		\begin{adjustbox}{width=1\textwidth}\begin{tabular}{l|c|c|c|c|c|c|c|c}
& {\textbf{Method}} & {\textbf{MOTA}} & {\textbf{IDF1}} & {\textbf{MT}} & {\textbf{ML}} & {\textbf{IDS}} & {\textbf{FRAG}} & {\textbf{Hz}} \\ \hline
1. & LMP \citep{Tang2017CVPR} & 71.0\% & 70.1\% & 46.9 \% & 21.9 \% & 434 & 587 & 0.5 \\
2. & KDNT  \citep{Yu2016ECCVWORK} & 68.2\% & 60.0\% & 41.0\%  & 9.0\% & 933 & 1,093 & 0.7 \\ 
3. & POI \citep{Yu2016ECCVWORK} & 66.1\% & 65.1\% & 34.0\%  & 20.8\%  & 805 & 3,093 & 9.9 \\ 
\hline
8. & NOMTwSDP16  \citep{Choi2015ICCV} & 62.2\% & 62.6\% & 32.5\%  & 31.1\%  & 406 & 642 & 3.1 \\ 
9. & DeepSORT\_2  \citep{Wojke2017ICIP} & 61.4 \%& 62.2\% & 32.8 \%  & 18.2\%  & 781 & 2,008 & 17.4 \\ 
10. & SORTwHPD16  \citep{Bewley2016ICIP} & 59.8\% & 79.6\% & 25.4\%  & 22.7\%  & 1,423 & 1,835 & 59.5 \\ 
11. & IOU \citep{Bochinski2017AVSS} & 57.1\% & 46.9\% & 23.6\%  & 32.9\%  & 2,167 & 3,028 & 3,004.6 \\ 
\end{tabular}\end{adjustbox}
		\caption{MOT16 Leaderboard using a Private Detector}
		\label{tab:mot_challenge_private_16}
		\vspace{0.2cm}
	\end{subtable}
	\caption{{\bf MOT16 Multi Object Tracking Leaderboard.} We report the Multiple Object Tracking Accuracy (MOTA), F1 score on identified detections (IDF1), the ratio of mostly tracked (MT) and mostly lost trajectories (ML), number of ID switches (IDS) and track segmentations (FRAG), and run time. The metrics are detailed in \citep{Milan2016ARXIV}. Methods below the horizontal line show older entries for reference. Accessed on: June 2019.}
	\label{tab:mot_challenge_16}
\end{table*}

\begin{table*}[t]
\begin{center}
\begin{adjustbox}{width=1\textwidth}\begin{tabular}{l|c|c|c|c|c|c|c|c}
& {\textbf{Method}} & {\textbf{MOTA}} & {\textbf{IDF1}} & {\textbf{MT}} & {\textbf{ML}} & {\textbf{IDS}} & {\textbf{FRAG}} & {\textbf{Hz}} \\ \hline
1. & JBNO \citep{Henschel2019CVPRWORK} &  52.6\% & 50.8\%  & 19.7 \% & 35.8 \% & 3,050 &  3,792 & 5.4  \\
2. &  FAMNet \citep{Chu2019ARXIV} &  52.0\% &  48.7\% &  19.1\% & 33.4 \% &  3,072 &  5,318 & - \\
3. & eTC  \citep{Wang2018ARXIV} & 51.9\%  & 58.1\% &  23.1\% & 35.5 \% &  2,288 & 3,071 & 0.7 \\
4. & eHAF17 \citep{Sheng2019TCSVT} & 51.8\% & 54.7\% & 23.4\% & 37.9\% &  1,834 & 2,739 & 0.7	 \\
5. & AFN \citep{Shen2018ARXIV} &  51.5\% &  46.9\% &  20.6\% &  35.5\% &  2,593 &  4,308 & 1.8 \\
6. & FWT \citep{Henschel2018CVPRWORK} & 51.3\% &  47.6\% &  21.4\% &  35.2\% &  2,648 & 4,279 & 0.2  \\
7. &  jCC \citep{Keuper2018PAMI} & 51.2\% &  54.5\% &  20.9\% & 37.0\% &  1,802 & 2,984 & 1.8  \\
\hline
9. & MHT\_DAM  \citep{Kim2015ICCV} &  50.7\% & 47.2\% &  20.8\% & 36.9\% & 2,314 &  2,865 & 0.9  \\
14. & DMAN  \citep{Zhu2018ECCV} & 48.2\% & 55.7\% &  19.3\% &  38.3\% & 2,194 & 5,378 & 0.3 \\
17. &  MHT\_bLSTM  \citep{Kim2018ECCV} &  47.5\% & 51.9\% & 18.2\% & 41.7\% &  2,069 &  3,124 & 1.9 \\
\end{tabular}\end{adjustbox}
\end{center}
\vspace{-0.4cm}
\caption{{\bf MOT17 Multi Object Tracking Leaderboard using Provided Detections.} We report the Multiple Object Tracking Accuracy (MOTA), F1 score on identified detections (IDF1), the ratio of mostly tracked (MT) and mostly lost trajectories (ML), number of ID switches (IDS) and track segmentations (FRAG), and run time. The metrics are detailed in \citep{Milan2016ARXIV}. Methods below the horizontal line show older entries for reference. Accessed on: June 2019.}
\label{tab:mot_challenge_17}
\end{table*}

\begin{table*}[t!]
	\begin{subtable}{\linewidth}
		\begin{adjustbox}{width=1\textwidth}\begin{tabular}{l | c | c | c | c | c | c | c | c}
& {\bf Method} & {\bf MOTA} & {\bf MOTP} & {\bf MT} & {\bf ML} & {\bf IDS} & {\bf FRAG} & {\bf Runtime}\\ \hline
1. & MOTBeyondPixels \citep{Sharma2018ICRA} & 84.24 \% & 85.73 \% & 73.23 \% & 2.77 \% & 468 & 944 & 0.3 s / 1 core \\
2. & IMMDP \citep{Xiang2015ICCV} & 83.04 \% & 82.74 \% & 60.62 \% & 11.38 \% & 172 & 365 & 0.19 s / 4 cores \\
3. & JCSTD \citep{Tian2019TITS} & 80.57 \% & 81.81 \% & 56.77 \% & 7.38 \% & 61 & 643 & 0.07 s / 1 core \\
4. & 3D-CNN/PMBM \citep{Scheidegger2018IV} & 80.39 \% & 81.26 \% & 62.77 \% & 6.15 \% & 121 & 613 & 0.01 s / 1 core \\
\hline
7. & NOMT* \citep{Choi2015ICCV} & 78.15 \% & 79.46 \% & 57.23 \% & 13.23 \% & 31 & 207 & 0.09 s / 16 cores \\
10. & DSM \citep{Frossard2018ICRA} & 76.15 \% & 83.42 \% & 60.00 \% & 8.31 \% & 296 & 868 & 0.1 s / GPU \\
11. & SCEA* \citep{Yoon2016CVPR} & 75.58 \% & 79.39 \% & 53.08 \% & 11.54 \% & 104 & 448 & 0.06 s / 1 core \\
12. & CIWT* \citep{Osep2017ICRA} & 75.39 \% & 79.25 \% & 49.85 \% & 10.31 \% & 165 & 660 & 0.28 s / 1 core \\
14. & SSP* \citep{Lenz2015ICCV} & 72.72 \% & 78.55 \% & 53.85 \% & 8.00 \% & 185 & 932 & 0.6 s / 1 core \\
18. & RMOT* \citep{Yoon2015WACV} & 65.83 \% & 75.42 \% & 40.15 \% & 9.69 \% & 209 & 727 & 0.02 s / 1 core \\
\end{tabular}\end{adjustbox}
		\caption{KITTI Car Tracking Leaderboard}
		\label{tab:kitti_tracking_car}
		\vspace{0.2cm}
	\end{subtable}
	\begin{subtable}{\linewidth}
		\begin{adjustbox}{width=1\textwidth}\begin{tabular}{l | c | c | c | c | c | c | c | c}
& {\bf Method} & {\bf MOTA} & {\bf MOTP} & {\bf MT} & {\bf ML} & {\bf IDS} & {\bf FRAG} & {\bf Runtime}\\ \hline
1. & IMMDP \cite{Xiang2015ICCV} & 47.22 \% & 70.36 \% & 24.05 \% & 27.84 \% & 87 & 825 & 0.9 s / 8 cores \\
2. & NOMT* \cite{Choi2015ICCV} & 46.62 \% & 71.45 \% & 26.12 \% & 34.02 \% & 63 & 666 & 0.09 s / 16 cores \\
4. & JCSTD \cite{Tian2019TITS} & 44.20 \% & 72.09 \% & 16.49 \% & 33.68 \% & 53 & 917 & 0.07 s / 1 core \\
5. & SCEA* \cite{Yoon2016CVPR} & 43.91 \% & 71.86 \% & 16.15 \% & 43.30 \% & 56 & 641 & 0.06 s / 1 core \\
6. & RMOT* \cite{Yoon2015WACV} & 43.77 \% & 71.02 \% & 19.59 \% & 41.24 \% & 153 & 748 & 0.02 s / 1 core \\
\hline
8. & CIWT* \cite{Osep2017ICRA} & 43.37 \% & 71.44 \% & 13.75 \% & 34.71 \% & 112 & 901 & 0.28 s / 1 core \\
10. & NOMT \cite{Choi2015ICCV} & 36.93 \% & 67.75 \% & 17.87 \% & 42.61 \% & 34 & 789 & 0.09 s / 16 core \\
11. & RMOT \cite{Yoon2015WACV} & 34.54 \% & 68.06 \% & 14.43 \% & 47.42 \% & 81 & 685 & 0.01 s / 1 core \\
13. & SCEA \cite{Yoon2016CVPR} & 33.13 \% & 68.45 \% & 9.62 \% & 46.74 \% & 16 & 717 & 0.05 s / 1 core \\
\end{tabular}\end{adjustbox}
		\caption{KITTI Pedestrian Tracking Leaderboard}
		\label{tab:kitti_tracking_pedestrian}
		\vspace{0.2cm}
	\end{subtable}
	\caption{{\bf KITTI Tracking Leaderboard.} We report the Multiple Object Tracking Accuracy (MOTA), Multiple Object Tracking Precision (MOTP), the ratio of mostly tracked (MT) and mostly lost trajectories (ML), number of ID switches (IDS) and track segmentations (FRAG), and run time. The metrics are detailed in \citep{Geiger2012CVPR}. Methods below the horizontal line show older entries for reference. Accessed on: June 2019.}
	\label{tab:kitti_tracking}
\end{table*}
\section{Datasets}
Early datasets for multi-object tracking include independent sequences such as PETS \citep{Ferryman2009PETS}, TUD \citep{Andriluka2008CVPR}, and ETHZ \citep{Ess2008CVPR}. The separate evaluation of these sequences led to tracking algorithms over-fitting to some of these sequences while performing worse on others.
The MOT Challenge \citep{Leal-Taixe2015ARXIV, Milan2016ARXIV} combines most of these sequences into one framework by providing a centralized evaluation and comparison.
While some sequences, \eg PETS and TUD, are captured from a static observer, other sequences that are more relevant for autonomous driving are acquired from a mobile platform. 
The KITTI dataset \citep{Geiger2012CVPR, Geiger2013IJRR} provides tracking data specific to autonomous driving with separate evaluations for tracking car and pedestrian classes. 
Recently, \citet{Chang2019CVPR} presented a novel 3D object tracking dataset collected by a fleet of autonomous vehicles. The dataset consists of $360^{\circ}$ images, forward-facing stereo imagery, LiDAR, and 6-DOF pose. The authors provide 290km of lane markings and 10k human-annotated tracked objects.

The MOT Benchmark published over three consecutive years, MOT15 \citep{Leal-Taixe2015ARXIV}, MOT16, and MOT17 \citep{Milan2016ARXIV}, consists of sets of sequences with tracking labels and provides an official evaluation protocol based on CLEAR metrics \citep{Stiefelhagen2007CLEAR}. The earliest MOT15 uses a classical object detector based on aggregated channel features (ACF) \citep{Dollar2014PAMI}. In MOT16, detections are obtained using Deformable Parts Model (DPM) \citep{Felzenszwalb2008CVPR} while in MOT17, three different sets of object detections are provided using DPM \citep{Felzenszwalb2008CVPR}, Faster R-CNN \citep{Ren2015NIPS}, and Scale Dependent Pooling (SDP) \citep{Yang2016CVPR}. Providing sets of detections allows comparing approaches based on their ability to track objects independent of errors caused by different detectors. For MOT16, the leaderboard with methods using the public (DPM) detections is provided in \tabref{tab:mot_challenge_public_16} and the methods using a private detector are shown in \tabref{tab:mot_challenge_private_16}. For MOT17, \tabref{tab:mot_challenge_17} shows the average results over three provided detectors on the same set of sequences.  

For autonomous driving application specifically, KITTI \citep{Geiger2012CVPR} provides two benchmarks, one for tracking of cars (KITTI car) in \tabref{tab:kitti_tracking_car} and the other for tracking of pedestrians in \tabref{tab:kitti_tracking_pedestrian}. Methods marked with an asterisk use Regionlet detections \citep{Wang2015PAMI} for an independent comparison of the tracking performance. 
The separate challenges for cars and pedestrians allow focusing on each class separately and investigating the problems specific to a class deeply. 

\section{Metrics}
In Tables \ref{tab:mot_challenge_public_16}, \ref{tab:mot_challenge_private_16}, \ref{tab:mot_challenge_17}, \ref{tab:kitti_tracking_car}, \ref{tab:kitti_tracking_pedestrian}, we consider the commonly used tracking measures, Multiple Object Tracking Accuracy (MOTA) and Multiple Object Tracking Precision (MOTP) introduced by \citep{Stiefelhagen2007CLEAR}, the ratio of mostly tracked (MT) and mostly lost trajectories (ML), number of ID switches (IDS) and track segmentations (FRAG). For the MOT leaderboards, following the benchmark page, we show the IDF1 score introduced by \citep{Ristani2016ECCVWORK} instead of MOTP. The IDF1 score is the F1 score of the identification precision and recall, \ie the ratio of correctly identified detections over the average number of ground truth and computed detections.
Mostly tracked and mostly lost trajectories show the percentage of trajectories that are covered by a hypothesis at least 80\% or at most 20\% of the time, respectively. For descriptions of the metrics and the detailed tables with additional metrics such as False Negatives, False Positives, ID recall, and ID precision, please check the KITTI \citep{Geiger2012CVPR} and MOT benchmarks \citep{Leal-Taixe2015ARXIV, Milan2016ARXIV} as well as \cite{Ristani2016ECCVWORK}.

\section{State of the Art on MOT \& KITTI}
\boldparagraph{MOT16 Benchmark}
Classical approaches such as near-online multi-target tracking approach \citep{Choi2015ICCV}, multiple hypothesis tracking approach \citep{Kim2015ICCV} and tracking based on Markov decision processes \citep{Xiang2015ICCV} still perform consistently well on MOT benchmarks in comparison to newly proposed methods. Their deep learning counterparts with better appearance models that are learned as explained in \secref{sec:tracking_DL} perform even better \citep{Sadeghian2017ICCV, Kim2018ECCV}. The comparison of \citep{Sadeghian2017ICCV} to \citep{Xiang2015ICCV} can be found on MOT15\footnote{\url{https://motchallenge.net/results/2D_MOT_2015/}} due to the date of the first publication preceding MOT16 and MOT17. The learning-based method proposed in \citep{Kim2018ECCV} is trained on ground truth detections and performs worse compared to \citep{Kim2015ICCV} on MOT17 (\tabref{tab:mot_challenge_17}) due to noisy DPM detections, which affect overall performance as explained in \citep{Kim2018ECCV}.

The success of single object trackers (SOT) triggered methods that combine several single object detectors for MOT (Tables \ref{tab:mot_challenge_public_16}, \ref{tab:mot_challenge_17}) by learning a tracker for each object \citep{Chu2017ICCV, Chu2019WACV}. These approaches initialize a new single object tracker whenever a new object is detected with high confidence. They assign detections of known objects to each single object tracker by restricting the search space according to a motion model and choosing the best detection candidate using a binary classifier. In \citet{Chu2017ICCV}, a spatial-temporal attention mechanism is proposed to handle the drift caused by occlusion and interactions among targets. \citet{Chu2019WACV} encode awareness both within and between object models and proposes an adaptive model refreshment strategy to eliminate noise in model initialization.

The customized tracker proposed by \citet{Ma2018ACCV} is the best-published method on MOT16 using the provided detections (\tabref{tab:mot_challenge_public_16}). The method is based on the formulation of tracking as a minimum cost lifted multi-cut problem similar to \citep{Tang2015CVPR, Tang2016ECCVWORK, Tang2017CVPR}. Despite being offline and therefore not directly applicable to autonomous driving, this type of graph-based clustering formulation performs very well on MOT. In contrast to previous methods, \citet{Ma2018ACCV} learn a sequence-specific tracker by fine-tuning a re-identification network using the test sequences. They use the assumption that non-overlapping tracklets represent different individuals to adapt a generic re-identification CNN on test sequences.

Comparing public and private detectors on MOT16 (Tables \ref{tab:mot_challenge_public_16}, \ref{tab:mot_challenge_private_16}) shows the importance of good object detectors. Tracking algorithms perform much better using private (usually better) object detections compared to public detections, \eg LMP \citep{Tang2017CVPR} 71.0\% versus 48.8\% and NOMT \citep{Choi2015ICCV} 62.2\% versus 46.4\% in MOTA.
Recent object detectors combined with simple tracking algorithms perform significantly better than any tracker with public detections such as the simple tracker IOU \citep{Bochinski2017AVSS} or SORT \citep{Bewley2016ICIP}, which are based on the Hungarian method in combination with Kalman filtering. \citet{Wojke2017ICIP} improve the performance of SORT further by incorporating deep features into the pipeline for appearance matching. Similarly, \citet{Yu2016ECCVWORK} also use a tracking algorithm based on the Hungarian algorithm and Kalman filtering in combination with deep features for appearance matching. However, their detector is trained on additional data including a self-collected surveillance dataset, which is not public.

\boldparagraph{MOT17 Benchmark}
Top-performing methods on MOT17 (\tabref{tab:mot_challenge_17}) follow a graph clustering scheme by associating tracklets, \ie a short sequence of detections, which can be easily and reliably associated, instead of detections. \citet{Wang2018ARXIV} first create tracklets based on IOU and epipolar geometry in the case of a moving camera. Tracklets represent nodes on a graph which are then clustered based on a greedy search-based clustering method. \citet{Shen2018ARXIV} incorporate the score of the tracklets into the learning-based network flow approach proposed in \citep{Schulter2017CVPR}. 
While these methods have a preprocessing step to generate tracklets, a more recent approach called FAMNet \citep{Chu2019ARXIV} combines feature extraction, affinity estimation and the assignment problem in a single network. Furthermore, single object tracking is incorporated into the tracking system in order to recover from missing detections.

Recently, several approaches \citep{Keuper2018PAMI, Henschel2018CVPRWORK, Henschel2019CVPRWORK} propose to use additional cues such as head detections and motion segmentation to improve tracking. Two of the best-performing methods on MOT17 (\tabref{tab:mot_challenge_17}) fuse head, body, and joint detectors into a tracking system \citep{Henschel2018CVPRWORK, Henschel2019CVPRWORK}. \citet{Keuper2018PAMI} address multi-object tracking with top-down clustering of bounding boxes and bottom-up motion segmentation by grouping point trajectories. 

\boldparagraph{KITTI Benchmark}
\begin{figure}[t]
	\centering
	\includegraphics[width=1.00\columnwidth]{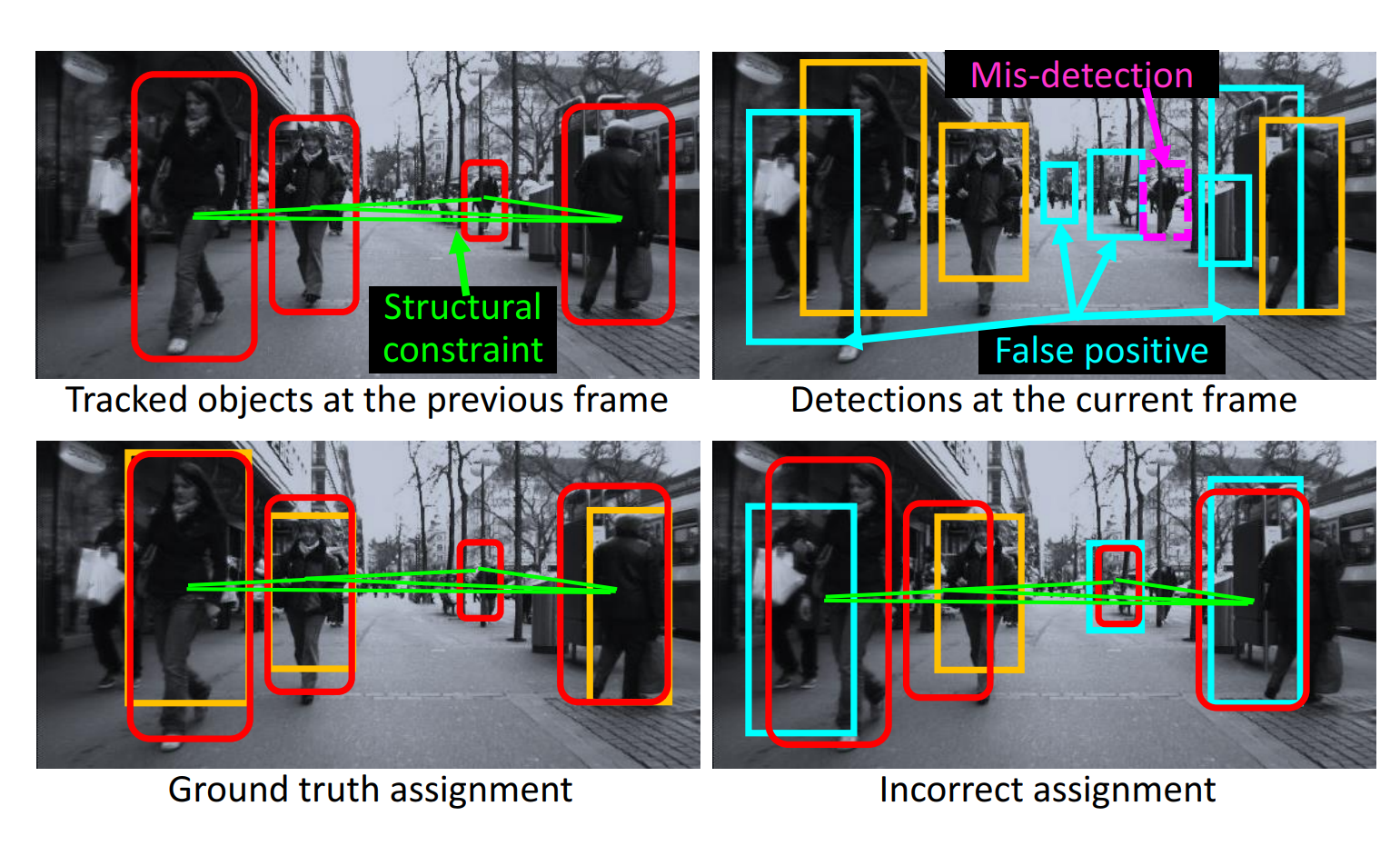}
	\caption[Multi-Object Tracking with Structural Motion Constraints]{\textbf{Tracking with Structural Motion Constraints.} Structural motion constraints introduced by \protect\citet{Yoon2016CVPR} to resolve errors caused by false positives. The correct detections are marked with red and yellow boxes. \figsourceC{\protect\citet{Yoon2016CVPR}}{2016}{IEEE}.}
	\label{fig:Yoon2016CVPR}
\end{figure}
In contrast to the MOT Challenge, the KITTI benchmark focuses on the challenging scenario of tracking pedestrians (\tabref{tab:kitti_tracking_pedestrian}) and cars (\tabref{tab:kitti_tracking_car}) in traffic scenes. Similarly to MOT, classical approaches perform reasonably well such as tracking based on Markov decision process (IMMDP) \citep{Xiang2015ICCV}, improved min-cost network flow \citep{Lenz2015ICCV}, or the near-online multi-target tracking algorithm (NOMT) \citep{Choi2015ICCV}. In IMMDP, a policy is learned using reinforcement learning, which corresponds to learning a similarity function for data association. An improved version with Region Proposal Network \citep{Ren2015NIPS} is the best performing method on the car tracking task. \citet{Lenz2015ICCV} propose a computational and memory bounded version of the min-cost network flow formulation presented in \citep{Zhang2008CVPR}. This approach achieves good accuracy and precision while being amongst the fastest approaches on KITTI car. NOMT \citep{Choi2015ICCV} proposes Aggregated Local Flow Descriptor (ALFD) which encodes relative motion patterns. Thanks to these features, distant detections can be robustly matched. Using multiple feature cues, their method outperforms all the online tracking approaches on KITTI car. 

Recent approaches leverage domain-specific information such as the motion of the car or the structure of the scene. 
\citet{Yoon2015WACV} factor out the camera motion by constructing a network to describe the relative motion between objects. 
They further improve in \citep{Yoon2016CVPR} by exploiting structural motion constraints defined by the location and velocity difference between two objects as illustrated in \figref{fig:Yoon2016CVPR}. Jointly reasoning about the structure allows them to alleviate problems that are common to 2D trackers (\eg occlusions) and outperform them, especially in the car tracking task. \citet{Frossard2018ICRA} propose to learn tracking in a network flow approach based on 3D detections. The structured hinge loss is adapted to backpropagate through the Integer Program. Other top-performing 3D algorithms are \citep{Osep2017ICRA} coupling image and world-space estimations using a novel 2D-3D Kalman filter and \citep{Scheidegger2018IV} proposing a Poisson multi-Bernoulli mixture (PMBM) tracker. 
\citet{Sharma2018ICRA} exploit the geometry of urban road scenes to infer 3D cues for tracking such as 3D pose and shape based on single view reconstruction of objects. 
This approach outperforms all others in accuracy (MOTA) and precision (MOTP) in the KITTI car leaderboard. 

\section{Discussion}
Reliable tracking-by-detection can only be achieved by using very accurate object detections. The impact of the detection system can be observed when comparing the methods marked with and without asterisks in the KITTI leaderboards (Tables \ref{tab:kitti_tracking_car},\ref{tab:kitti_tracking_pedestrian}). In the MOT16 leaderboards this can be observed when comparing the tables for methods using public detections in \tabref{tab:mot_challenge_public_16} and private object detectors in \tabref{tab:mot_challenge_private_16}. However, we discuss the problem of object detection in detail in \secref{sec:detection_discussion} and focus our attention in this section on the tracking problem. Similar to the detection, tracking pedestrians is typically more challenging than cars. The reason is the complex motion of pedestrians which is hard to predict, in contrast to the rigid motion of cars which are bound by the road region and follow a less erratic behavior due to their large mass and dynamical constraints. 3D reasoning can help to improve tracking performance, especially for cars, by identifying plausible solutions according to geometric relationships.

In traffic scenes, detectors frequently fail for partially or fully occluded objects. In these cases, the tracking system needs to re-identify the tracked objects later in time but this can be difficult due to changes in lighting conditions or similarity to other objects in the proximity. These problems cause a reinitialization of trajectories, which can be observed in the high number of fragmentations (FRAG) and ID switches (IDS) in the MOT and KITTI benchmarks. Furthermore, we note that most tracking systems comprise complex pipelines and very few end-to-end multiple target tracking algorithms have been proposed in the literature. Bridging this gap from detection to tracking with the goal of a generic and end-to-end trainable model will be an important direction for future research in this area.
\chapter{Semantic Segmentation} 
\begin{figure}[tb]
	\centering
	\includegraphics[width=1.00\columnwidth]{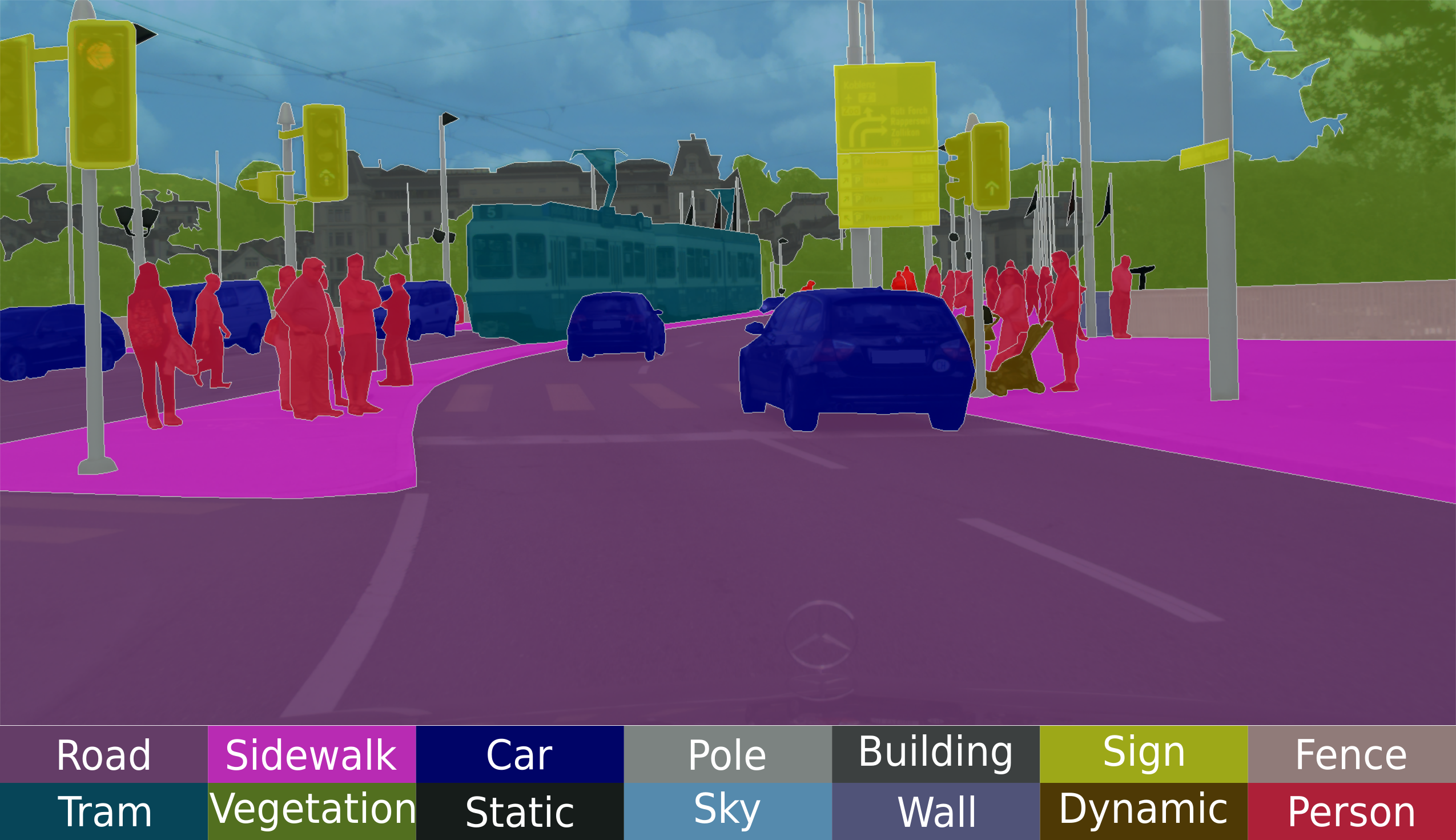}
	\caption[Semantic Segmentation Example from Cityscapes]{\textbf{Semantic Segmentation.} In semantic segmentation, the goal is to assign a semantic class label to each pixel in the image. Example from the Cityscapes dataset by \protect\citet{Cordts2016CVPR}. \courtesy{www.cityscapes-dataset.com}.}
	\label{fig:cityscapes}
\end{figure}
	
\section{Problem Definition}
Semantic segmentation is a fundamental problem in computer vision and an intermediate goal towards solving higher-level tasks such as scene understanding or sensorimotor control. The goal of semantic segmentation is to assign each pixel in the image a label from a predefined set of categories. The task is illustrated in \figref{fig:cityscapes} using an example from the Cityscapes dataset\footnote{\url{https://www.cityscapes-dataset.com/}} by \citet{Cordts2016CVPR}. Segmentation of images into semantic regions 
that are typically found in street scenes, such as cars, pedestrians, or road allows for a comprehensive understanding of the surrounding which is essential to autonomous navigation. The task is difficult due to the complexity of the scene, complicated object boundaries, small objects and the large size of the label space. 

\section{Methods}
\begin{figure}[t]
	\centering
	\includegraphics[width=1.00\columnwidth]{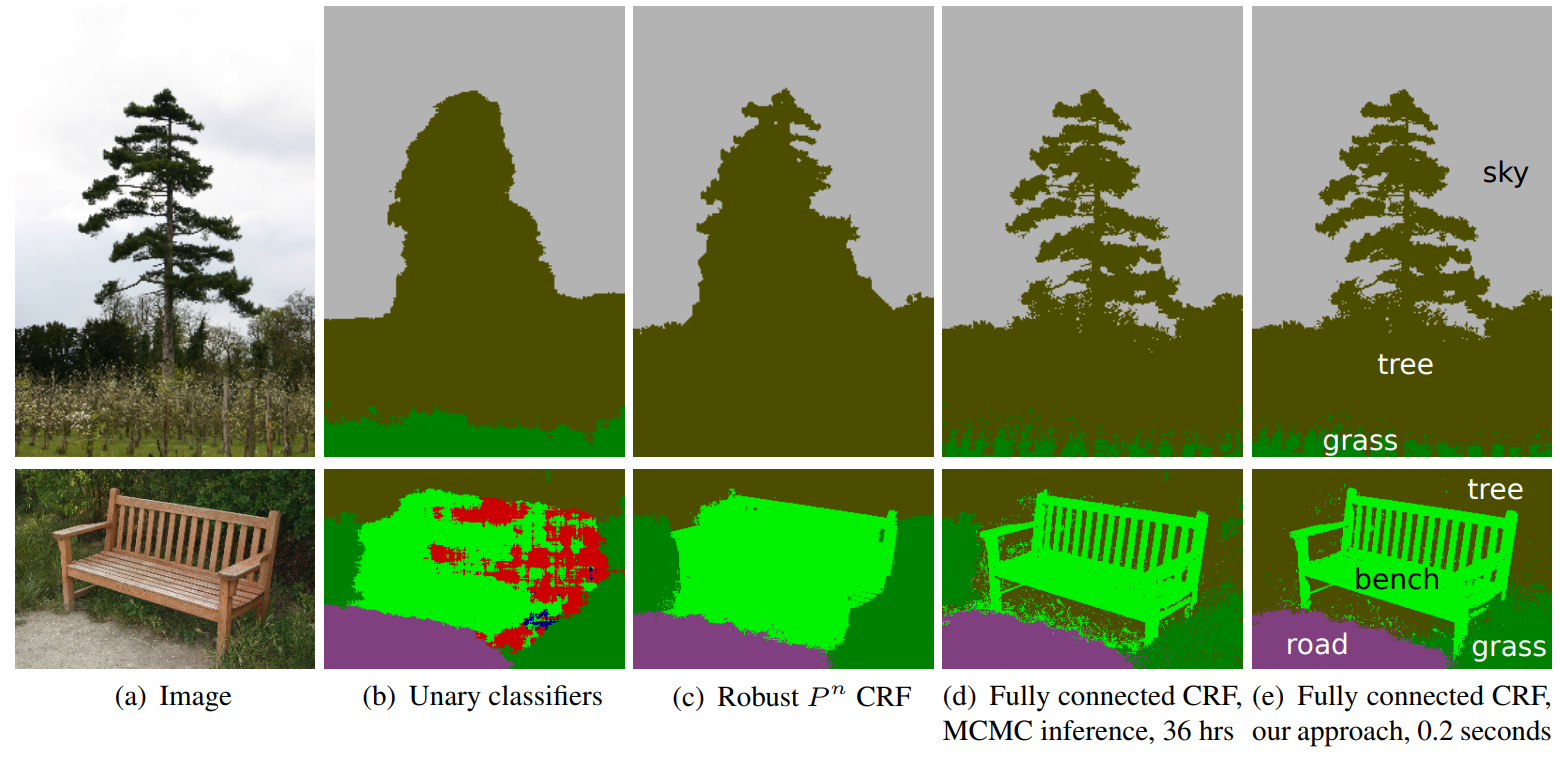}
	\caption[Fully Connected Conditional Random Field]{\textbf{Fully Connected Conditional Random Field.} Semantic segmentation results of a per-pixel classifier \protect\citep{Kraehenbuehl2011NIPS}, a superpixel-based CRF \protect\citep{Kohli2009IJCV} and a fully connected CRF \protect\citep{Kraehenbuehl2011NIPS}. \figsourceC{\protect\citet{Kraehenbuehl2011NIPS}}{2016}{NeurIPS}.}
	\label{fig:Kraehenbuehl2011NIPS}
\end{figure}

The goal of semantic segmentation is to assign a semantically meaningful class label (\eg road, sidewalk, pedestrian, sky) to each pixel of an image. 
Traditionally, the problem was posed as maximum-a-posteriori (MAP) inference in a conditional random field (CRF), defined over pixels \citep{He2004CVPR, Verbeek2007NIPS, Shotton2009IJCV} or superpixels \citep{He2006ECCV, Kohli2009IJCV}. Hierarchical \citep{He2004CVPR, Kumar2005ICCV, Ladicky2009ICCV, Ladicky2014PAMI} and long-range connectivity as well as higher-order potentials defined on image regions \citep{He2006ECCV, Kohli2009IJCV} have been exploited to compensate for limitations of CRFs with local connections and to model long-range interactions within the image. \citet{Kraehenbuehl2011NIPS} propose a tractable inference algorithm for fully connected CRF models which model pairwise potentials between all pairs of pixels in the image. While previous methods using fully connected CRFs \citep{Rabinovich2007ICCV, Galleguillos2008CVPR, Kohli2009IJCV} could only be applied to smaller image regions due to the computational and memory complexity of these algorithms, \cite{Kraehenbuehl2011NIPS} allows deploying fully connected CRF models at pixel-level. \figref{fig:Kraehenbuehl2011NIPS} illustrates the results of \cite{Kraehenbuehl2011NIPS} and compares them to pixel-wise classification and inference over superpixels \citep{Kohli2009IJCV}.

An alternative to inference in graphical models for the task of semantic segmentation is presented by \citet{Munoz2010ECCV}. They train a sequence of inference models in a hierarchical procedure that captures context over larger image regions. This allows them to bypass the difficulties of training structured prediction models when exact inference is intractable and yields a very efficient and accurate scene labeling algorithm.

While most previous approaches rely on very simple features such as color, edge and texture information, \citet{Shotton2009IJCV} observed that more powerful features have the potential to significantly boost performance. They propose an approach based on a novel feature type called texture-layout filter that exploits the textural appearance of objects, its layout as well as textural context. They combine texture-layout filters with lower-level image features in a CRF to obtain pixel-level segmentations. 

\boldparagraph{Co-occurrence of Object Classes}
The methods so far consider each object class independently. However, the co-occurrence of object classes is typically not random and can thus be an important cue for semantic segmentation, \eg cars are more likely to occur in a street scene than in an office scene and co-occur with other street scene objects such as traffic signs. 
\citet{Ladicky2010ECCV} propose to explicitly incorporate object class co-occurrence as global features into a CRF. They optimize the CRF using graph cuts and demonstrate better performance compared to pairwise models.
\citet{Zhang2012CVPR} extend this idea by encoding spatial arrangements of different object categories. \citet{Myeong2012CVPR} propose a retrieval-based approach which extracts contextual relationships from annotated region pairs.

\begin{figure}[t]
	\centering
	\includegraphics[width=1.00\columnwidth]{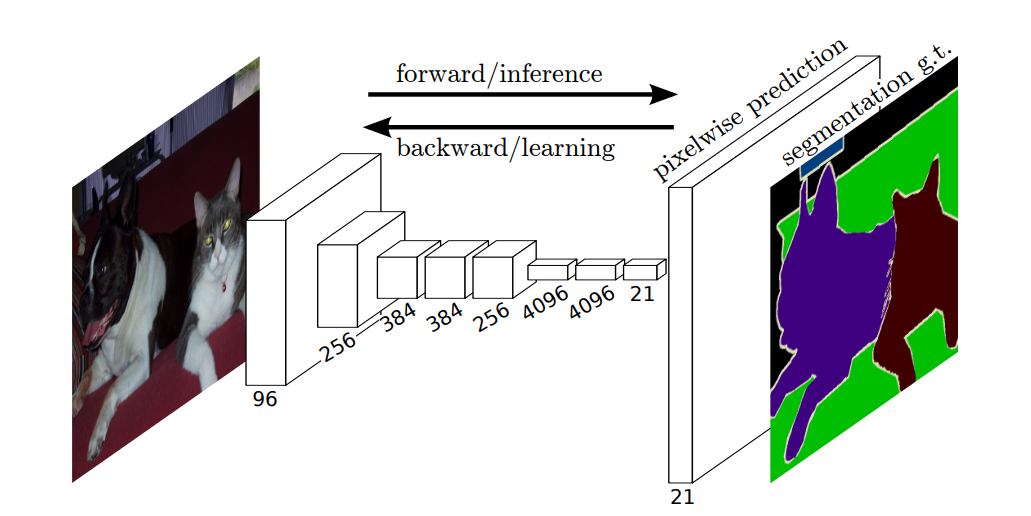}
	\caption[Convolutional Neural Network for Semantic Segmentation]{\textbf{Convolutional Neural Network.} Fully convolutional neural network for semantic segmentation proposed by \protect\citet{Long2015CVPR}. \figsourceC{\protect\citet{Long2015CVPR}}{2015}{IEEE}.}
	\label{fig:Long2015CVPR}
\end{figure}

\subsection{Deep Learning for Semantic Segmentation}
The success of deep convolutional neural networks for image classification and object detection has sparked interest in leveraging their potential for solving pixel-level tasks, in particular semantic segmentation.
The fully convolutional neural network \citep{Long2015CVPR} illustrated in \figref{fig:Long2015CVPR} is one of the earliest works which applies CNNs to the image segmentation problem. Modern convolutional neural networks for image classification combine multi-scale contextual information by consecutive pooling and sub-sampling layers that lower the resolution. However, semantic segmentation requires multi-scale contextual reasoning together with full-resolution predictions, \ie dense predictions.

Several methods \citep{Chen2015ICLR, Yu2016ICLR, Ghiasi2016ECCV, Badrinarayanan2017PAMI} have therefore been proposed to tackle the opposing needs of multi-scale inference and full-resolution outputs. Dilated convolutions \citep{Chen2015ICLR, Yu2016ICLR} enlarge the receptive field of neural networks without loss of resolution. The dilated convolution operation corresponds to a regular convolution that skips pixels while applying the filter. This allows for efficient multi-scale reasoning without increasing the number of model parameters. \citet{Chen2018PAMI} extend this idea by using multiple dilated convolutions with different sampling rates in parallel.

In contrast, \citet{Badrinarayanan2017PAMI} propose an encoder-decoder network with skip connections. Each decoder layer maps a low resolution feature map of an encoder (max-pooling) layer to a higher resolution feature map. In particular, the decoder in their model takes advantage of the pooling indices computed in the max-pooling (\ie downsampling) step of the corresponding encoder to implement the upsampling process. This eliminates the need to learn the upsampling and thus results in a smaller number of parameters. Furthermore, sharper segmentation boundaries can be obtained using this approach.

While activation maps at lower-levels of the CNN hierarchy lack information specific to object categories, they provide information of higher spatial resolution. \citet{Ghiasi2016ECCV} leverage this assumption and propose to construct a Laplacian pyramid based on a fully convolutional network. Aggregating information at multiple scales allows them to successively refine the boundary reconstructed from lower-resolution layers. They achieve this by using skip connections from higher resolution feature maps and multiplicative confidence gating, penalizing noisy high-resolution outputs in regions where low-resolution predictions have high confidence. 

\boldparagraph{Combining CNNs and CRFs}
A different way to address the needs of multi-scale inference and full resolution prediction is the combination of CNNs with CRF models. \citet{Chen2015ICLR, Chen2018PAMI} propose to refine the label map obtained using a convolutional neural network using a fully connected CRF model \citep{Kraehenbuehl2011NIPS}. The CRF allows them to capture fine details based on the raw RGB input which are missing in the CNN output due to the limited spatial accuracy of the CNN model. In a similar spirit, \citet{Jampani2016CVPR} generalize bilateral filters and backpropagate through the CRF inference \citep{Li2014ICMLWORK} which allows for end-to-end training of the (generalized) filter parameters from data. This effectively allows for reasoning over larger spatial regions within one convolutional layer by leveraging input features as a guiding signal. 

Inspired by higher-order CRFs for semantic segmentation, \citet{Gadde2016ECCV} propose a new Bilateral Inception module for CNN architectures as an alternative to structured CNNs and CRF techniques. They use the assumption that pixels which are spatially and photometrically similar are more likely to have the same label. This allows them to directly learn long-range interactions, thereby removing the need for post-processing using CRF models. Specifically, the proposed modules propagate edge-aware information between distant pixels based on their spatial and color similarity, incorporating the spatial layout of superpixels. Propagation of information is achieved by applying bilateral filters with Gaussian kernels at various scales.

\begin{figure}[t]
\centering
\includegraphics[width=0.60\columnwidth]{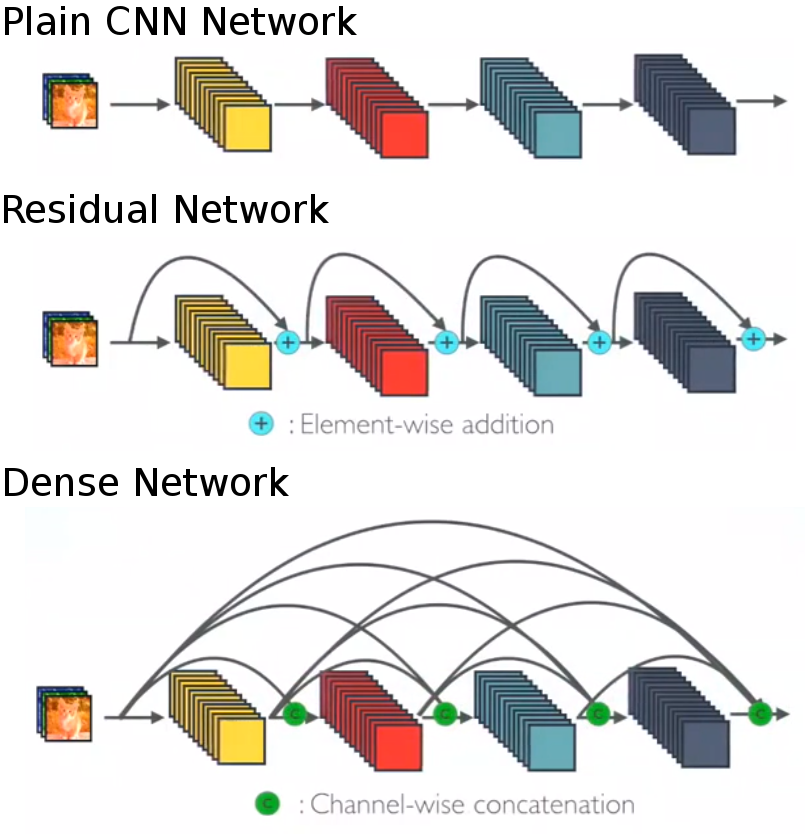}
\caption[Deeper Convolutional Neural Networks]{\textbf{Deep Convolutional Neural Networks.} Comparison of plain, Residual\protect\citep{He2016CVPR} and Dense\protect\citep{Huang2017CVPRb} convolutional neural networks. \figsource{\protect\citet{Huang2017CVPRPORAL}}.}
\label{fig:DeepArchitectures}
\end{figure}

\boldparagraph{Deeper CNNs} 
\citet{Simonyan2015ICLR} and \citet{Szegedy2015CVPR} have shown that the depth of a CNN is crucial to represent rich features. However, increasing the depth of a network leads to an increase in complexity as well as to saturation and degradation in accuracy. \citet{He2016CVPR} proposed the deep residual learning framework (ResNet) illustrated in \figref{fig:DeepArchitectures} to address this problem. In deep residual networks, each stacked layer learns a residual mapping instead of the original mapping. This facilitates the backpropagation of gradients and thus training and results in higher accuracy in comparison to regular deep networks.
\citet{Pohlen2017CVPR} present a ResNet-like architecture which preserves high-resolution information throughout the entire network by combining two different processing streams. One stream passes through a sequence of convolution and pooling layers, whereas the other stream processes feature maps at full image resolution by adding successive residuals from the other stream. Both processing streams are connected using full resolution residual units.

\begin{figure}[t]
	\centering
	\includegraphics[width=1.00\columnwidth]{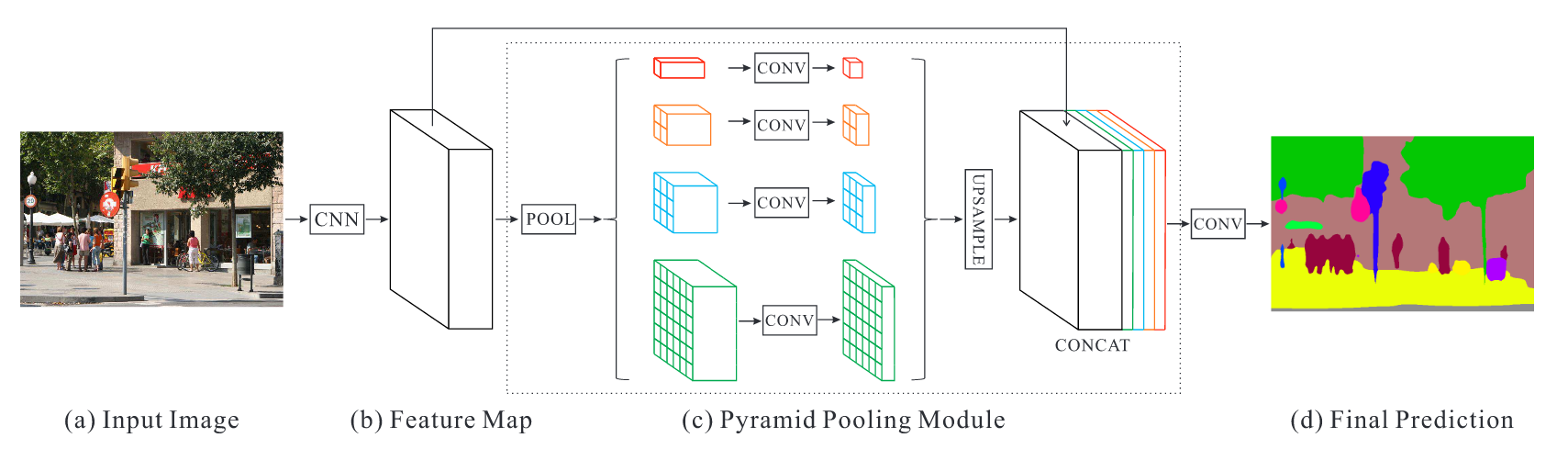}
	\caption[Pyramid Pooling Module]{\textbf{Pyramid Pooling Module.} Overview of the method proposed by \protect\citet{Zhao2017CVPR}. The pyramid parsing module (c) is applied to the output of a CNN feature map (b) and fed into a convolutional layer for per-pixel estimation of semantic class labels (d). \figsourceC{\protect\citet{Zhao2017CVPR}}{2017}{IEEE}.}
	\label{fig:Zhao2017CVPR}
\end{figure}

\citet{Wu2019PR} propose a more efficient ResNet architecture by analyzing the effective depth of residual units. They point out that ResNets behave as linear ensembles of shallow networks. Based on this understanding, they design a group of relatively shallow convolutional networks for the task of semantic image segmentation, which performs better.
To better incorporate global context information into the pixel-level prediction task, \citet{Zhao2017CVPR} propose a pyramid scene parsing network (PSPNet), illustrated in \figref{fig:Zhao2017CVPR}. They apply a pyramid parsing module to the last convolutional layer of a CNN which fuses features of several pyramid scales to combine local and global context information. The resulting representation is fed into a convolution layer to obtain the final per-pixel predictions. Inspired by this work, \citep{Chen2017ARXIV} revisited the Atrous Spatial Pyramid Pooling (ASPP) \citep{Chen2018PAMI} by experimenting with cascading and parallel application of dilated convolutions. This allows them to improve upon their previous work \citep{Chen2018PAMI} while achieving comparable results to PSPNet \citep{Zhao2017CVPR}.

Motivated by deeper architectures like ResNet, \citet{Huang2017CVPRb} propose dense convolutional networks that connect a layer with all preceding layers by concatenation. This allows maximal information throughput from lower to higher levels. In \figref{fig:DeepArchitectures}, plain, residual, and dense architectures are illustrated.  \citet{Jegou2017CVPRWORK} extend dense CNNs to the semantic segmentation problem by constructing a downsampling and upsampling path using dense modules and connecting them with skip connections \citep{Ronneberger2015MICCAI}.

\subsection{Videos} 
In robotic applications such as autonomous driving we usually have access to videos rather than single image frames. The temporal correlation between adjacent frames can be exploited to improve segmentation accuracy, efficiency and robustness. The scene usually changes only slightly between two adjacent frames. Thus, given correspondences between two frames, semantic labels can be propagated in time or corrected using temporal information. 

\citet{Floros2012CVPR} propose a graphical model for semantic segmentation operating on video sequences in order to enforce temporal consistency between frames. Specifically, they present a CRF where temporal consistency between consecutive video frames is ensured by linking corresponding image pixels to the inferred 3D scene points obtained by Structure-from-Motion (SfM). Compared to an image-only baseline, they achieve improved segmentation performance and observe better generalization to varying image conditions. 3D reconstruction works relatively well for static scenes but is still an open problem in dynamic scenes. The presence of both camera and object motion makes temporal association in videos a challenging task. In case of significant motion, Euclidean distance in the space-time volume is not a good measure for finding correspondences. In order to tackle this problem, \citet{Kundu2016CVPR} propose a method for optimizing the feature space of a dense CRF for spatio-temporal regularization. Specifically, the feature space is optimized such that distances between features associated with corresponding points are minimized using correspondences from optical flow. The resulting mapping is exploited by the CRF to achieve long-range regularization over the entire space-time volume.

\boldparagraph{Label Propagation}
Another way to explore temporal correlations in videos for semantic segmentation is label propagation. 
Creating large scale image datasets with highly accurate pixel-level annotations is labor-intensive, and thus obtaining the desired degree of quality is very expensive. Semi-supervised methods for annotating video sequences can help to reduce this cost. Compared to annotating individual images, video sequences offer the advantage of temporal consistency between consecutive frames. 
Label propagation techniques take advantage of this fact by propagating annotations from a small set of annotated keyframes to all unlabeled frames of the video by exploiting color and motion information.

Towards this goal, \citet{Badrinarayanan2010CVPR} propose a coupled Bayesian network which employs a propagation scheme based on correspondences obtained from patch-based similarities and semantically consistent regions. This allows them to transfer label information to unlabeled frames between annotated keyframes. \citet{Budvytis2010BMVC} extend this approach by proposing a hybrid model of the generative propagation introduced in \citep{Badrinarayanan2010CVPR} as well as a discriminative classification stage which tackles occlusions and disocclusions, and allows to propagate over larger time intervals. To correct erroneously propagated labels, \citet{Badrinarayanan2014IJCV} propose a superpixel based mixture-of-tree model for temporal correlation where each component of the mixture contains a tree-structured temporal linkage between superpixels of different frames. \citet{Vijayanarasimhan2012ECCV} tackle the problem of selecting the most promising keyframes for manual labeling such that the expected propagation error is minimized. 

While the aforementioned methods transfer annotations in 2D, \citet{Chen2014CVPRb,Xie2016CVPR} propose to annotate directly in 3D and then transfer these annotations into the image domain. Given 3D information (e.g., from stereo or LiDAR), these approaches are able to produce time coherent semantic labels with limited annotation costs. Towards this goal, \citet{Chen2014CVPRb} use annotations from KITTI \citep{Geiger2013IJRR} and leverage 3D CAD models of cars to infer separate figure-ground segmentations for all cars in the image. In contrast, \citet{Xie2016CVPR} reason jointly about all objects in the scene by also handling categories for which CAD models or 3D point measurements are not available. To this end, they propose a non-local CRF model which reasons jointly about semantic and instance labels of all 3D points and pixels in the image.

\boldparagraph{Scene Understanding}
Scene understanding approaches such as \citep{Ess2009BMVC, Geiger2014PAMI} discussed in \chpref{chap:scene_understanding} exploit semantic segmentation as a cue for reasoning about road topologies and traffic participants. While \citet{Ess2009BMVC} use semantic information to classify a scene into different road topologies based on a short video sequence, \citet{Geiger2014PAMI} formulate a probabilistic model which explains semantic segmentation together with vehicle trajectories, vanishing points, scene flow and occupancy information. However, both approaches do not leverage temporal correlations to improve the semantic segmentation itself.

\subsection{Street Side Views}
One specific application scenario of semantic segmentation which has important applications for autonomous vehicles is the segmentation of street-side images (\ie building facades) into their components (wall, door, window, vegetation, balcony, store, mailbox, \etc). Such semantic segmentations are useful for accurate 3D reconstruction \cite{Haene2013CVPR,Haene2014CVPR,Cherabier2018ECCV}, memory-efficient 3D mapping, robust localization \cite{Schoenberger2018CVPR} as well as path planning. As an example, in 3D reconstruction applications such side information allows for ignoring vegetation that is difficult to model and will change over time.

\begin{figure}[t]
	\centering
	\includegraphics[width=1.00\columnwidth]{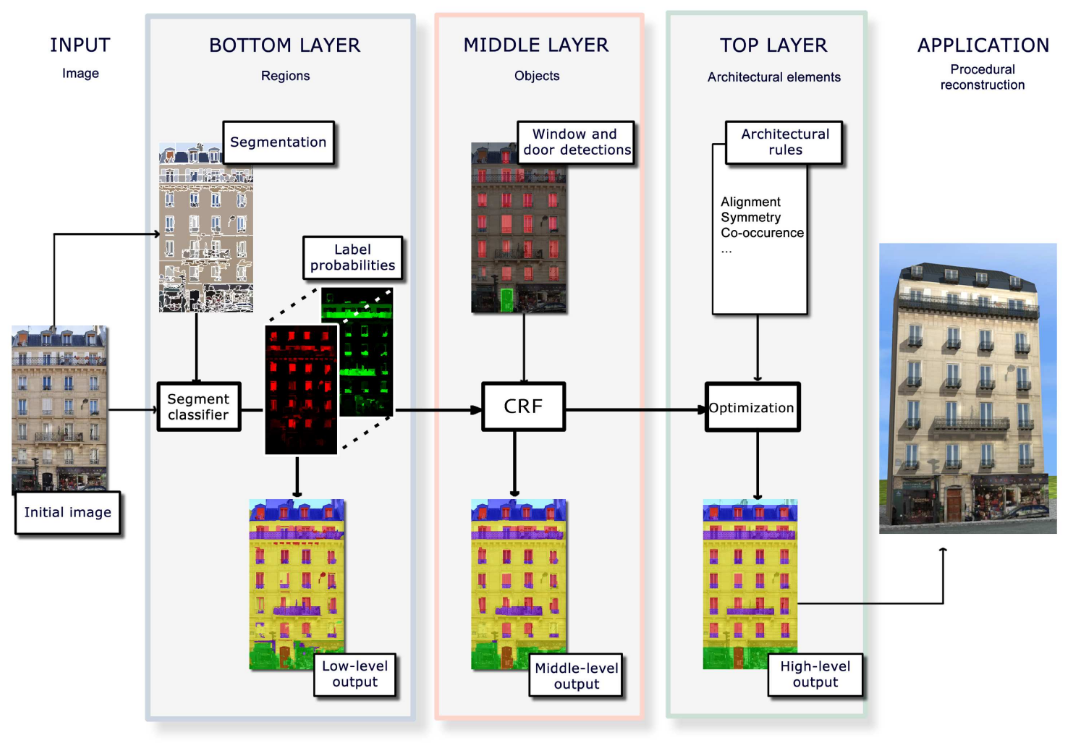}
	\caption[Three-layered Approach for Facade Parsing.]{\textbf{Facade Parsing.} The three-layered approach proposed by \protect\citet{Mathias2016IJCV} for facade parsing. They first segment the facade and assign probability distributions to semantic classes considering extracted visual features. In the next layer they use detectors for specific objects such as doors and windows to improve the classifier output. Finally, they incorporate weak architectural priors and search for the optimal facade labeling using a sampling-based approach. \figsourceSpringer{\protect\citet{Mathias2016IJCV}}{2016}{IJCV}.}
	\label{fig:Mathias2016IJCV}
\end{figure}
\citet{Xiao2009ICCV} propose a multi-view semantic segmentation framework for images captured by a camera mounted on a car driving along the street. Specifically, they define a pairwise MRF across superpixels in multiple views, where the unary terms are based on 2D and 3D features. Furthermore, they minimize color differences for spatial smoothness and use dense correspondences to enforce smoothness across different views. \citet{Xiao2009SIGGRAPH} go one step further and generate photo-realistic 3D models from images captured at ground level. In particular, they segment each image into semantically meaningful areas, such as building, sky, ground, vegetation or car. Then, they partition buildings into independent blocks exploiting architectural priors for inference. This allows them to cope with noisy and missing 3D data and produces visually compelling results. While \citet{Xiao2009ICCV, Xiao2009SIGGRAPH} represent facades with planes or simple geometric primitives, \citet{Mathias2016IJCV} propose a more flexible 3-layered method for segmentation of building facades. First, the facade is segmented into semantic classes which are combined with the output of detectors for architectural elements such as windows and door. Finally, weak architectural priors such as alignment, symmetry and co-occurrence are exploited to encourage the reconstruction to be architecturally consistent. The complete pipeline is illustrated in \figref{fig:Mathias2016IJCV}.

\subsection{3D Data}
While the problem of semantic object labeling has been studied extensively, most of these algorithms work in the 2D image domain where each pixel in the image is labeled with a semantic category such as car, road or pavement. However, 2D images lack important information such as the 3D shape and scale of objects, which are strong cues for object class segmentation and facilitate the detection and separation of individual object instances. 
Furthermore, semantic segmentation of 3D data enables autonomous systems to recognize their surroundings, identify and interact with objects of interest in physical 3D space.

\begin{figure}[t]
	\centering
	\includegraphics[width=1.00\columnwidth]{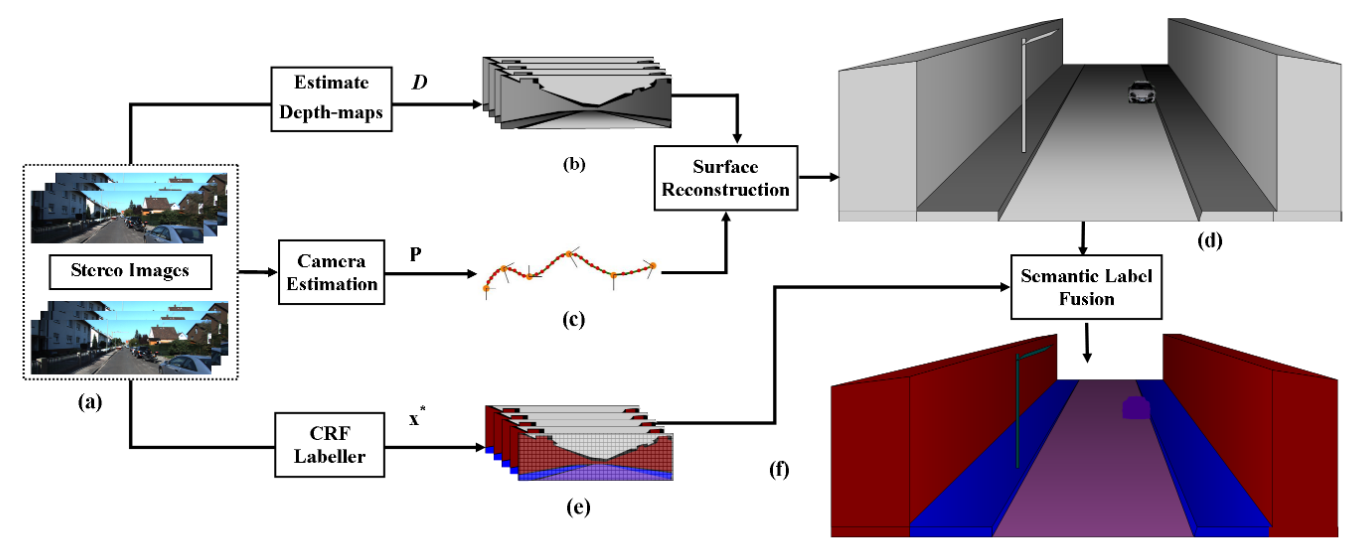}
	\caption[Semantic Segmentation of 3D Data]{\textbf{Semantic Segmentation of 3D Data.} From a stereo image pair (a) \protect\citet{Sengupta2013ICRA} compute the disparity map (b) and track the camera motion (c). They use both outputs to obtain a volumetric representation (d) and fuse the semantic segmentation of street images (e) into a 3D semantic model of the scene (f). \figsourceC{\protect\citet{Sengupta2013ICRA}}{2013}{IEEE}.}
	\label{fig:Sengupta2013ICRA}
\end{figure}

The problem of 3D semantic segmentation has been addressed using different input modalities, \ie monocular image sequences \citep{Martinovic2015CVPR}, stereo image sequences \citep{Valentin2013CVPR, Sengupta2013ICRA} or 3D point clouds \citep{Xiong2011ICRA, Hu2013ICRA, Hackel2016APRS, Gadde2018PAMI}. While \citet{Martinovic2015CVPR, Valentin2013CVPR} use multi-view reconstruction approaches which we discuss in \chpref{chap:mv_reconstruction} for estimating the 3D structure of the scene from monocular image sequences, \citet{Gadde2018PAMI, Hackel2016APRS} directly work with 3D point clouds, \eg from LiDAR.
\citet{Sengupta2013ICRA} propose to project 2D semantic segmentation into a 3D model obtained from depth map fusion using ego-motion estimation from visual odometry as illustrated in \figref{fig:Sengupta2013ICRA}. In parallel, the input images are semantically labeled using a CRF model. The results of this segmentation are then aggregated across the sequence to generate the final 3D semantic model.

\begin{figure}[t]
	\centering
	\includegraphics[width=1.00\columnwidth]{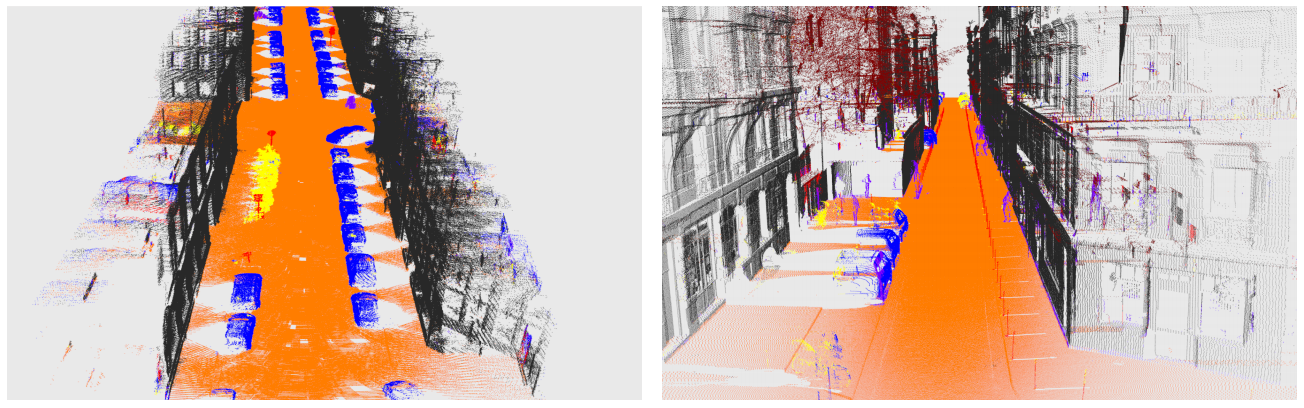}
	\caption[3D Semantic Segmentation]{\textbf{3D Semantic Segmentation.} Semantic segmentation of two 3D scenes using the method of \protect\citet{Hackel2016APRS} with facades (gray), ground (orange), cars (blue), motorcycles (yellow), traffic signs (red), pedestrians (violet) and vegetation (bordeaux). \figsourceC{\protect\citet{Hackel2016APRS}}{2016}{ISPRS}.}
	\label{fig:Hackel2016APRS}
\end{figure}

Several approaches \citep{Hu2013ICRA, Valentin2013CVPR, Martinovic2015CVPR, Gadde2018PAMI, Hackel2016APRS} tackle the problem of semantic scene reconstruction directly in 3D space as shown in \figref{fig:Hackel2016APRS}. 
\citet{Valentin2013CVPR} apply a cascaded classifier to learn geometric cues from the mesh and appearance cues from images.
In contrast, \citet{Martinovic2015CVPR} avoid time-consuming conversions between 2D and 3D representations by training Random Forest classifiers on 3D features. Afterwards, they separate individual facades based on their semantic structure and impose weak architectural priors. Instead of imposing architectural priors, \citet{Gadde2018PAMI} implement a sequence of boosted decision tree classifiers, stacked using auto-context features. They demonstrate that the system is fast at inference time and easily adapts to new datasets. \citet{Hackel2016APRS} propose a fast semantic segmentation approach for large 3D point clouds, which can also handle strongly varying densities. They construct approximate multi-scale neighborhoods by down-sampling the point cloud in order to generate a pyramid with decreasing density. This scheme allows extracting a rich feature representation that captures the geometry in a point's local neighborhood such as roughness, surface orientation, and height over ground. A random forest classifier finally predicts class-conditional probabilities. 

Image-based 3D semantic segmentation approaches like \citep{Sengupta2013ICRA} lead to redundant computations due to the overlap of images used for reconstruction of the 3D model. Therefore, approaches directly working in the 3D space are usually more efficient. \citet{Riemenschneider2014ECCV} exploit the inherent redundancy in the labeling of all overlapping images to further increase the efficiency of image-based 3D semantic segmentation. They propose an approach that exploits the geometry of a 3D mesh model obtained from multi-view stereo to predict the best view for each face of the mesh before inferring the semantic class label. This allows them to accelerate their pipeline by two orders of magnitude, however, with lower accuracy than \citet{Martinovic2015CVPR}.

\boldparagraph{Online Methods}
While all aforementioned methods work in batch mode, \ie they process all data at once, online methods allow the flexible incorporation of new measurements. This is particularly useful in the context of autonomous driving where new data arrives continuously.
Towards online 3D semantic segmentation, \citet{Xiong2011ICRA} train a sequence of classifiers to make predictions on different scales in a coarse-to-fine fashion (from regions to points). Predictions from the preceding scale are used as additional information for the current scale. They extend this work in \citep{Hu2013ICRA} with a hierarchical representation of the 3D data and an improved inference procedure. \citet{Vineet2015ICRA} propose an end-to-end system which processes data incrementally while performing real-time dense stereo reconstruction and semantic segmentation of outdoor environments. They achieve this using voxel hashing \citep{Niesner2013SIGGRAPH}, a hash-table-driven 3D volumetric representation that ignores unoccupied space in the target environment. Furthermore, they employ an online volumetric mean-field inference technique that incrementally refines the voxel labeling and achieve real-time rates by harnessing the processing power of modern GPUs. \citet{McCormac2017ICRA} present a pipeline for dense 3D semantic mapping designed to work online by fusing semantic predictions of a CNN with the geometric information from a SLAM system (ElasticFusion by \citet{Whelan2015RSS}). Specifically, ElasticFusion provides correspondences between 2D frames and a globally consistent map of surface elements or ``surfels''. Furthermore, they use a Bayesian update scheme which computes the class probabilities for each surfel based on the CNN's predictions.

\boldparagraph{3D CNN}
While convolutional networks have proven very successful in segmenting 2D images semantically, there exists comparably little work on labeling 3D data using convolutional networks. \citet{Maturana2015IROS} were one of the first to apply 3D Convolutional Neural Network (3D-CNN) for object recognition of volumetric 3D data. Their VoxNet approach classifies $32^3$ voxel volumes using a convolutional neural network. In contrast, \citet{Huang2016ICPR} propose a framework to directly label 3D point cloud data using a 3D-CNN. Specifically, they compute 3D occupancy grids of size $20^3$ centered at a set of randomly generated keypoints. The occupancy and the labels form the input to a 3D CNN which is composed of convolutional layers, max-pooling layers, a fully connected layer and a logistic regression layer. Towards processing larger volumes, \citet{Riegler2017CVPR} propose OctNets, a 3D convolutional network, that allows for training deep architectures at significantly higher resolutions. They build on the observation that 3D data (\eg point clouds, meshes) is often sparse in nature. OctNet exploits this sparsity property by hierarchically partitioning the 3D space into a set of octrees and applying pooling in a data-adaptive fashion. This leads to a reduction in computation and memory requirements as the convolutional network operations are defined on the structure of these trees. Thus, resources can be allocated dynamically depending on the structure of the input.

\subsection{Road Segmentation} 
Segmentation of road scenes is a crucial problem in computer vision for autonomous driving. For instance, in order to navigate, an autonomous vehicle needs to determine the drivable area ahead and determine its own position on the road with respect to the lane markings. However, the problem is challenging due to the presence of a variety of differently shaped objects such as cars and people, different road types and varying illumination and weather conditions.
Traditionally, the problem of autonomous driving has been tackled by detecting lane markings \citep{Borkar2012TITS, Wu2012IV, Lee2017ICCV, Li2017NNLS}.
However, as lane marking features are often not reliable (bad weather, construction sites, missing lane markings), more holistic approaches which consider the entire road area have been explored lately.

\citet{Alvarez2010CVPR} propose a Bayesian framework to classify road sequences by combining low-level appearance cues with contextual 3D road cues such as the horizon, vanishing points, the 3D scene layout and 3D road models.
In addition, they extract temporal cues for temporally smoothing the results.
In follow-up work, \citet{Alvarez2011TITS} convert the image into an illumination invariant feature space to make their method robust to shadows. \citet{Mansinghka2013NIPS} propose an inverse-graphics inspired method by employing generative probabilistic graphics programs  (GPGP) to infer roads in images taken from vehicle-mounted cameras. GPGPs consist of a stochastic scene generator for generating random samples from a road scene prior, a graphics renderer for rendering the image segmentation of each sample and a stochastic likelihood model linking the renderer's output and the data. 
\citet{Kuehnl2012IV} present a method to improve appearance-based classification by incorporating the spatial layout of the scene. Specifically, they suggest a two-stage approach for road segmentation. First, they represent the road surface and delimiting elements such as curbstones and lane-markings using confidence maps based on local visual features. From these confidence maps, they extract SPatial RAY (SPRAY) features that incorporate global properties of the scene and train a classifier on those features. Their evaluation shows that spatial layout helps especially in cases where there is a clear structural correspondence between properties at different spatial locations.

\boldparagraph{Deep Learning}
Recently, the problem of road segmentation has been addressed using convolutional neural networks \citep{Moh2014ARXIV, Oliveira2016IROS}.
\citet{Moh2014ARXIV} proposes a scene parsing system by using deconvolutional networks \citep{Zeiler2011ICCV} in combination with traditional CNNs for feature learning. Deconvolutional networks learn features that capture mid-level cues such as edge intersections, parallelism and symmetry in image data and thus obtain a more robust representation. \citet{Oliveira2016IROS} investigate the trade-off between segmentation quality and runtime using U-Nets \citep{Ronneberger2015MICCAI}. Specifically, they introduce a new mapping between classes and filters at the up-convolutional part of the network for reducing runtime. They further segment the entire image with a single forward pass, resulting in a more efficient approach compared to patch-based ones \citep{Moh2014ARXIV}.
However, as road segmentation is a subproblem of semantic segmentation, today most state-of-the-art results on road segmentation are achieved using generic off-the-shelf semantic segmentation networks.

\boldparagraph{Data Acquisition}
All existing algorithms for labeling road scenes are based on machine learning where the parameters of the respective model must be estimated from large annotated datasets. To alleviate the burden of annotating large datasets manually, \citet{Alvarez2012ECCV} propose a method for road segmentation where noisy training labels for road images are generated using a convolutional neural network trained on a general image database. \citet{AnkitLaddha2016IV} follow a different approach and obtain ground truth labels by exploiting OpenStreetMap information projected into the image domain using the vehicle pose provided by the GPS sensor.

\begin{figure}[t]
	\centering
	\includegraphics[width=0.80\columnwidth]{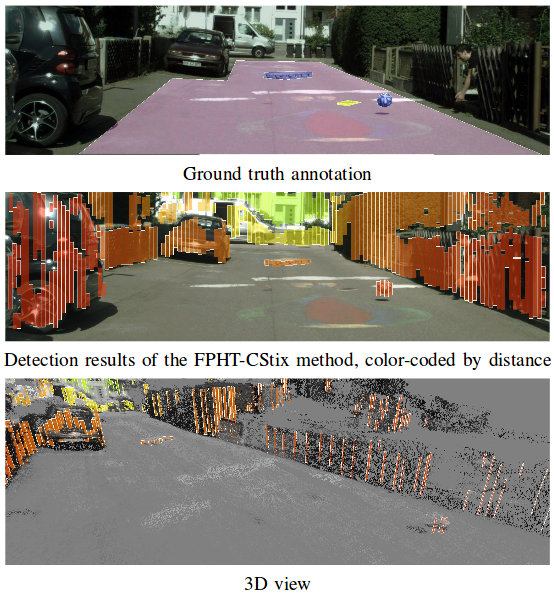}
	\caption[Free Space Estimation]{\textbf{Free Space Estimation.} Free space and detected obstacles on the Lost and Found dataset\protect\citep{Pinggera2016IROS}. \figsourceC{\protect\citet{Pinggera2016IROS}}{2016}{IEEE}}
	\label{fig:Pinggera2016IROS}
\end{figure}

\subsection{Free Space Estimation}
Accurate and reliable estimation of free space and the detection of obstacles are core problems that need to be solved for enabling autonomous driving. Free space is defined as the available space on the ground surface where navigation of vehicle is guaranteed without collision. Obstacles refer to structures that block the path of the vehicle by sticking out of the ground surface. In contrast to road segmentation approaches, methods estimating free-space in front of a vehicle often rely on geometric features which can be derived from a depth map computed from stereo sensors. However, both complementary approaches can be advantageously combined.

\citet{Badino2007ICCVWORK} propose a method for free space estimation by computing stochastic occupancy grids based on stereo information, where cells in a stochastic occupancy grid carry information about the likelihood of occupancy. Stereo information is integrated over time in order to reduce depth uncertainty. The boundary between free space and occupied space is robustly obtained using dynamic programming on the occupancy grid. This work laid the foundations for the Stixel representation, see \secref{sec:Stixels} for an in-depth discussion. While the original method of \citet{Badino2007ICCVWORK} makes the assumption of a planar road surface, this assumption is often violated in practice. In order to tackle more complicated road surfaces, \citet{Wedel2009TITS} propose an algorithm which models non-planar road surfaces using B-splines. The surface parameters are estimated from stereo measurements and tracked over time using a Kalman filter. In contrast, \citet{Suleymanov2016IROS} propose a complete pipeline to detect and drive on collision-free traversable paths, based on stereo information using a variational approach. In addition to free space detection, their approach also establishes a semantic segmentation of the scene, where labels include ground, sky, obstacles and vegetation.

Fisheye cameras discussed in \secref{sec:calibration_omnidirectional_cam} provide a wider field of view compared to regular cameras and allow for the detection of obstacles closer to the car. \citet{Haene2015IROS} propose a method for obstacle detection using monocular fisheye cameras. In order to reduce runtime, they avoid using visual odometry for accurate vehicle poses and instead, rely on less accurate pose estimates from wheel odometry. While they show good accuracy in the estimation of distances between objects, their experiments are limited to objects in close proximity to the sensor.

\boldparagraph{Long Range Obstacle Detection} 
The accuracy of obstacle detection at long-range is crucial for timely obstacle localization when the observer (\ie the ego-vehicle) moves at high speed, \eg in highways. Unfortunately, the error of stereo vision systems increases quadratically with depth in contrast to laser range sensors or radar sensors.
In order to tackle this problem, \citet{Pinggera2015IROS, Pinggera2016IROS} propose long range obstacle detection algorithms using stereo vision. They formulate obstacle detection as a statistical hypothesis test, exploiting geometric constraints on camera motion and planarity. Independent hypothesis tests are performed on small local patches distributed across the input images. Detection results for a scene from their dataset are illustrated in \figref{fig:Pinggera2016IROS}. 

\subsection{Stixels}
\label{sec:Stixels}
Stixels are a compact mid-level representation of 3D traffic scenes with the goal to bridge the gap between pixels and objects \citep{Badino2009DAGM}. The so-called ``Stixel World'' representation originates from the observation that free space in front of the vehicle is mostly limited by vertical surfaces. Stixels are represented by a set of rectangular sticks standing vertically on the ground to approximate these surfaces. Assuming a constant width, each stixel is defined by its height and its 3D position relative to the camera. The main goal of stixels is to gain efficiency through a compact, complete, stable, and robust representation. In addition, the stixel representation provides an encoding of the free space and the obstacles in the scene. 

\begin{figure}[t]
	\centering
	\includegraphics[width=1.00\columnwidth]{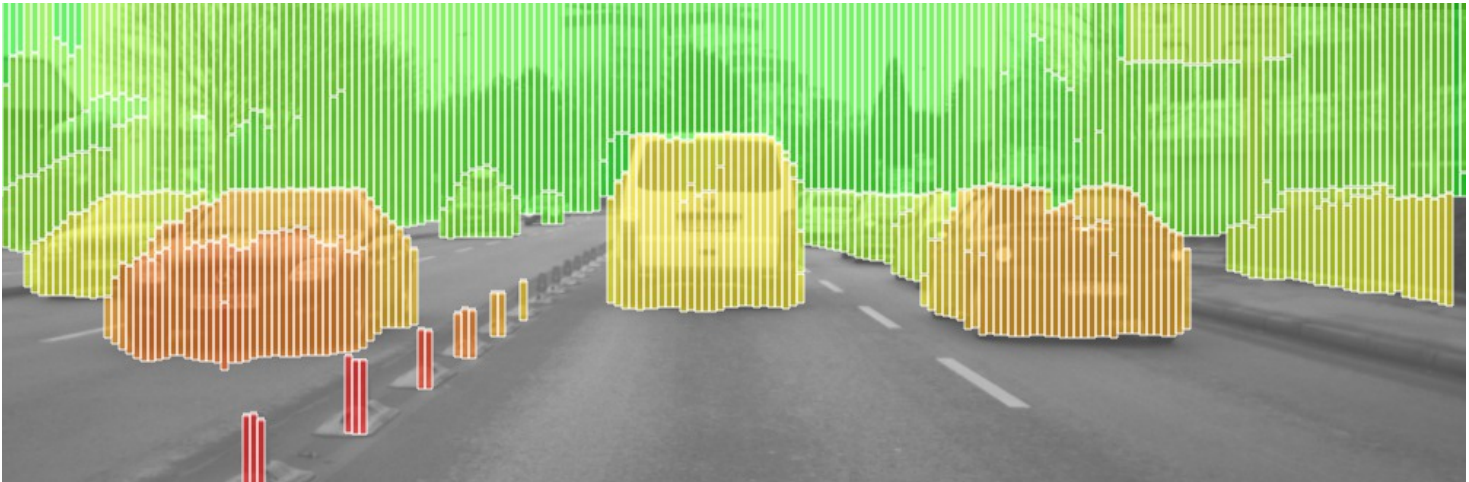}
	\caption[Multi-layer Stixel World]{\textbf{Multi-layer Stixel World.} The multi-layer Stixel World representation of \protect\citet{Pfeiffer2011BMVC}. The scene is segmented into planar segments termed ``Stixels''. In contrast to the Stixel World of \protect\citep{Badino2009DAGM}, objects are allowed to be located at multiple depths within a single image column. The color represents the distance to the obstacle with red being close and green far away. \figsourceC{\protect\citet{Pfeiffer2011BMVC}}{2016}{BMVA}.}
	\label{fig:multilayer_stixel_world}
\end{figure}
Using depth maps from SGM \citep{Hirschmueller2008PAMI} as input, \citet{Badino2009DAGM} use dynamic programming based on occupancy grids to compute free space, determining the stixels' lower positions. \citet{Pfeiffer2011BMVC} extend \citep{Badino2009DAGM} to a unified probabilistic scheme. They furthermore lift the constraint on stixels to touch the ground and allow multiple stixels for each image column, leading to a more flexible representation as illustrated in \figref{fig:multilayer_stixel_world}.

In the dynamic stixel world representation introduced by \citet{Pfeiffer2010IV} the stixel representation was extended to dynamic scenes by tracking stixels using 6D Kalman filters based on optical flow. In contrast, \citet{Guenyel2012ECCV} show that motion estimation for stixels can be reduced to a 1D problem and can be solved efficiently via 2D dynamic programming, avoiding costly dense optical flow computation. Based on the dynamic stixel world representation, \citet{Erbs2012BMVC,Erbs2013IV} present a CRF framework for semantically segmenting traffic scenes. 

Several approaches proposed to leverage high-level information for inferring stixel representations more robustly. \citet{Cordts2014GCPR} incorporate top-down object-level cues into the bottom-up stixel representation using a probabilistic approach. With the success of deep learning, \citet{Schneider2016IV} present a semantic stixel representation to jointly infer the semantic and geometric layout of the scene from a dense disparity map and pixel-level semantic scene labeling. Towards this goal, they used a deep learning-based scene labeling approach. In contrast, \citet{Levi2015BMVC} propose StixelNet to directly infer the foot point of each stixel from the input image.

\subsection{Aerial Images}
\label{sec:semantic_seg_aerial}
The aim of aerial image parsing is the automated extraction of urban objects from data acquired by airborne sensors.  The need for accurate and detailed information for urban objects such as roads is rapidly increasing because of its applications in the navigation of autonomous driving systems. For example, the output of aerial image parsing can be used to automatically build road maps (even in remote areas) and keep them up-to-date. Furthermore, information from aerial images can be used for localization. However, the problem is challenging because of the heterogeneous appearance of objects like buildings, streets, trees and cars which results in high intra-class variance but low inter-class variance. 
Furthermore, the complex structure of road networks and the difficulty of representing their geometry and topology accurately makes this problem hard. Roads must form a connected network of thin segments with slowly changing curvatures which meet at junctions. This type of prior knowledge is more challenging to formalize and integrate into a structured prediction formulation than standard smoothness assumptions.

\boldparagraph{Graphical Models}
Graphical models have been a very popular way of addressing the problem of semantic segmentation in aerial images \citep{Wegner2013CVPR, Wegner2015JPRS, Montoya2015CPIA, Verdie2014IJCV, Mattyus2015ICCV, Mattyus2016CVPR, Wegner2016CVPR}.
\citet{Wegner2013CVPR} propose a CRF formulation for road labeling in which the prior is represented by cliques that connect sets of superpixels along straight line segments. Specifically, they formulate the constraints as high-order cliques with asymmetric $P^N$-potentials which express a preference to assign all rather than just a few of their constituent superpixels to the road class. This allows the road likelihood to be amplified for thin chains while still being amenable to efficient inference using graph cuts.
\citet{Wegner2015JPRS} also model the road network using a CRF with long-range, higher-order cliques. However, unlike \citep{Wegner2013CVPR}, they allow for arbitrarily shaped segments which adapt to more complex road shapes by searching for putative roads with minimum cost paths based on local features.
\citet{Montoya2015CPIA} extend this formulation to multi-label classification of aerial images with class-specific
priors for buildings and roads. In addition to the road network prior of \citep{Wegner2015JPRS}, they introduce a second higher-order potential for cliques specific to buildings. In contrast, \citet{Verdie2014IJCV} propose the application of Markov point processes for recovering specific structures from images, including road networks.
Markov point processes are a generalization of traditional MRFs which can address object recognition problems by directly manipulating parametric entities such as line segments. 
Importantly, they implicitly solve the model-selection problem, \ie they allow for an arbitrary number of variables in the MRF which can be associated with the parameters of the objects of interest.

\boldparagraph{Aerial Image Parsing using Maps}
Instead of framing the problem of detecting topologically correct road networks as a semantic segmentation problem, \citet{Mattyus2015ICCV} exploit map information from the free and community-driven mapping project OpenStreetMap (OSM)\footnote{\url{https://www.openstreetmap.org/}}. 
Given a road map from OSM, \citet{Mattyus2015ICCV} propose an MRF which reasons about the location of the road centerline and its width for each road segment in OSM. In addition, they incorporate smoothness between consecutive line segments by encouraging their widths to be similar. This formulation has the advantage of being efficient at inference time due to the restriction of the road topology to the input maps.
However, it cannot recover from errors or missing information in the original map.
Very recently, Facebook has announced a new set of tools\footnote{\url{https://mapwith.ai/}} that leverage AI to help the OSM community to build maps more efficiently.

\boldparagraph{Fine-grained Image Parsing}
While aerial images provide full coverage of a significant portion of the world, they are of much lower resolution than ground images. In aerial imagery, the resolution relates to the ground area covered by one pixel. Whereas 1 meter resolution is already a high resolution for satellite imagery, the standard resolution for most publicly accessible image databases (\eg Google Earth\footnote{\url{https://www.google.com/earth/}}) is 0.30 meter. 
Resolutions of 0.15 to 0.03 meter are considered high resolutions for aerial imagery and are usually not made publicly available. 
This makes fine-grained segmentation from aerial images a challenging problem.
In contrast, ground images provide additional information which enables fine-grained semantic segmentation. 
Motivated by the complementary nature of these cues, several methods \citep{Mattyus2016CVPR, Wegner2016CVPR} for fine-grained segmentation have been recently proposed which jointly reason about co-located aerial and ground image pairs.

\citet{Mattyus2016CVPR} extend their approach \citep{Mattyus2015ICCV} by introducing a formulation that reasons about fine-grained road semantics such as lanes and sidewalks. To infer this information, they jointly consider monocular aerial images and high resolution stereo images captured from ground vehicles.
Specifically, they formulate the problem as energy minimization in a MRF, inferring the number and location of the lanes for each road segment, all parking spots and sidewalks as well as the alignment between the ground and aerial images. Towards this goal, they exploit deep learning to estimate semantics from aerial and ground images and define potentials exploiting both cues. 
\citet{Wegner2016CVPR} build a map of trees for urban planning applications from aerial images, street view images and semantic map data. They train CNN-based object detection algorithms on human-annotated data.

\begin{table*}[t]
	\centering
	\begin{adjustbox}{width=1\textwidth}\begin{tabular}{l l|c|c|c|c|c|c}
& \multirow{2}{*}{\textbf{Method}} & \multirow{2}{*}{\textbf{Coarse}} & \multirow{2}{*}{\textbf{Depth}} & \multirow{2}{*}{\textbf{IoU class}} & \multirow{2}{*}{\textbf{iIoU class}} & \textbf{IoU} & \textbf{iIoU} \\
 & & & & & & \textbf{category} & \textbf{category}  \\ \hline
1. & DRN\_CRL\_Coarse \citep{Zhuang2018ICIP} & \cmark &  & 82.8 & 61.1 & 91.8 & 80.7 \\
2. & DPC \citep{Chen2018NIPS} & \cmark &  & 82.7 & 63.3 & 92.0 & 82.5 \\
3. & RelationNet\_Coarse \citep{Zhuang2018ICPR} & \cmark &  & 82.4 & 61.9 & 91.8 & 81.4 \\
4. & SSMA \citep{Valada2018ARXIV} & \cmark & \cmark & 82.3 & 62.3 & 91.5 & 81.7 \\
5. & GFF-Net \citep{Li2019ARXIV} &  &  & 82.3 & 62.1 & 92.0 & 81.4 \\
\hline
10. & DeepLabv3 \citep{Chen2017ARXIV} & \cmark &  & 81.3 & 62.1 & 91.6 & 81.7 \\
11. & AdapNet++ \citep{Valada2018ARXIV} & \cmark &  & 81.3 & 59.5 & 91.0 & 80.1 \\
12. & PSPNet \citep{Zhao2017CVPR} & \cmark &  & 81.2 & 59.6 & 91.2 & 79.2 \\
14. & ResNet-38 \citep{Wu2019PR} & \cmark &  & 80.6 & 57.8 & 91.0 & 79.1 \\
\end{tabular}
\end{adjustbox}
	\caption{{\bf CITYSCAPES Semantic Segmentation Leaderboard.} Segmentation performance is measured by class intersection-over-union and instance-level intersection-over-union. All methods are trained on the dense dataset consisting of 5000 frames and methods trained on the coarse dataset consisting of additional 20000 frames are marked in the corresponding column. Methods below the horizontal line show older entries for reference. Accessed on: June 2019.}
	\label{tab:cityscapes_pixel_level}
\end{table*}

\section{Datasets}
There exist many large-scale realistic datasets for semantic segmentation as discussed in \chpref{chap:Datasets}. The most popular datasets are PASCAL VOC \citep{Everingham2010IJCV}, Microsoft COCO \citep{Lin2014ECCV} and Cityscapes \citep{Cordts2016CVPR}. Recently, several companies also created new datasets which focus on the autonomous driving scenario such as Mapillary \citep{Neuhold2017ICCV}, ApolloScape \citep{Huang2018CVPR} and Berkeley DeepDrive \citep{Yu2018ARXIV}. 
In addition, there exist several synthetic datasets for semantic segmentation, \eg SYNTHIA \citep{Ros2016CVPR} and Playing for data \citep{Richter2016ECCV}. Here, we focus on the comparison of different semantic segmentation approaches on the popular Cityscapes dataset\footnote{\url{https://www.cityscapes-dataset.com/}} by \citet{Cordts2016CVPR} as it is most relevant to the autonomous driving scenario. Cityscapes provides 5,000 images with high-quality dense annotations and 20,000 additional images with coarse labels obtained using a novel crowdsourcing platform.

In contrast to 2D semantic segmentation, there are only a few datasets that address the 3D semantic segmentation problem. Furthermore, these datasets are either very limited in size \cite{Munoz2009CVPR, Behley2012ICRA, Hackel2017APRS, Zhang2015ICRAa} or in the number of classes \citep{Geiger2013IJRR}. Recently, \citet{Behley2019ARXIV} presented a large-scale dataset for 3D semantic segmentation.

\section{Metrics}
The performance of methods for semantic segmentation is usually evaluated using the intersection-over-union metric (IoU) which is defined as the number of true positive pixels divided by the sum over true positive, false positive and false negative pixels. As classes with larger segments will have a larger effect on the IoU score, Cityscapes \citep{Cordts2016CVPR} also report the instance-level intersection-over-union (iIoU) metric which weights the contribution of each true positive and false negative pixel by the ratio of the average instance size of the respective class with respect to the respective ground truth instance size. Cityscapes \citep{Cordts2016CVPR} report the IoU and iIoU metrics for two semantic granularities, \ie classes and categories.

\section{State of the Art on Cityscapes}
\tabref{tab:cityscapes_pixel_level} shows the leaderboard of Cityscapes for the pixel-level semantic labeling task. All methods are trained on the dense dataset comprising 5,000 densely annotated frames. Methods that are additionally trained on the coarse dataset with additional 20,000 frames are marked in the table. The state of the art in semantic segmentation shows very similar accuracy in terms of IoU and iIoU. \citet{Li2019ARXIV} extend the pyramid scene parsing network (PSPNet) of \citep{Zhao2017CVPR} with an advanced fusion mechanism. They propose Gated Fully Fusion modules which enable for every pixel to fuse only the relevant information from different feature maps. This allows better accuracy on fine-level details than PSPNet\citep{Zhao2017CVPR}. In contrast, \citet{Valada2018ARXIV} present a multi-modal fusion approach that fuses features extracted from images and depth. The feature extraction network is based on the full pre-activation ResNet-50 \citep{He2016ECCV} with multi-scale residual units proposed in \citep{Valada2017ICRA} as well as an efficient variant of Atrous Spatial Pyramid Pooling (ASPP) \citep{Chen2017ARXIV}. \citet{Zhuang2018ICPR} follow a different approach and introduce a Relation Module that correlates features with their spatial neighborhood by shifting the features in four directions (left-right, top-down) using pre-defined offsets while passing them through Gated Recurrent Units. The features are extracted using a ResNet-like architecture \citep{Wu2019PR} modified with dilated and deformable convolutions \citep{Dai2017ICCV}. \citet{Zhuang2018ICIP} extend this idea by exploiting additional offsets to correlate features over larger neighborhoods. This allows them to outperform all other methods on Cityscapes (\tabref{tab:cityscapes_pixel_level}).

Most existing network architectures for semantic segmentation are designed by the developer. Recently, a new line of work proposes to search for novel architectures in a properly defined search space. \citet{Chen2018NIPS} address three dense prediction problems, \ie street scene parsing, person-part segmentation and semantic image segmentation, with an architecture search. They use Xception \citep{Chollet2017CVPR, Dai2017ICCV, Chen2018CVPR} as the backbone network and build a recursive search space from three popular operators, \ie 1x1 convolution, 3x3 atrous convolution and average spatial pyramid pooling. Finally, they adapt a random search algorithm to explore the recursive search space. By evaluating 28K architectures on 370 GPUs, they find an architecture that achieves state-of-the-art performance on Cityscapes. 

\section{Discussion}
The focus on multi-scale inference has led to impressive results in pixel-level semantic segmentation on Cityscapes.
Today, the top methods on Cityscapes (\tabref{tab:cityscapes_instance_level}) reach an impressive IoU of almost $83\%$ over classes and $92\%$ over categories. In contrast, the instance-weighted IoU still ranges around $63\%$ over classes and $82\%$ over categories. This indicates that semantic segmentation works well for instances covering large image areas but is still challenging for instances covering smaller regions which provide less information about the semantic label and require context reasoning. Furthermore, segmenting small, and possibly occluded objects is a challenging task which might benefit from accurate depth estimation. Recently, multi-modal fusion approaches leveraging depth data have shown great performance for indoor \citep{Hazirbas2016ACCV, Dai2018ECCV} and outdoor \citep{Valada2018ARXIV} semantic segmentation, the latter achieving state-of-the-art performance on Cityscapes as discussed in the previous section. Furthermore, exploiting temporal correlations as in \citep{Kundu2016CVPR} has the promise to further improve semantic segmentation accuracy and temporal consistency.
	\chapter{Semantic Instance Segmentation}
\label{chap:instance_segmentation}
\section{Problem Definition}
The goal of semantic instance segmentation is to simultaneously detect, segment and classify every individual object in an image. Unlike semantic segmentation, a solution to this task provides information about the position, semantics, shape, and count of individual objects, and therefore has many applications in autonomous driving. 

\section{Methods}
There exist two major lines of research for the task of semantic instance segmentation: Proposal-based and proposal-free instance segmentation. While proposal-based approaches usually consist of two steps, \ie proposal extraction and proposal classification, proposal-free methods predict pixel labels directly from the image.

\subsection{Proposal-based Approaches} 
\begin{figure}[t]
\centering
\includegraphics[width=1.00\columnwidth]{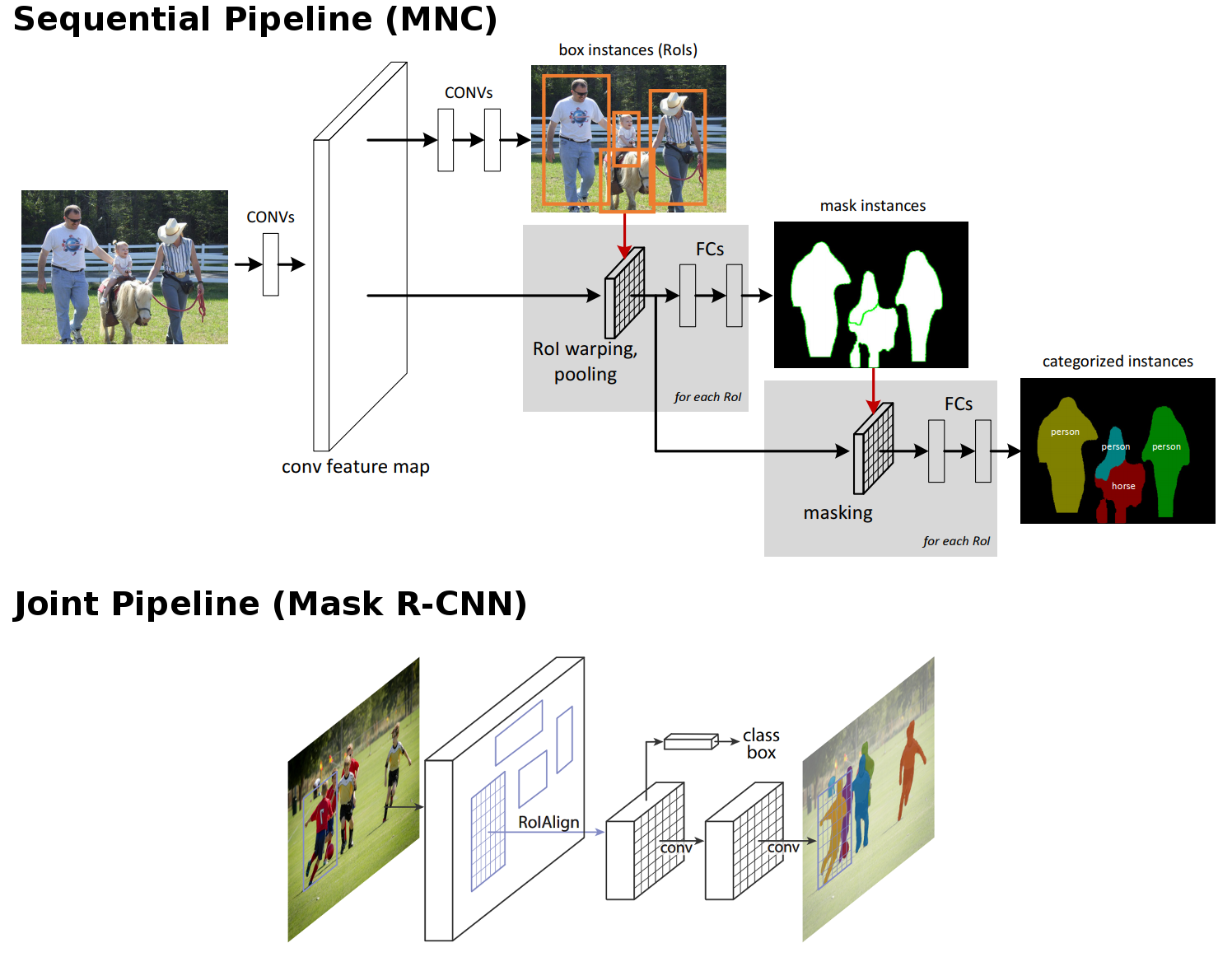}
\caption[Proposal-based Instance Segmentation Networks]{\textbf{Proposal-based Instance Segmentation Networks.} Architectures for proposal-based instance segmentation. Sequential pipelines as MNC \protect\citep{Dai2016CVPR} (upper row) use a detection and segmentation network sequentially. In contrast, joint formulations as Mask R-CNN \protect\citep{He2017ICCV} (bottom row) usually have the two networks in parallel, \ie an object mask prediction and bounding box recognition network. \figsourceC{\protect\citet{Dai2016CVPR, He2017ICCV}}{2016,2017}{IEEE}.}
\label{fig:proposal_instance_networks}
\end{figure}
Proposal-based instance segmentation methods extract class-agnostic proposals which are classified as an instance of a semantic class in order to obtain pixel-level instances. There exist several region proposal methods like Constrained Parametric Min-Cut (CMPC) \citep{Carreira2012PAMI}, Multiscale Combinatorial Grouping (MCG) \citep{Arbelaez2014CVPR}, DeepMask \citep{Pinheiro2015NIPS}, and SharpMask \citep{Pinheiro2016ECCV} returning generic class-agnostic region proposals which can be directly used as instance segments. Several object detection classifiers were proposed which simultaneously address object detection and semantic segmentation by leveraging region features from instance segments to improve the detection accuracy, \ie O$^2$P \citep{Carreira2012ECCV}, Simultaneous Detection, and Segmentation (SDS) \citep{Hariharan2014ECCV}, Convolutional Feature Masking (CFM) \citep{Dai2015CVPR}, HyperColumn \citep{Hariharan2015CVPR}. 

Proposal-based algorithms are slow at inference time due to the computationally expensive proposal generation step. To avoid this bottleneck, \citet{Dai2016CVPR} propose Multi-task Network Cascade (MNC) a fully convolutional network with three stages illustrated in \figref{fig:proposal_instance_networks}. They extract box proposals, use shared features to refine these to segments, and finally classify them into semantic categories. The causal relations between the outputs of the stages complicate training of the multi-task cascade. In order to overcome these difficulties, a fully differentiable mask prediction layer is presented to train the whole model in an end-to-end fashion. Box proposals can also induce errors into the proposal-based instance segmentation method due to wrongly scaled or shifted bounding boxes. In order to tackle this problem, \citet{Hayder2017CVPR} present  a shape aware object mask network that predicts a binary mask for each bounding box proposal, potentially extending beyond the box itself. They integrate the object mask network into the Multi-task Network Cascade framework of \citet{Dai2016CVPR} by replacing the original mask prediction stage. 

While earlier methods address the detection and segmentation problem with two sub-networks sequentially, recent work \citep{Li2017CVPR, He2017ICCV, Chen2018CVPRa} propose to jointly address these problems. We illustrate an example of a sequential and joint formulation in \figref{fig:proposal_instance_networks}.
All joint formulations use ResNet-like architectures \citep{He2016CVPR} for feature extraction. \citet{Li2017CVPR} propose FCIS, the first fully convolutional neural network for end-to-end instance semantic segmentation. They extend the fully convolutional mask proposal network \citep{Dai2016ECCV} by sharing the convolutional representation of the proposals with a detection and segmentation sub-network. In contrast to FCIS, Mask R-CNN \citep{He2017ICCV} and MaskLab \citep{Chen2018CVPRa} both build on Faster R-CNN \citep{Ren2015NIPS}. \citet{He2017ICCV} extend Faster R-CNN \citep{Ren2015NIPS} by an additional branch for predicting segmentation masks. \citet{Chen2018CVPRa} combine box predictions from Faster R-CNN with semantic segmentation logits for pixel-wise classification and direction prediction logits estimating the direction towards instance centers. The direction towards instance centers allows them eventually to separate instances from the same class.

\subsection{Proposal-free Approaches} 
Due to the problem of proposal-based approaches to inherit errors of the proposal generation, a number of alternative methods have been proposed recently.
These methods jointly infer the segmentation and the semantic category of individual instances by casting instance segmentation directly as a pixel labeling task. 

\begin{figure}[t]
	\centering
	\includegraphics[width=1.00\columnwidth]{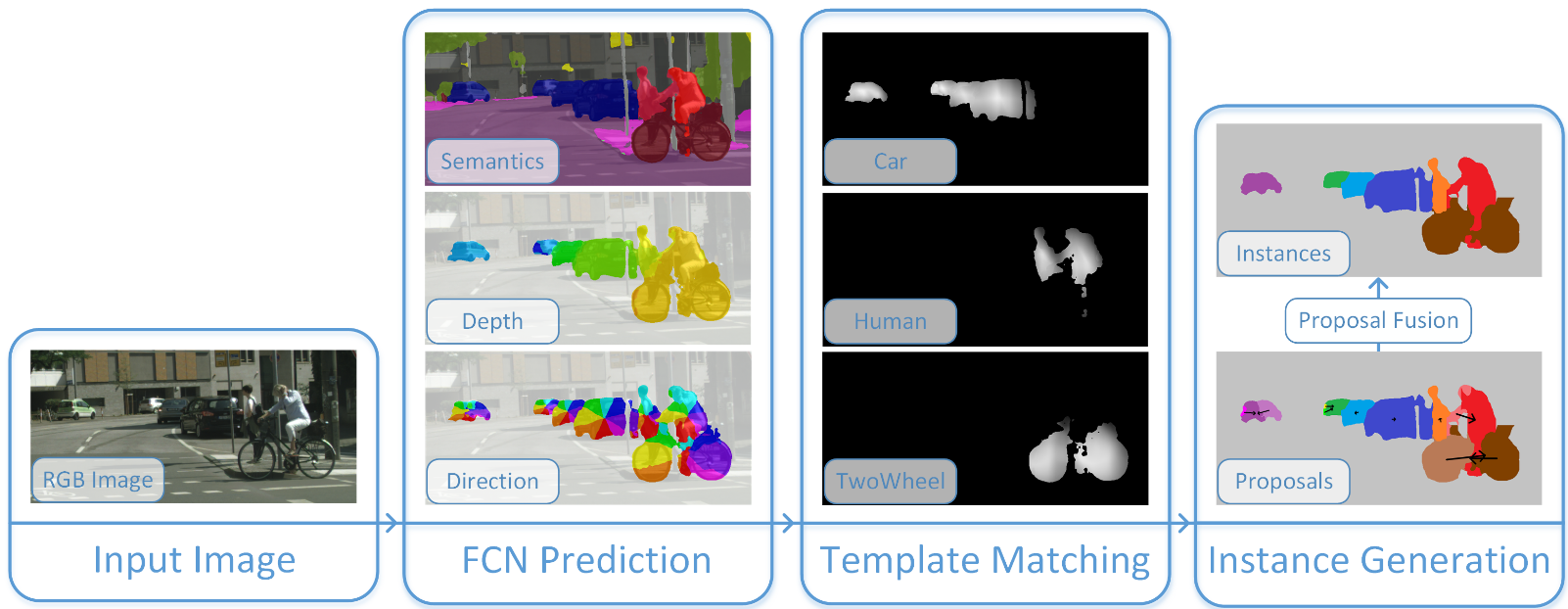}
	\caption[Proposal-free Instance Segmentation Pipeline]{\textbf{Proposal-free Instance Segmentation Pipeline.} \protect\citet{Uhrig2016GCPR} predict semantics, depth, and instance center direction from the input image to compute template matching scores for all semantic maps. They fuse them after generating instance proposals to obtain an instance segmentation. \figsourceSpringer{\protect\citet{Uhrig2016GCPR}}{2016}{GCPR}.}
	\label{fig:Uhrig2016GCPR}
\end{figure}
Several approaches \citep{Zhang2015ICCV, Zhang2016CVPR, Uhrig2016GCPR} show how depth information can be used to identify different object instances.
\citet{Zhang2015ICCV,Zhang2016CVPR} train a fully convolutional neural network (FCN) to directly predict pixel-level instance segmentations of densely sampled image patches while the instance ID encodes a depth ordering. They improve the predictions and enforce consistency with a subsequent Markov Random Field. \citet{Uhrig2016GCPR} propose an FCN to jointly predict semantic segmentation as well as depth and an instance-based direction relative to the centroid of each instance. This relative direction cue is then used for clustering pixels into individual instances. The instance segmentation pipeline is illustrated in \figref{fig:Uhrig2016GCPR}. However, all \citep{Zhang2015ICCV, Zhang2016CVPR, Uhrig2016GCPR} require ground-truth depth data for training their model. 

Instead of relying on depth information, concurrent work \citep{Kirillov2017CVPR, Bai2017CVPR, Arnab2017CVPR} present proposal-free approaches based on an initial semantic segmentation. \citet{Kirillov2017CVPR} combine semantic segmentation and object boundary detection via global reasoning in a multi-cut formulation to infer semantic instance segmentation. \citet{Bai2017CVPR} combine ideas from classical watershed transform with deep learning to create an energy map from an initial semantic segmentation and the input image where the basins correspond to object instances.  
This allows them to cut at a single energy level for obtaining a pixel-level instance segmentation. \citet{Arnab2017CVPR} propose to refine an initial semantic segmentation using an instance subnetwork. The initial category-level segmentation is used along cues from the output of an object detector within an unrolled Conditional Random Field \citep{Zheng2015ICCV} to predict pixel-level instances. 

A new line of work is presented by \citet{Liu2017ICCV}. They follow a sequential strategy with increasing semantic complexity. Several neural networks are applied in sequential order, each grouping pixels with different strategies starting by finding vertical and horizontal breakpoints, then connecting them to vertical and horizontal lines, grouping pixels in between these lines, and finally, extracting instances from the grouped pixels.

\subsection{Panoptic Segmentation}
\begin{figure}[t]
	\centering
	\includegraphics[width=0.70\columnwidth]{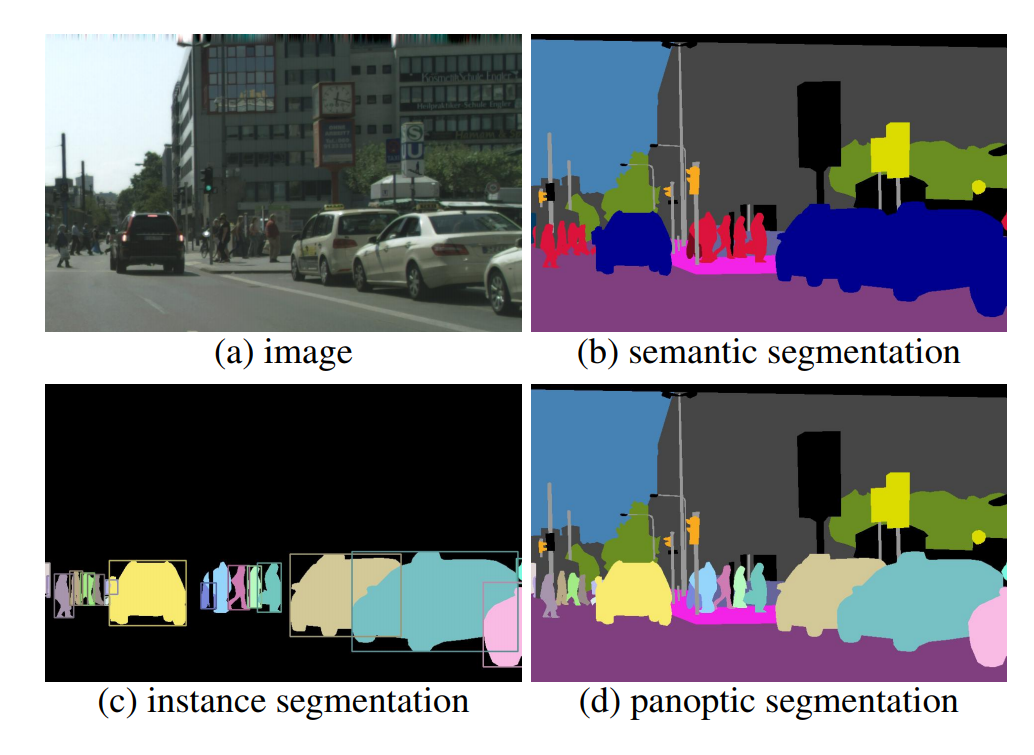}
	\caption[Panoptic Segmentation]{\textbf{Panoptic Segmentation.} Difference between semantic (b), instance (c) and panoptic segmentation (d). \figsourceC{\protect\citet{Kirillov2019CVPRa}}{2019}{IEEE}.}
	\label{fig:Kirillov2018ARXIV}
\end{figure} 
Instance segmentation focuses on instances of objects and usually ignores classes that are not amendable to this task like sky or road as illustrated in \figref{fig:Kirillov2018ARXIV}. In contrast, panoptic segmentation, first introduced by \citet{Kirillov2019CVPRa}, addresses the dense estimation of a semantic label and instance id. 
Several approaches \citep{Kirillov2019CVPRa, Li2018ECCVb, Costea2018ITSC, Xiong2019CVPR, Kirillov2019CVPR} have been proposed to address the problem.

Proposal-free instance segmentation approaches like \citep{Arnab2017CVPR, Bai2017CVPR, Kirillov2017CVPR} can be used directly to learn panoptic segmentation. However, ground truth for training them on this problem is very limited. Thus, \citet{Li2018ECCVb} use \citep{Arnab2017CVPR} in a semi-supervised fashion to learn panoptic segmentation. They use interactive foreground extraction (GrapCut) \citep{Rother2004SIGGRAPH}, proposal segmentation \citep{Pont-Tuset2017PAMI} and gradient-based localization of classes \citep{Selvaraju2017ICCV} to train the network in a semi-supervised fashion.

In contrast, several approaches \citep{Costea2018ITSC, Xiong2019CVPR, Kirillov2019CVPR} address panoptic segmentation with a joint semantic and instance segmentation formulation based on Mask R-CNN \citep{He2017ICCV}. \citet{Costea2018ITSC} propose to fuse object detections, semantic, and instance segmentation. They use semantic segmentation to distinguish between fore- and background regions. While the semantic class of background regions is directly obtained from the semantic segmentation, they use object detection, instance, and semantic segmentation to determine the class of foreground regions. In contrast, concurrent work \citep{Xiong2019CVPR, Kirillov2019CVPR} extends Mask R-CNN by an additional semantic segmentation branch removing the requirement of a heuristic fusion.
\citet{Xiong2019CVPR} combine a semantic segmentation network based on deformable convolutions with Mask R-CNN. They predict dense class labels by applying a softmax on concatenated channels of the semantic and instance segmentation networks. In contrast, \citet{Kirillov2019CVPR} train the semantic and instance segmentation networks simultaneously without concatenating the channels. They apply non-maximum suppression \citep{Kirillov2019CVPRa} to avoid overlapping instances.
 
\begin{table*}[t]
	\centering
	\begin{adjustbox}{width=1\textwidth}\begin{tabular}{l l|c|c|c|c|c|c|c}
 & \textbf{Method} & \textbf{Coarse} & \textbf{COCO} & \textbf{Depth} & \textbf{AP} & \textbf{AP 50\%} & \textbf{AP 100m} & \textbf{AP 50m} \\ \hline
1. & PANet \citep{Liu2018CVPR} &  & \cmark &  &  36.4 & 63.1 & 49.2 & 51.8 \\
2. & UPSNet \citep{Xiong2019CVPR} &  &  &  &  33.0 & 59.6 & 46.8 & 50.7 \\
3. & Mask R-CNN \citep{He2017ICCV} &  & \cmark &  &  32.0 & 58.1 & 45.8 & 49.5 \\
4. & PANet \citep{Liu2018CVPR} &  &  &  &  31.8 & 57.1 & 44.2 & 46.0 \\
5. & Mask R-CNN \citep{He2017ICCV} &  &  &  &  26.2 & 49.9 & 37.6 & 40.1 \\
6. & PolygonRNN++ \citep{Acuna2018CVPR} &  &  &  &  25.5 & 45.5 & 39.3 & 43.4 \\
7. & SGN  \citep{Liu2017ICCV} & \cmark &  &  &  25.0 & 44.9 & 38.9 & 44.5 \\
8. & Pixelwise Inst. Seg. with a DIN \citep{Arnab2017CVPR} & \cmark &  &  &  23.4 & 45.2 & 36.8 & 40.9 \\
9. & Multitask Learning \citep{Kendall2018CVPR} &  &  & &  21.6 & 39.0 & 35.0 & 37.0 \\
10. & Deep Watershed Transformation \citep{Bai2017CVPR} & &  &  &  19.4 & 35.3 & 31.4 & 36.8 \\
11. & Sem. Inst. Seg. with a DLF \citep{Brabandere2017CVPRWORK} &  &  &  &  17.5 & 35.9 & 27.8 & 31.0 \\
12. & Boundary-aware Inst. Seg. \citep{Hayder2017CVPR} & &  &  &  17.4 & 36.7 & 29.3 & 34.0 \\
13. & InstanceCut \citep{Kirillov2017CVPR} & \cmark &  &  &  13.0 & 27.9 & 22.1 & 26.1 \\
14. & Foveal Vis. for Inst. Seg. of Road Images \citep{Ortelt2018VISIGRAPP} &  &  & \cmark &  12.5 & 25.2 & 20.4 & 22.1 \\
15. & Joint Graph Decomp. \& Node Labeling \citep{Levinkov2017CVPR} &  &  & &  9.8 & 23.2 & 16.8 & 20.3 \\
16. & Pixel-level Encoding for Inst. Seg. \citep{Uhrig2016GCPR} &  &  & \cmark &  8.9 & 21.1 & 15.3 & 16.7 \\
17. & R-CNN + MCG convex hull \citep{Cordts2016CVPR} &  &  &  &  4.6 & 12.9 & 7.7 & 10.3 \\
\end{tabular}\end{adjustbox}
	\caption{{\bf CITYSCAPES Instance Segmentation Leaderboard.} Instance detection performance is measured in terms of several average precision variants. The coarse annotations only provide rough class-level labels and are thus only used by a few methods. Since the dense annotations are quite limited, Microsoft COCO \citep{Lin2014ECCV} is also sometimes used for training. More details in \citep{Cordts2016CVPR}. Accessed on: June 2019.}
	\label{tab:cityscapes_instance_level}
\end{table*}
\section{Datasets}
Only a few datasets for instance segmentation exist. Microsoft COCO \citep{Lin2014ECCV}, consisting of 328K dense annotations, and Cityscapes \citep{Cordts2016CVPR}, consisting of 5K dense and 25K sparse annotations, are the most popular datasets. 
While the KITTI \citep{Geiger2012CVPR} dataset also provides instance-level semantic annotations, the dataset consists of only 200 training and test scenes.  
The original PASCAL VOC \citep{Everingham2010IJCV} does not provide instance-aware annotations but \citet{Hariharan2011ICCV} extended the dataset by semantic contours which are instance-aware. Still, the extension of PASCAL VOC is rarely used. 
The new datasets Mapillary \citep{Neuhold2017ICCV}, ApolloScape \citep{Huang2018CVPR} and Berkeley DeepDrive \citep{Yu2018ARXIV} also provide instance-level annotations for 25K, 90K, and 10K images, respectively, but still need to prevail in the community.

Similar to semantic segmentation, we compare different methods on the Cityscapes dataset\footnote{\url{https://www.cityscapes-dataset.com/}} by \citet{Cordts2016CVPR} because of the autonomous driving context, the online leaderboard and its acceptance in the community.

\section{Metrics}
The performance of instance segmentation methods is typically assessed by measuring average precision (AP) on instance regions that reach a certain overlap with ground truth regions. Usually, different thresholds are considered for the overlap and comparisons are performed according to the average over these thresholds as well as all classes. Cityscapes \citep{Cordts2016CVPR} use the same metric reported in Microsoft COCO \citep{Lin2014ECCV} which considers 10 thresholds between 50\% and 95\%. In addition, the AP for an overlap value of 50 \% (AP 50\%) and for objects within 100 m and 50 m (AP 100m, AP 50m) are considered separately. 

\section{State of the Art on Cityscapes}
In \tabref{tab:cityscapes_instance_level}, we show the leaderboard of semantic instance segmentation methods on the Cityscapes dataset.
The state of the art in instance segmentation is dominated by proposal-based approaches \citep{Acuna2018CVPR, He2017ICCV, Xiong2019CVPR, Liu2018CVPR}.
However, they are closely followed by proposal-free approaches \citep{Bai2017CVPR, Arnab2017CVPR, Liu2017ICCV} with the sequential approach from \citet{Liu2017ICCV} being the best performing proposal-free approach. While the proposal-based approaches \citep{Acuna2018CVPR, He2017ICCV, Xiong2019CVPR} are built on Faster R-CNN \citep{Ren2015NIPS}, \citet{Liu2018CVPR} propose a new feature hierarchy to propagate features from all levels (\ia accurate localization signals from lower layers) to proposal sub-networks.
They outperform all other methods with additional training on the Microsoft COCO dataset \citep{Lin2014ECCV} since the dense annotations of Cityscapes are rather limited. The panoptic segmentation approach presented by \citet{Xiong2019CVPR} is the best performing method when training is restricted to the dense annotations of Cityscapes.

\section{Discussion}
The instance segmentation task is much harder than the semantic segmentation task. Each instance needs to be carefully labeled separately whereas in semantic segmentation groups of one semantic class can be labeled together when they occur next to each other. In addition, the number and size of instances vary greatly between different images. In the autonomous driving context, often a wide view is present. Therefore, a large number of instances appear rather small in the image, making them challenging to detect. In contrast to bounding box detections discussed in \secref{sec:detection_discussion}, the exact shape of each object instance needs to be inferred in this task. Thus, the state of the art is still struggling on the Cityscapes dataset (\tabref{tab:cityscapes_instance_level}) reaching an average precision of $36\%$ or less. Proposal-based approaches which jointly address detection and segmentation with parallel sub-networks are currently the most promising direction. The joint formulation allows improving the generation of small instance proposals, which is important for segmenting instances in the context of autonomous driving.
	\chapter{Stereo}
\label{chap:Stereo}
\begin{figure}[t]
	\centering
	\includegraphics[width=1.00\columnwidth]{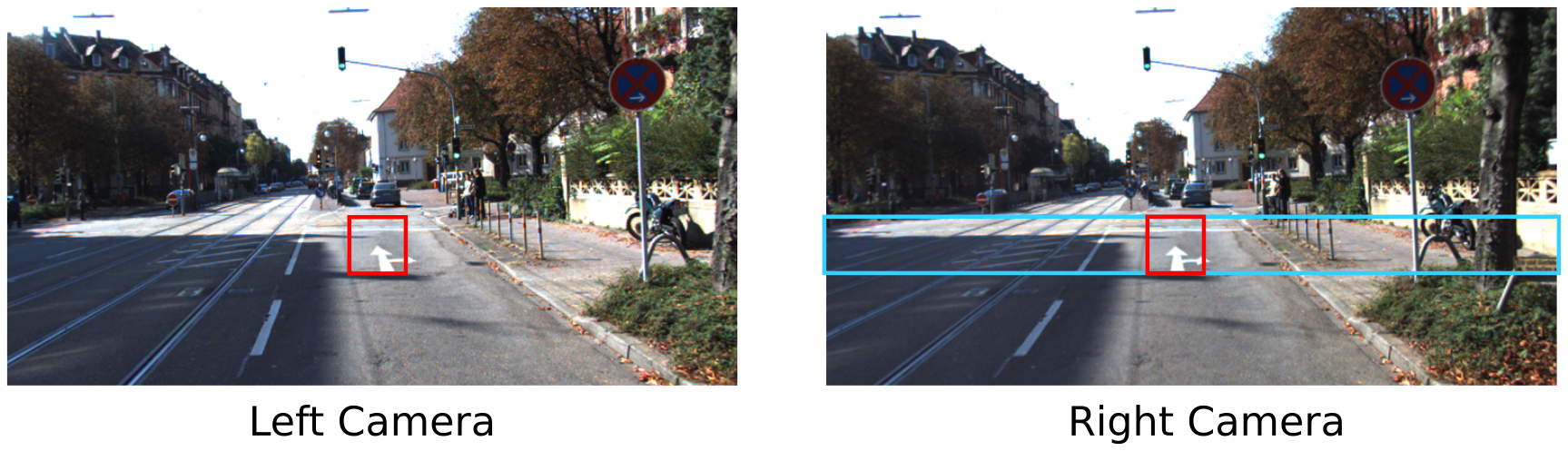}
	\caption[Stereo Matching Problem]{\textbf{Stereo Matching Problem.} Visualization of the stereo matching problem. Given two rectified images (from KITTI training \citep{Geiger2012CVPR}), stereo matching reduces to a 1D search problem along the epipolar line (blue rectangle). }
	\label{fig:stereo_matching}
\end{figure} 
\section{Problem Definition}
Stereo estimation is the process of extracting 3D information from passive 2D images captured by stereo cameras without the need for dedicated active light-emitting range measurement devices. In particular, stereo algorithms estimate depth information by finding correspondences between two images taken at the same point in time, typically by two cameras mounted next to each other on a fixed rig. These correspondences are projections of the same physical surface in the 3D world. Depth information is crucial for applications in autonomous driving or driver assistance systems. Accurate estimation of dense depth maps is a necessary step for 3D reconstruction, and many other problems such as obstacle detection, free space analysis, and tracking benefit from the availability of accurate depth estimates.   

\section{Methods}
In stereo matching, the images from two cameras are usually projected onto a common parallel during rectification. 
This reduces the matching problem to a 1D search along the epipolar line, as illustrated in \figref{fig:stereo_matching}, and the distance on this line is usually referred to as disparity.

The stereo literature can be separated into two groups. Feature-based methods \citep{Szeliski2011, Schauwecker2012IROS} provide only sparse depth maps, while dense methods generate dense outputs at the expense of computation time. In our survey, we focus on dense methods since they are more popular, and with the introduction of deep learning, they also became much more efficient. We further distinguish between local and global methods. Local methods compute the disparity by simply selecting the lowest matching cost, which is known as the winner takes all (WTA) solution \citep{Hirschmueller2007CVPR, Szeliski2011}. However, they usually result in very noise estimates caused by ambiguities. In contrast, global methods formulate disparity computation as an energy-minimization problem integrating smoothness assumptions between neighboring pixels or regions \citep{Hirschmueller2008PAMI, Geiger2010ACCV, Gallup2010CVPR, Haene2012THREEDIMPVT, Bredies2010JIS, Kuschk2013ICCVWORK, Ranftl2013SSVM}. 
Optimization can be carried out using variational approaches in the continuous domain and discrete approaches such as graph cuts or believe propagation for discrete label spaces.

\subsection{Matching Cost} 
Stereo matching is a correspondence estimation problem where the goal is to identify the matching points between the left and right image based on a cost function. The algorithms usually assume rectified images, and the search space is reduced to a horizontal line (\figref{fig:stereo_matching}).
The matching cost computation is the process of computing a cost function at each pixel for all possible disparities, which is minimal at the true disparity. However, it is hard to design such a cost function in practice. Therefore stereo algorithms typically use the assumption of constant appearance between matching points. This assumption is often violated in real-world situations, such as cameras with slightly different settings causing exposure changes, vignetting, image noise, non-Lambertian surfaces, illumination changes, etc. \citet{Hirschmueller2007CVPR} systematically investigate the effect of these radiometric changes on commonly used matching cost functions, namely absolute differences, filter-based costs (Laplacian of Gaussian, Rank and Mean), hierarchical mutual information (HMI), and normalized cross-correlation. They found that the performance of a cost function depends on the stereo method that uses it. On images with simulated and real radiometric differences, rank filter performed best for correlation-based methods. For global methods, in tests with global radiometric changes or noise, HMI performed best, while in the presence of local radiometric variations, Rank and Laplacian of Gaussian filters performed better than HMI. Qualitative results show that filter-based costs cause blurred object boundaries when used with global methods. None of the matching costs under consideration could succeed in handling strong lighting changes.

\subsection{Energy Optimization} 
The inherent ambiguity in appearance-based matching costs can be overcome by regularization, \ie introducing prior knowledge about the expected disparity map into the stereo estimation process. 
Therefore, an energy consisting of the matching cost and a smoothness constraint is usually optimized in contrast to WTA over the matching costs.
The simplest prior favors neighboring pixels to take on the same disparity value (local smoothness). 

\boldparagraph{Discrete Optimization} 
Discrete optimization methods optimize an energy with respect to a discrete set of disparities.
While the resulting minimization problem is NP-hard, good approximations can be obtained using belief propagation \citep{Felzenszwalb2006IJCV} and graph cuts \citep{Boykov1999PAMI}.

Semi-Global Matching (SGM) proposed by \citet{Hirschmueller2008PAMI} is the most prominent discrete optimization method for stereo matching.
They hierarchically compute the matching cost by considering Mutual Information. 
A global smooth energy is approximated with cost aggregation by summing costs along 1D paths from multiple directions towards each pixel using dynamic programming.
SGM became an influential stereo matching technique for autonomous driving due to its speed and high accuracy, as evidenced in various benchmarks such as Middlebury \citep{Scharstein2002IJCV} and KITTI \citep{Geiger2012CVPR}.

There are a few follow-up works investigating the practical and theoretical sides of SGM. \citet{Gehrig2009ICVS} propose a real-time, low-power implementation of SGM with algorithmic extensions for automotive applications on a reconfigurable hardware platform. \citet{Drory2014GCPR} offer a principled explanation for the success of SGM by clarifying its relation to belief propagation and Tree-Reweighted Message Passing \citep{Kolmogorov2006PAMI}. They show that SGM is equivalent to early stopping for a particular variant of belief propagation, effectively approximating the solution.

The performance of SGM can be further improved by incorporating a confidence measure. \citet{Seki2016BMVC} leverage CNNs to predict confidence for stereo estimations. Taking into account ideas from conventional methods, they design a two-channel disparity patch which is used as input to a CNN. The first channel uses local smoothness, and the second enforces left-right consistency (disparity estimation using the other image should yield corresponding results). The confidences are incorporated into SGM by weighting each pixel according to the estimated confidence. 

\boldparagraph{Continuous Optimization} 
Variational approaches optimize the energy function with respect to continuous disparities. Data costs using the image intensities are usually non-convex and, thus, the global optimum can only be approximated. 
Coarse-to-fine approaches are used to handle large disparities by going from a low to a high resolution solution of the matching problem. For each resolution, the previous lower resolution solution is used as initialization. Coarse-to-fine approaches are typically used for optical flow estimation and will be discussed in detail in \chpref{chap:optical_flow}. 

A commonly used smoothness prior is Total Variation (TV) \citep{Rudin1992} that penalizes the absolute difference between neighboring disparities. In the presence of weak and ambiguous observations, TV does not produce convincing results since it encourages piecewise constant disparities leading to stair-casing artifacts.

\subsection{Higher-Order Models} 
\label{sec:stereo_methods_superpixel}
\begin{figure}[t]
	\centering
	\includegraphics[width=1.00\columnwidth]{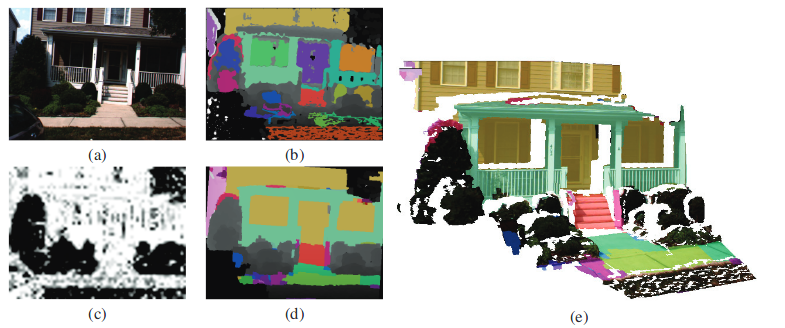}
	\caption[Piecewise Planarity]{\textbf{Piecewise Planarity.} \protect\citet{Gallup2010CVPR} enforces piecewise planarity on planar structures of the scene found with RANSAC and plane classifiers. Plane candidates obtained with RANSAC (b), planar class probabilities (c), final plane assignment (d) and the 3D model with highlighted planes (e) are shown. \figsourceC{\protect\citet{Gallup2010CVPR}}{2010}{IEEE}.}
	\label{fig:Gallup2010CVPR}
\end{figure} 
Pairwise smoothness priors fail to reconstruct poorly-textured and slanted surfaces, as they favor fronto-parallel planes. 
A more generic approach to handle arbitrary smoothness priors is to exploit high-order correlations between pixels. 
Higher-order priors are able to express more realistic assumptions about depth images, but usually at additional computational costs. 

\citet{Woodford2009PAMI} introduce second-order priors for a graph cut stereo formulation. While incorporating higher-order priors in discrete optimization has long been considered computationally infeasible, they propose an efficient optimization strategy for inference with triple cliques. In addition, they present an asymmetrical occlusion model that is combined with the second-order prior.

For continuous TV formulations, \citet{Haene2012THREEDIMPVT} introduce patch-based priors in the form of small, piecewise planar dictionaries. Total Generalized Variation (TGV) \citep{Bredies2010JIS} is argued to be a better prior than TV, since it does not penalize piecewise affine solutions. However, it is restricted to convex data terms in contrast to TV, where global solutions can be computed even in the presence of non-convex data terms. Coarse-to-fine approaches often end up with a loss of details. In order to preserve fine details, \citet{Kuschk2013ICCVWORK} integrate an adaptive regularization weight into the TGV framework by using edge detection and report improved results compared to coarse-to-fine approaches. \citet{Ranftl2013SSVM} obtain even better results by proposing a decomposition of the non-convex functional into two subproblems.

\subsection{Piecewise Planar Priors} 

One common way to deal with slanted surfaces in the literature is to assume piecewise planarity. \citet{Geiger2010ACCV} build a prior over the disparity space by forming a triangulation on a set of robustly matched correspondences, called support points. This reduces matching ambiguities and results in an efficient algorithm by restricting the search to plausible regions. \citet{Gallup2010CVPR}, illustrated in \figref{fig:Gallup2010CVPR}, first train a classifier to segment an image into piecewise planar and non-planar regions. Afterwards, they enforce a piecewise planarity prior only on planar regions using plane hypotheses obtained from RANSAC. Non-planar regions are modeled by the output of a standard multi-view stereo algorithm. 

\begin{figure}[t]
	\centering
	\includegraphics[width=0.80\columnwidth]{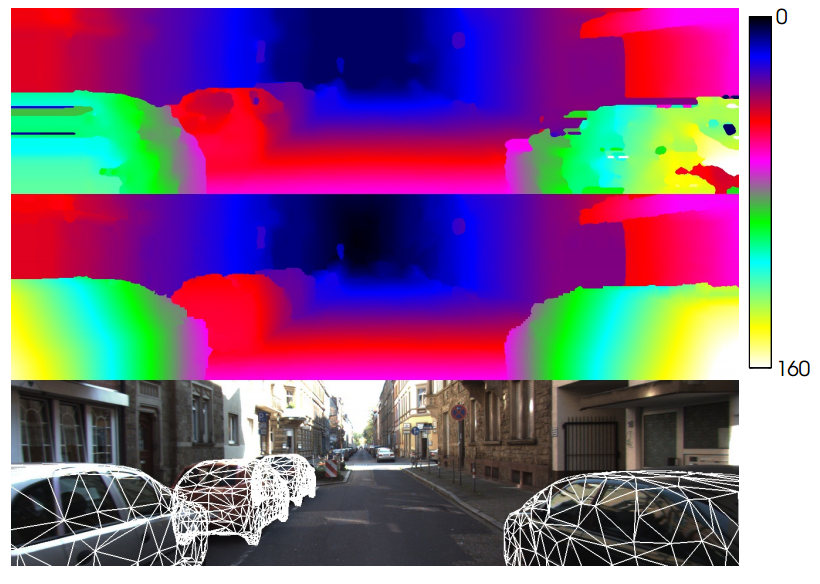}
	\caption[Stereo Matching using Object Knowledge]{\textbf{Stereo Matching using Object Knowledge.} Stereo methods often fail at reflecting, textureless or semi-transparent surfaces (top  \protect\citep{Zbontar2016JMLR}). By using object knowledge, \protect\citet{Guney2015CVPR} encourage disparities to agree with plausible surfaces (center). This improves results both quantitatively and qualitatively while simultaneously recovering the 3D geometry of the objects in the scene (bottom). The disparity is illustrated with a color coding shown on the right side. \figsourceC{\protect\citet{Guney2015CVPR}}{2015}{IEEE}.}
	\label{fig:stereo_displets}
\end{figure} 

\subsection{Segmentation-based Models} 
An alternative way of modeling piecewise planarity is to explicitly partition the image into superpixels (groups of pixels) and modeling the surface at each superpixel as a slanted plane \citep{Yamaguchi2012ECCV, Guney2015CVPR}. However, care must be taken to ensure that the superpixelization is indeed an oversegmentation of the image with respect to planarity, \ie no superpixel contains two surfaces that are not co-planar. \citet{Yamaguchi2012ECCV} jointly reason about occlusion boundaries and depth in a hybrid MRF composed of both continuous and discrete random variables. \citet{Guney2015CVPR} use a similar framework to incorporate object-category specific 3D shape proposals that regularize over larger distances. By leveraging semantic segmentation and 3D CAD models, they resolve ambiguities in reflective and textureless regions originating from highly specular surfaces of cars in the scene, as shown in \figref{fig:stereo_displets}.

\subsection{Deep Learning for Stereo Matching} 
\label{sec:stereo_methods_cnn}
\begin{figure}[t]
	\centering
	\includegraphics[width=0.80\columnwidth]{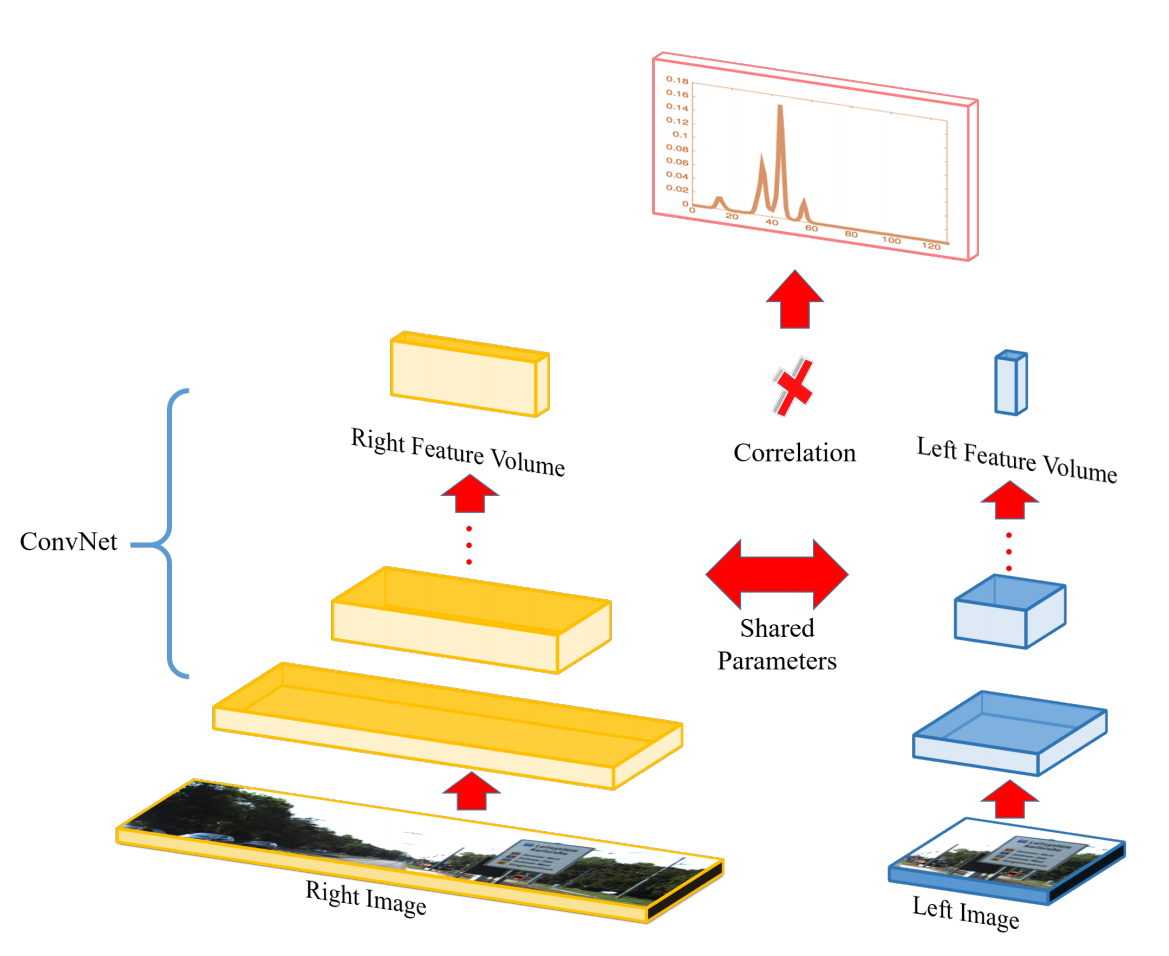}
	\caption[Deep Learning for Stereo Matching]{\textbf{Deep Learning for Stereo Matching.} A Siamese network is trained to extract marginal distributions over all possible disparities for each pixel. \figsourceC{\protect\citet{Luo2016CVPR}}{2016}{IEEE}.}
	\label{fig:stereo_deep_learning}
\end{figure}
In the last years, deep learning approaches gained popularity in stereo estimation. While some methods try to learn richer feature representations \citep{Zbontar2016JMLR, Luo2016CVPR}, others learn to directly predict a disparity map from the input stereo image pair \citep{Mayer2016CVPR, Kendall2017ICCV, Chang2018CVPR}. 

For richer feature representations, \citet{Zbontar2016JMLR, Luo2016CVPR} use a Siamese network that consists of two sub-networks with shared weights and a final score computation layer. The idea is to train the network for computing the matching cost by learning a similarity measure on small image patches. \citet{Zbontar2016JMLR} define positive/negative examples as matching and non-matching patches and use a margin loss to train either a fast architecture with a simple dot-product layer in the end or a slow but more accurate architecture which learns score computation with a set of fully connected layers. \citet{Luo2016CVPR} use a similar architecture, but formulate the problem as multi-class classification over all possible disparities to capture correlations between different disparities implicitly, as visualized in \figref{fig:stereo_deep_learning}. Both approaches rely on SGM  \citep{Hirschmueller2008PAMI} as a post-processing step to propagate information to neighboring pixels and estimate the final dense disparity map.

In contrast, \citet{Mayer2016CVPR} adapt the encoder-decoder architecture proposed by \citet{Dosovitskiy2015ICCV} for optical flow estimation (\chpref{chap:optical_flow}) to directly learn a model which predicts the entire disparity map at once without need for additional post-processing. The encoder computes abstract features while the decoder reestablishes the original resolution with additional cross-links between the contracting and expanding network parts for preserving the details. Post-processing and regularization are not necessary since the encoder-decoder architecture implicitly learns the entire mapping end-to-end.
However, this architecture has to learn the concept of matching from scratch. Thus, inspired by \citep{Dosovitskiy2015ICCV}, they also propose an alternative network (DispNetC) that first process each image independently and finally correlates extracted features from both images.
\citet{Kendall2017ICCV} combine ideas from previous methods, \ie Siamese feature extraction, and cost volume formation. More specifically, they propose to extract deep feature representations using a Siamese network and correlate these features to create a cost volume. After the cost volume, they use an encoder-decoder architecture to enlarge the receptive field and apply 3D convolutions on each encoder level. A differentiable soft argmin operation allows them to train the network end-to-end.
\citet{Chang2018CVPR} introduce a spatial pyramid pooling and 3D CNN module to exploit more context information. Spatial pyramid pooling allows the extraction of richer features by taking larger regions into account. The 3D CNN module with multiple stacked encoder-decoder networks enables them to leverage global context information and achieve state-of-the-art performance.

Recently, it has been demonstrated that semantic information can also be exploited in the context of deep learning-based stereo estimation. \citet{Yang2018ECCV} jointly formulate semantic segmentation and stereo estimation in one framework. This allows them to learn semantic cues and incorporate them into the disparity estimate by introducing a semantic softmax loss that regularizes the disparity with semantic cues. They show the benefit of their joint formulation in the unsupervised and in the supervised setting.

\subsection{Variable Baseline} 
Stereo estimates can be fused to yield a more complete reconstruction of the static parts of the three-dimensional scene. However, assuming a fixed baseline, focal length, and field of view might not always be the best strategy. \citet{Gallup2008CVPR} point out two problems with traditional stereo methods: dropping accuracy in the far range and unnecessary computation time spent in the near range. They, therefore, propose to use a multi-camera rig and to dynamically select the best cameras with the appropriate baseline for accurate estimation. In addition, they reduce the resolution to speed up the computation in the near range. In contrast to traditional fixed-baseline stereo, the proposed variable baseline stereo algorithm achieves constant accuracy over the reconstructed volume by evenly spreading the computation throughout the volume.

\subsection{Omnidirectional Cameras} 
Omnidirectional sensors discussed in \ref{sec:calibration_omnidirectional_cam} allow to significantly increase the field of view for stereo matching. However, only limited work on stereo estimation using omnidirectional sensors exist. \citet{Haene2014THREEDV} extend the plane-sweeping stereo matching for fisheye cameras by incorporating the unified projection model for fisheye cameras \citep{Geyer2000ECCV} directly into the plane-sweeping stereo matching algorithm \citep{Gallup2007CVPR}. This kind of approach allows for producing dense depth maps directly from fisheye images in real-time using GPUs. \citet{Schoenbein2014IROS} consider the stereo matching problem for catadioptric omnidirectional cameras.

\section{Datasets}
The most popular datasets for stereo estimation are the Middlebury \citep{Scharstein2002IJCV, Scharstein2003CVPR, Scharstein2014GCPR} and KITTI \citep{Geiger2012CVPR} datasets. The ETH3D \citep{Schoeps2017CVPR} also provides a two-view benchmark but is relatively new and does not focus on the autonomous driving scenario.
Since only the KITTI dataset considers the autonomous driving context, we focus our attention on the KITTI benchmark. 

Larger datasets are necessary for training deep models. In this case, the community relies on synthetic datasets such as SYNTHIA \citep{Ros2016CVPR}, Virtual KITTI \citep{Gaidon2016CVPR}, Flying Things \citep{Mayer2016CVPR} and Sintel \citep{Butler2012ECCV}. However, the models trained on synthetic datasets are usually not generalizing to real datasets and need further fine-tuning on real datasets. 

\section{Metrics}
\label{sec:stereo_metrics}
Multiple metrics have been proposed to measure the performance of stereo approaches. The most popular measures are the root-mean-squared error (RMS) and outlier ratio, \ie percentage of bad pixels (pixel with an error larger than a threshold). Typically, the average RMS is reported while the outlier ratio is often evaluated using several thresholds. Middlebury reports results for 0.5, 1, 2, and 4 pixels thresholds. In contrast, the KITTI benchmark reports the percentage of pixels with an error larger than 3 pixels or 5\%. In addition, they separately evaluate the percentage of bad pixels over background and foreground regions.

\begin{table*}[t!]
	\begin{center}
		\begin{adjustbox}{width=1\textwidth}\begin{tabular}{l l | c | c | c | c }
& {\bf Method} & {\bf D1-bg} & {\bf D1-fg} & {\bf D1-all} & {\bf Runtime}\\ \hline
1. & EdgeStereo-V2 \citep{Song2018ACCV} & 1.84 \% & 3.30 \% & 2.08 \% & 0.32s / GPU \\
2. & Stereo-fusion-SJTU \citep{Song2018ACCV} & 1.87 \% & 3.61 \% & 2.16 \% & 0.7 s / GPU \\
3. & SegStereo \citep{Yang2018ECCV} & 1.88 \% & 4.07 \% & 2.25 \% & 0.6 s / GPU \\
4. & PSMNet \citep{Chang2018CVPR} & 1.86 \% & 4.62 \% & 2.32 \% & 0.41 s / GPU \\
5. & PDSNet \citep{Tulyakov2018NIPS} & 2.29 \% & 4.05 \% & 2.58 \% & 0.5 s / 1 core \\
6. & SCV \citep{Lu2018RS} & 2.22 \% & 4.53 \% & 2.61 \% & 0.36 s / GPU \\
7. & CRL \citep{Pang2017ICCV} & 2.48 \% & 3.59 \% & 2.67 \% & 0.47 s / GPU \\
8. & GC-NET \citep{Kendall2017ICCV} & 2.21 \% & 6.16 \% & 2.87 \% & 0.9 s / GPU \\
\hline
15. & Displets v2 \citep{Guney2015CVPR} & 3.00 \% & 5.56 \% & 3.43 \% & 265 s / >8 cores \\
19. & MC-CNN-acrt \citep{Zbontar2016JMLR} & 2.89 \% & 8.88 \% & 3.89 \% & 67 s / GPU \\
20. & PRSM \citep{Vogel2015IJCV} & 3.02 \% & 10.52 \% & 4.27 \% & 300 s / 1 core \\
21. & DispNetC \citep{Mayer2016CVPR} & 4.32 \% & 4.41 \% & 4.34 \% & 0.06 s / GPU \\
41. & SGM\_ROB \citep{Hirschmueller2008PAMI} & 5.06 \% & 13.00 \% & 6.38 \% & 0.11 s / GPU \\
\end{tabular}\end{adjustbox}
	\end{center}
	\vspace{-0.2cm}
	\caption{{\bf KITTI 2015 Stereo Leaderboard.} Numbers correspond to percentages of bad pixels according to the 3px/5\% criterion defined in \cite{Menze2015CVPR} in background (bg), foreground (fg) or all regions. The methods below the horizontal line are older entries, serving as reference. Accessed on: June 2019.}
	\label{tab:kitti_stereo_2015}
\end{table*}
\section{State of the Art on KITTI}
In \tabref{tab:kitti_stereo_2015} we show the ranking of stereo methods on the KITTI stereo 2015 benchmark. 
\citet{Tulyakov2018NIPS} combine similar to \citep{Dosovitskiy2015ICCV, Kendall2017ICCV, Chang2018CVPR} learning of the feature extraction, correlation, and regularization in an end-to-end trainable model. They extract deep features with a bottleneck architecture in contrast to a regular encoder-decoder network and propose a novel sub-pixel maximum a posteriori (MAP) approximation for inference based on the weighted mean around the disparity with maximum posterior probability. While the bottleneck architecture allows reducing the memory footprint, the sub-pixel MAP approximation enables to handle different disparity ranges than used for training. They achieve competitive results on KITTI.
However, the spatial pyramid pooling and 3D CNN proposed by \citet{Chang2018CVPR}, as discussed in \secref{sec:stereo_methods_cnn}, is computationally more efficient and improves significantly in the background regions.
\citet{Yang2018ECCV} jointly address the semantic segmentation problem to incorporate more contextual information as discussed in \secref{sec:stereo_methods_superpixel}. While they reach similar performance on the background regions, the joint formulation enables to improve also on the foreground regions.
The best performance in foreground and background regions is achieved by \citet{Song2018ACCV}. Similar to \citep{Yang2018ECCV}, they use a joint formulation, but instead of semantic segmentation, they jointly learn image edges using an edge-aware smoothness loss. In combination with a context pyramid to extract multi-scale features and one-stage residual pyramid returning a full-size disparity map, they outperform all other methods, as shown in \tabref{tab:kitti_stereo_2015}. 
However, DispNetC presented by \citet{Mayer2016CVPR} remains one of the fastest approaches while achieving competitive results on the foreground.

\section{Discussion}
\begin{figure*}[p]
\includegraphics[width=0.5\linewidth]{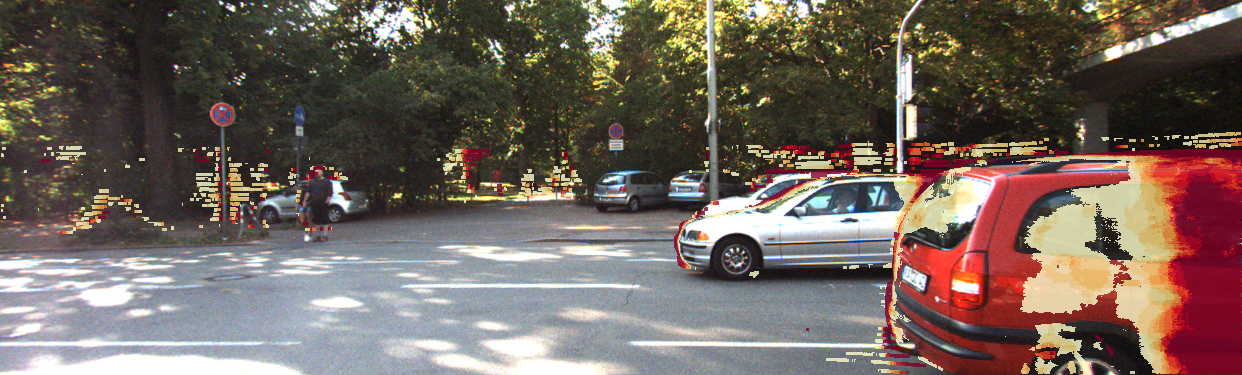}%
\includegraphics[width=0.5\linewidth]{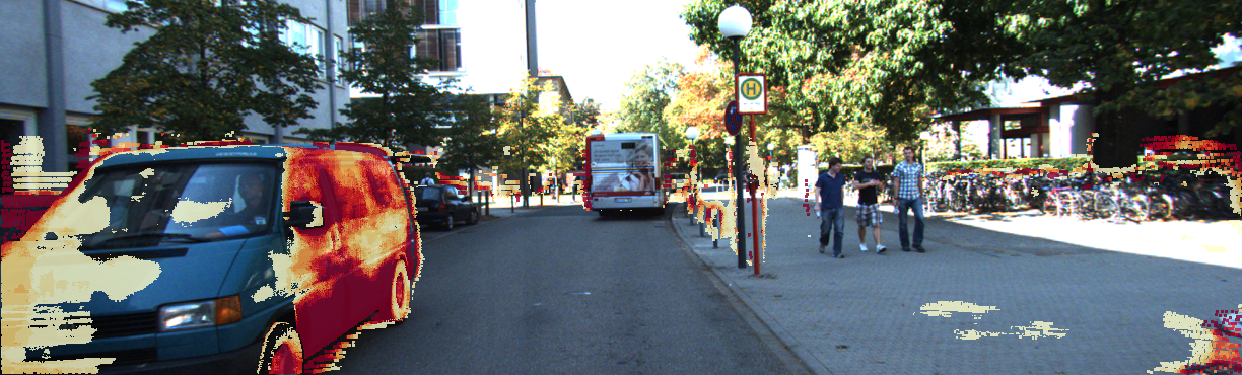}\\%
\includegraphics[width=0.5\linewidth]{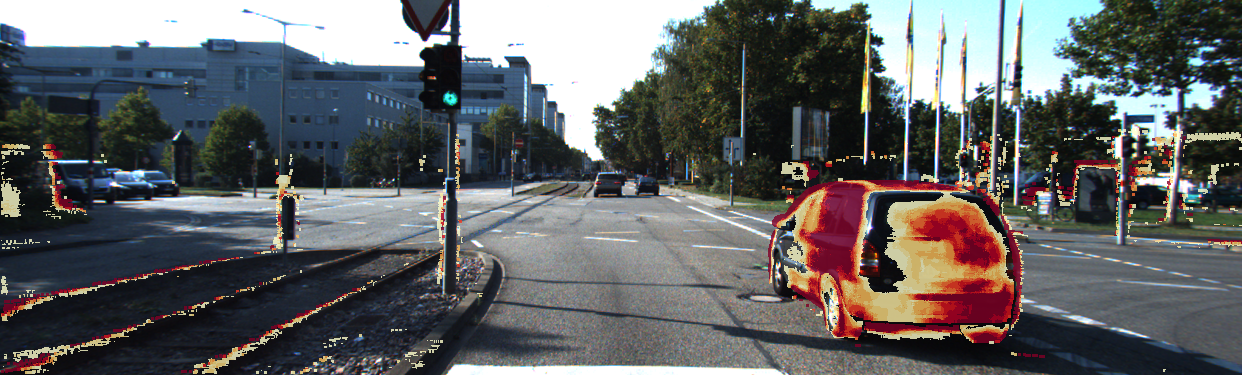}%
\includegraphics[width=0.5\linewidth]{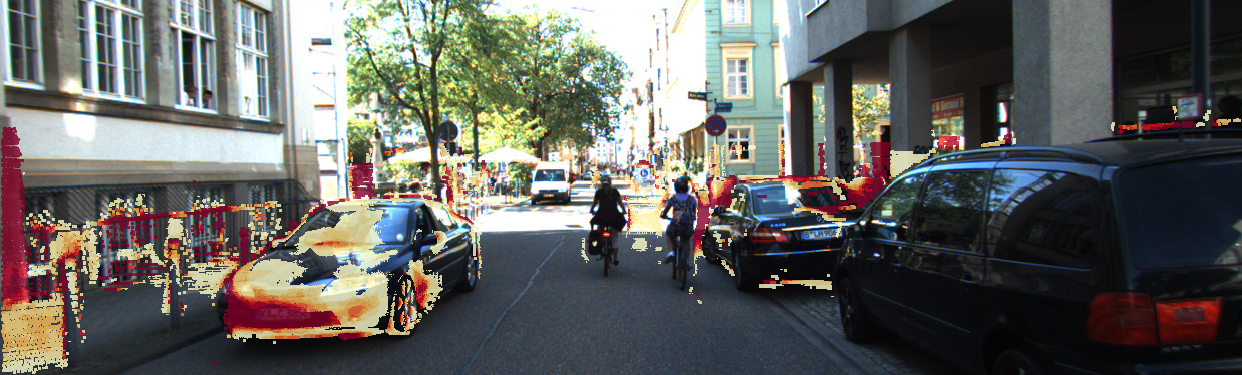}\\%
\includegraphics[width=0.5\linewidth]{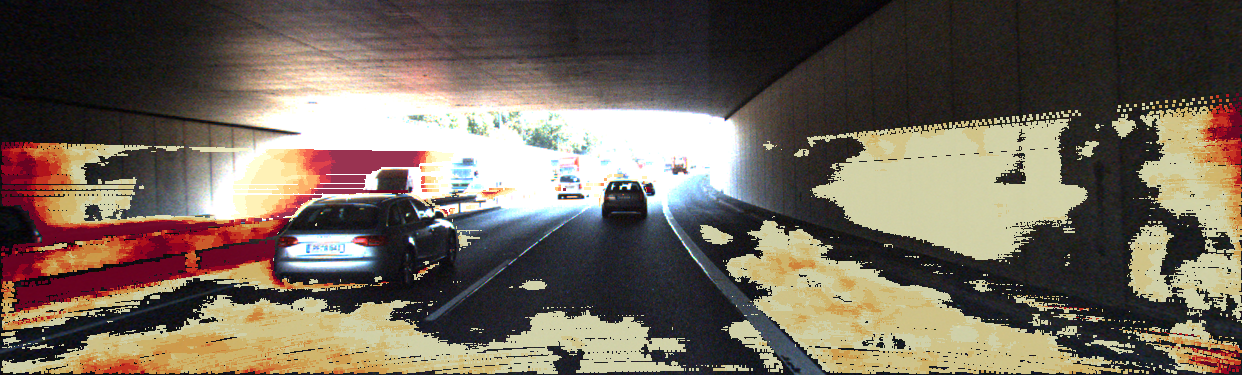}%
\includegraphics[width=0.5\linewidth]{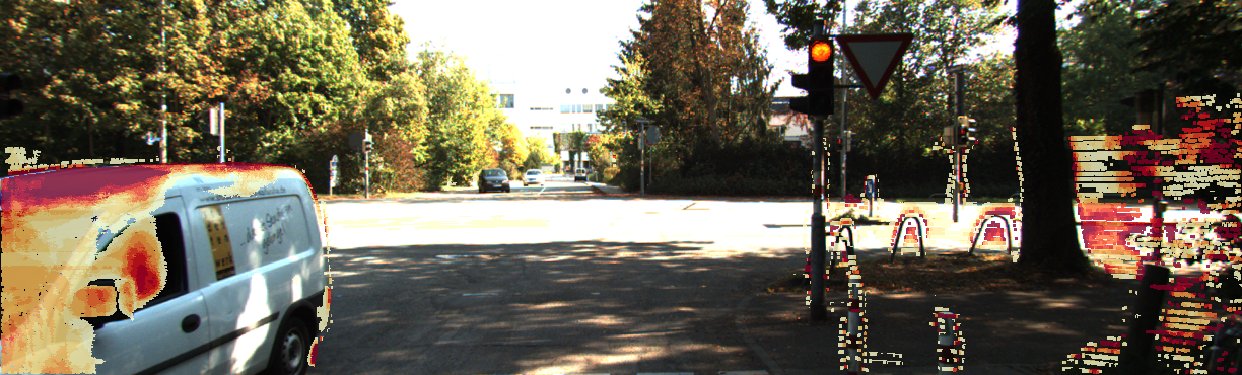}\\%
\includegraphics[width=0.5\linewidth]{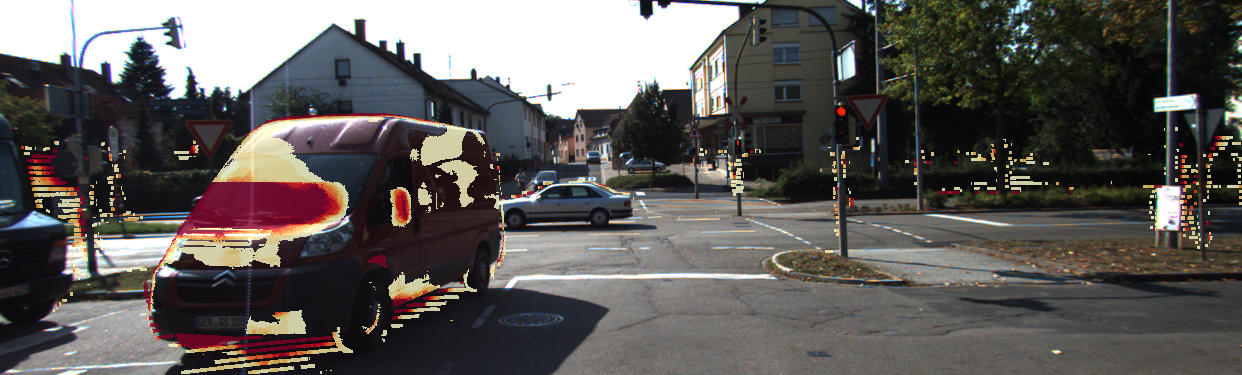}%
\includegraphics[width=0.5\linewidth]{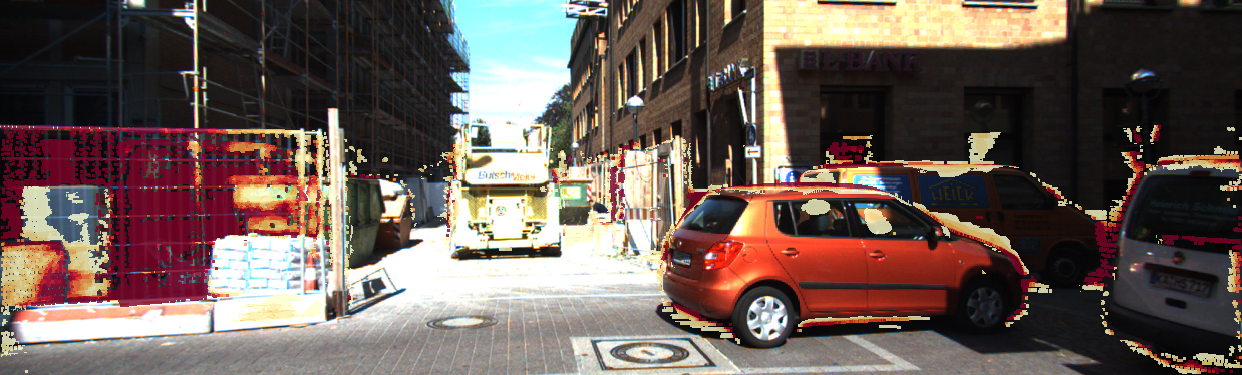}\\%
\includegraphics[width=0.5\linewidth]{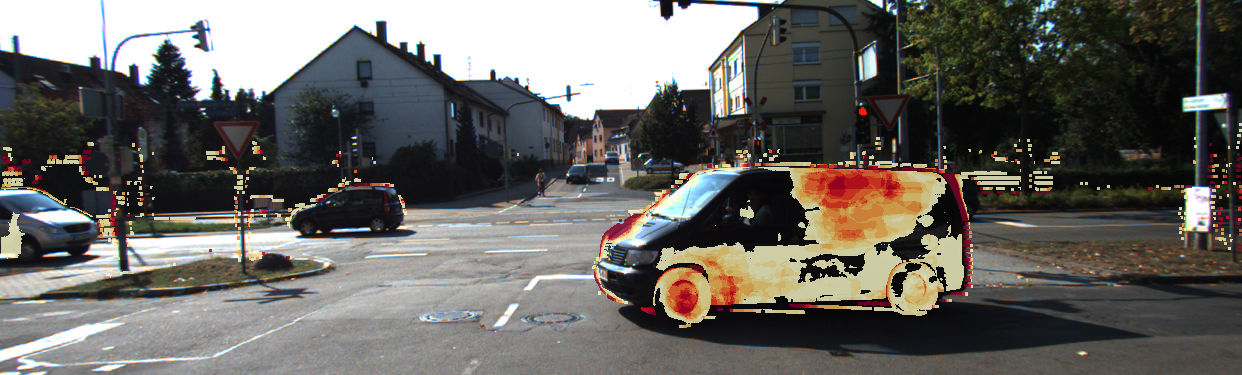}%
\includegraphics[width=0.5\linewidth]{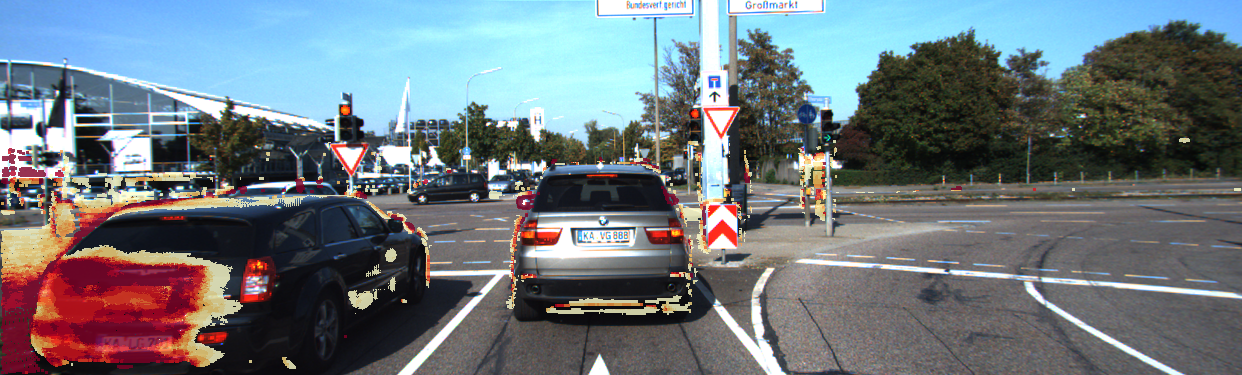}\\%
\includegraphics[width=0.5\linewidth]{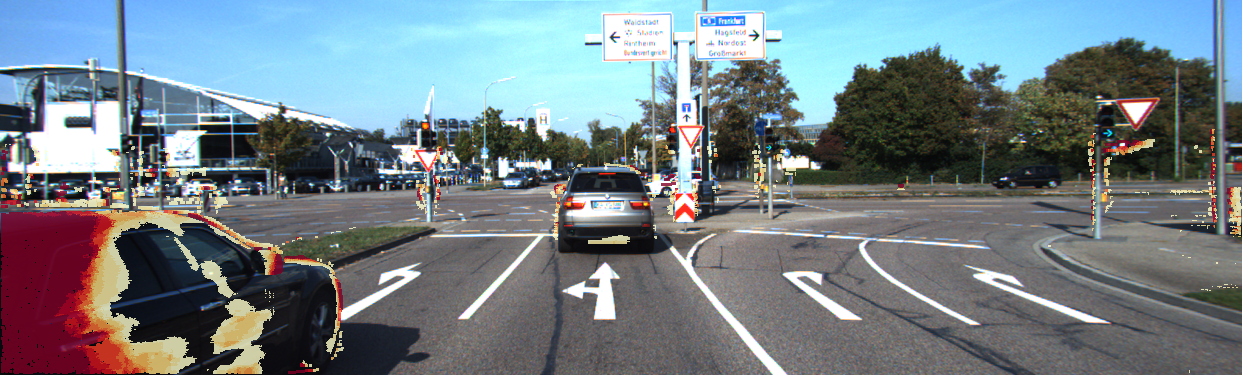}%
\includegraphics[width=0.5\linewidth]{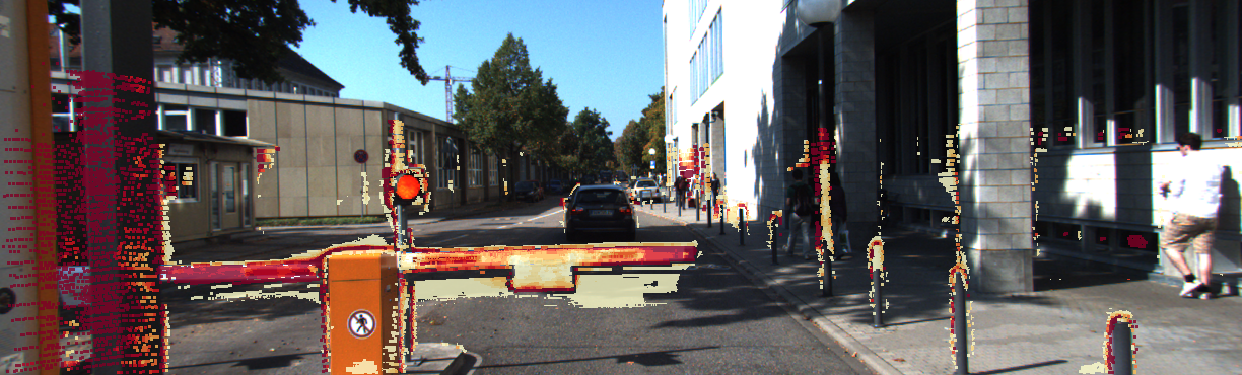}\\%
\includegraphics[width=0.5\linewidth]{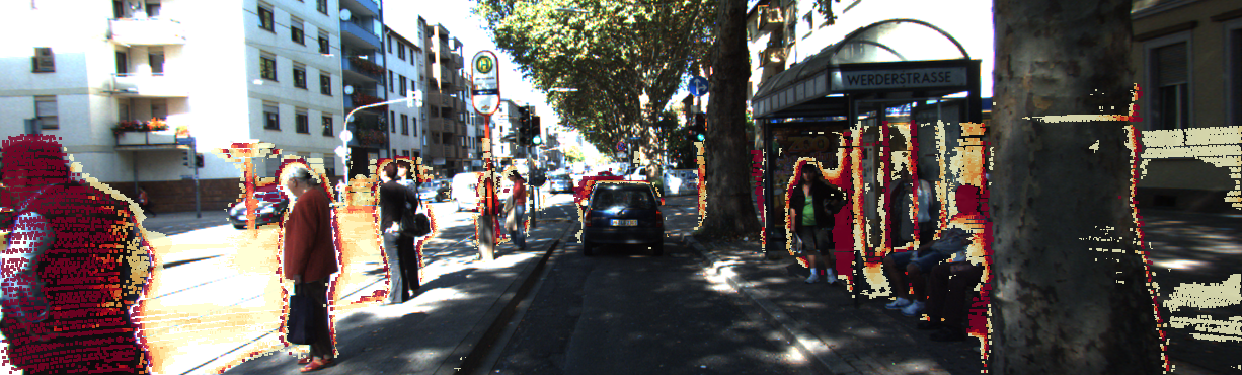}%
\includegraphics[width=0.5\linewidth]{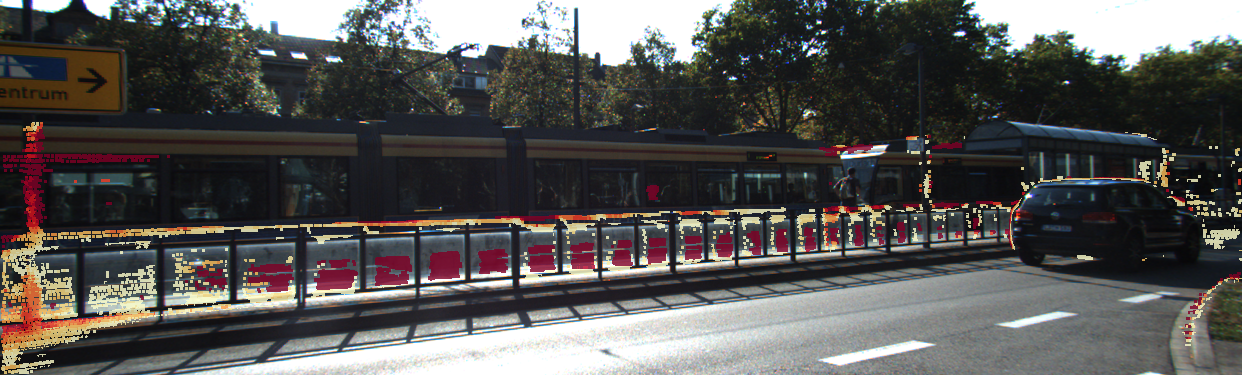}\\%
\includegraphics[width=0.5\linewidth]{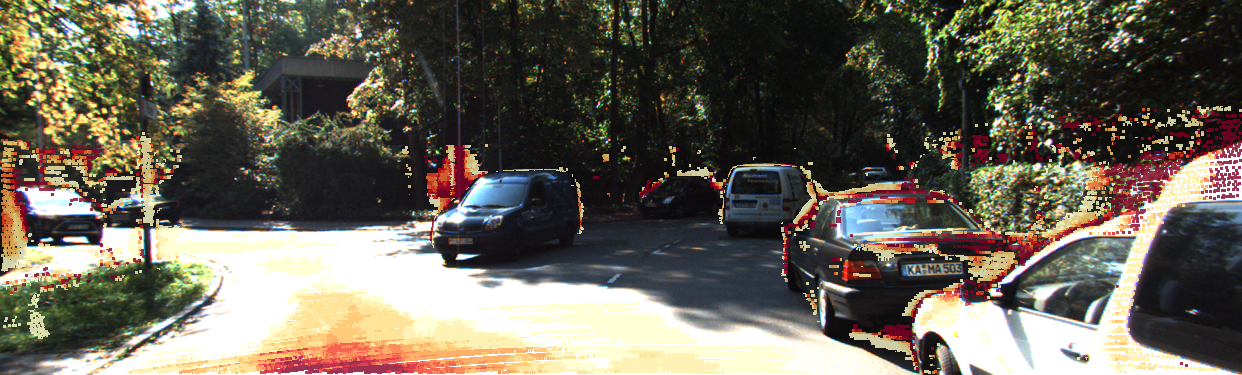}%
\includegraphics[width=0.5\linewidth]{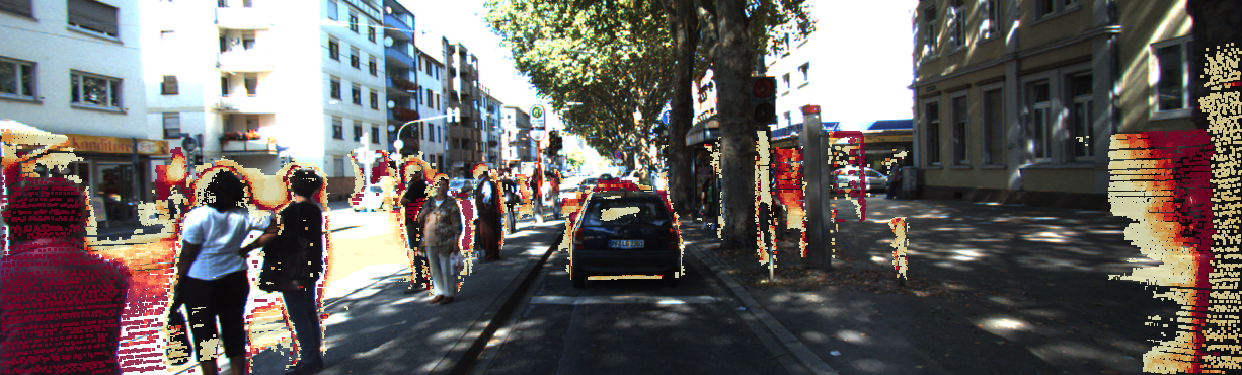}\\%
\caption{{\bf KITTI 2015 Stereo Analysis.} The averaged errors of the 15 best-performing stereo methods published on the KITTI 2015 Stereo benchmark. Red colors correspond to regions where the majority of methods fail according to the 3px/5\% criterion defined in \cite{Menze2015CVPR}. Yellow colors correspond to regions where some of the methods fail. Regions that are correctly estimated by all methods are transparent.}
\label{fig:stereo_qualitative_results}
\end{figure*}

Stereo estimation has shown great progress in the last years both in terms of accuracy and efficiency. However, some inherent problems prevent it from being considered solved. Stereo matching is equivalent to searching for correspondences in two images based on the assumption of constant appearance. However, appearance frequently changes due to non-rigidity or illumination changes. Furthermore, saturated pixels, occluded regions, or pixels leaving the frame cannot be matched. Therefore, failure in those cases is inevitable for methods that solely rely on appearance matching without any other prior assumptions about the geometry. We show the accumulated errors of the top 15 methods on the KITTI stereo benchmark \citep{Geiger2012CVPR} in \figref{fig:stereo_qualitative_results}. 
The most common examples of failure cases in the autonomous driving context are car surfaces that cause appearance changes due to their shiny and reflective nature. This problem can be addressed by leveraging more context information, \eg using joint formulations \citep{Yang2018ECCV,Song2018ACCV}. Similarly, windows that are reflective and transparent cannot be matched reliably. Occlusions are another source of error and require geometric reasoning beyond matching. Other examples of problematic regions include thin structures like traffic signs or repetitions as caused by fences. 
In these cases, continuous disparity estimation and the incorporation of more context information could be promising future directions. 

	\chapter{Multi-view 3D Reconstruction}
\label{chap:mv_reconstruction}
\section{Problem Definition}
\begin{figure}[t]
	\centering
	\includegraphics[width=1.00\columnwidth]{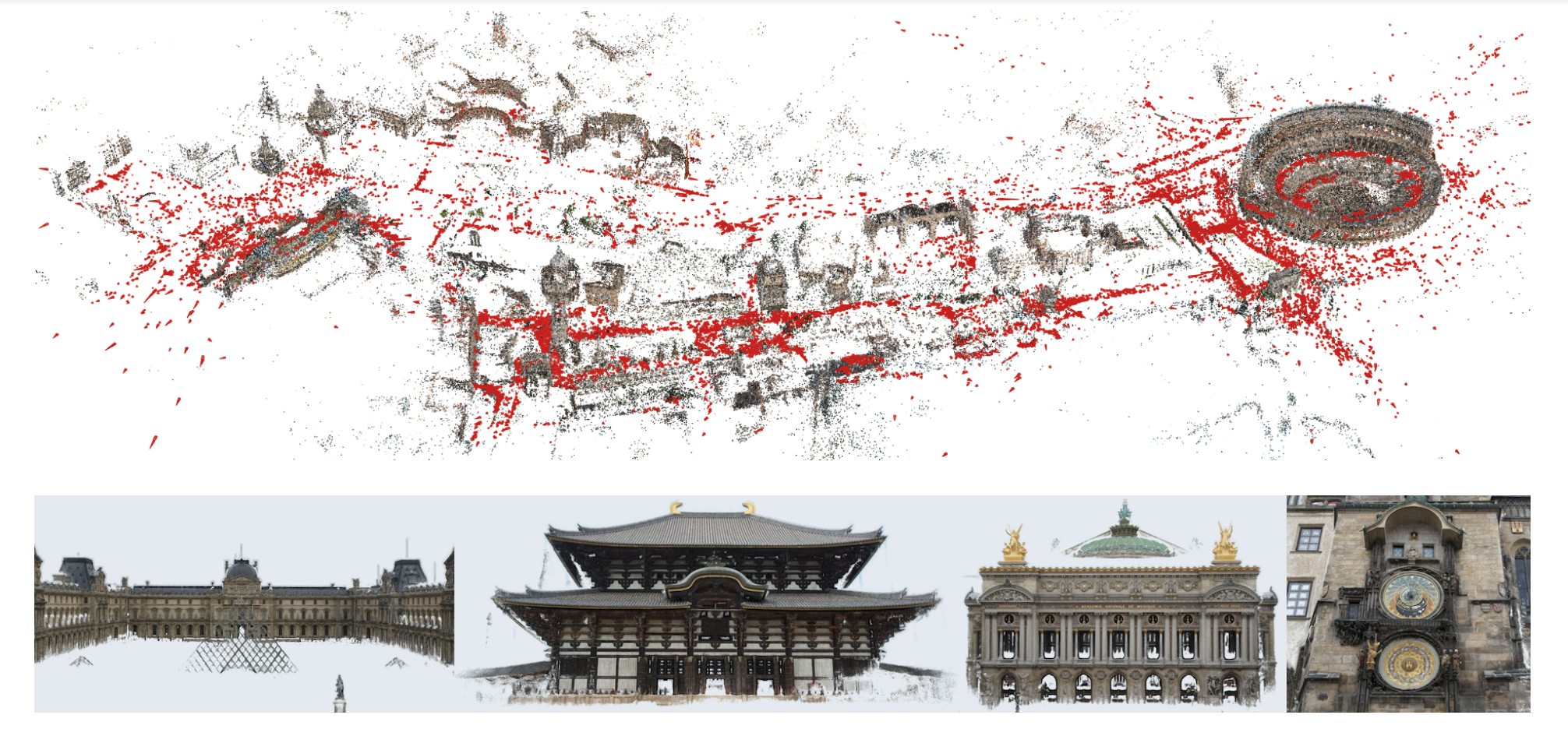}
	\caption[Large-Scale 3D Reconstruction]{\textbf{Large-Scale 3D Reconstruction.} \textbf{Upper row:} Coarse reconstruction and camera pose estimation (red) from the SfM \protect\citep{Schoenberger2016CVPR} pipeline of COLMAP. \figsourceC{\protect\citet{Schoenberger2016CVPR}}{2016}{IEEE}.
	\textbf{Bottom row:} Fine reconstruction with the MVS \protect\citep{Schoenberger2016ECCV} pipeline of COLMAP. \figsourceSpringer{\protect\citet{Schoenberger2016ECCV}}{2016}{ECCV}}
	\label{fig:Schoenberger2016ECCV}
\end{figure}
The goal of multi-view 3D reconstruction is to infer 3D geometry from a set of 2D images by inverting the image formation process using appropriate prior assumptions. In contrast to two-view stereo, multi-view reconstruction algorithms recover the complete 3D shape of an object by inferring shape from many viewpoints.

In this survey, we focus on multi-view reconstruction from an autonomous driving perspective which mainly concerns the reconstruction of urban areas. The goal of urban reconstruction algorithms is to produce fully automatic, high-quality, dense reconstructions of urban areas by addressing inherent challenges such as lighting conditions, occlusions, appearance changes, high-resolution inputs, and large scale outputs. In the context of autonomous driving, 3D reconstructions can be used for static obstacle detection (traffic lights, road signs, \etc) and avoidance or precise localization as discussed in \secref{sec:Localization}. 

\citet{Musialski2013CGF} provide a survey of urban reconstruction approaches by following an output-based ordering which considers buildings and semantics, facades and images, and finally, city blocks and cities. They list ground, aerial, and satellite imagery, as well as Light Detection and Ranging (LiDAR) scans as the most commonly used sensor modality for urban reconstruction. Ground-level imagery is the most prevalent one due to its ease of acquisition, storage, and exchange. However, more and more aerial and satellite images become available today as well. In contrast to aerial or multi-view imagery, satellite imagery provides worldwide coverage at low costs, but also with low resolution. LiDAR delivers semi-dense 3D point clouds at high precision, both ground-level and aerial, but the sensor is expensive and the data is sparse. Some approaches \citep{Frueh2005IJCV, Bodis-Szomoru2016ICPR} also combine these data types in order to leverage their complementary strengths. 
Several methods \citep{Vosselman2001ISPRS, Haala1997SPIE} leverage additional information, like Digital Surface Models (DSMs), which capture the Earth’s surface, to deal with challenging outdoor conditions. DSMs are $2.5D$ representations of an urban scene that provide a height for each surface point.

\section{Structure from Motion}
\label{sec:structure_from_motion}

In Structure from Motion (SfM), the camera parameters (intrinsic and extrinsic) need to be estimated jointly with the 3D structure while in Multi-View Stereo (MVS), the camera parameters are assumed to be known. Furthermore, while MVS approaches create a dense 3D model of the object or scene of interest, SfM approaches typically recover a sparse 3D point cloud of the scene. Solving for the camera parameters and 3D geometry of the scene is equivalent to solving the correspondence problem based on a photo-consistency function that measures the agreement between different viewpoints.
Typically, 3D reconstruction pipelines consist of an SfM method to estimate a coarse 3D reconstruction while recovering the camera parameters followed by an MVS method to obtain a finer reconstruction, as illustrated in \figref{fig:Schoenberger2016ECCV} using COLMAP\citep{Schoenberger2016CVPR, Schoenberger2016ECCV}. 

Classical SfM pipelines \citep{Snavely2006TG, Snavely2008IJCV, Cornelis2008IJCV, OpenMVG,  Agarwal2009ICCV, Wu2011VSFM, Fuhrmann2014GCH, Sweeney2016Theia} first extract and match sparse features. Usually, an initial transformation between pairs of cameras (essential matrix) is estimated with RANSAC. Given the initial camera transformations, a geometric verification stage evaluates photometric consistency between re-projected sparse features and excludes outliers.
Starting from an initial two-view reconstruction, an incremental reconstruction is performed based on best view selection, triangulation, and bundle adjustment. 
Due to this incremental approach, SfM pipelines are usually not very efficient and need to be applied offline. Simultaneous Localization and Mapping (SLAM) methods discussed in \secref{sec:slam} also address the problem of joint camera estimation (ego-motion) and 3D scene reconstruction. However, SLAM techniques focus primarily on accurate ego-motion estimation and real-time performance, typically sacrificing geometric accuracy for these goals.

The web provides large amounts of publicly available imagery from cities taken by tourists that can be used to reconstruct popular buildings or even entire cities. This task requires a different approach than the ones mentioned earlier because of the large amount of images and the unknown geometric properties of the cameras the images have been taken with. \citet{Agarwal2009ICCV} address this problem considering Flickr images of Rome. They use SIFT feature matching in combination with an efficient image retrieval approach to reduce the number of comparisons. Afterwards, a fast bundle adjustment method on minimal subsets of images captures the geometry of a scene. Finally, they optimize the whole pipeline in parallel, which allows them to reconstruct cities from 150K images in less than a day using 500 computing nodes. \citet{Frahm2010ECCV, Frahm2010JPRS} present a highly efficient system for city-scale reconstruction from millions of images on a single computer by leveraging the high parallelization capabilities of graphics hardware. Recently, \citet{Schoenberger2016CVPR} proposed a structure-from-motion pipeline with better completeness and accuracy while better reducing drift in comparison to previous methods \citep{Snavely2006TG, Snavely2008IJCV, Agarwal2009ICCV, Frahm2010ECCV, Frahm2010JPRS, Wu2011VSFM, Fuhrmann2014GCH}. They further propose a more robust best view selection and triangulation method, producing more complete structures. Finally, a novel iterative Bundle Adjustment, re-triangulation, and outlier filtering step lead to significantly more complete and accurate 3D models.

\section{Multi-view Stereo}
Multi-view stereo approaches can be classified according to their scene representation into depth map-, point cloud-, mesh-, and volumetric-based methods. We first introduce and discuss classical approaches by grouping them based on their scene representation (Depth Maps, Point Clouds, Volumetric) and the final representation of the reconstruction (Mesh or Surfaces). 

\boldparagraph{Depth Map}
The depth map representation summarizes a 3D scene using one 2.5D depth map for each input view. These depth maps can later be fused into a single coherent 3D reconstruction using 3D fusion techniques \cite{Zach2007ICCV,Curless1996SIGGRAPH,Riegler2017THREEDV}. One strategy which is particularly effective for recovering depth maps from urban scenes is the Plane Sweeping Stereo algorithm \citep{Collins1996CVPR}. This algorithm ``sweeps'' a family of parallel hypothetical planes through the scene, projects images into each other via the homography induced by these planes, and evaluates photo-consistency. In very large scenes, one of the primary challenges is to handle large amounts of data efficiently. \citet{Pollefeys2008IJCV} proposes a large scale, real-time MVS system based on the depth map representation by exploiting the parallel processing capabilities of modern GPUs.

\begin{figure}[t]
	\centering
	\includegraphics[width=1.00\columnwidth]{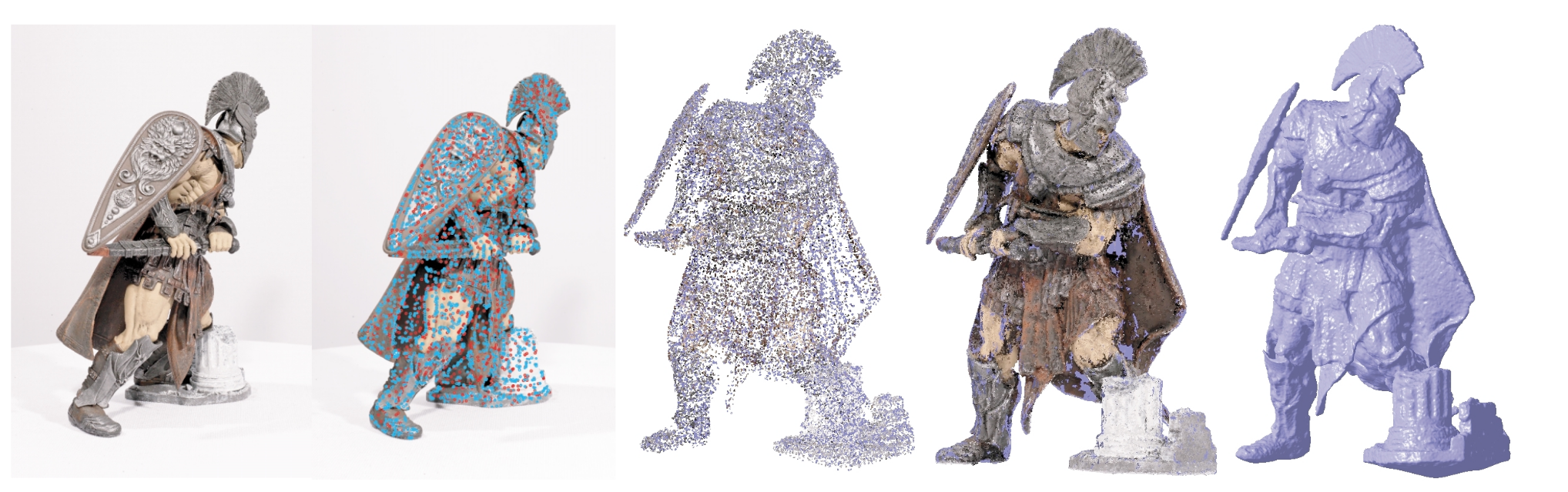}
	\caption[Point Cloud and Surface Representation]{\textbf{Point Cloud and Surface Representation.} The different steps of Patch-based Multi-View Stereo (PMVS) \protect\citep{Furukawa2010PAMI}. The input image, extracted features, reconstructed patches from the initial matching, reconstruction after expansion and filtering, and the final polygonal surface representation. \figsourceC{\protect\citet{Furukawa2010PAMI}}{2010}{IEEE}.}
	\label{fig:Furukawa2010PAMI}
\end{figure}

\boldparagraph{Point Cloud}
The reconstruction problem can also be addressed with a 3D point cloud representation \citep{Furukawa2010PAMI,Schoenberger2016CVPR}. Patch-based Multi-View Stereo (PMVS) \citep{Furukawa2010PAMI} starts with a feature matching step to generate a sparse set of patches and then iterates between a greedy expansion step and a filtering step to make patches dense and remove erroneous matches. The steps of PMVS are visualized in \figref{fig:Furukawa2010PAMI}.

\boldparagraph{Volumetric Representation}
Volumetric approaches represent geometry using a regularly sampled 3D grid, \ie volume, either as a discrete occupancy function \citep{Kutulakos2000IJCV} or a function encoding distance to the closest surface (level-set) \citep{Faugeras1998TIP}. More recent approaches use a probability map defined at regular voxel locations to encode the probability of occupancy \citep{Bhotika2002ECCV, Pollard2007CVPR, Ulusoy2015THREEDV}. The amount of memory required is the main limitation of volumetric approaches. There exists a variety of proposals for dealing with this problem, such as voxel hashing \citep{Niesner2013SIGGRAPH}, data-adaptive discretization of the space in the form of Delaunay triangulation \citep{Labatut2007ICCV}, or using octrees \citep{Haene2013CVPR, Riegler2017CVPR}.

\boldparagraph{Mesh or Surfaces}
The final representation of a 3D reconstruction algorithm is typically a triangular mesh-based surface (right image in \figref{fig:Furukawa2010PAMI}). Volumetric surface extraction techniques can fuse multiple 2.5D measurements (MVS depth maps or laser scans) into a single, coherent 3D mesh model. Seminal work by \citet{Curless1996SIGGRAPH} proposes an algorithm to accumulate surface evidence into a voxel grid using signed distance functions. The surface is implicitly represented as the zero crossing of the aggregated signed distance functions and can be extracted using the Marching Cube algorithm \citep{Lorensen1987SIGGRAPH} to label each voxel as either interior or exterior. Other approaches directly start from images \citep{Cootes1995CVIU, Delaunoy2011IJCV, Delaunoy2014CVPR} and refine a mesh model using an energy function composed of a data and a regularization term.

\begin{figure}[t]
	\centering
	\includegraphics[width=0.70\columnwidth]{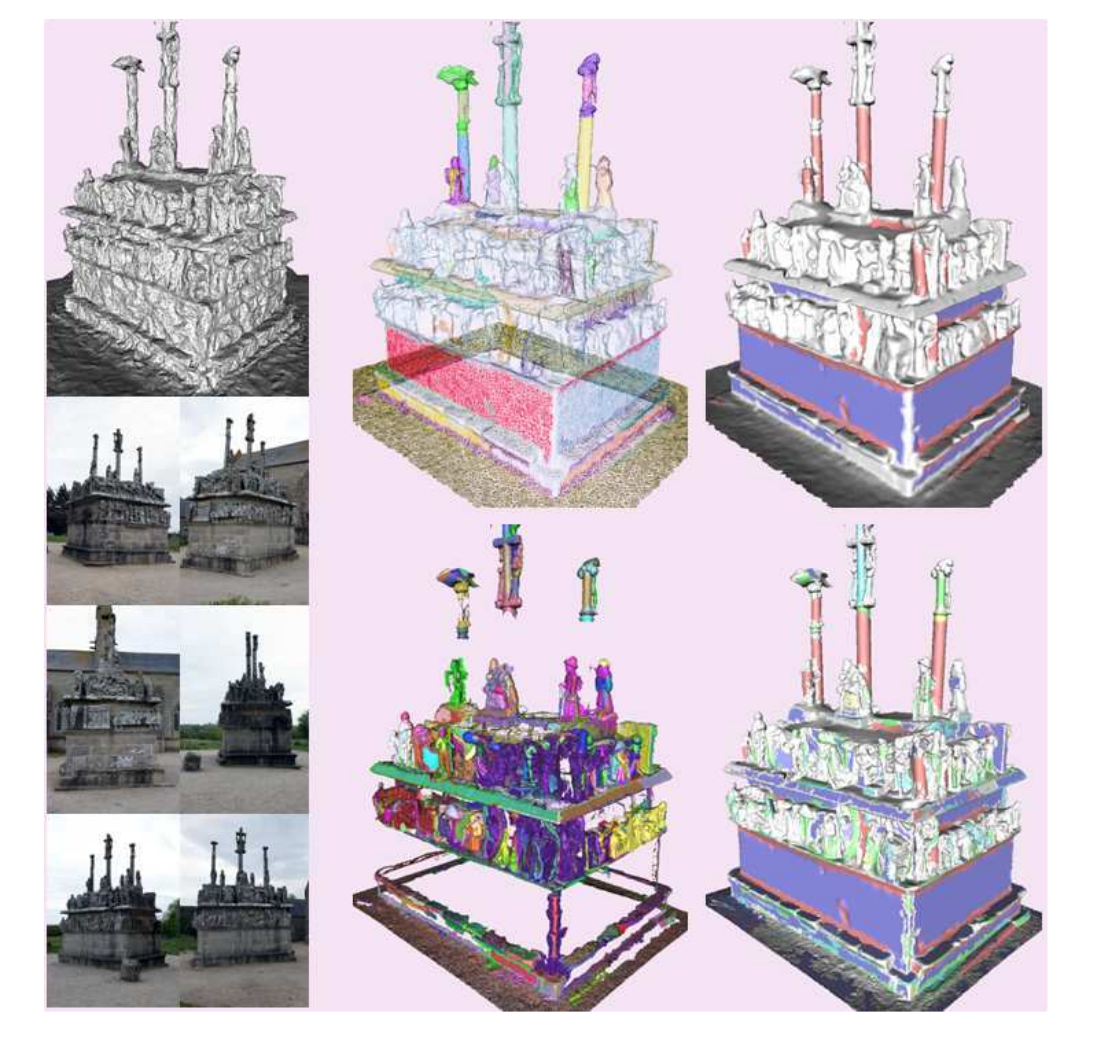}
	\caption[Primitive-based Reconstruction]{\textbf{Primitive-based Reconstruction.} The hybrid reconstruction approach of \protect\citet{Lafarge2013PAMI} uses primitives for regular structures (top right) and meshes for irregular structures (bottom left) to compactly represent a coarse initial mesh (top left). \figsourceC{\protect\citet{Lafarge2013PAMI}}{2013}{IEEE}.}
	\label{fig:Lafarge2013PAMI}
\end{figure}

\subsection{Planarity and Primitives}
Man-made environments usually consist of regular structures. The introduction of appropriate priors, therefore, allows for more accurate and dense reconstructions. \citet{Micusik2009CVPR} present a method exploiting image segmentation cues as well as the presence of dominant scene orientations and piecewise planar structures. In particular, they adopt a super-pixel-based dense stereo reconstruction method exploiting the Manhattan world assumption in their MRF formulation. Another way of exploiting piecewise planar structures and repetitive shapes is to detect primitives such as planes, spheres, cylinders, cones, and tori \citep{Lafarge2010PAMI, Lafarge2012IJCV, Lafarge2013PAMI}. Primitive-based approaches lead to compact and memory-efficient representations. However, their representations are often simplistic and fail to model fine details and irregular shapes. Therefore, \citet{Lafarge2013PAMI} propose a hybrid approach that is both compact and detailed. Starting from an initial mesh-based reconstruction, they use primitives for regular structures such as columns and walls, while irregular elements are described using triangular meshes for preserving architectural details (\figref{fig:Lafarge2013PAMI}). 

\subsection{Shape Priors} 
Advances in sensors to acquire 3D shapes and the performance of object detection algorithms have encouraged the use of 3D shape priors in multi-view stereo approaches. Dimensionality reduction is an effective and popular way of representing shape knowledge. Early approaches \citep{Tsai2003MedicalImaging} use linear dimensionality reduction such as Principal Component Analysis (PCA) to capture shape variance in low dimensional latent shape spaces. 
More recent approaches, like \citet{Dame2013CVPR}, who investigate the importance of shape priors in a monocular SLAM approach, use non-linear dimensionality reduction techniques such as Gaussian Process Latent Variable Models (GP-LVM). In parallel with depth estimation, they refine an object’s pose, shape, and scale to match an initial segmentation and depth cues. Their experiments show improvements on transparent and specular surfaces, and even in unobserved parts of the scene.

In addition to the mean shape, \citet{Bao2013CVPR} propose to learn a set of anchor points to represent object shape across several instances. 
They first perform an initial alignment of the mean shape to the point cloud from SfM using 2D object detectors. Finally, they warp and refine the mean shape to approximate the actual shape. Their evaluation demonstrates that the model is general enough to learn semantic priors for different object categories by handling large shape variations across instances.

An alternative to using latent space representations is to directly leverage 3D CAD models provided by free 3D model repositories. \citet{Guney2015CVPR} propose a model for jointly inferring disparity maps and the geometry, pose, and type of 3D car models in urban scenes. \citet{Ulusoy2017CVPR} extend this approach to the volumetric multi-view case.
While the approaches mentioned earlier \citep{Dame2013CVPR, Bao2013CVPR} fit a parametric shape model to input data, \citet{Haene2013CVPR,Haene2014CVPR} model the local distribution of normals for an object. They also propose an object class-specific shape prior in the form of spatially varying anisotropic smoothness terms.

\citet{Zhou2015ICCV} propose to jointly learn volumetric shape models for 3D reconstruction of street scenes from a sequence of fisheye cameras. Motivated by recurring objects of similar 3D shapes in outdoor scenes, they first localize buildings and vehicles using 3D object detectors and then jointly reconstruct them while learning a volumetric model of their shape. This allows the reduction of noise while completing missing surfaces as objects of similar shape benefit from all observations of the respective category. Instead of modeling a semantic prior for each object explicitly, \citet{Wei2014THREEDV} propose a data-driven regularization to transfer shape information from semantically matched patches in the training database using the SIFT flow algorithm.

\begin{figure}[t]
	\centering
	\includegraphics[width=1.00\columnwidth]{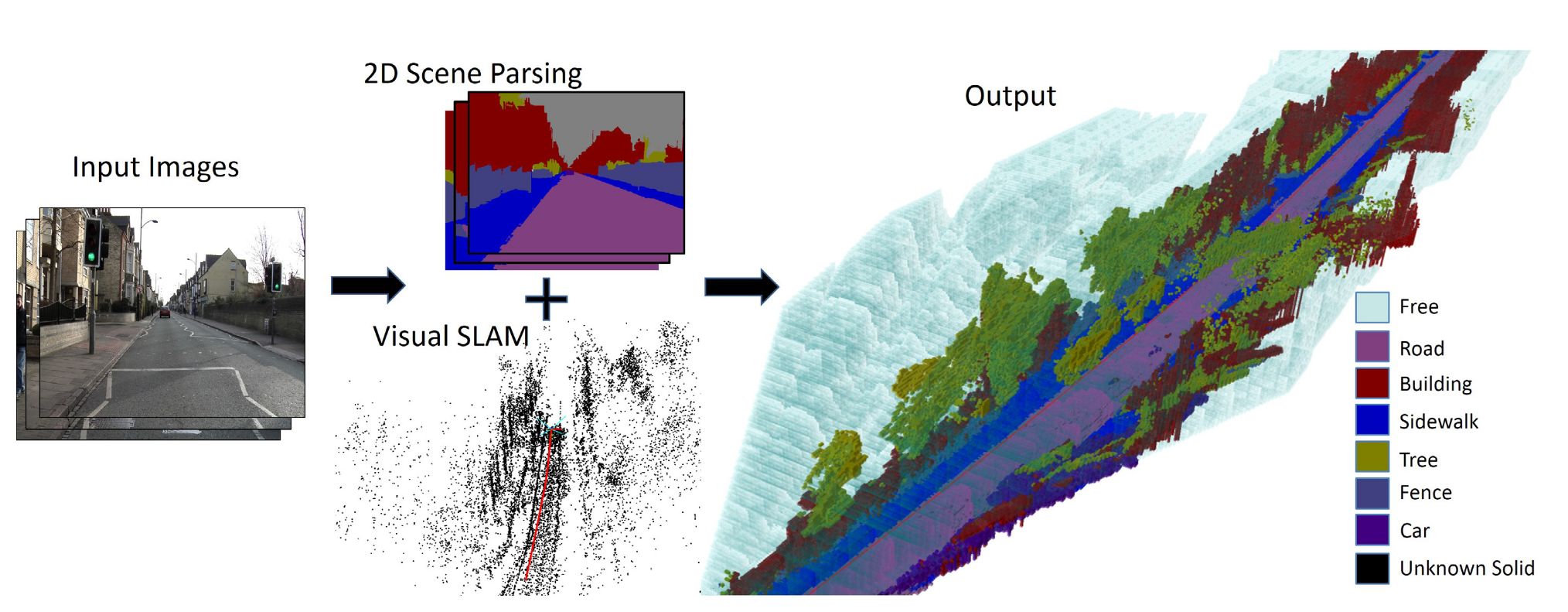}
	\caption[Joint Reconstruction and Semantic Segmentation]{\textbf{Joint Reconstruction and Semantic Segmentation.} Joint 3D scene reconstruction and segmentation by \protect\citet{Kundu2014ECCV}. \figsourceSpringer{\protect\citet{Kundu2014ECCV}}{2014}{ECCV}.}
	\label{fig:semantic_reconstruction}
\end{figure}

\subsection{Semantics} 
Similar to stereo, semantic information allows multi-view stereo approaches to recover from potential failures of photo-consistency in case of imperfect and ambiguous image information, \eg specularities, lack of texture, repetitive structures, or strong lighting changes. Semantic labels provide geometric cues about likely surface orientations at a certain location and help to resolve inherent ambiguities as illustrated in \figref{fig:semantic_reconstruction} by the joint reconstruction and semantic segmentation approach of \citet{Kundu2014ECCV}.

Volumetric scene reconstruction typically segments the volume into occupied and free-space regions. \citet{Haene2013CVPR} present the mathematical framework to extend this approach to multi-label volumetric segmentation, assigning object classes or a free-space label to voxels. They first learn appearance likelihoods and class-specific geometry priors for surface orientations from the training data. Afterwards, these data-driven priors are used to define unary and pairwise potentials in a continuous formulation for volumetric segmentation.

\citet{Haene2013CVPR} require dense depth measurements, which can be difficult to obtain because of textureless regions and low parallax. Thus, \citet{Kundu2014ECCV} propose another approach working on sparse 3D point clouds. They model the problem using a higher-order Conditional Random Field in 3D, which allows them to impose realistic scene constraints and priors such as 3D object support. In addition, they explicitly model-free space, which provides cues to reduce ambiguities, especially along weakly supported surfaces. Their evaluation on the CamVid and Leuven datasets shows improved 3D structure compared to traditional SfM and state-of-the-art MVS pipelines as well as better segmentation quality over video segmentation methods.

Previous works on semantic reconstruction \citep{Haene2013CVPR, Kundu2014ECCV} are limited to small scenes and low resolutions due to their large memory footprint and computational cost. 
In order to scale to larger scenes, \citet{Blaha2016CVPR} note that high resolution is not required for large regions such as free space, parts under the ground, or inside the building. They propose an extension of \citet{Haene2013CVPR} and employ an adaptive octree data structure with coarse-to-fine optimization to generate 3D city models from terrestrial and aerial images. Starting from a coarse voxel grid, they solve a sequence of problems in which the solution is gradually refined near the predicted surfaces. The adaptive refinement saves memory and runs much faster while still being as accurate as the fixed voxel discretization at the highest target resolution, both in terms of geometric reconstruction and semantic labeling. Besides the spatial extent, the number of different semantic labels is also problematic for scalability due to the increasing memory requirements. \citet{Cherabier2016THREEDV} propose to divide the scene into blocks in which only a set of relevant labels is active. Thus, the absence of semantic classes from a specific block can be determined early on. Accordingly, they deactivate labels from the beginning of the optimization, which leads to more efficient processing.

\subsection{Efficient Reconstruction}
The extraction of detailed 3D information from video streams leads to high computational costs for multi-view stereo algorithms. \citet{Cornelis2008IJCV} focus on creating compact, memory-efficient 3D city models from a stereo pair at high frame rates based on simplified geometry assumptions such as ruled surfaces for building facades. Since objects such as cars violate these assumptions, they integrate the detection and localization of cars into the reconstruction. In contrast, \citet{Geiger2011IV} propose an efficient stereo matching algorithm to generate accurate piece-wise planar 3D reconstructions in real-time.

\begin{figure}[t]
	\centering
	\includegraphics[width=1.00\columnwidth]{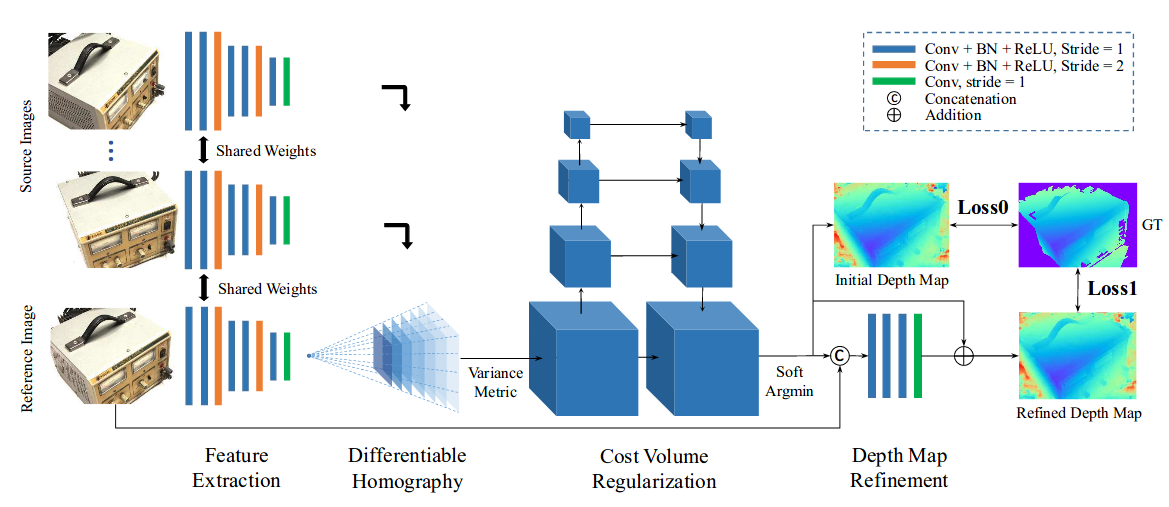}
	\caption[Deep Learning for Multi-View Stereo]{\textbf{Deep Learning for Multi-View Stereo.} MVSNet by \protect\citet{Yao2018ECCV} comprising feature extraction networks and a differentiable homography warping stage for constructing cost volumes. The final depth map is obtained using a refinement network. \figsourceSpringer{\protect\citet{Yao2018ECCV}}{2018}{ECCV}.}
	\label{fig:Yao2018ECCV}
\end{figure}

\subsection{Deep Learning for Multi-View Stereo}
\label{sec:mvs_deep_learning}
Several learning-based approaches have been proposed to address the Multi-View Stereo problem. In early works, learning was mainly used to obtain more robust feature representations for establishing better correspondences \citep{Han2015CVPR, Zbontar2016JMLR, Luo2016CVPR}. Pairwise similarities obtained from these approaches are usually averaged in order to match features from multiple images. In contrast, \citet{Hartmann2017ICCV} propose to directly learn a matching function using multiple images as input.

Recently, several pipelines for end-to-end learning of multi-view reconstruction \citep{Kar2017NIPS, Ji2017ICCV, Paschalidou2018CVPR, Huang2018CVPRa} have been presented combining learned high-level information with classical constraints. \citet{Kar2017NIPS, Ji2017ICCV} propose to unproject features along the viewing rays onto a 3D feature grid for matching. Afterwards, both approaches use 3D convolutional networks to smooth the 3D feature grid. In contrast, \citet{Huang2018CVPRa} propose to learn disparities by combining a plane-sweep approach with an end-to-end trained CNN for feature extraction. Considering a reference view, they create plane-sweep volumes consisting of neighboring views warped according to the hypothetical depth. Afterwards, they extract features using a network on patch pairs (patches from the reference view and plane-sweep volumes). An encoder-decoder network with skip connections is used to combine features over larger regions and, eventually, the disparity map is estimated with a max-pooling layer. While previous approaches leverage physical constraints by projecting features according to the camera transformation, they do not model occlusion relationships. \citet{Paschalidou2018CVPR} combine learning-based feature extraction with a Markov Random Field that employs high-order ray-potentials \citep{Ulusoy2015THREEDV} to model the image formation process and occlusions. 

Inference with 3D feature grids \citep{Kar2017NIPS, Ji2017ICCV},using a voxel grid \citep{Paschalidou2018CVPR} or plane-sweep volumes \citep{Huang2018CVPRa}, is computationally expensive. A more efficient MVS reconstruction approach was presented by \citet{Yao2018ECCV}, see \figref{fig:Yao2018ECCV}. Similar to \citet{Huang2018CVPRa}, they decouple the MVS reconstruction problem into a depth prediction problem for each view. A differentiable homography warping operation allows them to encode the camera geometry and to build a 3D cost volume. A 3D convolutional network predicts the depth from the 3D cost volume, and the final reconstruction is obtained with a depth map fusion approach \citep{Merrell2007ICCV} which minimizes depth occlusions and differences between viewpoints.

\subsection{Omnidirectional Cameras} 
While omnidirectional cameras, as discussed in \secref{sec:calibration_omnidirectional_cam}, provide a larger field of view compared to traditional perspective cameras, their special geometric properties need to be addressed during 3D reconstruction. The epipolar geometry of central catadioptric systems was explored by \citet{Svoboda2002IJCV}, who showed that correspondences lie on epipolar conics and who proposed a rectification procedure for this setup. In contrast, \citet{Bunschoten2003TRA,Gonzalez-Barbosa2005ICRA} propose to project the omnidirectional image to a panoramic view and use standard stereo matching methods to search for correspondences. While \citet{Bunschoten2003TRA} search on sinusoidal shaped epipolar curves, \citet{Gonzalez-Barbosa2005ICRA} rectify the panoramic view to obtain straight epipolar lines.
\citet{Schoenbein2014IROS} propose a method for 3D reconstruction through joint optimization of disparity estimates from two temporally and two spatially adjacent omnidirectional views in a unified omnidirectional space using plane-based priors.

\section{Datasets}
Several datasets have been proposed to evaluate multi-view stereo algorithms. 
Popular datasets include Middlebury \citep{Scharstein2002IJCV} and DTU MVS \citep{Jensen2014CVPR}. However, these datasets provide only a few or no examples for urban reconstruction. While EPFL Multi-View \citep{Strecha2008CVPR}, \citet{Restrepo2014JPRS}, ETH3D \citep{Schoeps2017CVPR} and Tanks and Temples \citep{Knapitsch2017SIGGRAPH} provide urban scenes, they do not focus on the autonomous driving task. 

Large-scale reconstruction methods \citep{Agarwal2009ICCV, Frahm2010ECCV, Frahm2010JPRS, Schoenberger2016CVPR} typically use the BigSFM dataset \footnote{\url{http://www.cs.cornell.edu/projects/bigsfm/}}, a collection of smaller datasets from Cornell University which consists of Vienna \citep{Irschara2009CVPR}, Dubrovnik \citep{Li2010ECCV}, Rome and Quad datasets \citep{Crandall2011CVPR}. However, these datasets do not have ground truth data and, therefore, a quantitative evaluation of methods is not possible.

As ETH3D and Tanks and Temples are the MVS datasets closest to the autonomous driving scenario and also provide an online evaluation server, we focus our discussion on these two datasets. As opposed to ETH3D \citep{Schoeps2017CVPR}, Tank and Temples \citep{Knapitsch2017SIGGRAPH} does not provide camera poses and thus an additional structure-from-motion pipeline \citep{Snavely2008IJCV, Wu2011VSFM, OpenMVG, Fuhrmann2014GCH, Sweeney2016Theia, Schoenberger2016CVPR} is necessary to estimate camera poses.

\section{Metrics}
In MVS, the accuracy and completeness of the output reconstruction are standard measures for evaluation. Accuracy is defined as the percentage of estimated points with a distance smaller than a predefined threshold to the closest ground truth points. Completeness is defined as the percentage of ground truth points with a distance smaller than a predefined threshold to the closest estimated points. Some benchmarks also report the mean (Chamfer) or harmonic mean (F1-measure) of accuracy and completeness. 

\section{State of the Art on ETH3D \& Tanks and Temples}
\begin{table*}[t]
\begin{center}
\begin{adjustbox}{width=1\textwidth}\begin{tabular}
{@{}cl@{\hspace{8mm}}|c|ccc|ccc@{}}
	&  &  & \multicolumn{3}{c|}{\bf Low-Res Many-View} & \multicolumn{3}{c}{\bf High-Res Multi-View} \\
	& {\bf Method} & {\bf All} & {\bf All} & {\bf Indoor} & {\bf Outdoor} & {\bf All} & {\bf Indoor} & {\bf Outdoor} \\
	\hline
1. & ACMM \citep{Xu2019CVPR}  &  & & & & 80.78 & 79.84 & 83.58 \\
2. & OpenMVS \citep{OpenMVS} & 72.83 & 56.18 & 45.66 & 63.19 & 79.77 & 78.33 & 84.09 \\
3. & LTVRE\_ROB \citep{Kuhn2017IJCV} & 69.57 & 53.52 & 45.46 & 58.89 & 76.25 & 74.54 & 81.41 \\
4. & ACMH \citep{Xu2019CVPR} & 67.68 & 47.97 & 38.24 & 54.45 & 75.89 & 73.93 & 81.77 \\
5. & COLMAP\_ROB \citep{Schoenberger2016CVPR,Schoenberger2016ECCV} & 66.92 & 52.32 & 42.45 & 58.89 & 73.01 & 70.41 & 80.81 \\
6. & OpenMVS\_ROB \citep{OpenMVS} & 64.09 & 48.56 & 38.68 & 55.15 & 70.56 & 68.19 & 77.65 \\
7. & CMP-MVS \citep{Jancosek2011CVPR} & 51.72 & 7.38 & 0.03 & 12.27 & 70.19 & 68.16 & 76.28 \\
8. & Gipuma \citep{Galliani2015ICCV} &  &  & & & 45.18 & 41.86 & 55.16 \\
9. & PMVS \citep{Furukawa2010PAMI} & 37.38 & 21.09 & 11.49 & 27.48 & 44.16 & 40.28 & 55.82 \\
10. & MVE \citep{Fuhrmann2014GCH} & 26.22 & 16.26 & 16.97 & 15.79 & 30.37 & 25.89 & 43.81
\end{tabular}
\end{adjustbox}
\end{center}
\vspace{-0.4cm}
\caption{{\bf ETH3D Leaderboard.} Evaluation results on two ETH3D \citep{Schoeps2017CVPR} challenges: low-resolution multi-view stereo from video data (many-view) and high-resolution multi-view stereo on few images recorded with a DSLR. The average F-measure is reported. Accessed on: May 2019.}
\label{tab:eth3d}
\end{table*}

\begin{table*}[t]
\begin{center}
\begin{adjustbox}{width=1\textwidth}\begin{tabular}
{@{}l|cc|cc@{}}
 & \multicolumn{2}{c}{\bf Intermediate} & \multicolumn{2}{c}{\bf Advanced} \\
{\bf Method} & {\bf Rank} & {\bf F-measure}  & {\bf Rank} & {\bf F-measure}\\
\hline
ACMM \citep{Xu2019CVPR} &  1. & 57.27 & 1.  & 34.02  \\
ACMH \citep{Xu2019CVPR} & 2. &  54.82 & 2. & 33.73 \\
Dense R-MVSNet \citep{Yao2019CVPR} & 3. &  50.55 & 3.  & 29.55  \\
R-MVSNet \citep{Yao2019CVPR} & 4. & 48.40 & 5. & 24.91  \\
MVSNet \citep{Yao2018ECCV} & 5. & 43.48 &   \\
COLMAP \citep{Schoenberger2016CVPR,Schoenberger2016ECCV} & 6. & 42.14 & 4. & 27.24  \\
\hline
VisualSfM \citep{Wu2011VSFM} + PMVS \citep{Furukawa2010PAMI} & 14. & 27.80 & 15. & 10.22  \\
VisualSfM \citep{Wu2011VSFM} + CMP-MVS \citep{Jancosek2011CVPR} & 18. & 22.40 & 17. &  7.57 \\
Bundler \citep{Snavely2006TG,Snavely2008IJCV} + PMVS \citep{Furukawa2010PAMI} & 19. &  12.86 & 18. &  5.61
\end{tabular}
\end{adjustbox}
\end{center}
\vspace{-0.4cm}
\caption{{\bf Tanks and Temples Leaderboard.} Evaluation results for intermediate and advanced scenes from Tanks and Temples \citep{Knapitsch2017SIGGRAPH}. The rank and average F-measure are reported. Methods below the horizontal line show older entries for reference. Accessed on: June 2019.}
\label{tab:tanks_and_temples}
\end{table*}

In \tabref{tab:eth3d}, \tabref{tab:tanks_and_temples}, we show the leaderboards for the intermediate as well as advanced scenes of ETH3D \citep{Schoeps2017CVPR} and Tanks and Temples \citep{Knapitsch2017SIGGRAPH}, respectively. Both benchmarks use the F1-measure for comparison.

COLMAP \citep{Schoenberger2016ECCV, Schoenberger2016CVPR} jointly models pixel-level view selection and depth estimation using a graphical model. They incorporate geometric as well as temporal priors for improved view selection and a geometric consistency for simultaneous depth/normal estimation with a PatchMatch sampling scheme. COLMAP achieves competitive results on both benchmarks and is considered one of the leading open MVS methods today, serving as the backbone for several other techniques. In contrast to COLMAP, MVSNet \citep{Yao2018ECCV} discussed in \secref{sec:mvs_deep_learning} learns depth map inference with an end-to-end deep learning architecture. For unstructured image sequences like Tanks and Temples, they obtain the depth range and camera trajectory using OpenMVG \citep{OpenMVG}. \citet{Yao2019CVPR} extend this work with a recurrent version called R-MVSNet which replaces 3D convolutions applied on the cost volume for regularization. While both methods were not evaluated on ETH3D, R-MVSNet improves on the intermediate and advanced scenes from Tanks and Temples. 
The best performing reconstruction method, ACMM \citep{Xu2019CVPR}, proposes an adaptive checkerboard sampling scheme and a multi-hypothesis joint view selection approach (ACMH) for improved propagation of hypotheses and pixel-wise view selection. In addition, they propose a multi-scale geometric consistency guidance scheme (ACMM) for improved depth estimation in low textured regions. In contrast to \citet{Yao2018ECCV, Yao2019CVPR}, they use COLMAP's structure-from-motion method \citep{Schoenberger2016CVPR} to obtain the camera trajectory.

The runtime of methods improved significantly with the introduction of learning-based methods. MVSNet \citep{Yao2018ECCV} is currently the fastest approach, with 230 seconds per scan on the DTU MVS evaluation set \citep{Jensen2014CVPR}. The authors used the DTU dataset to compare the runtime and report a large speedup in comparison to COLMAP \citep{Schoenberger2016ECCV, Schoenberger2016CVPR}. ACMM \citep{Xu2019CVPR} compares their runtime to COLMAP on Tanks and Temples and achieves a 3-fold speed up.

\section{Discussion}
In the last decade, great advances have been made in multi-view reconstruction, as can be observed from Tables \ref{tab:eth3d} and \tabref{tab:tanks_and_temples}. The current state of the art significantly improves upon classical approaches like PMVS \citep{Furukawa2010PAMI} and CMP-MVS \citep{Jancosek2011CVPR} on all benchmarks. However, the performance on low-resolution images of ETH3D and all scenes from Tank and Temples is still far from perfect. While great advances have also been made for large-scale reconstruction, a unified benchmark that considers the autonomous driving/mapping task is still missing.

An open question that remains for the autonomous driving problem is what kind of accuracy and completeness are necessary to realize safe mapping, localization, and navigation. For localization (\secref{sec:Localization}) and loop-closure detection (\secref{sec:LoopClosure}) high accuracy is required. In contrast, for obstacle avoidance, high completeness is necessary in order not to miss any obstacle.
	\chapter{Optical Flow}
\label{chap:optical_flow}
\section{Problem Definition}
\begin{figure}[t]
\centering
\includegraphics[width=1.00\columnwidth]{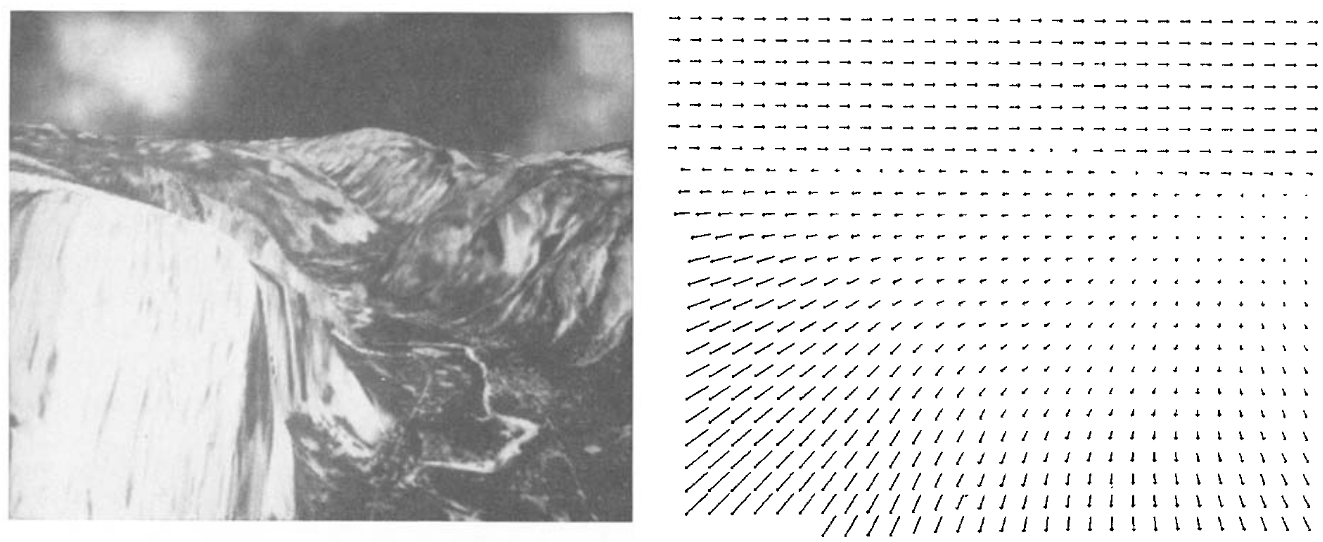}
\caption[Yosemite Optical Flow Sequence]{\textbf{Optical Flow Problem.} The Yosemite sequence generated by \protect\citet{Quam1984IUW} and the corresponding ground truth flow created by \protect\citet{Heeger1988IJCV}. The sequence was later incorporated into the Middlebury dataset of \protect\citet{Baker2011IJCV}. \figsourceSpringer{\protect\citet{Heeger1988IJCV}}{1988}{IJCV}.}
\label{fig:optical_flow_example}
\end{figure}
\label{sec:optical_flow}
Optical flow is defined as the two-dimensional motion of brightness patterns between two images. This definition only represents the motion of intensity patterns in the image plane but not the 3D motion of the objects in the scene. Recovering the 3D motion itself is the goal in Scene Flow discussed in \chpref{sec:SceneFlow}. \figref{fig:optical_flow_example} shows the synthetic Yosemite sequence with the optical flow ground truth generated by texture mapping aerial images of Yosemite valley onto an approximate mesh model. 

Optical flow provides essential information about the scene and serves as input for several tasks such as ego-motion estimation (\chpref{chap:EgoMotionEstimation}), structure from motion (\chpref{chap:EgoMotionEstimation}), and tracking (\chpref{chap:tracking}). 
Research on this problem started several decades ago with the variational formulation by \citet{Horn1981AI}, assuming the brightness of a pixel to be constant over time. Despite the long history of the optical flow problem, occlusions, large displacement, and fine details are still challenging for modern methods. A fundamental problem with the optical flow definition is that besides the actual motion of interest, illumination changes, reflections, and transparency can also cause intensity changes. 
In contrast to stereo, the search space for finding correspondences is two-dimensional in the case of optical flow.

\section{Methods}
Traditionally, the optical flow problem has been approached with a variational formulation. 
Variational methods minimize an energy comprising a data term, assuming little appearance change over time, and a smoothness term, encouraging similarity between spatial neighbors. \citet{Horn1981AI} introduced the brightness constancy assumption which models the intensity value of a pixel as constant over time. Considering a single pixel in isolation, this assumption yields one equation with two unknowns, which does not result in a unique solution (known as the aperture problem). Additional constraints must, therefore, be introduced in order to solve the aperture problem and estimate optical flow.
A common way of regularizing variational optical flow estimation is to encourage similarity of spatially neighboring flow vectors. This prior assumption is motivated by the fact that flow fields are often smooth and discontinuities typically occur only at object boundaries. 

The original formulation \citep{Horn1981AI} uses a quadratic penalty function in the data and smoothness term. However, a quadratic penalty cannot handle frequent violations of brightness constancy assumption, \eg due to varying illumination conditions. One way to alleviate this problem is to use a robust penalty function, as proposed by \citet{Black1993ICCV}. In addition, several different data terms have been proposed that are less affected by illumination changes. \citet{Vogel2013GCPR} systematically evaluate pixel- and patch-based data costs on the KITTI dataset \citep{Geiger2012CVPR}. On real data, they found patch-based terms to perform better than pixel-based terms. 

Flow discontinuities frequently occur near motion boundaries caused by objects moving in front of each other. The original formulation by \citet{Horn1981AI}, cannot handle these discontinuities due to a homogeneous, non-robust smoothness term. Total Variation regularization used in \citet{Zach2007DAGM} replaces the quadratic penalization by the $L_1$ norm to preserve discontinuities in the flow field. 
However, like the original formulation by Horn and Schunck, this model also biases the solution towards fronto-parallel surfaces leading to artifacts in the estimation results, in particular in the presence of strongly slanted planes (\eg the road surface). Thus, higher-order regularizations like the Total Generalized Variation (TGV) model have been proposed \citep{Bredies2010JIS}. TGV priors can better represent real data as they leverage a piecewise affine motion model. The non-local Total Generalized Variation \citep{Ranftl2014ECCV} is an extension of this model that enforces the piecewise affine assumption in a local neighborhood. This allows them to improve performance in regions where the data term is ambiguous in comparison to TGV which considers only direct neighbors. \citet{Zimmer2011IJCV} provide a detailed assessment of image- and flow-driven regularizers for the variational formulation and discuss the qualities of different data terms. 

Besides the model specifications, the choice of the optimization method and its implementation are additional factors that influence the performance of variational optical flow estimation algorithms. A detailed study of optical flow methods is provided by \citet{Sun2014IJCV}. They investigate the most critical factors for the success of optical flow methods and propose an approach optimizing a classical formulation with modern techniques.

\begin{figure}[t]
	\centering
	\includegraphics[width=1.00\columnwidth]{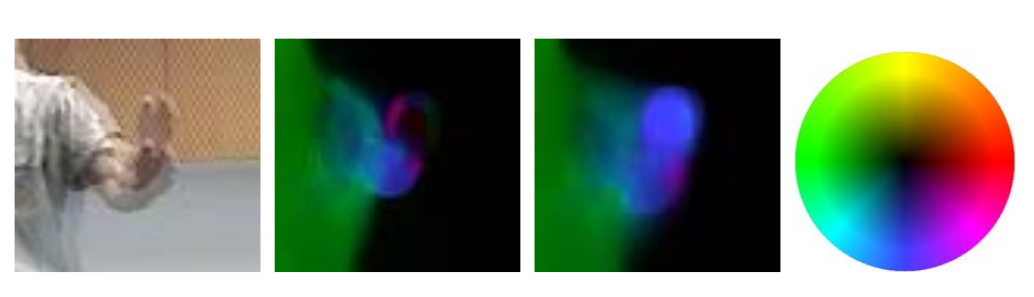}
	\caption[Sparse Matches for Optical Flow]{\textbf{Sparse Matching Guidance.} The fast hand motion (left) is an example where classical warping methods fail (center left), but sparse matches introduced by \protect\citet{Brox2011PAMI} help to estimate the flow (center right). The color encoding with the hue and intensity representing the orientation and magnitude of the flow, accordingly, is visualized in the right image. \figsourceC{\protect\citet{Brox2011PAMI}}{2011}{IEEE}.}
	\label{fig:optical_flow_sparse}
\end{figure}
\subsection{Sparse Matches}
Linear approximations that are used to obtain the optical flow equation hold only for pixel motion. Therefore, variational methods cannot handle large displacements without an additional strategy.
In variational formulations, this problem is typically addressed with a coarse-to-fine strategy, estimating the flow on a coarser resolution to initialize the estimation on a finer resolution. While this strategy works for large structures of little complexity by capturing the dominant motion in the scene, fine geometric details are often lost in the process. Besides, textural details important for correspondence estimation are lost at coarse resolutions, hence leading the optimizer to a local minimum. One example of the loss of fine details is illustrated in \figref{fig:optical_flow_sparse}, which shows the optical flow field of a fast-moving hand.
These problems can be alleviated by integrating sparse feature correspondences into the variational formulation, as proposed by \citet{Brox2011PAMI}. The feature matches, obtained from a nearest neighbor search on a coarse grid, are used as a soft constraint in a coarse-to-fine optimization. In \figref{fig:optical_flow_sparse}, the classical formulation fails to recover the optical flow for the hand, while integrating feature matches guides the optimizer to a better solution.
 
Another solution for large displacements is proposed by \citet{Revaud2015CVPR}. They replace the coarse-to-fine strategy with an interpolation of sparse matches to initialize a dense optimization at full resolution. Sparse matches are obtained using DeepMatching, a deep neural network matching approach introduced by \citet{Weinzaepfel2013ICCV}. In contrast to DeepMatching, \citet{Menze2015GCPR} use approximate nearest neighbor search to generate a set of proposals as candidates to be used in a discrete optimization framework. The inference is made feasible by restricting the number of matches to the most likely ones and by exploiting the truncated form of the pairwise potentials. Motivated by the success of Siamese networks in stereo \citep{Zbontar2016JMLR} (\chpref{chap:Stereo}), \citet{Guney2016ACCV} extend this work to learning features for 2D patch matching. They further investigate the importance of the receptive field size exploiting dilated convolutions as proposed by \citet{Yu2016ICLR} for semantic segmentation. \citet{Chen2016CVPR} argue that the heuristic pruning used to make inference feasible destroys the highly regular structure of the space of mappings and propose a discrete optimization over the full space.  Min-convolutions are used to reduce the complexity and to effectively optimize the large label space using a modified version of Tree-Reweighted Message Passing \cite{Kolmogorov2006PAMI}.

\citet{Wulff2015CVPR} present a different approach to obtain dense optical flow from sparse matches. In their approach, the optical flow field is represented as a weighted sum of basis flow fields learned from reference flow fields which have been estimated from Hollywood movies. They estimate the optical flow by finding the weights which minimize the error with respect to the detected sparse feature correspondences. While this results in overly smooth flow fields, the so-called PCA Flow approach is very fast compared to variational and discrete optimization methods. A slower but more accurate version is also proposed to better handle flow discontinuities by using a layered approach.

\begin{figure}[t]
	\centering
	\includegraphics[width=1.00\columnwidth]{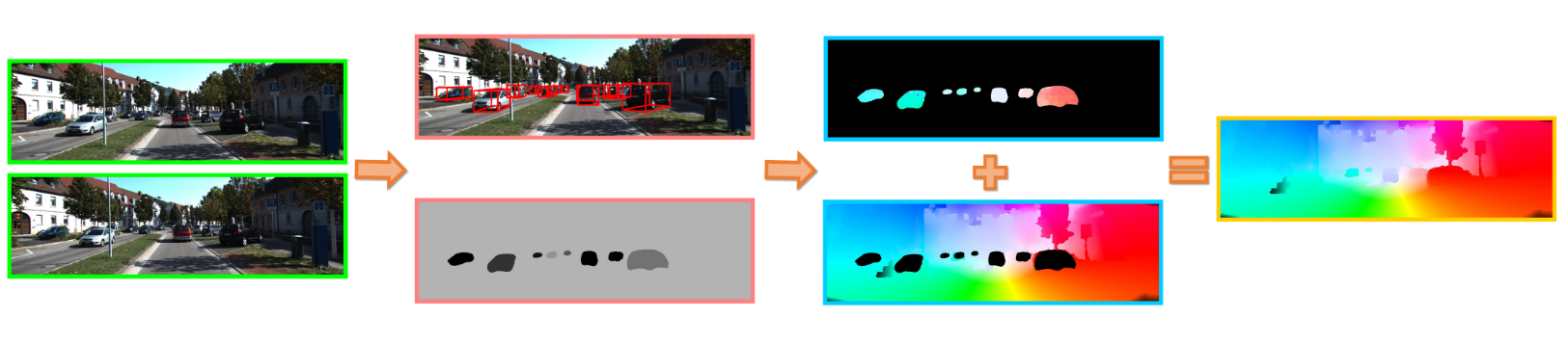}
	\caption[Epipolar Flow]{\textbf{Epipolar Flow.} The full pipeline of \protect\citet{Bai2016ECCV} first segments the scene into dynamic objects (cars) and the static background. Afterwards the motion is estimated for each object and background, independently, and finally combined to one flow field. \figsourceSpringer{\protect\citet{Bai2016ECCV}}{2016}{ECCV}.
	}
	\label{fig:Bai2016ECCV}
\end{figure}
\subsection{Epipolar Flow}
In the context of autonomous driving, application-specific assumptions can be made to simplify the optical flow estimation.
The assumption of a static scene or the decomposition of a scene into rigidly moving objects allows for treating optical flow as a matching problem along epipolar lines radiating from the focus of expansion. \citet{Yamaguchi2013CVPR} propose a slanted-plane Markov random field that represents the epipolar flow of each segment with slanted planes. This formulation needs a time-consuming optimization, which can be avoided with the joint stereo and flow formulation of \citet{Yamaguchi2014ECCV}. They assume the scene to be static and present a new semi-global block matching algorithm using the joint evidence of stereo and video.

\subsection{Semantic Segmentation}
Scenes in the context of autonomous driving are usually composed of a static background and dynamically moving traffic participants. This observation can be exploited by splitting the scene into independently moving objects. \citet{Bai2016ECCV} extract traffic participants using instance-level segmentation and estimate the optical flow independently for different instances. 
Similar to \citep{Yamaguchi2013CVPR,Yamaguchi2014ECCV}, they use the slanted plane model but only for background flow estimation. 
For each moving object, an independent epipolar flow estimation is performed, as illustrated in \figref{fig:Bai2016ECCV}. 
\citet{Sevilla-Lara2016CVPR} use semantic segmentation for optical flow estimation. 
First, semantics provide information on object boundaries and spatial relationships between objects that can be exploited to reason about depth ordering, which in turn determines occlusion relationships in optical flow. 
Second, the division of the scene into semantic units allows them to exploit different motion models according to the respective object type, similar to \citep{Bai2016ECCV}. The motion of planar regions is modeled with homographies, whereas independently moving objects, \eg cars, are modeled by affine motions. Complex objects like vegetation are modeled with a classical spatially varying dense flow field. Finally, the constancy of object identities over time is used to encourage the temporal consistency of the optical flow. 

\begin{figure}[t]
	\centering
	\includegraphics[width=1.00\columnwidth]{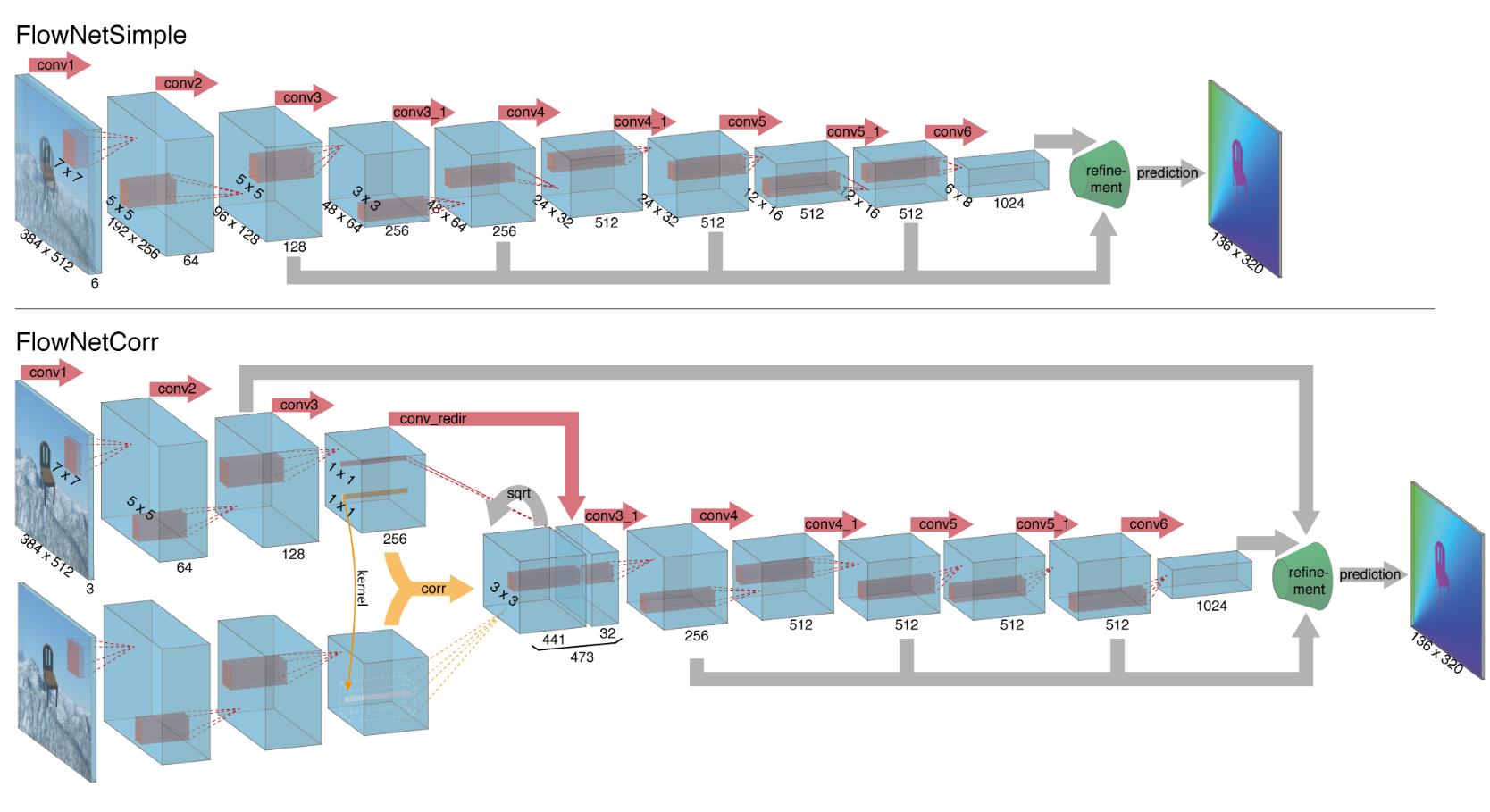}
	\caption[FlowNet Optical Flow Network]{\textbf{FlowNet Optical Flow Network.} The encoder of the FlowNetSimple and FlowNetCorr architecture proposed by \protect\citet{Dosovitskiy2015ICCV}. In FlowNetSimple the images are stacked before the first convolutional layer. In contrast, FlowNetCorr processes both images separately and correlates the extracted feature maps. The refinement model is a decoder consisting of deconvolutional layers that get informed by the encoder using skip connections. \figsourceC{\protect\citet{Dosovitskiy2015ICCV}}{2015}{IEEE}.
	}
	\label{fig:Dosovitskiy2015ICCV}
\end{figure}
\subsection{Deep Learning for Optical Flow}
Most optical flow approaches do not incorporate high-level information making it hard to overcome ambiguities that require reasoning about larger image regions. The recent success of convolutional neural networks has led to an attempt to use them for the optical flow problem. 

\citet{Dosovitskiy2015ICCV} presented FlowNet to learn optical flow end-to-end using a CNN. FlowNet consists of a contracting part that extracts important features and an expanding part that produces the high-resolution optical flow field as output. They propose two different architectures illustrated in \figref{fig:Dosovitskiy2015ICCV}: a simple network (FlowNetSimple) stacking the images and a complex network (FlowNetCorr) correlating features of the separately processed images. One problem in learning optical flow is the limited amount of training data. KITTI 2012 \citep{Geiger2012CVPR} and KITTI 2015 \citep{Menze2015CVPR} only provide around 200 training examples each while Sintel \citep{Butler2012ECCV} has 1041 training image pairs. Since these datasets are too small to train large CNNs, \citet{Dosovitskiy2015ICCV} created the Flying Chairs dataset by rendering 3D chair models on top of images from Flickr. This first attempt to end-to-end optical flow learning demonstrated that it was possible to learn optical flow estimation from data, despite not yet reaching the performance of state-of-the-art traditional methods on KITTI or Sintel. However, due to the parallel GPU implementation, FlowNet was able to run in real-time as opposed to most of the classical algorithms implemented on the CPU.

\begin{figure}[t]
	\centering
	\includegraphics[width=1.00\columnwidth]{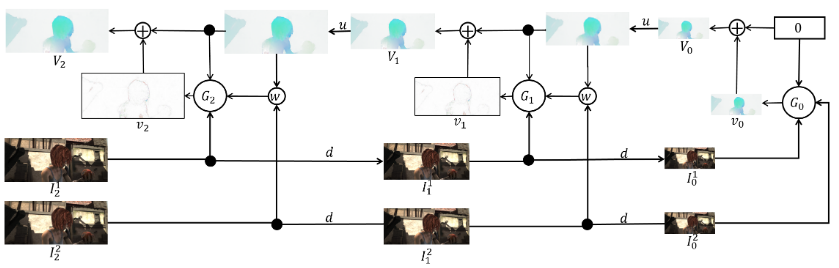}
	\caption[SpyNet Optical Flow Network]{\textbf{SpyNet Optical Flow Network.} The SpyNet architecture by \protect\citet{Ranjan2017CVPR} inspired from classical coarse-to-fine approaches. An image pyramid is created and for each resolution a network is trained to predict the residual flow \wrt the previous layer. \figsourceC{\protect\citet{Ranjan2017CVPR}}{2017}{IEEE}.
	}
	\label{fig:Ranjan2017CVPR}
\end{figure}
\begin{figure}[t]
	\centering
	\includegraphics[width=1.00\columnwidth]{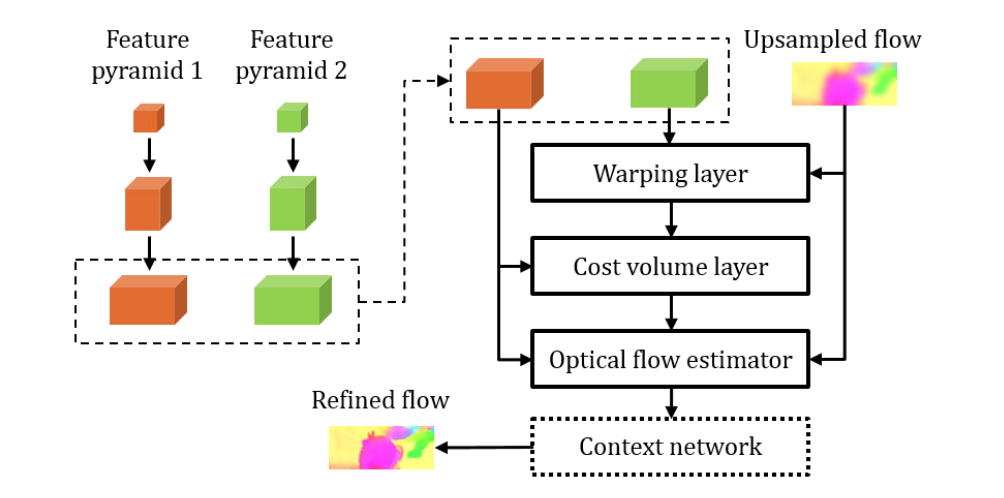}
	\caption[PWCNet Optical Flow Network]{\textbf{PWCNet Optical Flow Network.} \protect\citet{Sun2018CVPR} combine coarse-to-fine estimation with cost-volume filtering using a Siamese network for feature extraction. This figure shows one level of the architecture that uses one level of the feature pyramid and the upsampled flow from the previous level for residual flow estimation. \figsourceC{\protect\citet{Sun2018CVPR}}{2018}{IEEE}.
	}
	\label{fig:Sun2018CVPR}
\end{figure}

In contrast to the contracting and expanding networks of \citet{Dosovitskiy2015ICCV}, \citet{Ranjan2017CVPR} present the SpyNet architecture which is inspired by the coarse-to-fine matching strategy leveraged in traditional optical flow estimation techniques. As shown in \figref{fig:Ranjan2017CVPR}, each layer of the network represents a different scale and only estimates the residual flow with respect to the image warped according to the flow of the previous layer. This formulation allowed them to achieve similar performance as FlowNet while being faster and 96 \% smaller in terms of network weights, making it attractive for embedded systems with limited compute capabilities. 
\citet{Ilg2017CVPR} present FlowNet2, an improved version of FlowNet, by stacking the architectures and fusing the stacked network with a subnetwork specialized in small motions. Similar to SpyNet, they also input the warped image into the stacked networks. Each stacked network estimates the flow between the original frames instead of the residual flow, as in SpyNet. In contrast to FlowNet and SpyNet, they use the FlyingThings3D dataset \citep{Mayer2016CVPR} consisting of 22k renderings of static 3D scenes with moving 3D models from the ShapeNet dataset \citep{Savva2015CVPRWORK}. Recently, PWC-Net \citep{Sun2018CVPR} illustrated in \figref{fig:Sun2018CVPR} was proposed that combines the classical ideas of coarse-to-fine warping \citep{Ranjan2017CVPR} and cost volume filtering \citep{Dosovitskiy2015ICCV} with a Siamese network that proved to learn rich feature representations \citep{Zbontar2016JMLR}. 

\boldparagraph{Unsupervised Learning}
Because large annotated datasets for supervised learning of optical flow are rather limited, several recent works \citep{Yu2016ECCV, Meister2018AAAI, Wang2018CVPR, Janai2018ECCV} address the problem of unsupervised learning of optical flow. Typically, these approaches train one of the standard networks with a photometric loss and a smoothness loss. The photometric loss compares the first image with the second image warped according to the predicted flow. The smoothness loss encourages a similar motion between neighboring pixels. Recently, several approaches \citep{Meister2018AAAI, Wang2018CVPR, Janai2018ECCV} noticed that occluded regions introduce errors in the photometric loss that cause misleading gradients during training. They propose to mask out occluded regions in order to avoid this problem. While \citet{Meister2018AAAI, Wang2018CVPR} both rely on heuristics for estimating occlusions, \citet{Janai2018ECCV} use a three frame formulation to jointly learn occlusions and optical flow in an unsupervised fashion. Even though occlusion handling results in large improvements and performance comparable to the first fully supervised approaches \citep{Dosovitskiy2015ICCV,Ranjan2017CVPR}, they are not yet able to compete with the state-of-the-art supervised approaches \citep{Ilg2017CVPR,Sun2018CVPR} that dominate the leaderboards today.

\begin{figure}[t]
	\centering
	\includegraphics[width=1.00\columnwidth]{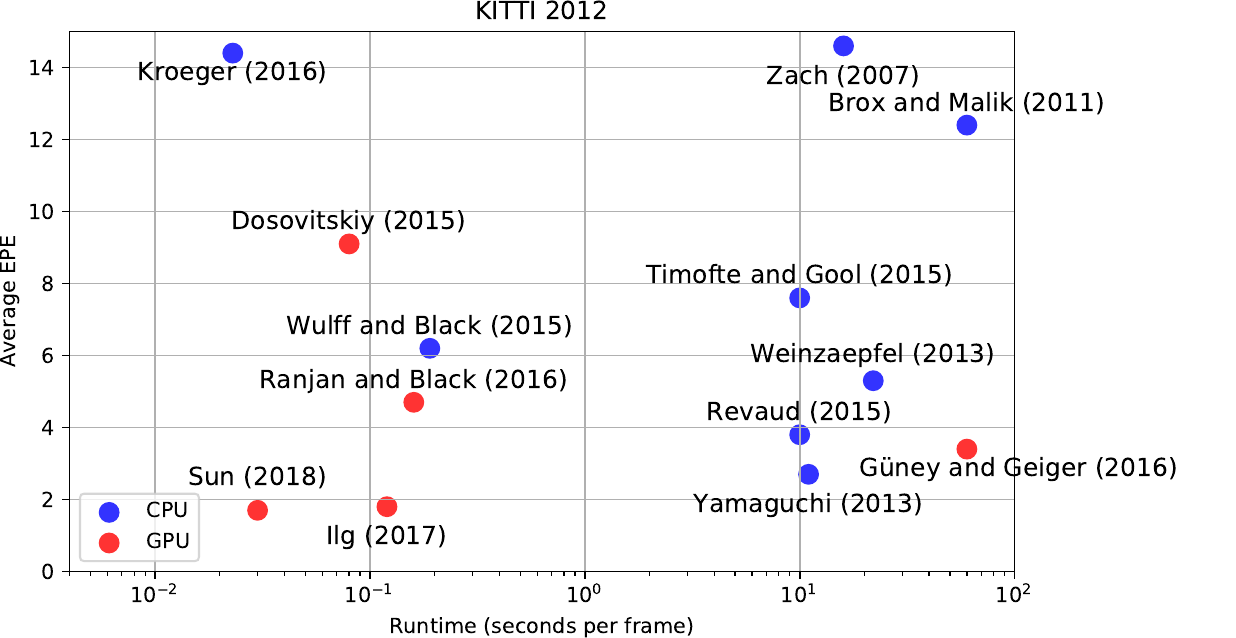}
	\caption[Accuracy vs Efficiency for Optical Flow Methods]{\textbf{Accuracy vs Efficiency.} The trade-off between performance and speed on KITTI 2012 \citep{Geiger2012CVPR}. 
	}
	\label{fig:optical_flow_high_speed}
\end{figure}
\subsection{High-Speed Flow}
With some exceptions (\citet{Wulff2015CVPR, Timofte2015WACV, Weinzaepfel2013ICCV, Farneback2003SCIA, Zach2007DAGM}), most of the classical optical flow approaches are very inefficient and cannot be applied in real-time which is necessary for applications in autonomous driving. The trade-off between accuracy and speed for different algorithms on the KITTI 2012 benchmark \citep{Geiger2012CVPR} is illustrated in \figref{fig:optical_flow_high_speed}. 
While variational approaches yielded a good precision, they belonged to the slowest set of methods for motion estimation. 
The duality-based approach for total variation optical flow proposed by \citet{Zach2007DAGM} allows an efficient GPU implementation that performs in real-time (30 Hz) on a resolution of $320 \times 240$. Sparse matching approaches are usually more efficient than variational formulations but often need variational refinement as a post-processing step to achieve subpixel precision. 

The recent introduction of deep learning to the optical flow problem yielded several near real-time approaches (\citet{Dosovitskiy2015ICCV,Ranjan2017CVPR}) including (\citet{Ilg2017CVPR, Sun2018CVPR}) which achieve state-of-the-art performance on popular datasets. The approach proposed by \citet{Kroeger2016ECCV} allows to trade-off accuracy and runtime. They obtain fast patch correspondences with inverse search resulting in a dense flow field when aggregating patches across multiple scales. This allows them to estimate optical flow at up to 600 Hz, but at the cost of accuracy. 

\subsection{Confidences}
Considering the remaining challenges in optical flow, a confidence measure to assess the quality of the estimated flow is desirable. In the autonomous driving application, for instance, the importance of optical flow estimates can be adjusted with a good confidence measure. In the case of low confidences, the system could heighten the attention to other sources of information, \eg other sensors. 

Several measures based on spatial and temporal gradients have been proposed \citep{Uras1988BC,Anandan1989IJCV,Simoncelli1991CVPR} to quantify the uncertainty in the optical flow estimate. In contrast, algorithm-specific measures propose confidence estimates for a specific group of methods, \ie variational methods \citep{Bruhn2006GPID} and methods for pixel-based minimization problems \citep{Kybic2011CVIU}. While \citet{Bruhn2006GPID} propose a confidence measure based on the energy function optimized by the variational method, \citet{Kybic2011CVIU} uses bootstrap resampling, which repeatedly run the optical flow computation while randomly replacing the contributions of some pixels to the energy. 

Learning-based measures \citep{Kondermann2007GCPR, Kondermann2008ECCV, MacAodha2013PAMI} learn a model that relates the success of flow algorithm success to spatio-temporal image data or the computed flow field. \citet{Kondermann2007GCPR} use linear subspace projection of the optical flow and define a confidence based on the reconstruction error using the linear basis.  In contrast, \citet{Kondermann2007GCPR} learn a probabilistic motion model from annotated training data and use hypothesis testing of flow estimates based on the derived model to compute confidences. \citet{MacAodha2013PAMI} learn a classifier to directly measure the quality of the optical flow predictions based on multiple feature types, such as temporal features, texture or distance from image edges. In addition, they provide a detailed evaluation of different confidence measures.

Several approaches \citep{Wannenwetsch2017ICCV, Ilg2018ECCV, Gast2018CVPR, Yin2019CVPR} proposed to estimate the optical flow and confidences simultaneously. \citet{Wannenwetsch2017ICCV} formulate a probabilistic method based on general energy formulations. The optical flow is estimated by minimizing the expected loss over the posterior, while confidences are measured using the marginal entropy of the posterior. They rely on a mean-field approximation to make inference tractable. \citet{Gast2018CVPR} propose lightweight probabilistic CNNs. Instead of learning a two-dimensional optical flow field, they learn the mean and standard deviation of a Gaussian distribution. Furthermore, they suggest to learn the distribution in each layer and describe how to propagate the probabilistic activations in forward and backward direction. In concurrent work, \citet{Ilg2018ECCV} suggest two approaches for learning uncertainties using CNNs. In a simple approach, they train a set of different models and estimate uncertainty empirically. Since training several models is expensive, they propose an extension of FlowNet \citep{Dosovitskiy2015ICCV} in the spirit of \citep{Gast2018CVPR} by replacing some optical flow layers by the mean and standard deviation of a Gaussian. Recently, \citet{Yin2019CVPR} presented a probabilistic formulation based on discrete distributions over possible correspondences. With a general model representing the matching probabilities, they do not need to rely on any parametric distribution assumption. They decompose the match distribution into multiple scales to make the computations feasible.

\section{Datasets}
Sintel \citep{Butler2012ECCV} and KITTI \citep{Geiger2012CVPR, Geiger2013IJRR} discussed in \chpref{chap:Datasets} are the most popular datasets for the evaluation of optical flow algorithms. However, in this survey, we focus on the autonomous driving application. Therefore, we will only refer to the KITTI leaderboard when comparing methods.

\section{Metrics}
\label{sec:optical_flow_metrics}
The performance of methods is usually assessed considering the endpoint error (Euclidean distance) between the estimated flow vectors and the ground truth. While Sintel reports the average endpoint error for different velocities, occluded and non-occluded regions, the KITTI dataset uses outliers which are computed as the percentage of flow vectors with the absolute endpoint error (EPE) exceeding 3 pixels and 5\% of its true values. The percentage of outliers is averaged over background (Fl-bg), foreground (Fl-fg), and all regions (Fl-all), resulting in three different evaluation metrics.

\section{State of the Art on KITTI}
\begin{table*}[t!]
	\begin{adjustbox}{width=1\textwidth}\begin{tabular}{l l | c | c | c | c}
 & {\bf Method} & {\bf Fl-bg} & {\bf Fl-fg} & {\bf Fl-all} & {\bf Runtime}\\ \hline
1. & PWC-Net+ \citep{Sun2018ARXIV} & 7.69 \% & 7.88 \% & 7.72 \% & 0.03 s / GPU \\
2. & LiteFlowNet \citep{Hui2018CVPR} & 9.66 \% & 7.99 \% & 9.38 \% & 0.0885 s / GPU \\
3. & PWC-Net \citep{Sun2018CVPR} & 9.66 \% & 9.31 \% & 9.60 \% & 0.03 s / GPU \\
4. & ContinualFlow\_ROB (MF) \citep{Neoral2018ACCV} & 8.54 \% & 17.48 \% & 10.03 \% & 0.15 s / GPU \\
5. & MirrorFlow \citep{Hur2017ICCV} & 8.93 \% & 17.07 \% & 10.29 \% & 11 min / 4 core \\
6. & FlowNet2 \citep{Ilg2017CVPR} & 10.75 \% & 8.75 \% & 10.41 \% & 0.1 s / GPU \\
7. & SDF \citep{Bai2016ECCV} & 8.61 \% & 23.01 \% & 11.01 \% & TBA / 1 core \\
8. & UnFlow \citep{Meister2018AAAI} & 10.15 \% & 15.93 \% & 11.11 \% & 0.12 s / GPU \\
\hline
26. & RicFlow \citep{Hu2017CVPR} & 18.73 \% & 19.09 \% & 18.79 \% & 5 s / 1 core \\
27. & FlowFields+ \citep{Bailer2015ICCV} & 19.51 \% & 21.26 \% & 19.80 \% & 28s / 1 core \\
28. & PatchBatch \citep{Gadot2016CVPR} & 19.98 \% & 26.50 \% & 21.07 \% & 50 s / GPU \\
29. & DDF \citep{Guney2016ACCV} & 20.36 \% & 25.19 \% & 21.17 \% & ~1 min / GPU \\
36. & Back2FutureFlow (MF) \citep{Janai2018ECCV} & 22.67 \% & 24.27 \% & 22.94 \% & 0.12 s / GPU \\
37. & MotionSLIC \citep{Yamaguchi2013CVPR} & 14.86 \% & 64.44 \% & 23.11 \% & 30 s / 4 cores \\
39. & FullFlow \citep{Chen2016CVPR} & 23.09 \% & 24.79 \% & 23.37 \% & 4 min / 4 cores \\
45. & EpicFlow \citep{Revaud2015CVPR} & 25.81 \% & 28.69 \% & 26.29 \% & 15 s / 1 core \\
50. & SPyNet \citep{Ranjan2017CVPR} & 33.36 \% & 43.62 \% & 35.07 \% & 0.16 s / 1 core \\
51. & HS \citep{Sun2014IJCV} & 39.90 \% & 51.39 \% & 41.81 \% & 2.6 min / 1 core \\
52. & DB-TV-L1 \citep{Zach2007PRL} & 47.52 \% & 48.27 \% & 47.64 \% & 16 s / 1 core \\
\end{tabular}\end{adjustbox}
	\caption{{\bf KITTI 2015 Optical Flow Leaderboard.} Numbers correspond to percentages of bad pixels according to the 3px/5\% criterion defined in \cite{Menze2015CVPR} averaged over background (bg), foreground (fg), or all regions. Methods followed by (MF) use multiple frames as input. Methods below the horizontal line show older entries for reference. Accessed on: June 2019.}
	\label{tab:kitti_flow_2015}
\end{table*}
In \tabref{tab:kitti_flow_2015}, we show the leaderboard for the KITTI 2015 benchmark. In addition to the estimation error, the density of the output flow field and the runtime are also provided. 

\citet{Bai2016ECCV} achieve great accuracy in background regions by leveraging semantic segmentation and epipolar geometry. However, their performance drops on foreground regions with dynamic objects that do not follow their assumptions. \citet{Hur2017ICCV} formulate a symmetric optimization problem to jointly reason about optical flow and occlusions. Using an alternating optimization of forward-backward flow and occlusions, they obtain similar results in background regions while improving on foreground regions.

The best performing methods learn optical flow end-to-end \citep{Sun2018CVPR, Hui2018CVPR, Neoral2018ACCV, Ilg2017CVPR, Meister2018AAAI, Sun2018ARXIV}. FlowNet2 \citep{Ilg2017CVPR} provides different network variants for the spectrum between 8fps and 140fps, allowing the trade-off between accuracy and computation. The most accurate network achieves comparable results to the state of the art. \citet{Neoral2018ACCV} use a three frame formulation to jointly learn optical flow and occlusions. This allows them to perform well on background regions that are often occluded by foreground objects. \citet{Hui2018CVPR} follow a similar approach to PWC-Net by using a Siamese network, coarse-to-fine warping, and computing correlations. \citet{Sun2018ARXIV} propose a new learning rate schedule consisting of several disruptions (a strong increase of the learning rate) and show how this improves the training of the original PWC-Net. PWC-Net \citep{Sun2018CVPR} with the adapted training protocol \citep{Sun2018ARXIV} outperforms all methods on KITTI 2015 (\tabref{tab:kitti_flow_2015}) and Sintel in both background and foreground objects while being one of the fastest methods on both benchmarks.

\section{Discussion}
\begin{figure*}[p]
\includegraphics[width=0.5\linewidth]{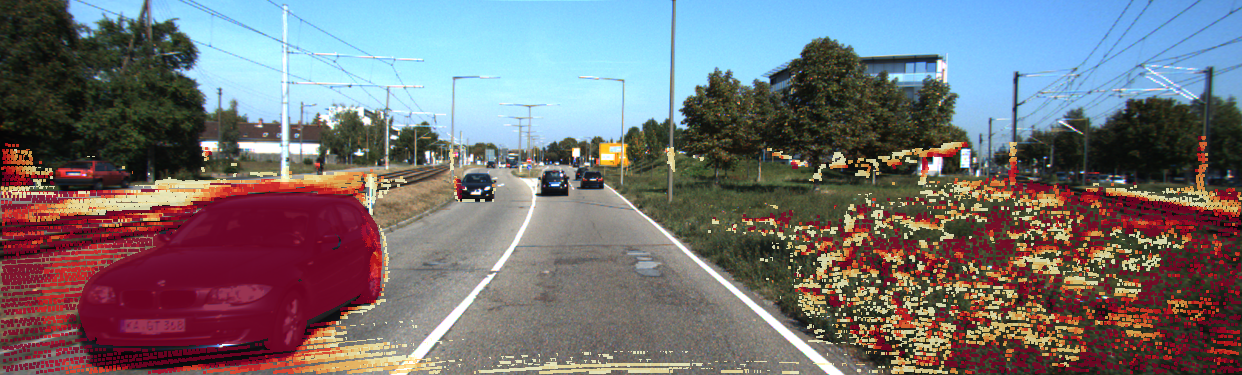}%
\includegraphics[width=0.5\linewidth]{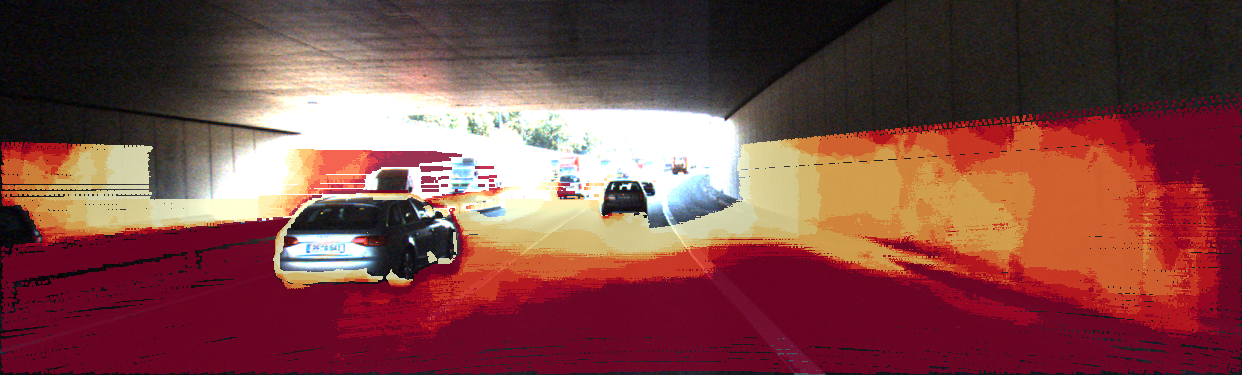}\\%
\includegraphics[width=0.5\linewidth]{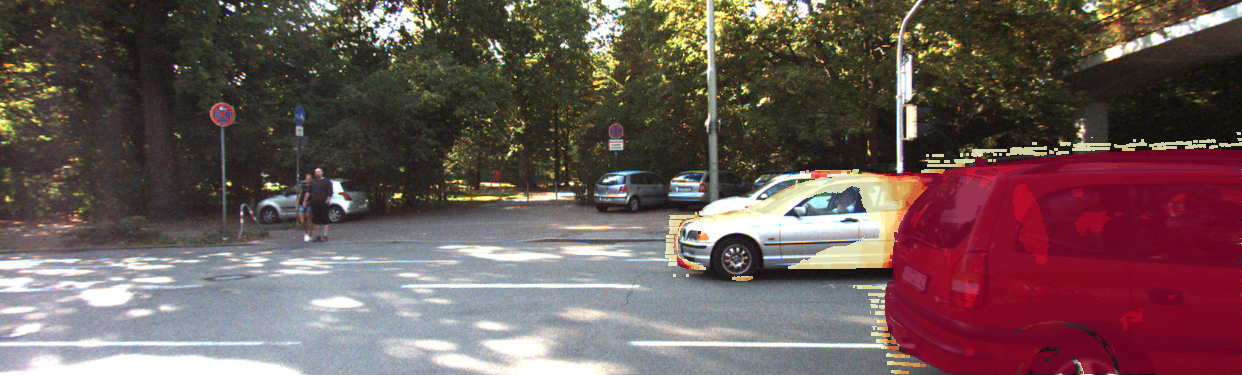}%
\includegraphics[width=0.5\linewidth]{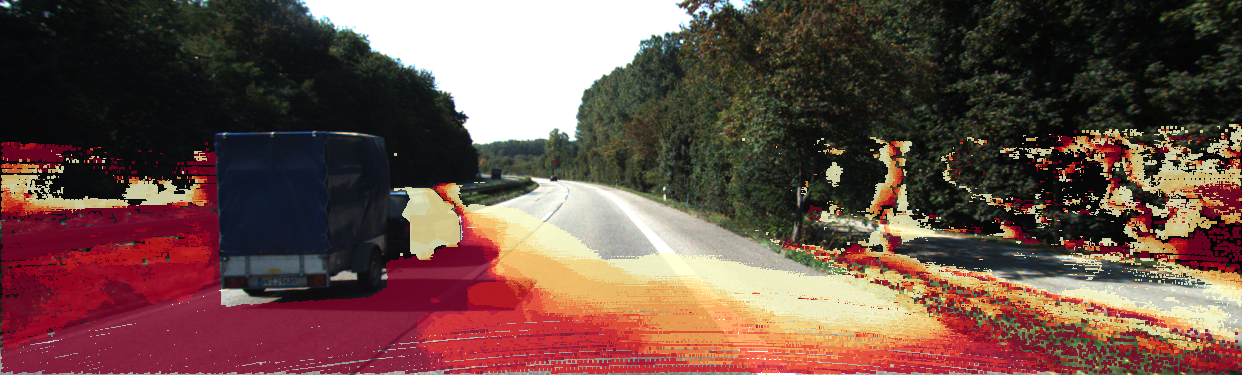}\\%
\includegraphics[width=0.5\linewidth]{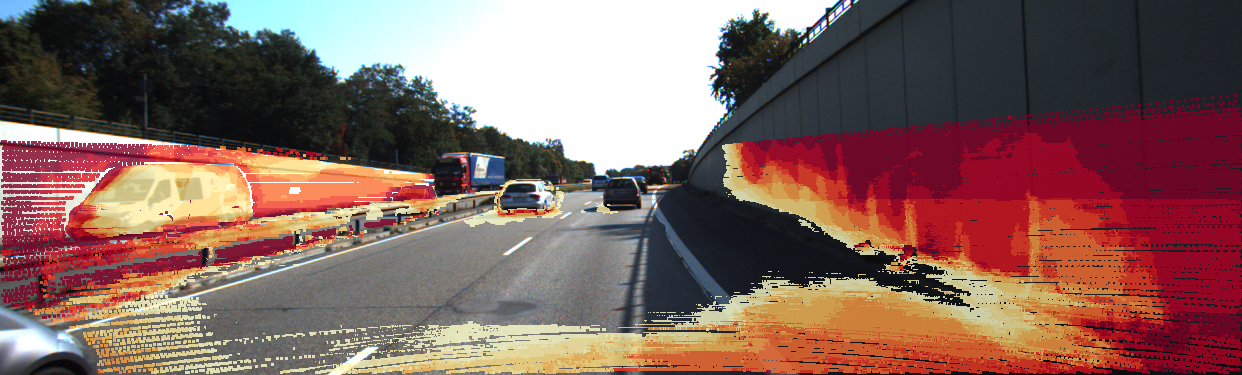}%
\includegraphics[width=0.5\linewidth]{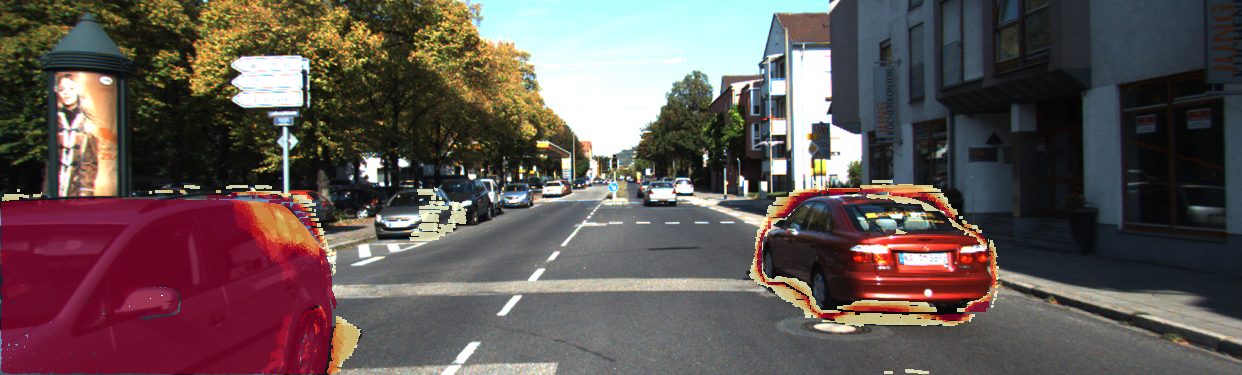}\\%
\includegraphics[width=0.5\linewidth]{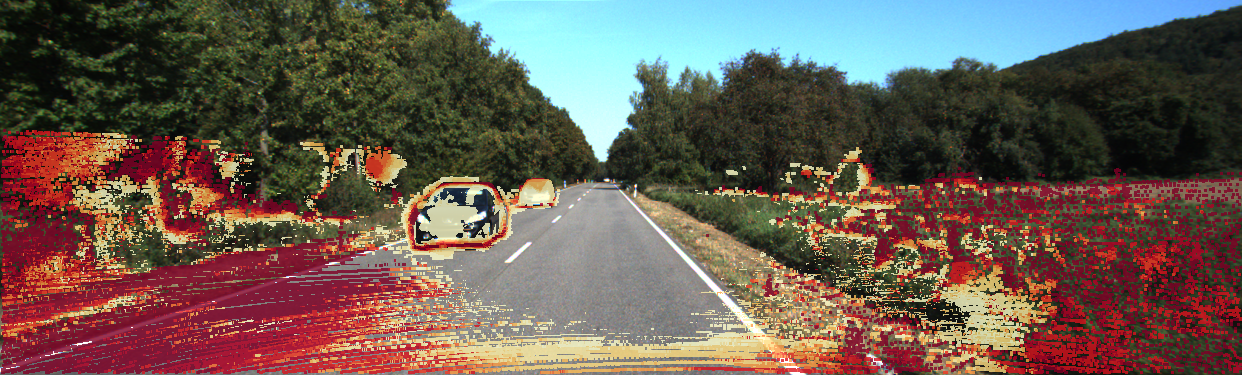}%
\includegraphics[width=0.5\linewidth]{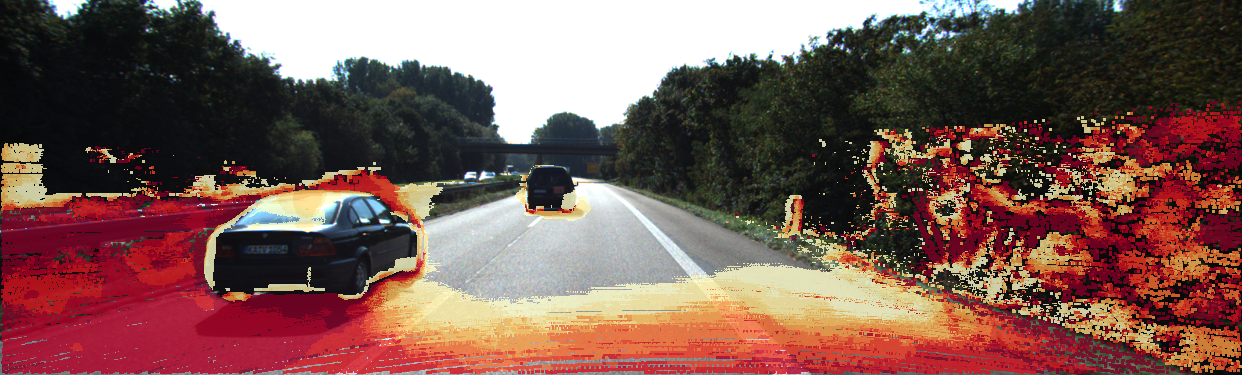}\\%
\includegraphics[width=0.5\linewidth]{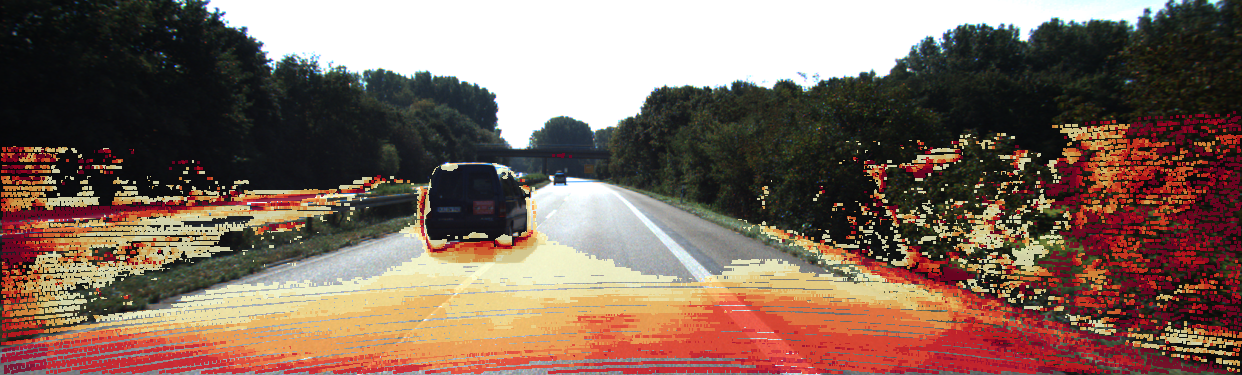}%
\includegraphics[width=0.5\linewidth]{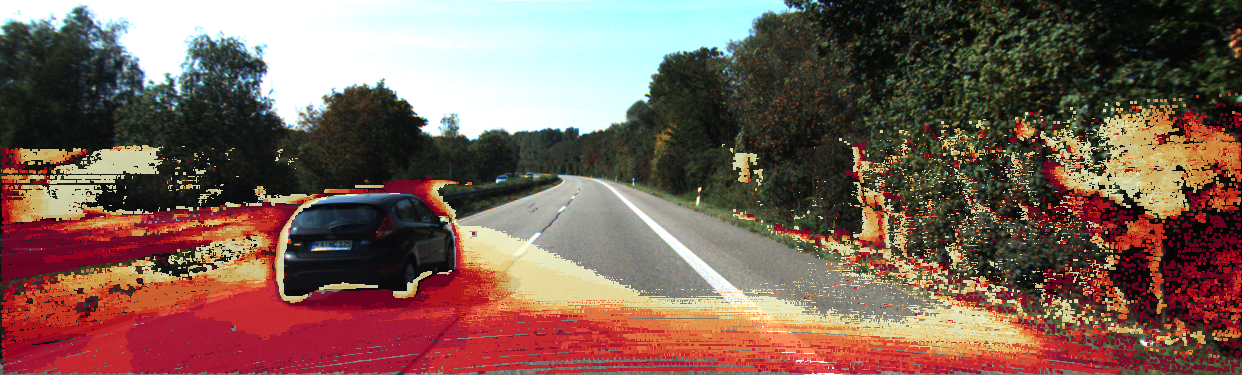}\\%
\includegraphics[width=0.5\linewidth]{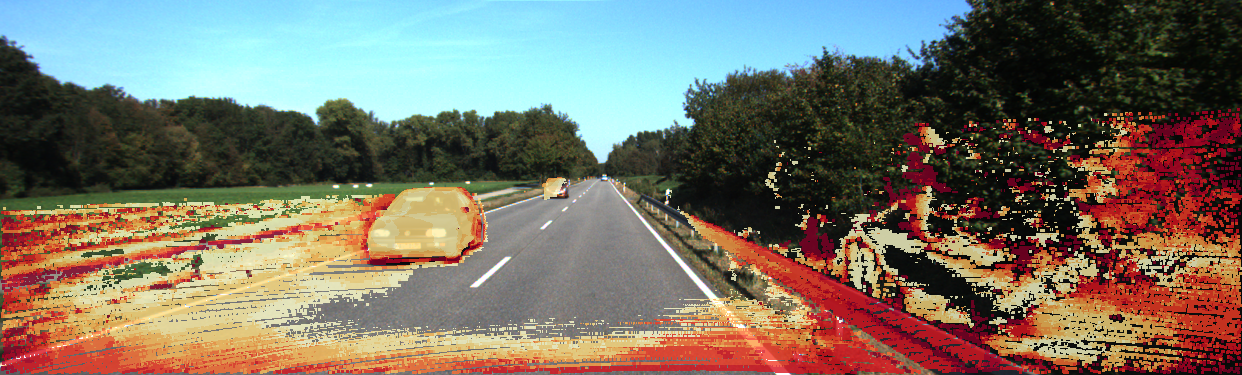}%
\includegraphics[width=0.5\linewidth]{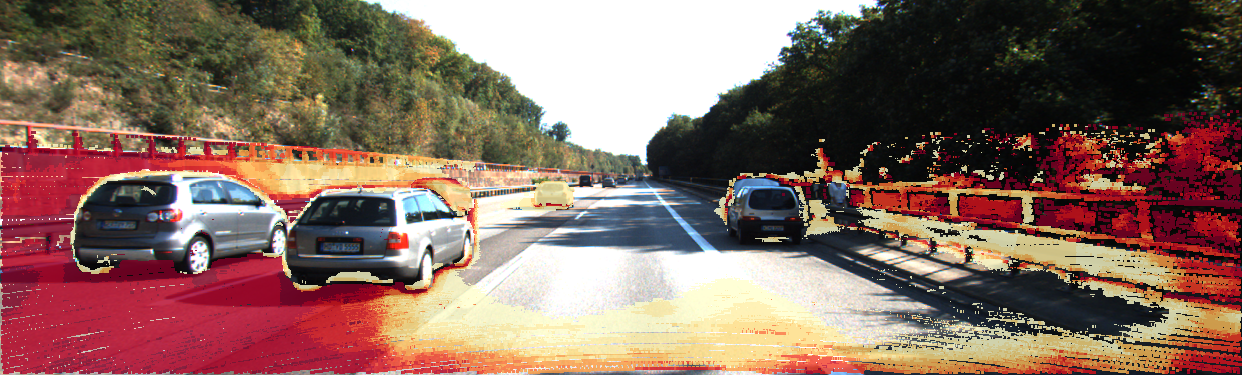}\\%
\includegraphics[width=0.5\linewidth]{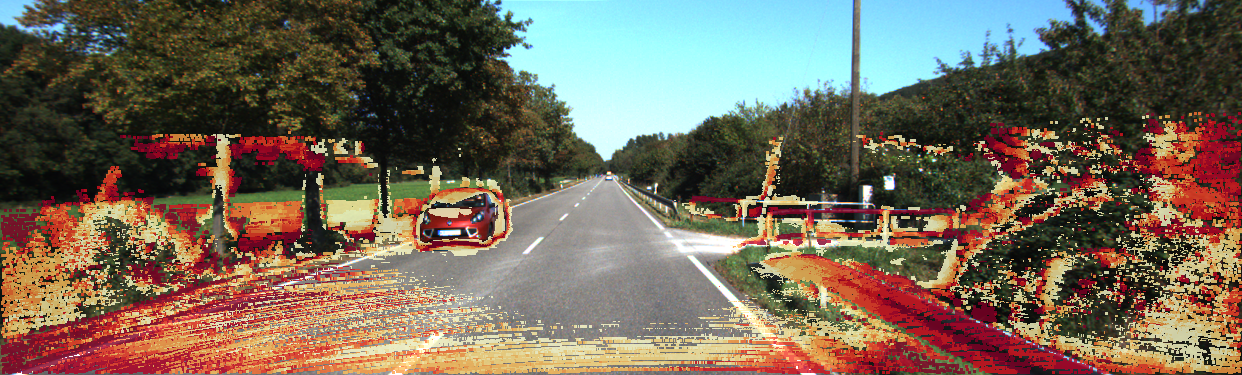}%
\includegraphics[width=0.5\linewidth]{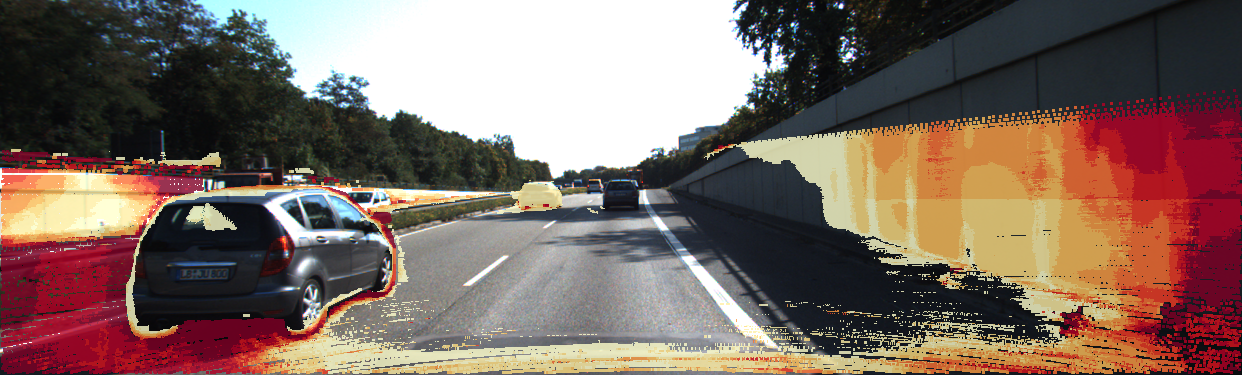}\\%
\includegraphics[width=0.5\linewidth]{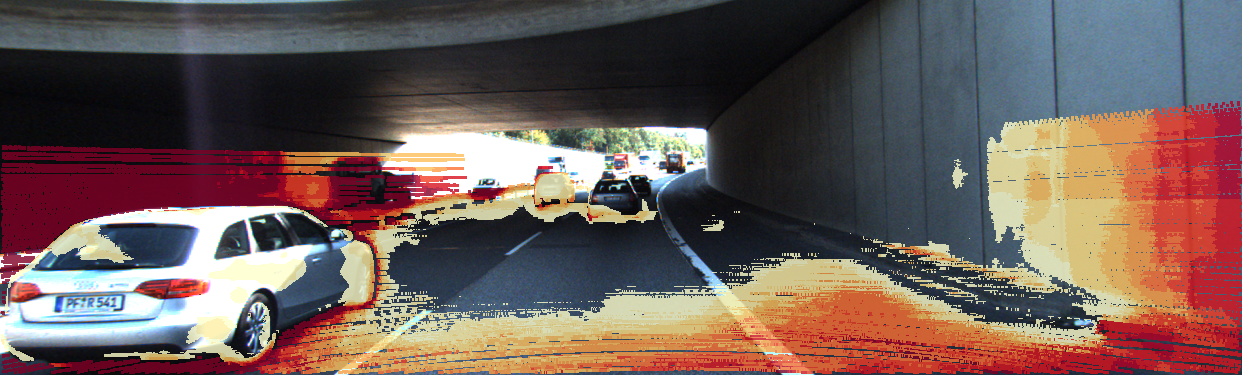}%
\includegraphics[width=0.5\linewidth]{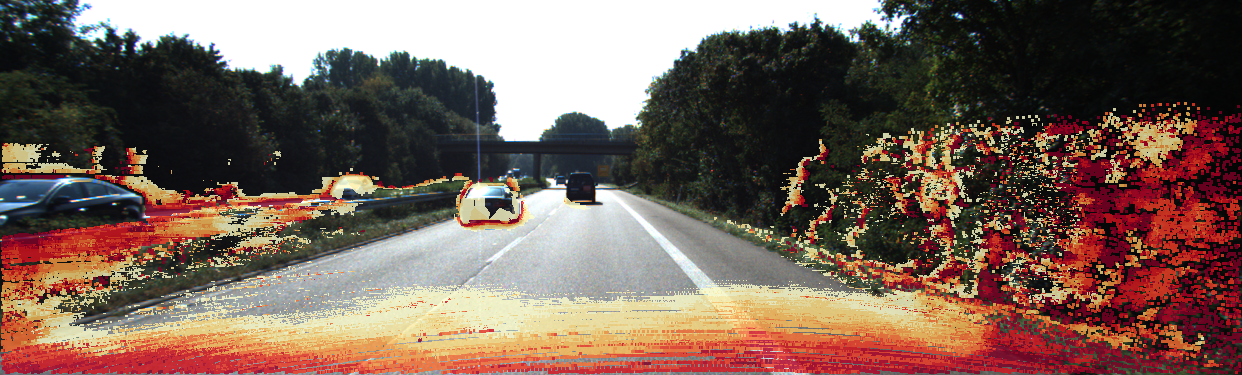}\\%
\caption{{\bf KITTI 2015 Optical Flow Analysis.} The averaged errors of the 15 best-performing optical flow methods published on the KITTI 2015 Flow benchmark. Red colors correspond to regions where the majority of methods fail according to the 3px/5\% criterion defined in \cite{Menze2015CVPR}. Yellow colors correspond to regions where some of the methods fail. Regions that are correctly estimated by all methods are transparent.}
\label{fig:flow_qualitative_results}
\end{figure*}

Robust optical flow methods need to handle intensity changes not caused by the actual motion of interest but by illumination changes, reflections, and transparency. In real-world scenes, repetitive patterns, textureless surfaces, saturated image regions, and occlusions are frequent sources of errors. While illumination changes have been addressed with novel data terms \citep{Black1993ICCV,Vogel2013GCPR}, the problems caused by reflection, transparency, ambiguities, and occlusions remain mostly unsolved. In \figref{fig:flow_qualitative_results}, we show the accumulated error of the 15 best-performing methods on KITTI 2015 \citep{Menze2015CVPR}. The highest error can be observed for regions moving outside the image domain for which the optical flow has to be guessed, as observations are not available. Untextured, reflective, and transparent regions also result in large errors in many cases. A better understanding of the world is necessary to tackle these problems. Semantics \citep{Bai2016ECCV} and learned high-capacity models \citep{Sun2018CVPR, Hui2018CVPR, Neoral2018ACCV, Ilg2017CVPR, Meister2018AAAI, Sun2018ARXIV} have already proven to improve optical flow estimation by resolving ambiguities in the data. Joint optical flow and occlusion formulations have also shown great potential to alleviate these problems for optimization-based \citep{Hur2017ICCV} as well as learning-based \citep{Meister2018AAAI, Wang2018CVPR, Janai2018ECCV} methods. 

	\chapter{3D Scene Flow}
\label{sec:SceneFlow}
\section{Problem Definition}
\begin{figure}[t]
	\centering
	\includegraphics[width=1.00\columnwidth]{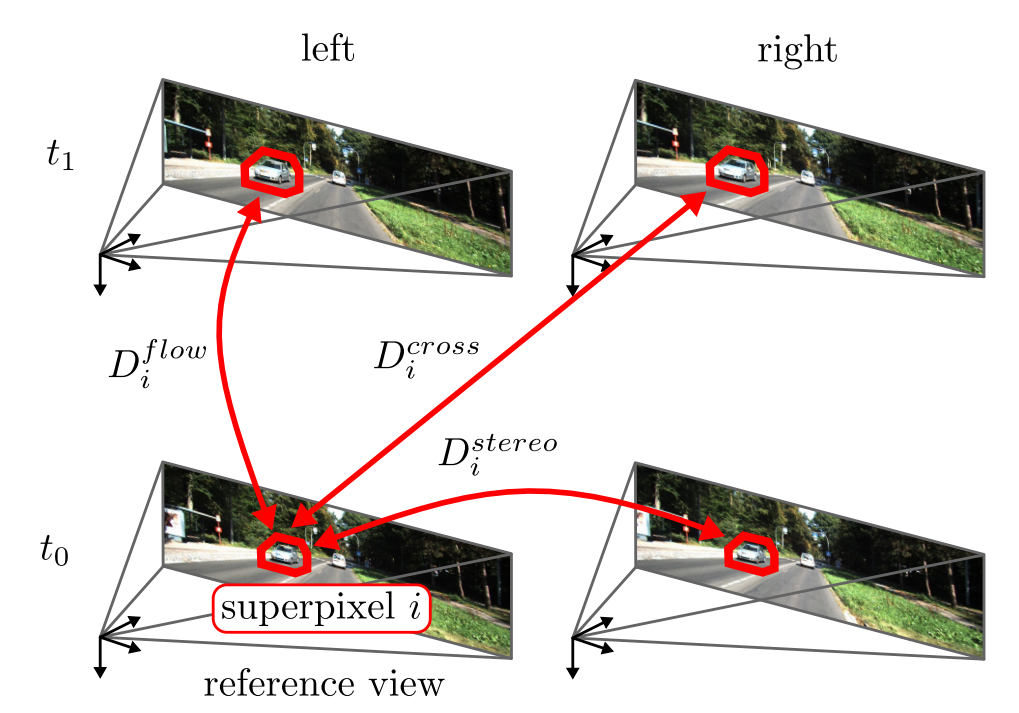}
	\caption[Scene Flow]{\textbf{Scene Flow.} The minimal setup for image-based scene flow estimation is given by two consecutive stereo image pairs. \figsourceC{\protect\citet{Menze2015CVPR}}{2015}{IEEE}.}
	\label{fig:scene_flow}
\end{figure}
Humans are able to effortlessly integrate depth and motion cues from observations over time. That kind of reasoning is essential for many tasks in autonomous driving, such as segmentation of moving objects in the 3D world. Scene flow generalizes optical flow to 3D, or equally, dense stereo to dynamic scenes. Given stereo image sequences, the goal is to estimate the three-dimensional motion field that is a 3D motion vector for every point on every visible surface in the scene. The minimal setup for image-based scene flow estimation is given by two consecutive stereo image pairs, as visualized in \figref{fig:scene_flow}. Establishing correspondences between the four images results in the 3D location of the surface point in both frames and hence fully describes the 3D motion of that surface point. A dense output is preferred, although some early works focused on establishing sparse correspondences \citep{Franke2005DAGM}. Scene flow shares some of the challenges with stereo and optical flow, such as matching ambiguities in weakly textured regions and the aperture problem, but integrating observations from four images and solving both tasks jointly leads to a better-constrained problem. 

\section{Methods}
Following the seminal work by \citet{Vedula1999CVPR}, the problem is traditionally formulated in a variational setting where optimization proceeds in a coarse-to-fine manner, and local regularizers are leveraged
to encourage spatial smoothness of depth and motion. Wedel \etal \citep{Wedel2008ECCV, Wedel2011IJCV} propose a variational framework by decoupling the motion estimation from the disparity estimation while maintaining stereo constraints. Starting from a precomputed disparity map at each time step, optical flow for the reference frame and disparity for the other view are estimated. The motivation for this decoupling is mainly computational efficiency by choosing the optimal technique for each task. In addition, \citet{Wedel2011IJCV} propose a solution for varying lighting conditions based on residual images and provide an uncertainty measure which they showed to be useful for object segmentation. \citet{Rabe2010ECCV} integrate a Kalman filter to the decoupling approach for temporal smoothness and robustness.

\begin{figure}[t]
	\centering
	\includegraphics[width=1.00\columnwidth]{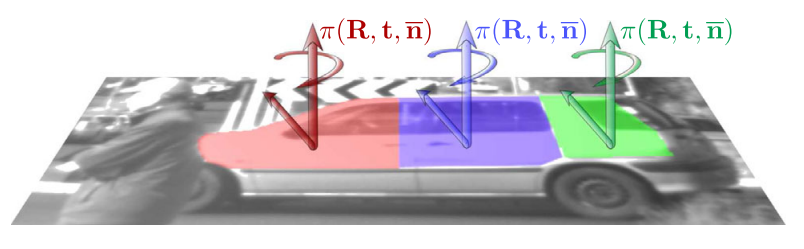}
	\caption[Scene Flow with Piecewise Rigidity]{\textbf{Piecewise Rigidity.} In \citet{Vogel2015IJCV} the scene is modeled as a collection of rigidly moving planar segments. \figsourceSpringer{\protect\citet{Vogel2015IJCV}}{2015}{IJCV}.}
	\label{fig:scene_flow_rigidity}
\end{figure}
\subsection{Piecewise Rigidity}
Similar to stereo and optical flow, prior assumptions about the geometry and motion can be exploited to better handle the challenges of the scene flow problem. \citet{Vogel2015IJCV} and \citet{Lv2016ECCV} represent the dynamic scene as a collection of rigidly moving planar regions, as shown in \figref{fig:scene_flow_rigidity}. \citet{Vogel2015IJCV} jointly recover this segmentation while inferring the shape and motion parameters of each region. They use a discrete optimization framework and incorporate occlusion reasoning as well as other scene priors in the form of spatial regularization of geometry, motion, and segmentation. In addition, they reason over multiple frames by constraining the segmentation to remain stable over a temporal window. Their experiments show that their view-consistent multi-frame approach significantly improves accuracy for challenging scenarios. Using the same representation, \citet{Lv2016ECCV} focus on an efficient solution to the problem. They assume a fixed superpixel segmentation and perform optimization in the continuous domain for faster inference. Starting from an initialization based on Deep Matching \citep{Weinzaepfel2013ICCV}, they independently refine the geometry and motion of the scene, and finally perform a global non-linear refinement using the Levenberg-Marquardt algorithm.

\boldparagraph{Piecewise Rigidity at the Object Level}
\citet{Menze2015CVPR, Behl2017ICCV} also follow a slanted plane approach, but in addition to previous methods \citep{Vogel2015IJCV, Lv2016ECCV}, they model the decomposition of the scene into a small number of independently moving objects and the background. By conditioning on a superpixelization, they jointly estimate this decomposition as well as the rigid motion of the objects and the plane parameters of each superpixel in a discrete-continuous Conditional Random Field (CRF). Compared to \citep{Vogel2015IJCV, Lv2016ECCV}, they leverage a more compact representation, by implicitly regularizing over larger distances. They also present a new scene flow dataset by annotating dynamic scenes from the KITTI raw data collection using detailed 3D CAD models. \citet{Menze2015ISA} propose an extension of this model where the pose and 3D shape of the objects are inferred in addition to the rigid motion and segmentation. In particular, they incorporate a deformable 3D active shape model of vehicles into the scene flow approach. 

\subsection{Semantic Segmentation}
\begin{figure}[t]
	\centering
	\includegraphics[width=1.00\columnwidth]{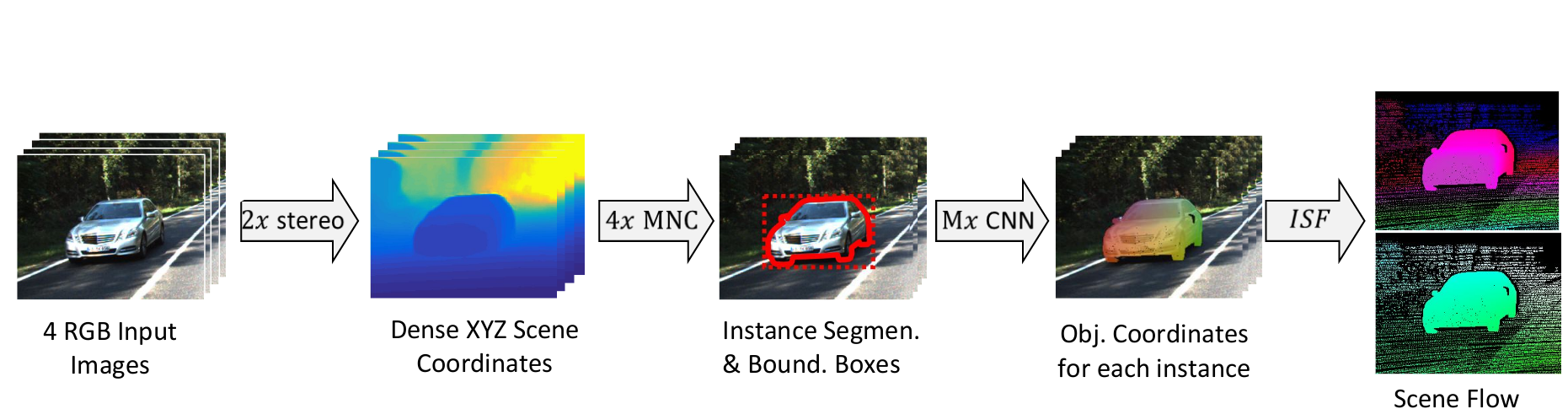}
	\caption[Semantic Segmentation for Scene Flow]{\textbf{Semantic Segmentation for Scene Flow.} \citet{Behl2017ICCV} leverage instance segmentation and bounding boxes from an instance segmentation pipeline to independently model the motion of each object and the background. \figsourceC{\protect\citet{Behl2017ICCV}}{2017}{IEEE}.}
	\label{fig:Behl2017ICCV}
\end{figure}

Semantic information allows constraining the space of possible rigid body motions. For instance, in an autonomous driving scenario, pixels which are grouped together in the segmentation are likely to move as a single rigid object in the case of vehicles. Furthermore, a pixel on a vehicle instance in one frame should be mapped to a vehicle instance in the other frame. 
\citet{Behl2017ICCV} investigate the impact of bounding box detection, instance segmentation, and 3D object coordinates on scene flow estimations and show which one is most beneficial for scene flow. They obtain the bounding boxes and instance segmentation from the proposal-based instance segmentation method MNC \citep{Dai2016CVPR} discussed in \chpref{chap:instance_segmentation}. 3D object coordinates are predicted with a CNN trained on the 2D instance segmentations as illustrated in \figref{fig:Behl2017ICCV}. Using a CRF based on \citep{Menze2015CVPR}, they show that semantic cues lead to significant improvements. However, the benefit of 3D object coordinates over instance segmentations is negligible.
Recently, \citet{Ma2019CVPR} leverage multiple cues consisting of CNNs for instance segmentation (Mask R-CNN \citep{He2017ICCV}), optical flow (PWC-Net \citep{Sun2018CVPR}) and stereo (PSM-Net \citep{Chang2018CVPR}) to address the scene flow problem. They formulate an energy combining all the cues with a photometric term. By unrolling the optimization as a recurrent network, they are able to train the whole pipeline end-to-end.

\subsection{Scene Flow from 3D Point Clouds}
The image-based methods discussed earlier estimate scene flow based on two consecutive image pairs of a calibrated stereo camera rig. However, stereo-based scene flow methods suffer from the ``curse of two-view geometry'', \ie the depth error grows quadratically with the distance to the observer. Furthermore, most modern self-driving car platforms rely on LiDAR technology for 3D geometry perception. In contrast to cameras, laser scanners do not suffer from the quadratic error behavior of stereo cameras. In addition, laser scanners provide a 360-degree field of view with just one sensor and are generally unaffected by lighting conditions. Therefore, there have been several methods proposed recently for estimating 3D scene flow from pairs of unstructured 3D point clouds. \citet{Dewan2016IROS} propose a 3D scene flow approach where local SHOT descriptors \citep{Tombari2010ECCV} are associated via a CRF that incorporates local smoothness and rigidity assumptions. However, local shape representations such as SHOT often fail in the presence of noisy or ambiguous inputs. In contrast, \citet{Behl2019CVPR} address the scene flow problem using a generic end-to-end trainable model that is able to learn local and global statistical relationships directly from data. In order to apply the standard 3D convolution operations, they discretize the point cloud to a grid of voxels. However, because of the sparse nature of the LiDAR data, most of the space is empty, which makes the approach computationally and memory inefficient. To alleviate this problem, \citet{Wang2018CVPRb} propose a novel continuous convolution that operates over non-grid structured data.

\section{Datasets}
Only a few datasets exist for scene flow \citep{Geiger2012CVPR, Mayer2016CVPR, Butler2012ECCV}. Similar to flow and stereo, the KITTI scene flow benchmark \citep{Geiger2012CVPR} is the most popular dataset allowing the comparison of methods on an online evaluation server. For deep learning, the Flying Things datasets \citep{Mayer2016CVPR} is often used for pre-training since KITTI is too small. Recently, MPI Sintel \citep{Butler2012ECCV} published stereo sequences for the training dataset\footnote{\url{http://sintel.is.tue.mpg.de/stereo}} and is since used to show the generalization of scene flow approaches to other scenes than street scenes from KITTI. 

\section{Metrics}
Scene flow methods are usually evaluated by jointly measuring the accuracy of the stereo (\secref{sec:stereo_metrics}) and optical flow estimates (\secref{sec:optical_flow_metrics}). The KITTI benchmark considers the percentage of erroneous pixels. A pixel is erroneous if the Euclidean distance to the ground truth exceeds a 3 pixels or 5\% threshold. The percentage of stereo disparity outliers in the first frame (D1), the percentage of stereo disparity outliers in the second frame (D2), the percentage of optical flow outliers (Fl), and the percentage of scene flow outliers (SF), \ie outliers in either D0, D1 or Fl are reported. The outlier ratio for foreground/background regions can be found separately on the website of the benchmark\footnote{\url{http://www.cvlibs.net/datasets/kitti/eval_scene_flow.php}}, but it is omitted here for space reasons.

\section{State of the Art on KITTI}
\begin{table*}[t]
\begin{center}
\begin{adjustbox}{width=1\textwidth}%
\begin{tabular}{l l | c | c | c | c | c }
& {\bf Method} & {\bf D1} & {\bf D2} & {\bf Fl} & {\bf SF} & {\bf Runtime}\\ \hline
1. & UberATG-DRISF \citep{Ma2019CVPR} & 2.55 \% & 4.04 \% & 4.73 \% & 6.31 \% & 0.75 s / CPU+GPU \\
2. & ISF \citep{Behl2017ICCV} & 4.46 \% & 5.95 \% & 6.22 \% & 8.08 \% & 10 min / 1 core \\
3. & PRSM \citep{Vogel2015IJCV} (MF) & 4.27 \% & 6.79 \% & 6.68 \% & 8.97 \% & 300 s / 1 core \\
4. & OSF+TC \citep{Neoral2017CVWW} (MF) & 5.03 \% & 6.84 \% & 7.02 \% & 9.23 \% & 50 min / 1 core \\
5. & OSF 2018 \citep{Menze2018JPRS} & 5.28 \% & 7.06 \% & 7.41 \% & 9.66 \% & 390 s / 1 core \\
6. & SSF \citep{Ren2017THREEDV} & 4.42 \% & 7.02 \% & 7.14 \% & 10.07 \% & 5 min / 1 core \\
7. & OSF \citep{Menze2015CVPR} & 5.79 \% & 7.77 \% & 7.83 \% & 10.23 \% & 50 min / 1 core \\
8. & FSF+MS \citep{Taniai2017CVPR} (MF) & 6.74 \% & 9.85 \% & 11.30 \% & 14.96 \% & 2.7 s / 4 cores \\
9. & PWOC-3D \citep{Saxena2019IV} & 5.13 \% & 8.46 \% & 12.96 \% & 15.69 \% & 0.13 s / GPU \\
\hline
18. & SGM+C+NL \citep{Hirschmueller2008PAMI} & 6.84 \% & 28.25 \% & 35.61 \% & 40.33 \% & 4.5 min / 1 core \\
19. & SGM+LDOF \citep{Hirschmueller2008PAMI} & 6.84 \% & 28.56 \% & 39.33 \% & 43.67 \% & 86 s / 1 core \\
20. & DWBSF \citep{Richardt2016THREEDV} & 20.12 \% & 34.46 \% & 39.14 \% & 45.48 \% & 7 min / 4 cores \\
21. & GCSF \citep{Cech2011CVPR} & 14.21 \% & 33.41 \% & 46.40 \% & 53.54 \% & 2.4 s / 1 core \\
22. & VSF \citep{Huguet2007ICCV} & 26.38 \% & 57.08 \% & 49.28 \% & 66.90 \% & 125 min / 1 core
\end{tabular}\end{adjustbox}
\end{center}
\vspace{-0.4cm}
\caption{{\bf KITTI 2015 Scene Flow Leaderboard.} Numbers correspond to percentages of bad pixels according to the 3px/5\% criterion defined in \citep{Menze2015CVPR} for disparity in the first frame (D1), disparity in the second frame (D2), optical flow between both frames (Fl) as well as the combination of all criteria yielding the final scene flow metric (SF). Approaches using more than 2 frame pairs are marked by (MF). Methods below the horizontal line show older entries for reference. Accessed on: June 2019.}
\label{tab:kitti_sceneflow_2015}
\end{table*}

\tabref{tab:kitti_sceneflow_2015} shows the ranking of methods on the KITTI Scene Flow 2015 benchmark \citep{Menze2015CVPR}. 

All top-performing methods use either semantic cues \citep{Behl2017ICCV, Ma2019CVPR, Ren2017THREEDV} or the assumption of rigidly moving segments \citep{Vogel2015IJCV, Menze2015CVPR, Neoral2017CVWW, Menze2018JPRS}. Modeling the motion of objects using a rigid transformation \citep{Menze2015CVPR, Menze2015ISA, Menze2018JPRS, Neoral2017CVWW} achieves impressive results on the KITTI dataset. However, this is a very strong assumption even in street scenes as the non-rigidity of pedestrians cannot be handled. In contrast, the segmentation of the scene into superpixels \citep{Vogel2015IJCV} alleviates this problem and allows better performance since non-rigid objects can be modeled by multiple superpixels. However, the best performance is achieved by integrating semantic information \citep{Behl2017ICCV, Ma2019CVPR}. While most scene flow approaches are very inefficient, the method of \citet{Ma2019CVPR} is a notable exception, requiring only 0.75 seconds. They achieve this efficiency by combining CNNs for instance segmentation (Mask R-CNN \citep{He2017ICCV}), optical flow (PWC-Net \citep{Sun2018CVPR}) and stereo (PSM-Net \citep{Chang2018CVPR}) to address the scene flow problem and exploiting a GPU for inference (in contrast to classical scene flow approaches which typically run on the CPU).

\section{Discussion}
The scene flow problem shares many challenges with stereo and optical flow while integrating more information than each task alone and consequently leading to better results. Ideally, methods should exploit depth and motion cues together to reason about dynamic 3D scenes. However, considering the optical flow (\tabref{tab:kitti_flow_2015}) and stereo matching leaderboards (\tabref{tab:kitti_stereo_2015}), the joint formulation is more advantageous for the optical flow problem leading to significant improvements as for instance UberATG-DRISF \citep{Ma2019CVPR} reaches an outlier ratio of 4.73\% in comparison to PWC-Net+ \citep{Sun2018ARXIV} reaching 7.72 \%. In contrast, the stereo matching performance is comparable with an outlier ratio of 2.55\% (UberATG-DRISF \citep{Ma2019CVPR}) in comparison to 2.08 \% (EdgeStereo-V2 \citep{Song2018ACCV}).

\begin{figure*}[p]
\includegraphics[width=0.5\linewidth]{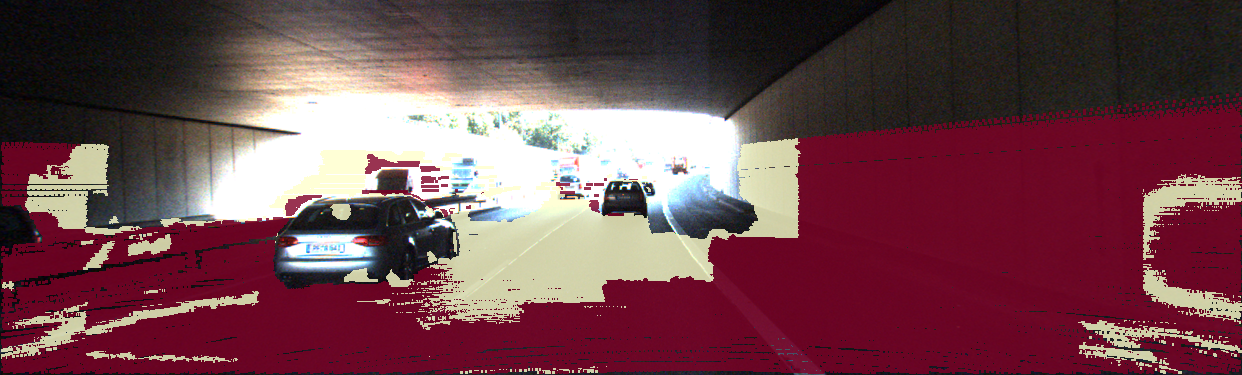}%
\includegraphics[width=0.5\linewidth]{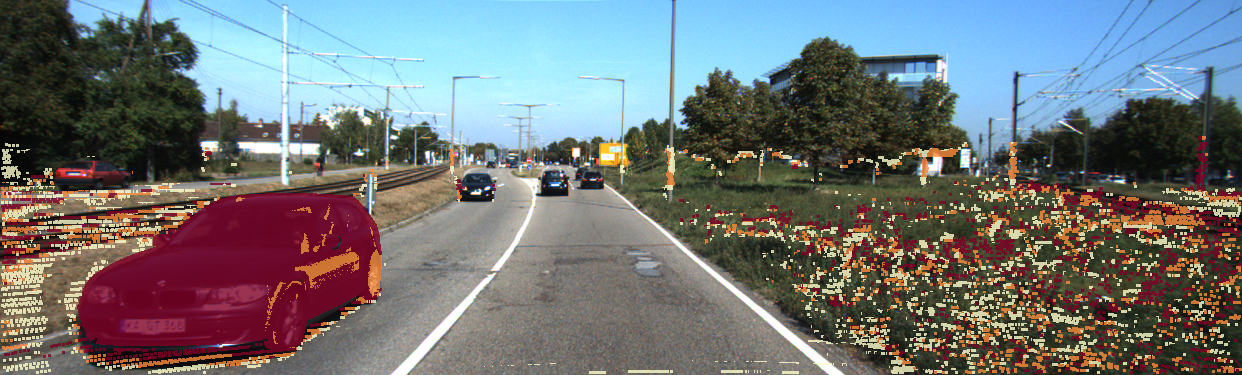}\\%
\includegraphics[width=0.5\linewidth]{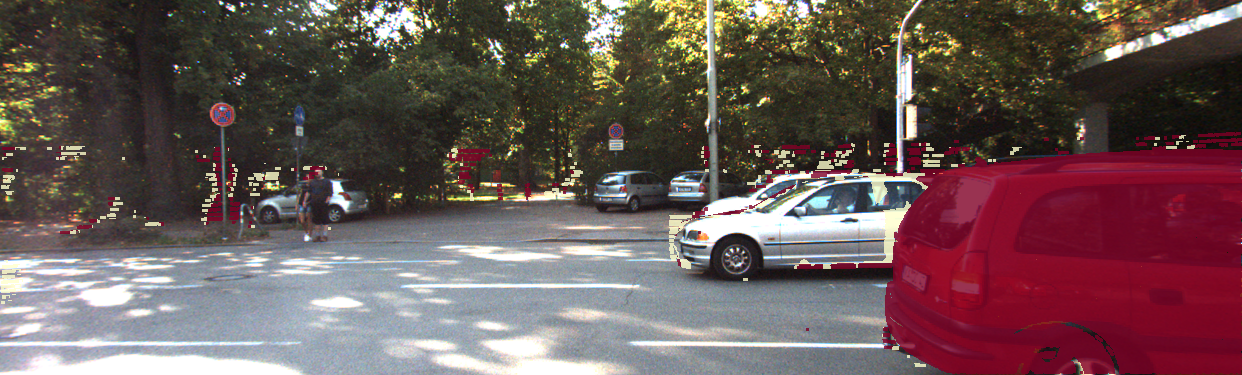}%
\includegraphics[width=0.5\linewidth]{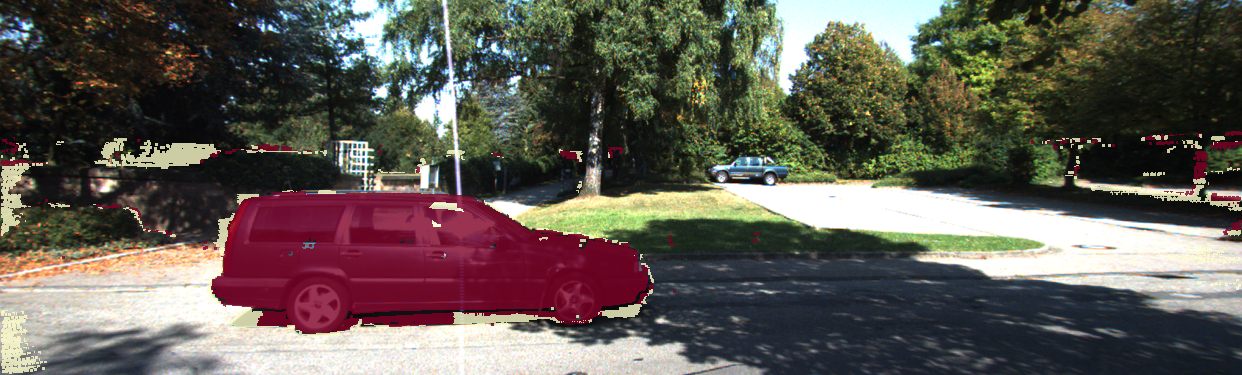}\\%
\includegraphics[width=0.5\linewidth]{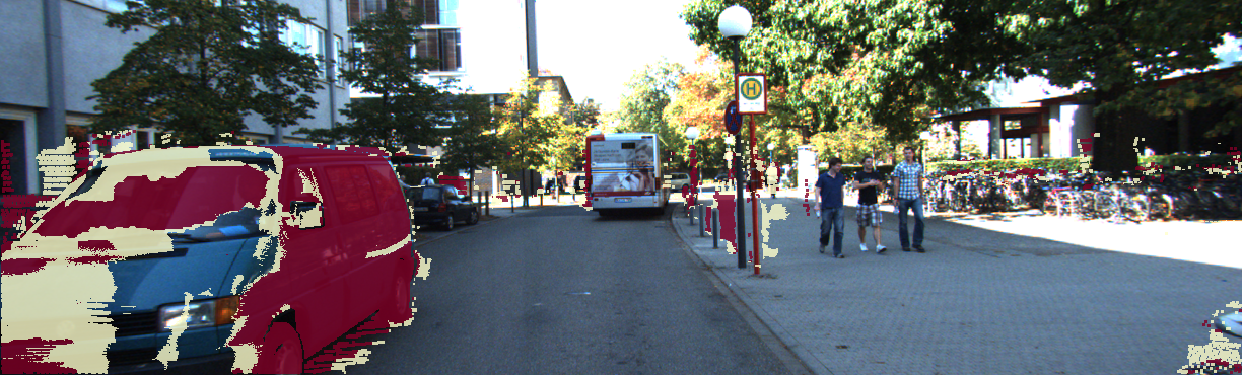}%
\includegraphics[width=0.5\linewidth]{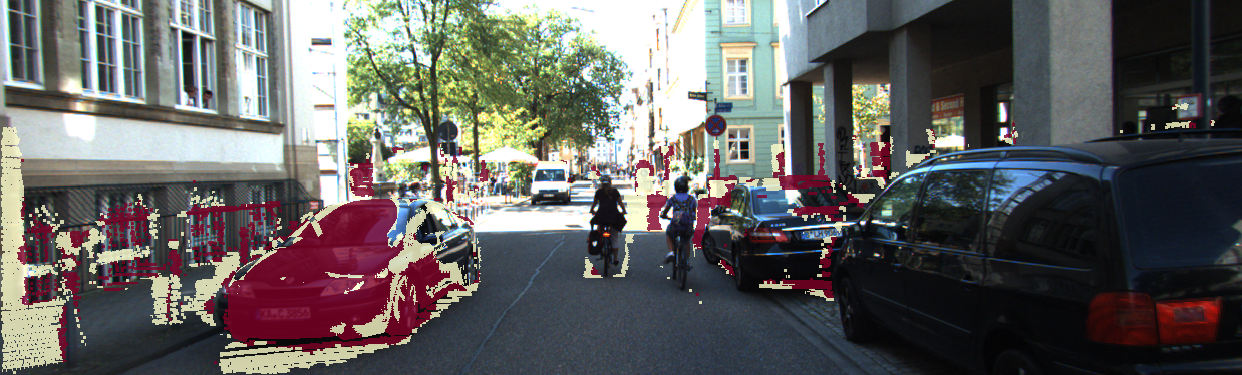}\\%
\includegraphics[width=0.5\linewidth]{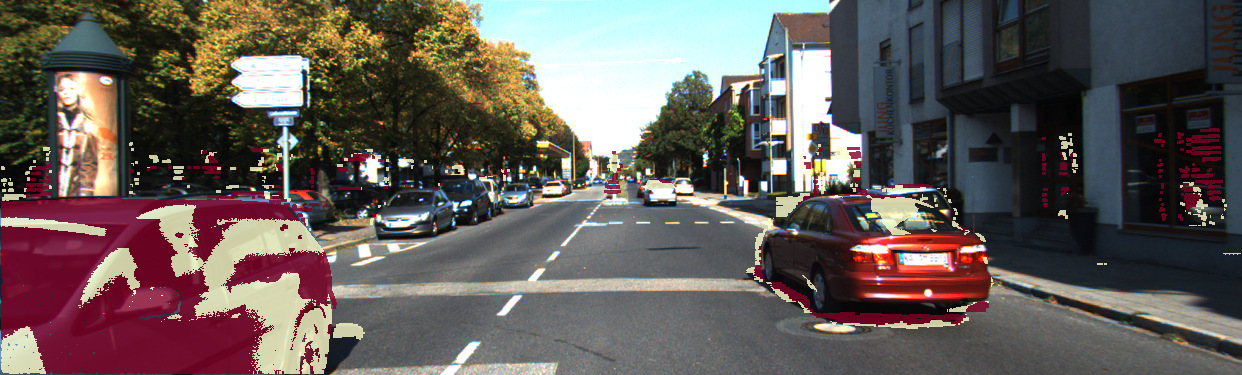}%
\includegraphics[width=0.5\linewidth]{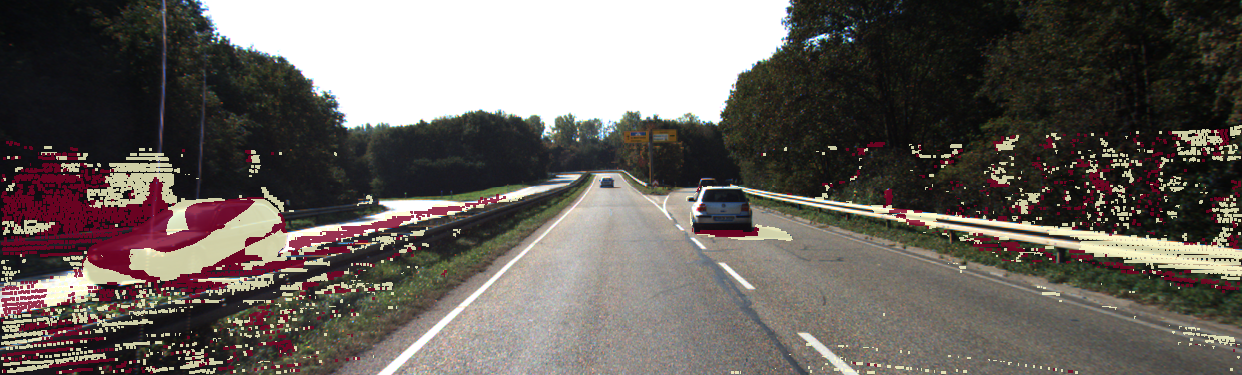}\\%
\includegraphics[width=0.5\linewidth]{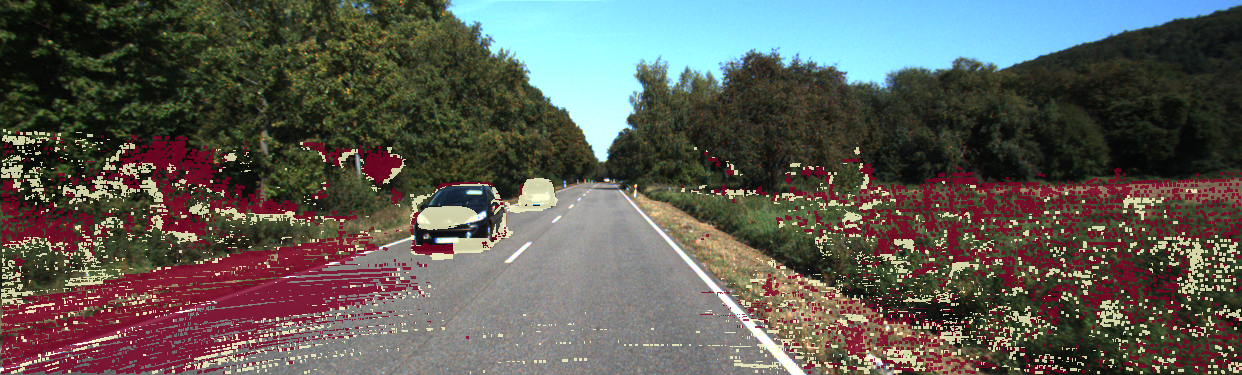}%
\includegraphics[width=0.5\linewidth]{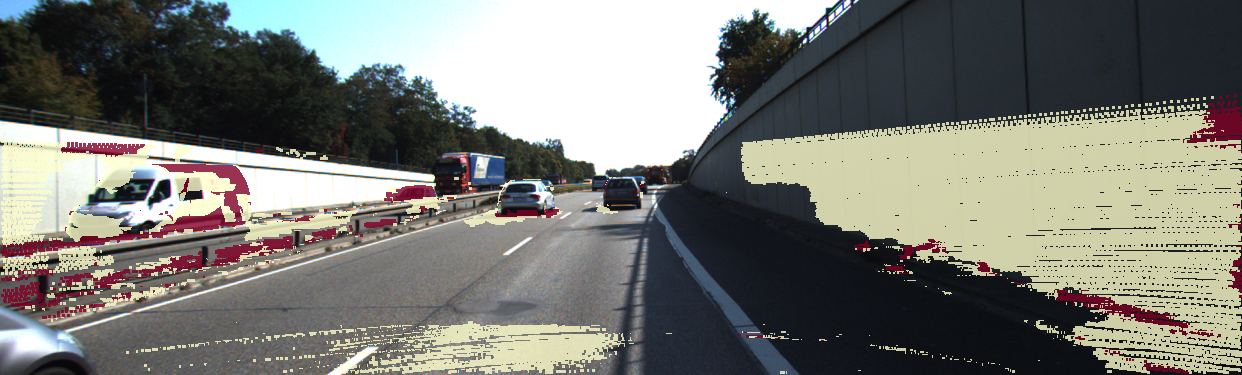}\\%
\includegraphics[width=0.5\linewidth]{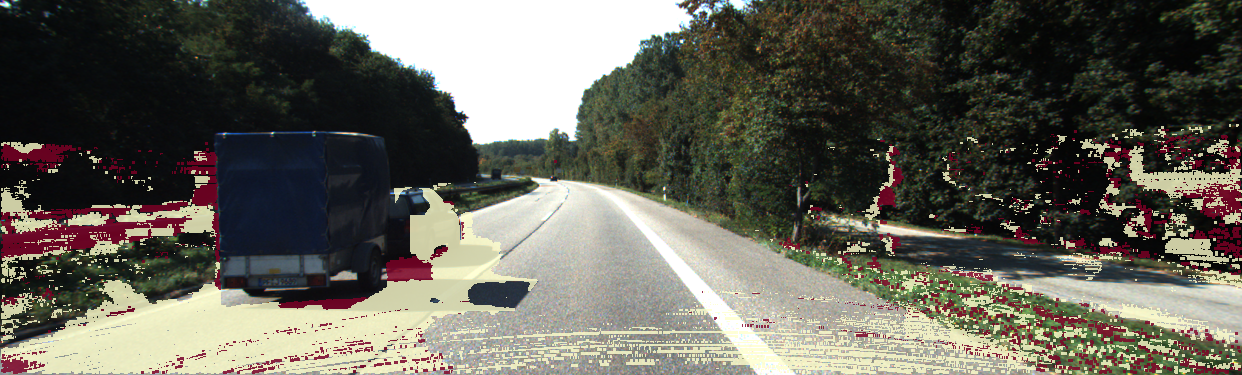}%
\includegraphics[width=0.5\linewidth]{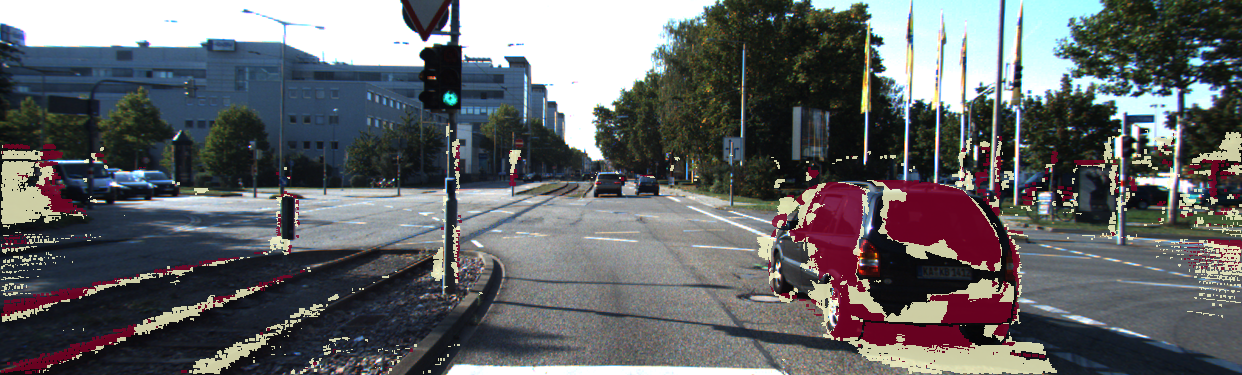}\\%
\includegraphics[width=0.5\linewidth]{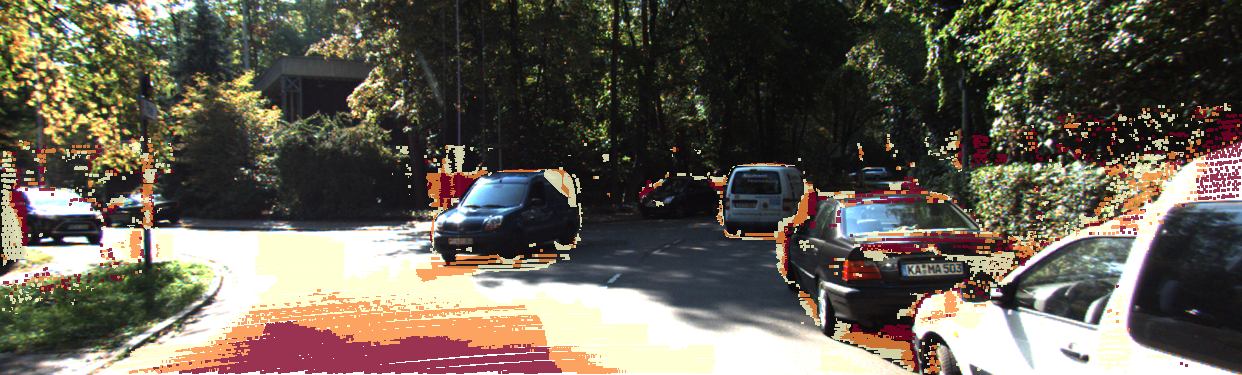}%
\includegraphics[width=0.5\linewidth]{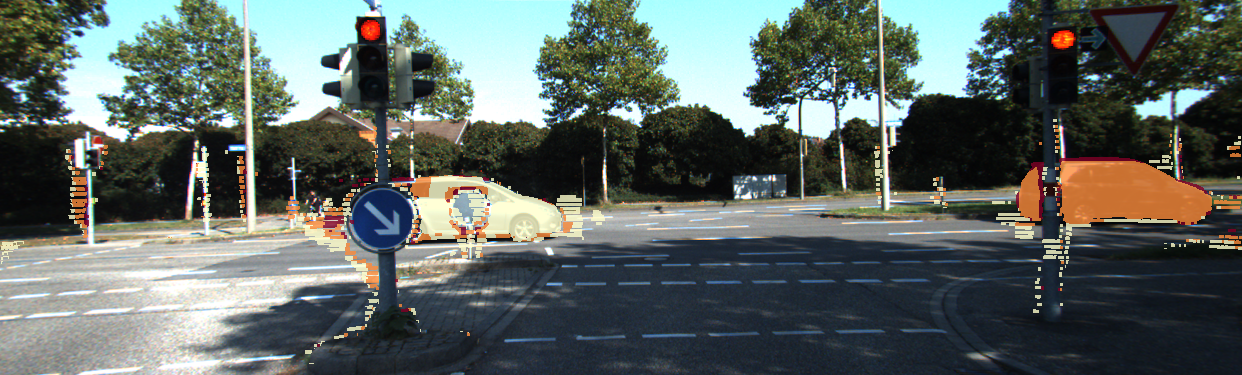}\\%
\includegraphics[width=0.5\linewidth]{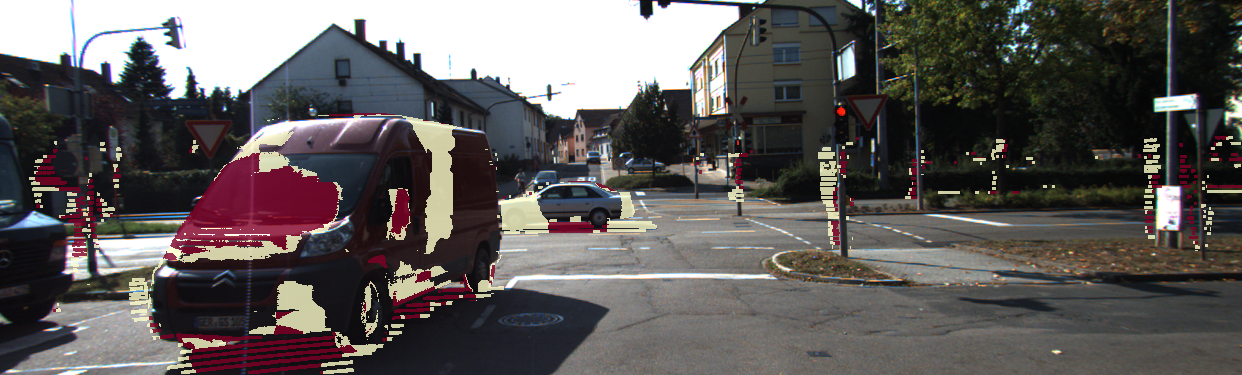}%
\includegraphics[width=0.5\linewidth]{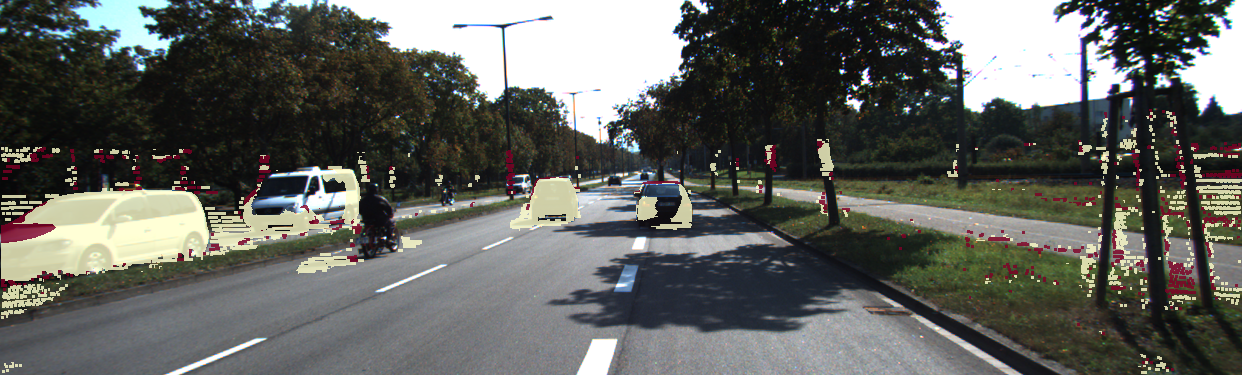}\\%
\caption{{\bf KITTI 2015 Scene Flow Analysis.} The averaged errors of the 15 best-performing scene flow methods published on the KITTI 2015 Scene Flow benchmark. Red colors correspond to regions where the majority of methods yield bad pixels according to the 3px/5\% criterion defined in \cite{Menze2015CVPR}. Yellow colors correspond to regions where some of the methods fail. Regions that are correctly estimated by all methods are transparent.}
\label{fig:sceneflow_qualitative_results}
\end{figure*}

We show the accumulated errors of the top 5 methods on the KITTI scene flow benchmark in \figref{fig:sceneflow_qualitative_results}. Car surfaces are the most problematic regions due to matching problems and the independent motion of cars. Pixels close to the image boundary are another common source of error, especially on the road surfaces in front of the car, where large scale changes occur. Although local planarity and rigidity assumptions alleviate the problem, they are often violated due to complex geometric objects like vegetation, pedestrians, or bicycles. Superpixels grouping different surfaces due to wrong estimation of planes cause additional problems, especially at the boundaries of objects. Semantic image understanding seems a promising direction \citep{Ren2017THREEDV, Behl2017ICCV, Ma2019CVPR}, especially at the object level, by segmenting car instances. 
However, an additional network has to be trained for obtaining this information, and prediction errors can lead to irreversible errors in the final scene flow estimation.
Leveraging temporal information \citep{Vogel2015IJCV, Neoral2017CVWW} also leads to improvements and should be exploited whenever possible. Especially, long-term temporal interactions could allow to alleviate ambiguities and improve. However, obtaining a robust, accurate, and real-time multi-frame scene flow estimate remains an open problem that requires further work.

	\chapter{Mapping, Localization \& Ego-Motion Estimation}
\chaptermark{Mapping, Localization \& Ego-Motion}
\label{chap:EgoMotionEstimation}

\begin{figure}[t]
	\centering
	\includegraphics[width=1.00\columnwidth]{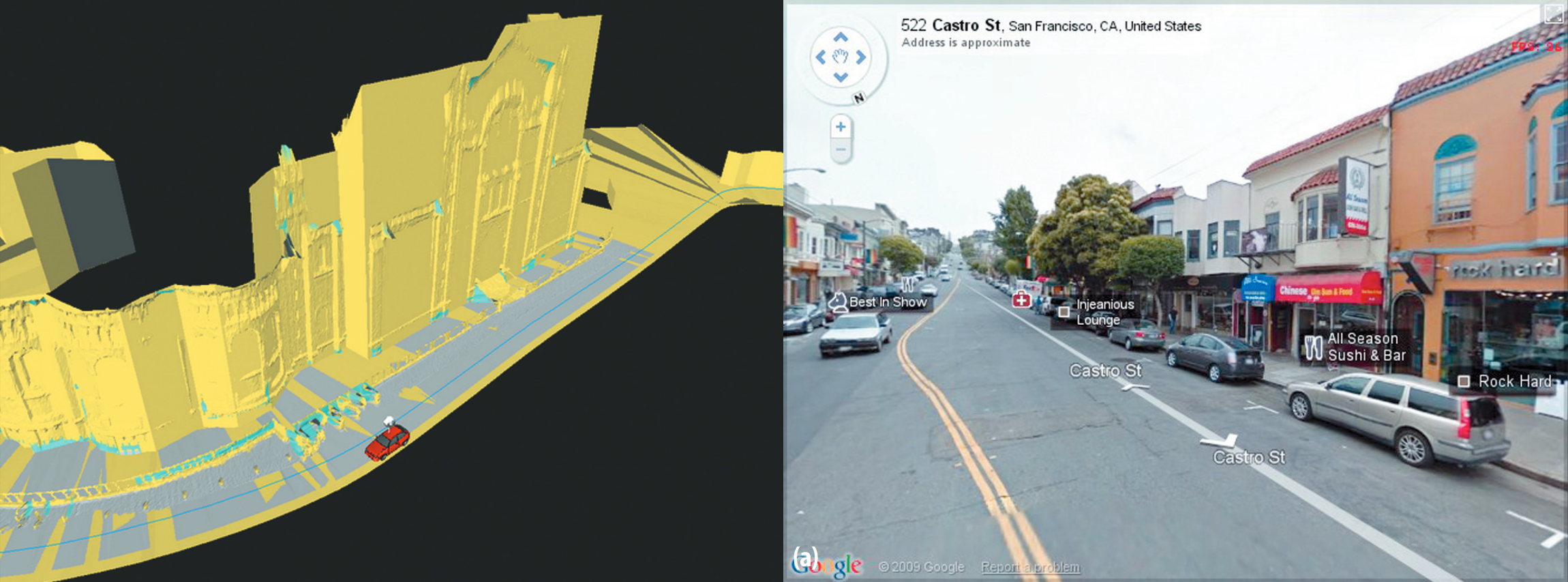}
	\caption[Google Street View]{\textbf{Google Street View.} Dominant scene surfaces reconstructed from images and laser range data (left) and a scene from the Google Street View project \protect\citep{Anguelov2010COMPUTER}. \figsourceC{\protect\citet{Anguelov2010COMPUTER}}{2010}{IEEE}.}
	\label{fig:street_view}
\end{figure}
\section{Problem Definition}
Navigating a vehicle requires a precise understanding of the position and orientation of the car. Localization is a well-studied problem in both robotics and vision, covering a broad range of techniques from indoor localization using noisy sensory measurements to locating where a picture was taken. From an autonomous driving perspective, the main task is to localize the vehicle on a map in order to exploit static features provided by the map. The task of generating a map of the world is defined as the mapping problem.
In this chapter, we discuss approaches for generating both metric as well as semantic maps. While metric maps allow for accurate localization, semantic maps provide problem-specific information such as the location of parking areas.
Maps for localization can be generated offline, exploiting accurate, but computationally expensive optimization techniques. 

In contrast to localization, the ego-motion estimation problem considers the change in position and orientation of the vehicle.
While this problem can be addressed more efficiently than the localization problem (the previous position is assumed to be known), small inaccuracies quickly accumulate to larger drifts.
Approaches for Simultaneous Localization and Mapping (SLAM) address this problem by detecting loop closures to correct for drift.

\section{Mapping}
Street, aerial, and satellite imagery enable the generation of precise metric and semantic maps. Depending on the required level of detail, various computer vision techniques, \ie multi-view reconstruction, scene understanding, or semantic segmentation, are typically employed for generation maps.

\subsection{Metric Maps} 
For autonomous driving, 2D metric maps (\ie representing information in bird's-eye view) are usually sufficient for localization.
Methods for scene understanding, such as \citep{Ess2009BMVC, Wojek2008ECCV, Wojek2010ECCV, Geiger2014PAMI, Topfer2015TITS, Seff2016ARXIV, Zhang2013ICCV} discussed in \chpref{chap:scene_understanding} can also be used to extract road features.
3D information can be obtained using multi-view reconstruction techniques operating on street-level  \citep{Agarwal2009ICCV, Frahm2010ECCV, Frahm2010JPRS} (\secref{sec:structure_from_motion}) or aerial \citep{Frueh2005IJCV, Bodis-Szomoru2016ICPR, Duan2016ECCV} images. 

The Google Street View project \citep{Anguelov2010COMPUTER} is a prominent example of a large collection of panoramic images that are registered with respect to each other to form a world map, see \figref{fig:street_view}. For registering the dataset, \citet{Anguelov2010COMPUTER} estimate the pose of the vehicle using a Kalman filter, fusing data from GPS, wheel encoder, and inertial navigation \citep{Klingner2013ICCV}. The pose estimates are refined with a probabilistic graphical model, and the 3D scene geometry is recovered by robustly fitting coarse meshes to the 3D measurements.

\citet{Levinson2007RSS} propose to construct a map based on aggregated reflectance measurements from a LiDAR scanner. They exploit these maps for centimeter-accurate LiDAR-based localization during the DARPA Urban Challenge. In contrast, \citet{Geiger2009IV} presents an approach for road mosaicing in dynamic environments with the goal of creating obstacle-free bird's-eye views. The road surface is extracted using optical flow on Harris corners and approximated by a plane. Afterwards, multiple road reconstructions are combined using multi-band blending.

\subsection{Semantic Maps} 
Metric maps ignore semantic information, which is important for some tasks such as automated parking. 
Semantic maps are necessary to address this problem. Several approaches
address the creation of semantic maps \citep{Wegner2013CVPR, Wegner2015JPRS, Montoya2015CPIA, Verdie2014IJCV, Mattyus2015ICCV, Mattyus2016CVPR, Wegner2016CVPR, Mattyus2016CVPR} . Scene understanding approaches like \citep{Ess2009BMVC, Geiger2014PAMI} also estimate semantic classes to extract road topologies but do not create a semantic map. 

\citet{Sengupta2012IROS} present an approach to generate a semantic overhead map of an urban scene from street-level images. They formulate the problem using two CRFs. The first is used for semantic image segmentation of the street view images treating each image independently. Each street view image is then projected into an overhead view. These views are then aggregated over many images to form the input for a second CRF producing a semantic labeling of the ground plane. 

In contrast, \citet{Grimmett2015ICRA} fuse semantic and metric maps for vision-only automated parking. They update the map with static and dynamic labels and use active learning for lane, parking space, and pedestrian crossings detection. 

\section{Localization}
\label{sec:Localization}
\begin{figure}[t]
	\centering
	\includegraphics[width=1.00\columnwidth]{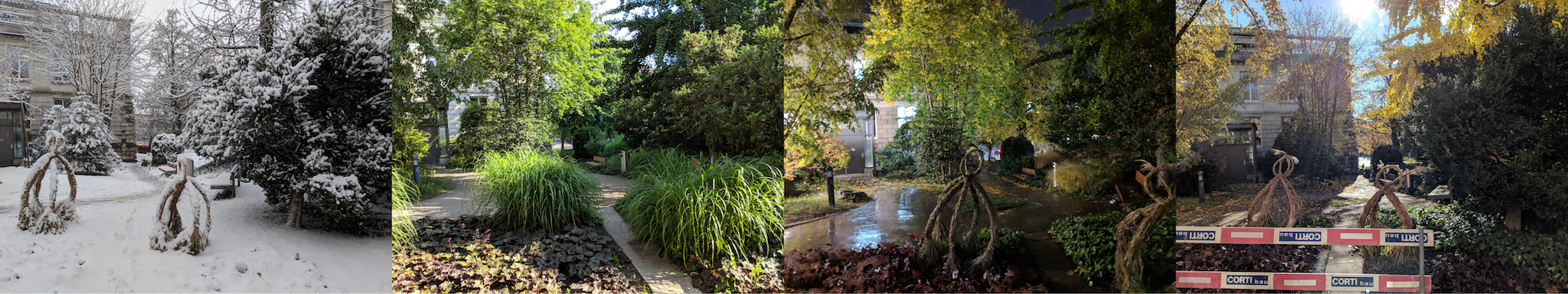}
	\caption[Appearance Changes in Localization]{\textbf{Appearance Changes in Localization.} Examples for different weather conditions, seasons, and day times for a scene from the Workshop organized by \protect\citet{Hammarstrand2019CVPRWORK}. \figsource{\protect\citet{Hammarstrand2019CVPRWORK}}. 
	}
	\label{fig:LocHammarstrand2019CVPRWORK}
\end{figure}
Localization can be performed using either a sensor like GPS or visual information based on images. Using GPS alone typically provides an accuracy of around 5 meters. Although centimeter-level precision is possible in unobstructed environments using correction signals and a combination of several sensors as in the KITTI car \citep{Geiger2012CVPR}, it is often rendered infeasible in traffic scenes with several disturbing effects such as occlusions by vegetation and buildings or multi-path effects due to reflections. Therefore, image-based localization independent of satellite systems remains highly relevant.

Visual localization techniques are commonly classified into metric and topological methods. Metric localization \citep{Dellaert1999ICRA, Oh2004IROS} is achieved by computing the 3D pose with respect to a map. Topological localization approaches \citep{Li2009ICCV, Zheng2009CVPR, Hays2008CVPR} provide a coarse estimate from a finite set of possible locations that are represented as nodes in a graph and connected by edges that link them according to some distance or appearance criteria. Metric localization can be very accurate, but is usually not suitable for very long sequences, while topological localization may be more reliable, but only provides rough estimates. 

\boldparagraph{Metric Localization}
The problem of metric map localization has been traditionally addressed using Monte Carlo methods which recover the probability distribution over the agent's pose by drawing a set of samples. \citet{Dellaert1999ICRA} define indoor localization in two steps, global position estimation and tracking of the local position over time. Instead of modeling the probability density function itself, they represent uncertainty by a set of samples and update the representation over time using Monte Carlo methods. This allows them to model arbitrary multi-modal distributions in a memory-efficient way. 

Outdoor localization is, in general, more challenging compared to the indoor localization task due to its scale and often unreliable sensor information, \eg GPS failures. \citet{Oh2004IROS} use semantic information available in maps to compensate for the failure cases of GPS sensors. By exploiting knowledge about the environment, they assign low probabilities to implausible map locations, \eg inside buildings. They incorporate these map-based priors into their particle filter formulation to bias the motion model towards areas of higher probability.

\boldparagraph{Topological Localization}
Early image-based techniques \citep{Li2009ICCV, Zheng2009CVPR} approach the problem of localizing in topological maps as classification into one of a predefined set of places which are often referred to as ``landmarks''. Others \citep{Hays2008CVPR, Cummins2008IJRR, Paul2010ICRA, Torii2015CVPR, Arandjelovic2016CVPR} create a database of images with known locations and formulate localization as an image retrieval problem. These methods require a similarity measure to compare images based on local or global appearance cues. The larger the database, the more difficult the localization task becomes. Challenges include appearance changes, similar-looking places, and changes due to viewpoint or position. 
In \figref{fig:LocHammarstrand2019CVPRWORK}, we show an example for the appearance change of a scene over different seasons from the Workshop organized by \citet{Hammarstrand2019CVPRWORK}.

\citet{Lowry2016TR} provide a comprehensive review of visual place recognition techniques. Given a map of the environment, the goal of place recognition is to decide whether the current observation is a place already included in the map, and if so, which one.

\boldparagraph{Topometric Localization}
In contrast to purely topological methods, the graph of a topometric localization model is more fine-grained: each node corresponds to a metric location without semantic meaning. 
Towards this goal, \citet{Badino2012ICRA} propose to construct a graph using the vehicle's position from GPS at fixed distance intervals while associating visual or 3D features to the corresponding graph node. At runtime, real-time localization is performed using a Bayes filter to estimate the probability distribution of the vehicle position along the route by matching features extracted from the sensor data to the map’s feature database. \citet{Brubaker2016PAMI} leverage a graph-based representation. In contrast to traditional localization approaches, however, they do not require a visual feature database of the environment, but instead, directly build this graph from road networks extracted from OpenStreetMap. They further propose a probabilistic model that allows inferring a distribution over the vehicle location along the edges of this road graph using visual odometry measurements. For tractability in very large environments, they leverage several analytic approximations for efficient inference yielding higher stability compared to particle-based filtering techniques.

\boldparagraph{Scale and Accuracy}
The scale of the target area is a distinctive property to compare different approaches and is related to the accuracy achieved. Both scale and accuracy depend on the methodology used, such as map-based approaches \citep{Brubaker2016PAMI} which cover a large area but might suffer from the errors on the map compared to descriptor-based approaches \citep{Badino2012ICRA, Schreiber2013IV} on a smaller area. While the descriptor-based method of \citet{Badino2012ICRA} achieves an average localization accuracy of 1 m over an 8 km route, the localization approach of \citet{Brubaker2016PAMI} which requires only road networks as input attains an accuracy of 4 m on a 18 km$^2$ map containing 2,150 km of drivable roads.

\citet{Schreiber2013IV} point out that the required precision for autonomous driving and future driver assistance systems is in the range of a few centimeters and present a feature- and map-matching-based localization algorithm which can achieve centimeter-level accuracy on approximately 50 km of rural roads. They approach the problem from the perspective of lane recognition. First, they create a highly accurate map that contains road markings and curbs. Then while driving, they detect and match them to the map in order to determine the position of the vehicle relative to the markings.

\begin{figure}[t]
	\centering
	\includegraphics[width=1.00\columnwidth]{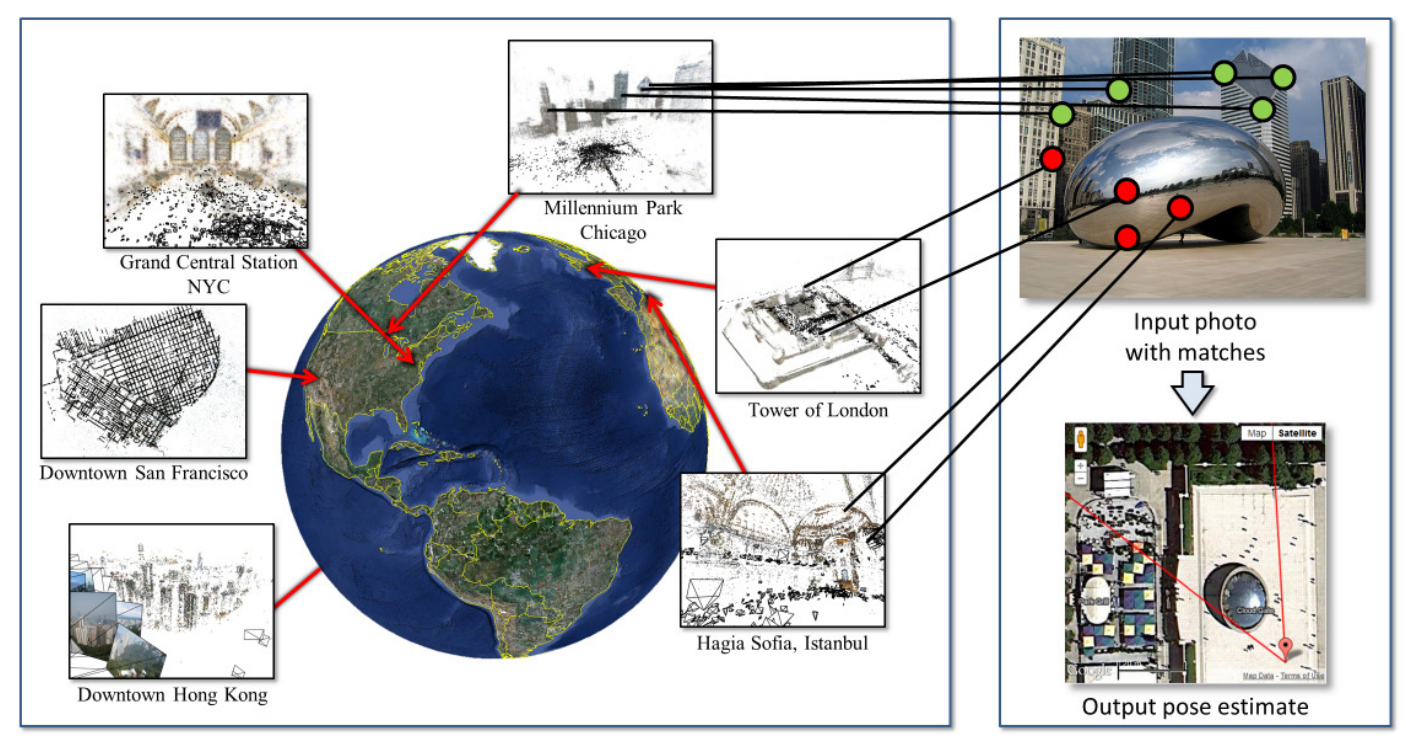}
	\caption[Structure-based Localization]{\textbf{Structure-based Localization.} A query image is matched to a database of geo-referenced structure-from-motion point clouds assembled from photos of places around the world (left). In structure-based approaches, the goal is to compute the geo-referenced pose of new query images by matching to a large database of feature descriptors (right). \figsourceSpringer{\protect\citet{Li2012ECCV}}{2012}{ECCV}.}
	\label{fig:localization_structure_based}
\end{figure}

\subsection{Structure-based Localization} 

While the output of the aforementioned localization approaches is either a rough camera position or a distribution over positions, another line of work which is known as ``structure-based localization'', aims to estimate all camera matrix parameters, including position, orientation, and sometimes also camera intrinsics. Estimating the intrinsics usually enables more accurate results. Localization is realized as a 2D-to-3D matching problem where the 2D points on the images are matched to a large, geo-registered 3D point cloud, and the pose is estimated with respect to correspondences as shown in \figref{fig:localization_structure_based}.

Direct matching by approximate nearest neighbor search using SIFT features usually results in many incorrect matches. Therefore, many approaches rely on the SIFT ratio test \citep{Lowe2004IJCV} to detect and reject ambiguous matches. This works well on small to medium scale scenes.  However, with growing model size, the discriminative power of the descriptors decreases, and many matches will be rejected by the ratio test. On the other hand, relaxing the ratio test leads to many ambiguous and wrong matches.

Several approaches \citep{Irschara2009CVPR,Sarlin2019CVPR,Li2012ECCV,Svarm2014CVPR, Zeisl2015ICCV} address this problem by restricting the search space.
\citet{Irschara2009CVPR,Sarlin2019CVPR} use image retrieval techniques to identify parts of the scene which likely include the query image. Afterwards, 2D-3D matching is performed to 3D points visible in the retrieved images.
In contrast, \citet{Li2012ECCV} find statistical co-occurrences of 3D model points in images and then use them as a sampling prior for RANSAC to exploit co-visibility relations. In addition, they employ a bidirectional matching scheme, forward from features in the image to points in the database and inverse from points to image features. They show that the bidirectional approach performs better than forward or inverse matching alone. 
\citet{Svarm2014CVPR, Zeisl2015ICCV} propose to use geometric cues to obtain matches that are likely to be inliers. They also exploit the gravity direction obtained from gravitational sensors and an approximation of the camera height to reduce the search space. 

Besides ambiguities, the efficiency of the matching stage and memory requirements to store the large number of descriptors contained in the model are also problems related to large scale. Therefore, several approaches use only a subset of the 3D points \citep{Lynen2015RSS} or present compression schemes for the descriptors \citep{Lynen2015RSS, Sattler2015ICCV, Sattler2016PAMI, Liu2017ICCVa, Camposeco2019CVPR} for more efficient matching or memory reduction.  \citet{Sattler2015ICCV, Sattler2016PAMI} use quantization into a fine vocabulary to accelerate the matching stage where each descriptor is represented by its word ID. \citet{Sattler2015ICCV} separate the difficult problem of finding a unique 2D-3D matching into two simpler ones. They first establish locally unique 2D-3D matches using a fine visual vocabulary and a visibility graph which encodes the visibility relation between 3D points and cameras. Then, they disambiguate these matches by using a simple voting scheme to enforce the co-visibility of the selected 3D points. Their experiments show that matching based on a visual vocabulary leads to state-of-the-art results. \citet{Sattler2016PAMI} propose a prioritized matching scheme based on quantization, focusing on efficiency. They significantly accelerate 2D-to-3D matching by considering more likely features first and terminating the correspondence search as soon as enough matches are found.
A hybrid approach combining the idea of working on a subset of 3D points and the compression of the descriptors is presented by \citet{Camposeco2019CVPR}. For a small subset of 3D points, they keep the full appearance information, while for a larger set of points, they store a compressed descriptor. This enables them to obtain a more complete representation of the scene with a memory consumption similar to the previous approaches.

\begin{figure}[t]
	\centering
	\includegraphics[width=1.00\columnwidth]{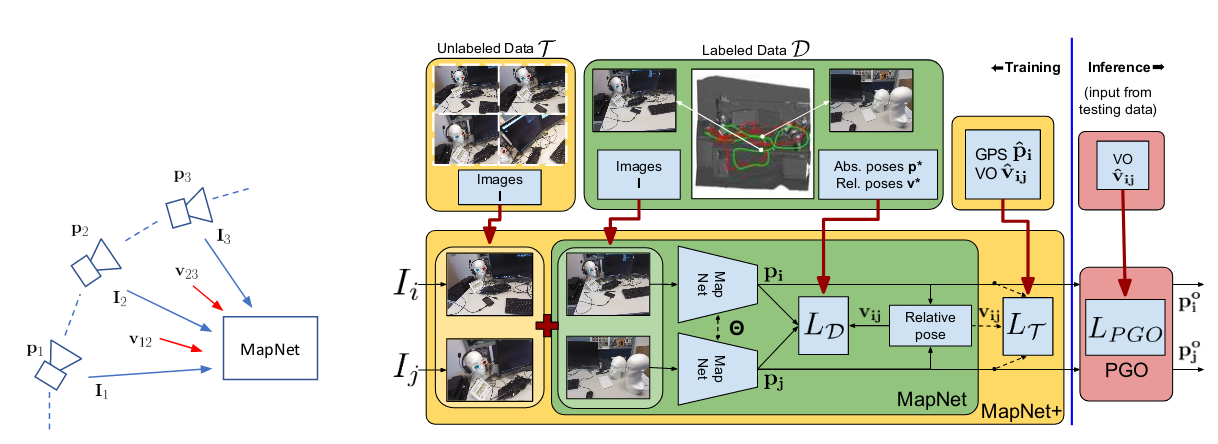}
	\caption[Learning Structure-based Localization]{\textbf{Learning Structure-based Localization.} MapNet proposed by \citet{Brahmbhatt2018CVPR} learns a map representation from images, visual odometry, and GPS (left). During inference (right) visual odometry is used to update the maps in a self-supervised fashion and pose graph optimization (GPO) allows for further refinement.  \figsourceC{\protect\citet{Brahmbhatt2018CVPR}}{2018}{IEEE}.}
	\label{fig:Brahmbhatt2018CVPR}
\end{figure}

\boldparagraph{Deep Learning}

The motivation for using CNNs for structure-based localization is to learn high-level information which might help to handle problems like textureless areas, motion blur, and illumination changes. In contrast to classical localization approaches whose runtime depends on several factors such as the number of features found in a query image or the number of 3D points in the model, the runtime of CNN-based approaches only depends on the size of the network.

\citet{Kendall2015ICCV} and \citet{Walch2017ICCV} use a convolutional neural network to regress the camera pose from a single RGB image in an end-to-end manner. \citet{Kendall2015ICCV} modify GoogLeNet \citep{Szegedy2015CVPR} by replacing the softmax classifiers with affine regressors and inserting another fully connected layer before the final regressor, which can be used as a localization feature vector for further analysis. The final architecture, dubbed PoseNet, is initialized by using the weights of classification networks trained on giant datasets such as ImageNet \citep{Deng2009CVPR} and Places \citep{Zhou2014NIPS}. The network is further fine-tuned on a new pose dataset which was automatically created using SfM to generate camera poses from a video of the scene. \citet{Walch2017ICCV} use a similar approach, but in addition, they spatially correlate each element of the output of the CNN using Long Short-Term Memory (LTSM) units. This way, the network is able to capture more contextual information and outperform PoseNet in different localization tasks, including large-scale outdoor, small-scale indoor, and a newly proposed large-scale indoor localization benchmark. 

Recently, \citet{Brahmbhatt2018CVPR} proposed MapNet for representing maps as deep neural networks. They exploit visual odometry and GPS in addition to images for image-based localization and formulate geometric constraints as additional loss terms. Thus, the model can be updated in a self-supervised fashion using unlabeled data. This allows them to significantly improve in comparison to PoseNet-based approaches. The model is illustrated in \figref{fig:Brahmbhatt2018CVPR}.

While previous methods \citep{Kendall2015ICCV,Walch2017ICCV,Brahmbhatt2018CVPR} regress the absolute pose in a given scene, another line of work \citep{Saha2018BMVC, Balntas2018ECCV} proposes to learn the relative pose with respect to an image retrieved from a database. Eventually, the absolute pose is obtained from the known pose of the retrieved image and the relative pose. 

\citet{Sattler2019CVPR} notice that PoseNet-based approaches \citep{Kendall2015ICCV,Walch2017ICCV} are not able to outperform simple image retrieval approaches \citep{Torii2015CVPR} and learning-based approaches are in general still inferior to structure-based approaches such as \citep{Svarm2017PAMI}. 

\begin{figure}[t]
	\centering
	\includegraphics[width=0.80\columnwidth]{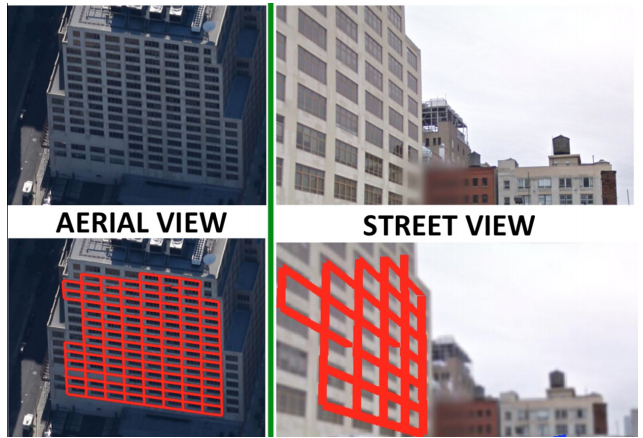}
	\caption[Aerial to Street-View Matching]{\textbf{Aerial to Street-View Matching.} Repeating patterns of buildings are exploited by regularity-driven approaches for aerial to street-view matching. \figsourceC{\protect\citet{Wolff2016CVPR}}{2016}{IEEE}.}
	\label{fig:localization_facade}
\end{figure}

\subsection{Cross-view Localization}
It is a very difficult endeavor to keep an up-to-date repository of ground-level imagery around the world. In contrast, establishing live maps from aerial and satellite images is comparably easier. This motivated the development of geo-localization approaches that try to register ground-level images to aerial imagery. The underlying idea is to learn a mapping between ground-level and aerial image viewpoints to localize a ground-level query in an aerial reference image database. 

\citet{Lin2013CVPR} match ground-level queries to other ground-level reference photos as in traditional geo-localization, but then use the overhead appearance and land cover attributes of those ground-level matches to build sliding-window classifiers in the aerial and land cover domain. In contrast to previous methods, they are able to localize a query even if it has no corresponding ground-level image in the database by learning the co-occurrence of features in different views. Inspired by the success of face verification algorithms using deep learning, \citet{Lin2015CVPR}  train a Siamese network to match cross-view pairs of the same location. Towards this goal, they collect a cross-view patch dataset using range data and camera parameters from Google Street View. Finally, they warp the dominant building surface plane to appear approximately as a 45\% aerial view. In contrast, \citet{Workman2015ICCV} use CNNs for extracting ground-level image features and predict these features from aerial images of the same location. This way, the CNN is able to extract semantically meaningful features from aerial images without manually specifying semantic labels. They conclude that the cross-view localization approach can obtain a precise estimate of the geographic locations which are distinctive from above. Otherwise, it can be used as a pre-processing step to a more expensive matching process.

\boldparagraph{Buildings Facades}
Several methods have been developed which specialize in building facades from cross-view matching. The repeating patterns yield valuable matching cues, as illustrated in \figref{fig:localization_facade}. By combining satellite and oblique bird's-eye views, \citet{Bansal2011ICM} first extract building outlines as well as facades and then match the ground image to oblique aerial images based on a statistical description of the facade pattern. \citet{Wolff2016CVPR} define a matching cost function to compare street-view motifs to aerial view motifs based on the similarity of color, texture, and edge-based context features.

\boldparagraph{Geo-Referenced Reconstruction}
Another line of work addresses the problem of geo-referencing a reconstruction by automatic alignment with a satellite image, floor plan, map, or other overhead views. \citet{Kaminsky2009CVPRWORK} compute the optimal alignment between SfM reconstructions and overhead images using an objective function that matches 3D points to image edges and imposes free space constraints based on the visibility of points in each camera. Matching ground and aerial images directly is a difficult endeavor due to the large differences in camera viewpoints, occlusions, and imaging conditions. Instead of seeking invariant feature detections, \citet{Shan2014THREEDV} propose a viewpoint-dependent matching technique by exploiting approximate alignment information and the underlying 3D geometry.

\subsection{Semantic Alignment from LiDAR}
Several companies acquire LiDAR data from scanners mounted on cars driving through cities to acquire  3D models of real-world urban environments. However, the accuracy of the 3D point positions acquired by the 3D scanners depends on the scanner poses predicted by GPS, inertial sensors, and structure-from-motion, which often fail in urban environments. These misalignments cause problems for point cloud registration methods. \citet{Yu2015CVPR} propose to align semantic features that can be matched robustly at different scales. By following a coarse-to-fine approach, they first successively align roads, facades, and poles which can be matched robustly. Afterwards, they match cars and other small objects which require better initial alignments to find correct correspondences. The use of semantic features provides a globally consistent alignment of LiDAR scans, and their evaluation shows improvement over the initial alignments.

\begin{figure}[t!]
	\centering
	\includegraphics[width=1.00\columnwidth]{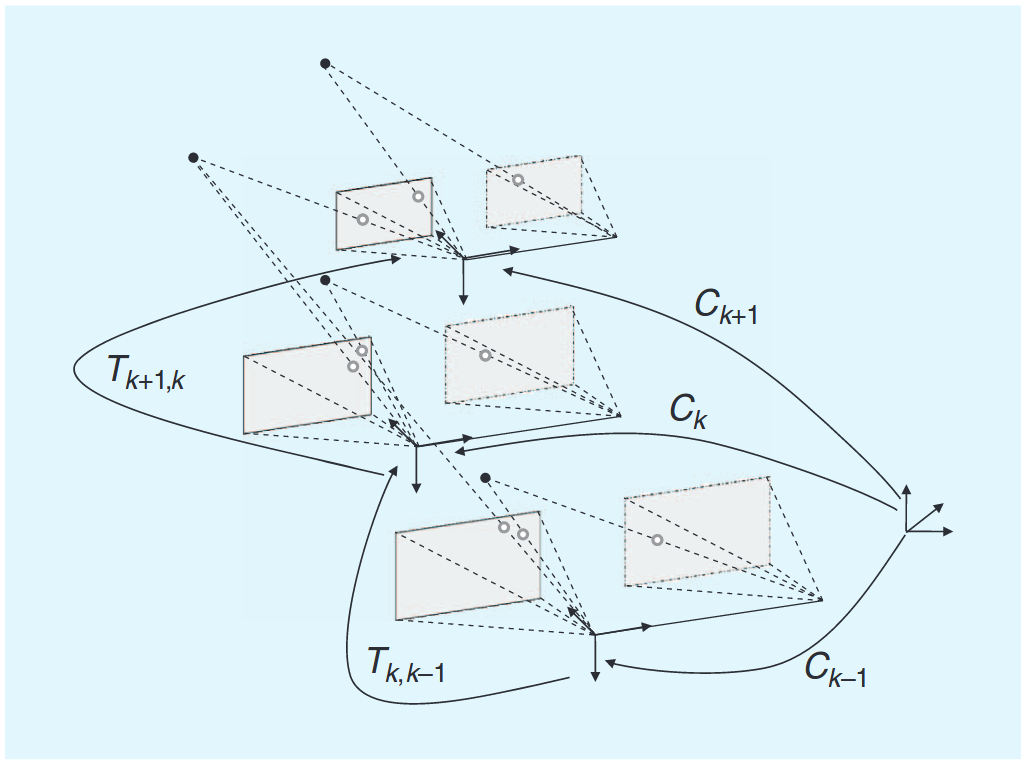}
	\caption[Visual Odometry]{\textbf{Visual Odometry.} Illustration of the incremental visual odometry approach by \protect\citet{Scaramuzza2011RAM}. The transformation $T_{k,k-1}$ between two adjacent camera systems is obtained using visual features. The accumulation of all transformations yields the absolute pose $C_k$ with respect to the initial coordinate frame $k=0$. \figsourceC{\protect\citet{Scaramuzza2011RAM}}{2011}{IEEE}.}
	\label{fig:visual_odometry_example}
\end{figure}

\section{Ego-Motion Estimation}
One of the simplest ways of estimating the ego-motion of a vehicle is to use the wheel angle in combination with the output of wheel encoders which measure the rotation of the wheel. These methods suffer from wheel slip in uneven terrain or adverse conditions and can not recover from errors in the measurements. Visual odometry and LiDAR-based odometry techniques that estimate ego-motion from visual observations (images or laser range measurements) are more robust in many situations and can correct for drift by loop closure detection, \ie by recognizing re-visited places (\secref{sec:LoopClosure}). In this section, we provide a summary of the most relevant visual odometry techniques for autonomous driving. For a more detailed survey on visual odometry techniques, we refer the reader to \citet{Scaramuzza2011RAM} and \citet{Fraundorfer2011RAM}.

In visual odometry, the goal is to recover a trajectory (\ie a sequence of poses) of one camera or a camera system comprising multiple cameras from images. Most approaches incrementally estimate the relative transformation between two frames and integrate this information over time to recover the full trajectory. The incremental approach is illustrated in \figref{fig:visual_odometry_example}. Methods on visual odometry can be roughly divided into two main categories: feature-based methods \citep{Longuet-Higgins1981Nature, Nister2004PAMI, Scaramuzza2009ICRA, Lee2013CVPR, Kitt2010IV, Mur-Artal2015TR} that extract features from key points to optimize a geometric error, and direct formulations \citep{Newcombe2011ICCV, Kerl2013ICRA, Engel2013ICCV, Engel2014ECCV, Fanani2016IV, Fanani2017IVC, Zhu2017IJCAI, Yang2018ECCVb, Engel2018PAMI} which directly operate on raw measurements by optimizing the photometric error.  

\begin{figure}[t!]
	\centering
	\includegraphics[width=1.00\columnwidth]{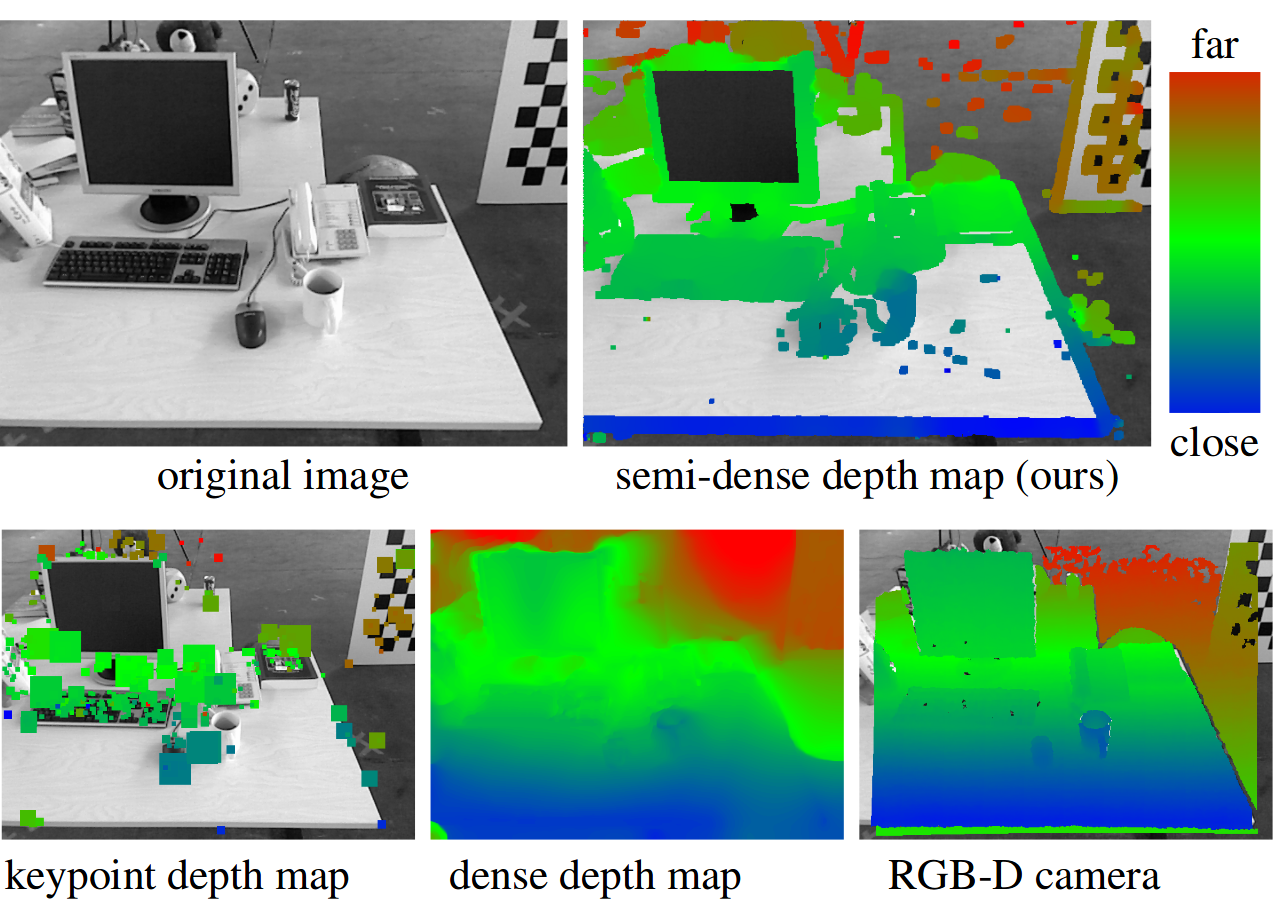}
	\caption[Semi-Dense Depth Maps]{\textbf{Semi-Dense Depth Maps.} The semi-dense depth map representation of \protect\citet{Engel2013ICCV} (top right) in comparison to key points \protect\citep{Klein2007ISMAR} (bottom left), a dense depth \protect\citep{Matthies1988CVPR} (bottom middle), and the output of a dedicated RGB-D camera \protect\citep{Sturm2012IROS} (bottom right). \figsourceC{\protect\citet{Engel2013ICCV}}{2013}{IEEE}.}
	\label{fig:semi_dense_depth}
\end{figure}

Feature-based methods typically detect corners in the image and match the corresponding feature descriptors across different images. While these approaches are very efficient, they discard valuable information, \eg straight or curved edges, that are very common in man-made environments. 
In contrast, direct methods leverage structural information in the entire image. Therefore, these methods usually achieve higher accuracy and robustness in environments with fewer key points.
In addition, they allow to simultaneously estimate semi-dense \citep{Engel2013ICCV, Engel2014ECCV} and even dense depth maps \citep{Sturm2012IROS, Newcombe2011ICCV}, as illustrated in \figref{fig:semi_dense_depth}.
However, direct methods suffer more from local minima in the optimization problem compared to feature-based methods, in particular when the pose initialization is far from the true solution.
Initially, the field was dominated by feature-based methods since they are typically more efficient, but direct formulations have recently grown in popularity due to their increased accuracy \citep{Newcombe2011ICCV, Sturm2012IROS, Kerl2013ICRA, Engel2013ICCV, Engel2014ECCV, Fanani2016IV, Fanani2017IVC, Zhu2017IJCAI, Yang2018ECCVb, Engel2018PAMI}.

\subsubsection{Feature-based Methods} 

\boldparagraph{2D-to-2D Matching} 
Depending on how corresponding points between two time steps are represented (2D or 3D), different methods must be used to obtain the camera transformation. The essential matrix (or fundamental matrix), which represents the epipolar geometry between the two cameras and contains relative pose information, can be recovered from 2D feature matches (2D-to-2D). One of the most popular algorithms for estimating the essential or fundamental matrix is the eight-point algorithm \citep{Hartley1995ICCV}.
The five-point algorithm \citep{Nister2004PAMI} is a minimal solution that only applies to the scenario of calibrated cameras. \citet{Scaramuzza2009ICRA} estimate the essential matrix from monocular images with only one 2D feature correspondence using non-holonomic constraints of wheeled vehicles imposing a restrictive motion model.

In general, visual odometry with monocular images cannot recover the metric scale due to the inherent scale ambiguity. \citet{Lee2013CVPR} extend \citep{Scaramuzza2009ICRA} to a novel two-point minimal solution that is able to obtain the metric scale using a multi-camera system. In contrast to the non-holonomic constraints, \citet{Lee2014CVPR} assume the vertical directions to be known (from an Inertial Measurement Unit) and propose a minimal four-point and linear eight-point algorithm for a multi-camera system. \citet{Kitt2010IV} estimate the ego-motion using trifocal geometry, which relates features between three images. 
Most algorithms employ RANSAC for robust estimation. The number of iterations necessary to guarantee that a correct solution is found with RANSAC depends on the number of points from which the model can be instantiated. 
Minimal solvers allow to the reduction of the number correspondences leading to a reduced number of iterations and runtime of the approach. 

Omnidirectional cameras discussed in \secref{sec:calibration_omnidirectional_cam} enable feature-based approaches that extract and match interest points from all around the car.
The increased field of view makes the visual odometry problem more constrained and consequently allows for more accurate visual odometry. \citet{Scaramuzza2008TR} exploit this observation and estimate the ego-motion of the vehicle relative to the road from a single, central omnidirectional camera using a homography-based tracker for the ground plane and an appearance-based tracker for the rotation of the vehicle.

\boldparagraph{3D-to-2D Matching} 
If stereo or RGB-D information is available, a simple solution to the visual odometry problem is to project 3D features from one image into the other view and optimize for the pose by minimizing reprojection errors. Following this idea, \citet{Geiger2011IV} present a real-time visual odometry and sparse 3D reconstruction method. They detect sparse features in stereo images using blob and corner detectors and estimate the vehicle's ego-motion by minimizing the reprojection error of the projected 3D features. In addition, they propose a real-time stereo reconstruction algorithm \cite{Geiger2010ACCV} and fuse disparity maps over time into a coherent city-scale 3D reconstruction.

\boldparagraph{3D-to-3D Matching} 
When dealing with 3D correspondences (3D-to-3D), the relative transformation between two time steps can be obtained by aligning the two sets of 3D features, for instance, using the iterative closest point (ICP) algorithm \cite{Besl1992PAMI}. In visual odometry, the features extracted from images are projected into 3D using depth, whereas LiDAR-based approaches such as \citet{Zhang2014RSS,Zhang2015ICRA} directly obtain the 3D points from the sensor. However, the triangulated 3D points from stereo will exhibit a large anisotropic uncertainty due to the small baseline and the quadratic increase of errors \wrt distance. Thus it is more natural to minimize reprojection errors in the images where error statistics can be approximated more easily. Laser-based approaches do not suffer from this problem and thus typically optimize in 3D space.

\subsubsection{Direct Methods} 
In contrast to feature-based methods that optimize reprojection errors, direct approaches optimize the photometric error for estimating motion. \citet{Engel2013ICCV} estimate a semi-dense inverse depth map for whole-image alignment of monocular images. Depth is estimated using multi-view stereo for pixels with non-negligible gradients and is represented by a Gaussian probability distribution. They propagate depth information from frame to frame and obtain camera poses by minimizing the photometric error. With this semi-dense formulation, they achieve comparable performance to fully dense methods \citep{Newcombe2011ICCV} while not requiring a depth sensor \citep{Kerl2013ICRA}. \citet{Engel2018PAMI} present a direct sparse approach for monocular visual odometry. They use a probabilistic model and jointly optimize all model parameters (camera poses, camera intrinsics, and inverse depth) in real-time. 

\subsection{Drift} 
The incremental approach to ego-motion estimation greatly suffers from drift caused by the accumulation of estimation errors of the individual transformations. One way of alleviating the drift problem is to use an iterative refinement over several images that are observed most recently. In feature-based approaches, this is done by reprojecting image points into 3D by triangulation and minimizing the sum of squared reprojection errors (sliding window bundle adjustment or windowed bundle adjustment). 
However, simpler techniques such as a proper selection of the extracted features can also reduce drift. \citet{Kitt2010IV} use bucketing to obtain well distributed corner-like feature matches, whereas \citet{Deigmoeller2016GCPR} use various heuristics on flow and depth estimation to reject non-stable features. 

The drift problem can also be addressed with simultaneous localization and mapping (SLAM) discussed in \secref{sec:slam}, which jointly estimates the location and a map of the environment to recognize places that have been visited before. The detection of already mapped places is also known as ``loop closure detection''.
If a loop has been detected, additional constraints can be added to the bundle adjustment problem, which leads to globally consistent maps and vehicle poses.
However, poses are only corrected in hindsight, and thus, the drift problem persists during longer periods in which no loop closure can be detected. Furthermore, as loop closure detection is computationally expensive and computation increases with the length of the trajectory, such techniques are often only executed sporadically and not with every new incoming frame.

\begin{figure}[t]
	\centering
	\includegraphics[width=1.00\columnwidth]{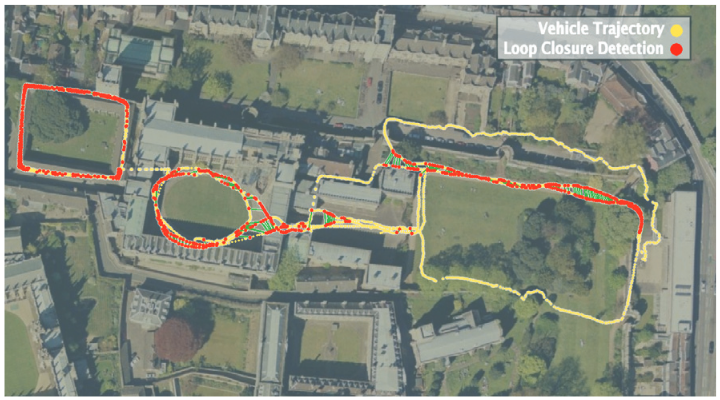}
	\caption[Loop Closure Detection]{\textbf{Loop Closure Detection.} Loop closure with appearance-based matching overlaid on an aerial image by \protect\citet{Cummins2008IJRR}. Images that are matched with a probability larger than $99\%$ are marked in red. \figsourceSage{\protect\citet{Cummins2008IJRR}}{2008}{IJRR}.}
	\label{fig:slam_loop_closure}
\end{figure}

\subsection{Loop Closure Detection} 
\label{sec:LoopClosure}
The relocalization in already mapped areas is an important subproblem of SLAM, known as loop closure detection. Relocalization is used to correct drift in the trajectory and inaccuracies in the map caused by drift. 

\citet{Cummins2008IJRR} present a probabilistic approach for the recognition of places based on their appearance. They learn a generative model of appearances using a bag-of-words model as distinctive combinations of visual words will often arise from common objects. The generative model is robust and works even in visually repetitive environments. The performance of the approach is demonstrated on a self-recorded dataset and visualized in \figref{fig:slam_loop_closure}. \citet{Paul2010ICRA} extend this idea by incorporating pairwise distances between words coupled to the observation of visual words using a random graph. The random graph models the pairwise distance between words besides their distribution of occurrences. In contrast, \citet{Lee2013IROS} consider a pose graph with vertices representing camera poses and edges representing constraints between the poses. They show that the relative pose with metric scale between two loop-closing vertices can be obtained from the epipolar geometry of a multi-camera system with overlapping views.

Image-based loop closure detection can become unreliable in case of strong illumination or viewpoint changes. In contrast, LiDAR-based localization is not affected by changes in illumination and does not suffer as much from changes in viewpoint due to the captured 3D geometry and the large field of view. \citet{Dube2017ICRA} propose a loop closure detection algorithm based on matching 3D segments. Segments from the point cloud are extracted and described using a combination of descriptors. Matching of segments is performed by obtaining candidates with k-d tree search in feature space and estimating matching scores using a random forest. 

\subsection{Simultaneous Localization and Mapping (SLAM)}
\label{sec:slam}
A detailed map of the environment simplifies planning and navigation in autonomous vehicles. However, in places for which no map is provided or the map is outdated or incomplete, the autonomous car must locate itself while generating the map. Further, the map needs to be updated continuously to reflect environmental changes over time. In this context, SLAM refers to the task of simultaneous estimation of the location of an agent while continuously constructing a map of the environment. While SLAM addresses a similar problem as structure-from-motion techniques discussed in \secref{sec:structure_from_motion}, SLAM approaches focus particularly on large-scale environments, loop-closure detection, and real-time performance. 

Traditionally, a map is represented by a set of landmarks that may correspond to semantically meaningful parts or detected image features. Early approaches to SLAM have addressed the problem with Bayesian formulations using extended Kalman filters \citep{Smith1987ICRA} or particle filters \citep{Montemerlo2002AI}. Given the last state and new observations, the current state, represented by pose, velocity, and the locations of the landmarks is recursively updated. However, this formulation is not applicable to large environments since the belief state and time complexity of the filter update grow quadratically with the number of landmarks in the map ($n$). 

One solution for reducing complexity is to leverage filtering techniques that maintain a tractable approximation of the belief state as proposed by \citet{Paskin2003IJCAI}. However, filtering may lead to inconsistent maps when applied to non-linear SLAM problems \citep{Julier2001ICRA}. In contrast, full SLAM approaches, such as graph-based or least-squares formulations, provide more accurate solutions as they consider all poses at once. \citet{Kaess2008TR} propose an incremental smoothing and mapping approach based on fast incremental matrix factorization. They extend their earlier work \citep{Dellaert2006IJRR} on factorizing the matrix of a non-linear least-squares problem to an incremental approach that only recalculates entries which change in the matrix.
\citet{Kaess2012IJRR} introduce the Bayes tree, a novel data structure, which allows for a better understanding of the connection between inference in graphical models and sparse matrix factorization. Factored probability densities are encoded in the Bayes tree which naturally maps to a sparse matrix. 
Recently, \citet{Lenac2018IJRR} proposed a filtering-based SLAM method that is able to compete with graph-based optimization techniques.

\begin{figure}[t!]
	\centering
	\includegraphics[width=1.00\columnwidth]{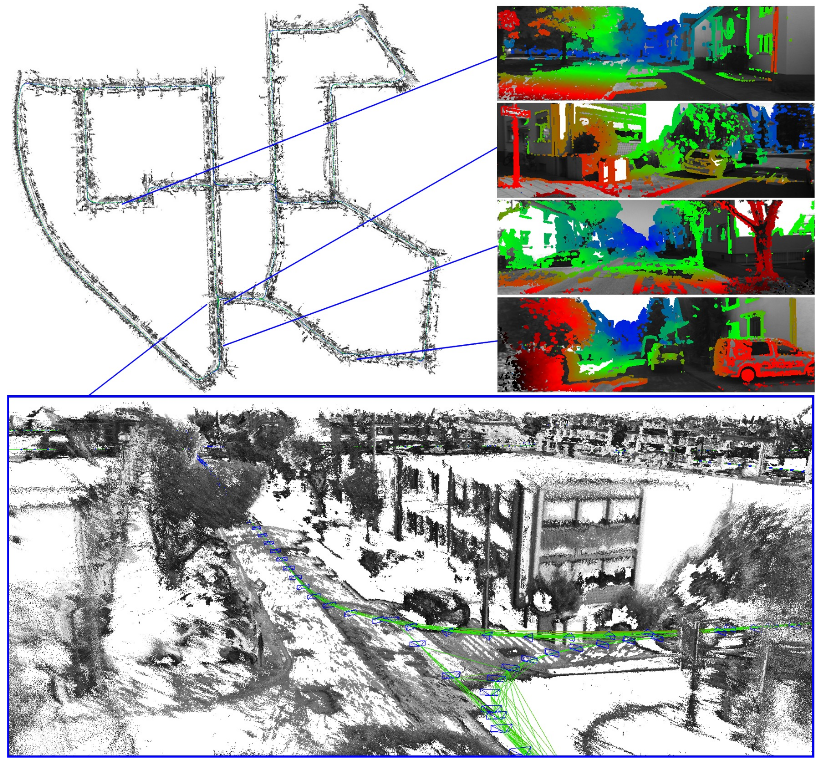}
	\caption[Stereo LSD-SLAM]{\textbf{Stereo LSD-SLAM.} \protect\citet{Engel2015IROS} compute accurate camera movement as well as semi-dense probabilistic depth maps in real-time. The depth visualization uses blue for far away scene points and red for close objects. \figsourceC{\protect\citet{Engel2015IROS}}{2015}{IEEE}.}
	\label{fig:visual_odometry_lsd_slam}
\end{figure}

\boldparagraph{Stereo SLAM}
Stereo cameras are a popular choice for tackling the SLAM problem since they allow to estimate the depth while simultaneously providing detailed information of an objects' appearance (in contrast to LiDAR sensors).
\citet{Lategahn2011ICRA} propose a dense stereo visual SLAM method that estimates a dense 3D map. Using a sparse visual SLAM system, they obtain the pose and a sparse map. For the dense 3D map, they compute a dense representation from stereo in a local coordinate system and continuously update the map by tracking the local coordinate systems with the sparse SLAM system. 
\citet{Engel2015IROS} propose LSD-SLAM, a real-time large-scale direct SLAM algorithm that couples static stereo from a camera setup with temporal multi-view stereo (\figref{fig:visual_odometry_lsd_slam}). This allows them to estimate the depth of pixels that are under-constrained in static stereo while avoiding scale-drift that occurs using multi-view stereo. The images are directly aligned based on the photoconsistency of high contrast pixels. \citet{Mur-Artal2015TR} use the ORB features proposed by \citet{Rublee2011ICCV} for tracking, mapping, relocalization, and loop closure. They combine methods from loop detection \citep{Galvez-Lopez2012TR}, loop closing \citep{Strasdat2010RSS,Strasdat2011ICCV}, and pose graph optimization \citep{Kuemmerle2011ICRA} into a single system which they call ORB-SLAM and which became one of the most widely used SLAM systems today.

A fusion approach is proposed by \citet{Leutenegger2013RSS} in order to take advantage of the complementary nature of visual and inertial cues. They use a non-linear optimization approach and integrate IMU measurements with reprojection errors into a joint cost function. 
Similarly, \citet{Usenko2016ICRA} also propose a joint visual-inertial SLAM method. However, they present a fully direct method based on \citep{Engel2015IROS} that estimates geometry from semi-dense depth maps in contrast to sparse key points.

\boldparagraph{Environmental Changes}
Changes in the environment that might not be represented in the map are a major challenge in SLAM. \citet{Levinson2007RSS} alleviate this problem by creating a map comprising of features that are very likely to be static over time. Using 3D LiDAR, they retain only flat surfaces and obtain an infrared reflectivity map of overhead views of the road surface. The map is then used to locate a vehicle with a particle filter in real-time. \citet{Levinson2010ICRA} extend this work considering maps as probability distributions over environment properties instead of a fixed representation. Specifically, every cell of the probabilistic map is represented as its own Gaussian distribution. This allows them to represent the world more accurately and localize with fewer errors. In addition, they use offline SLAM to align multiple passes of the same environment at different times to establish an increasingly robust understanding of the world.

\section{Datasets}
Several datasets have been considered in the localization and ego-motion estimation literature. 
The popular dataset 7 Scenes from \citet{Shotton2013CVPR} focuses only on indoor scenes. 
Large-scale reconstruction datasets such as Vienna \citep{Irschara2009CVPR},  Dubrovnik \citep{Li2010ECCV}, Rome \citep{Crandall2011CVPR} are very popular in particular for structure-based localization methods. With the introduction of deep learning to structure-based localization, \citep{Kendall2015ICCV} presented a new outdoor localization dataset (Cambridge Landmarks dataset), which became popular for CNN-based approaches. Most of the aforementioned datasets are limited in their variety in terms of weather conditions and seasons, which are important factors for evaluating the robustness of localization systems. To address this issue, \citet{Carlevaris-Bianco2016IJRR} proposed a new long-term vision and LiDAR dataset created on the campus of the University of Michigan comprising 27 sessions. Recently, \citet{Sattler2018CVPR} presented three datasets for the same problem: Aachen Day-Night, RobotCar Seasons, and CMU Seasons. \citep{Badino2012ICRA} also recorded a dataset in different weather conditions, seasons, and during the night as well as day. 

Only few datasets exist which particularly address the visual odometry problem. Most of these datasets are either small \citep{Smith2009IJRR, Pandey2011IJRR, Blanco-Claraco2014IJRR}, provide only low-quality images \citep{Guivant2006WEB}, or are not yet established \citep{Maddern2016IJRR, Huang2018CVPR}. A notable exception is the KITTI benchmark \citep{Geiger2012CVPR} discussed in \chpref{chap:Datasets}, which provides a large dataset of challenging sequences and evaluation metrics as well as an online evaluation server. We list the current leading monocular, stereo, and LiDAR methods on the KITTI benchmark in \tabref{tab:kitti_odometry_mono}, \tabref{tab:kitti_odometry_stereo}, and \tabref{tab:kitti_odometry_lidar}, respectively.

\section{Metrics}
For image-retrieval approaches, a popular metric is the percentage of recognized queries (Recall at N). A place is considered recognized if at least one of the top $N$ retrievals are within 25 meters from the query. For autonomous driving, this precision is not satisfactory since a higher accuracy is necessary to navigate through the environment. Consequently, localization approaches for loop closure detection \citep{Cummins2008IJRR,Paul2010ICRA,Lee2013IROS} strive for higher accuracy and typically consider the precision-recall metric. 

Structure-based localization approaches consider the position error (Euclidean distance between the estimated pose and the ground truth pose) as well as the orientation error. \citet{Sattler2018CVPR} report the percentage of localized query images that differ from the ground truth pose using high (0.25m, 2$\deg$), medium (0.5, 5$\deg$), and low (5m, 10$\deg$) accuracy thresholds.

The performance of methods for visual odometry is often measured using the Absolute Trajectory Error (ATE) or Relative Pose Error (RPE). The  APE estimates the absolute distance between the estimated and ground truth trajectory. The RPE considers a fixed time interval and measures the local accuracy of the translational and rotational component. The KITTI dataset reports the average translational and rotational error measured for all possible subsequences of length $(100,200,\ldots,800)$ meters.

\section{State of the Art on KITTI}
\label{sec:visual_odometry_sota}

\boldparagraph{Localization}
A unified and established benchmark for localization methods is still missing which makes the comparison of different approaches difficult. However, several newly introduced datasets \citep{Carlevaris-Bianco2016IJRR, Sattler2018CVPR}, reveal open challenges to the community. \citet{Sattler2018CVPR} compare two structure-based methods \citep{Sattler2016PAMI, Svarm2017PAMI} and three image retrieval approaches \citep{Torii2015CVPR, Arandjelovic2016CVPR, Cummins2008IJRR} on their dataset. While the structure-based methods significantly outperform the image retrieval approaches and show better robustness, all methods fail in more challenging conditions, particularly at night, when foliage changes as well as in suburban and park regions.

\begin{table*}[t]
\begin{center}
\begin{adjustbox}{width=1\textwidth}\begin{tabular}{l l | c | c | c}
	& {\bf Method} & {\bf Translation} & {\bf Rotation} & {\bf Runtime}\\ \hline
1. & DVSO \citep{Yang2018ECCVb} & 0.90 \% & 0.0021 [deg/m] & 0.1 s / GPU \\
2. & BVO \citep{Pereira2017WVC} & 1.76 \% & 0.0036 [deg/m] & 0.1 s / 1 core \\
3. & PMO / PbT-M2 \citep{Fanani2017IVC} & 2.05 \% & 0.0051 [deg/m] & 1 s / 1 core \\
4. & FTMVO \citep{Mirabdollah2015GCPR} & 2.24 \% & 0.0049 [deg/m] & 0.11 s / 1 core \\
5. & PbT-M1 \citep{Fanani2016IV,Fanani2017IV} & 2.38 \% & 0.0053 [deg/m] & 1 s / 1 core \\
6. & MLM-SFM \citep{Song2014CVPR} & 2.54 \% & 0.0057 [deg/m] & 0.03 s / 5 cores \\
7. & RMCPE+GP \citep{Mirabdollah2014GCPR} & 2.55 \% & 0.0086 [deg/m] & 0.39 s / 1 core \\
8. & EB3DTE+RJMCM \citep{Boukhers2018MTA} & 5.45 \% & 0.0274 [deg/m] & 1 s / 1 core \\
9. & VISO2-M + GP \citep{Song2014CVPR} & 7.46 \% & 0.0245 [deg/m] & 0.15 s / 1 core \\
10. & VISO2-M \citep{Geiger2011IV} & 11.94 \% & 0.0234 [deg/m] & 0.1 s / 1 core \\
11. & OABA \citep{Frost2016ICRA} & 20.95 \% & 0.0135 [deg/m] & 0.5 s / 1 core
\end{tabular}
\end{adjustbox}
\end{center}
\vspace{-0.4cm}
\caption{{\bf KITTI Monocular Odometry Leaderboard.} The numbers show relative translational errors and relative rotational errors, averaged over all subsequences of length 100 meters to 800 meters. Accessed on: April 2019.}
\label{tab:kitti_odometry_mono}
\end{table*}

\boldparagraph{Monocular Visual Odometry}
Monocular visual odometry methods are able to recover motion only up to a scale factor. The absolute scale can be determined by computing the size of objects in the scene, from motion constraints, or by integrating other sensors. 

\citet{Fanani2016IV} follow a direct approach and propagate 3D key points into the next frame using relative pose predictions. Combined with the scale estimation method proposed in \citep{Fanani2017IV} which uses dense and sparse ground plane estimates for scale correction, they achieve competitive results in \tabref{tab:kitti_odometry_mono}. However, their approach is not applicable in real-time. In contrast, \citet{Mirabdollah2015GCPR} follow a robust feature-based monocular visual odometry approach capable of real-time estimation using the iterative five-point method. They obtain the location of landmarks using a probabilistic triangulation method and estimate the scale of the motion from sparse low-quality features on the ground plane. 
\citet{Fanani2017IVC} improve the scale correction of \citep{Fanani2017IV} by utilizing street pixels detected with a convolutional neural network for ground plane pose estimation. Furthermore, they extend the keypoint propagation method presented in \citep{Fanani2016IV} which allows them to improve on previous work.

In contrast to other approaches, \citet{Pereira2017WVC} consider backward motion with a backward-facing camera or by processing the images of a forward facing-camera in reverse order. They argue that initial depth estimation of sparse feature matching approaches is not very accurate since usually, new features are initialized the first time they have been observed in the far distance. By considering the reverse order for a forward-facing camera, new features will be detected in the nearest frame, which allows more accurate depth estimates in case of forward motion. 

Recently, \citet{Yang2018ECCVb} propose to use deep monocular depth predictions for monocular visual odometry by incorporating depth predictions into a windowed direct bundle adjustment. With this direct approach, they outperform all monocular visual odometry methods in \tabref{tab:kitti_odometry_mono}. However, we remark that the KITTI dataset requires metric output, thus scale drift and scale estimation have a strong impact on the performance of the approaches.

\begin{table*}[t]
\begin{center}
\begin{adjustbox}{width=1\textwidth}\begin{tabular}{l l | c | c | c}
& {\bf Method} & {\bf Translation} & {\bf Rotation} & {\bf Runtime}\\ \hline
1. & SOFT2 \citep{Cvivsic2017JFR} & 0.65 \% & 0.0014 [deg/m] & 0.1 s / 2 cores \\
2. & LG-SLAM \citep{Lenac2018IJRR} & 0.82 \% & 0.0020 [deg/m] & 0.2 s / 4 cores \\
3. & RotRocc+ \citep{Buczko2016ITSC, Buczko2018IV} & 0.83 \% & 0.0026 [deg/m] & 0.25 s / 2 cores \\
4. & GDVO \citep{Zhu2017IJCAI} & 0.86 \% & 0.0031 [deg/m] & 0.09 s / 1 core \\
5. & SOFT \citep{Cvisic2015ECMR} & 0.88 \% & 0.0022 [deg/m] & 0.1 s / 2 cores \\
6. & RotRocc \citep{Buczko2016ITSC} & 0.88 \% & 0.0025 [deg/m] & 0.3 s / 2 cores \\
7. & Stereo DSO \citep{Wang2017ICCVb} & 0.93 \% & 0.0020 [deg/m] & 0.1 s / 1 core \\
8. & ROCC \citep{Buczko2016IV} & 0.98 \% & 0.0028 [deg/m] & 0.3 s / 2 cores \\
9. & cv4xv1-sc \citep{Persson2015IV} & 1.09 \% & 0.0029 [deg/m] & 0.145 s / GPU \\
10. & MonoROCC \citep{Buczko2017IV} & 1.11 \% & 0.0028 [deg/m] & 1 s / 2 cores \\
\hline
31. & VISO2-S \citep{Geiger2011IV} & 2.44 \% & 0.0114 [deg/m] & 0.05 s / 1 core \\
\end{tabular}
\end{adjustbox}
\end{center}
\vspace{-0.4cm}
\caption{{\bf KITTI Odometry Stereo Leaderboard.} The numbers show relative translational errors and relative rotational errors, averaged over all subsequences of length 100 meters to 800 meters. Methods below the horizontal line show older entries for reference. Accessed on: April 2019.}
\label{tab:kitti_odometry_stereo}
\end{table*}

\boldparagraph{Stereo Visual Odometry}
Stereo visual odometry methods exploit the known baseline between the cameras of the stereo camera rig for estimating scale. Therefore, stereo methods are typically able to outperform monocular methods on the KITTI dataset (see \tabref{tab:kitti_odometry_mono} and \tabref{tab:kitti_odometry_stereo}).

\citet{Cvisic2015ECMR} decouple estimation of rotation and translation as translation is dependent on the scene depth while rotation is not. They estimate rotation using the five-point algorithm \cite{Nister2004PAMI} and translation using the three-point method. \citet{Buczko2016ITSC} exploit the same idea and propose to use an initial rotation estimation to decouple rotational and translational optical flow.
In contrast, \citet{Wang2017ICCVb} tackle the visual odometry problem with a direct method by combining static stereo with multi-view stereo as in \citep{Engel2014ECCV,Engel2015IROS}. In contrast to \citep{Engel2014ECCV,Engel2015IROS}, they extend the energy function instead of relying on filtering approaches to update the geometry and provide an efficient bundle adjustment procedure for real-time optimization. 
One weakness of direct methods is that they often get stuck in local optima, especially in case of large motions.
\citet{Zhu2017IJCAI} addresses this problem with a dual Jacobian scheme for multi-scale pyramid optimization. This allows them to avoid local optima and obtain more accurate camera pose estimations that are closer to the optimal solution. In addition, they introduce a gradient-based feature representation, which improves robustness against illumination changes.

\citet{Lenac2018IJRR} propose a filtering-based SLAM approach that leverages a novel filtering solution on Lie groups. Combined with the visual odometry method proposed in \citep{Cvisic2015ECMR}, they are ranked second in stereo visual odometry.
\citet{Cvivsic2017JFR} improve the feature selection approach suggested in \citep{Cvisic2015ECMR} with an age-based weighting factor suggested in \citep{Geiger2011IV} that gives higher weight to features that are horizontally closer to the image center. This allows them to better handle calibration errors and outperform all stereo-based methods (\tabref{tab:kitti_odometry_stereo}) while obtaining results competitive with LiDAR-based techniques.

\citet{Kreso2015VISAPP} observed that camera calibration is critical for visual odometry and that the remaining calibration errors in pre-calibrated systems like KITTI have adversarial effects on the estimation results. They, therefore, propose to explicitly correct the calibration of the camera by exploiting ground truth motion which they use to recover a deformation field by optimizing the reprojection error of point feature correspondences in neighboring stereo frames.

\begin{table*}[t]
\begin{center}
\begin{adjustbox}{width=1\textwidth}\begin{tabular}{l l | c | c | c}
& {\bf Method} & {\bf Translation} & {\bf Rotation} & {\bf Runtime}\\ \hline
1. &  V-LOAM \citep{Zhang2015ICRA} & 0.56 \% & 0.0013 [deg/m] & 0.1 s / 2 cores \\
2. &  LOAM \citep{Zhang2014RSS} & 0.59 \% & 0.0014 [deg/m] & 0.1 s / 2 cores \\
3. &  IMLS-SLAM \citep{Deschaud2018ICRA} & 0.69 \% & 0.0018 [deg/m] & 1.25 s / 1 core \\
4. &  MC2SLAM \citep{Neuhaus2018GCPR} & 0.69 \% & 0.0016 [deg/m] & 0.1 s / 4 cores \\
5. &  LIMO2\_GP \citep{Grater2018IROS} & 0.84 \% & 0.0022 [deg/m] & 0.2 s / 2 cores \\
6. &  LIMO2 \citep{Grater2018IROS} & 0.86 \% & 0.0022 [deg/m] & 0.2 s / 2 cores \\
7. &  CPFG-slam \citep{Ji2018IV} & 0.87 \% & 0.0025 [deg/m] & 0.03 s / 4 cores \\
8. &  LIMO \citep{Grater2018IROS} & 0.93 \% & 0.0026 [deg/m] & 0.2 s / 2 cores \\
9. &  DEMO \citep{Zhang2014IROS} & 1.14 \% & 0.0049 [deg/m] & 0.1 s / 2 cores \\
10. &  STEAM-L WNOJ \citep{Tang2019RAL} & 1.22 \% & 0.0058 [deg/m] & 0.2 s / 1 core \\
\end{tabular}\end{adjustbox}
\end{center}
\vspace{-0.4cm}
\caption{{\bf KITTI Odometry LiDAR Leaderboard.} The numbers show relative translational errors and relative rotational errors, averaged over all subsequences of length 100 meters to 800 meters. Accessed on: April 2019.}
\label{tab:kitti_odometry_lidar}
\end{table*}

\boldparagraph{LiDAR-based Odometry}
Motivated by the impact of small calibration errors on the depth estimation of stereo-based methods \citep{Kreso2015VISAPP}, \citet{Grater2018IROS} leverages depth information obtained from LiDAR for monocular visual odometry. Rejecting outliers based on a local plane assumption and fusing depth similar to \citep{Cvisic2015ECMR,Buczko2016IV}, they obtain competitive results (\tabref{tab:kitti_odometry_lidar}).

In contrast, \citet{Neuhaus2018GCPR} directly address the SLAM problem by integrating LiDAR data with inertial measurements. The integration of IMU data allows them to cope with high-frequency motion, \eg in off-road environments.

Inspired by RGB-D methods \citep{Newcombe2011ISMAR}, \citet{Deschaud2018ICRA} uses an implicit surface representation \citep{Curless1996SIGGRAPH} of the map for aligning new scans in a LiDAR SLAM approach. In combination with a specific sampling strategy based on LiDAR scans, they achieve results similar to \citep{Neuhaus2018GCPR}.

\begin{figure}[t!]
	\centering
	\includegraphics[width=1.00\columnwidth]{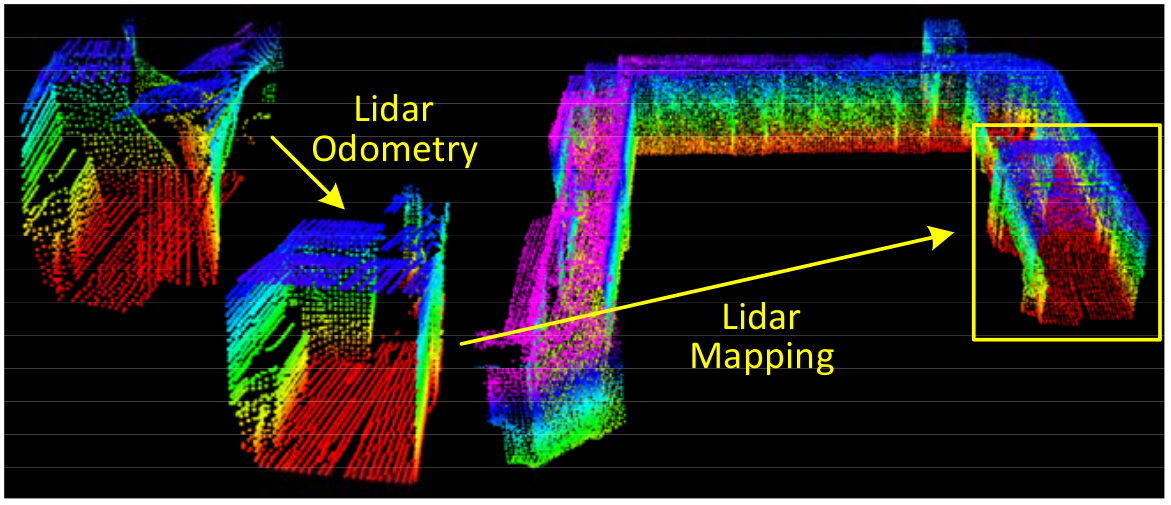}
	\caption[LOAM LiDAR-based SLAM]{LOAM by \protect\cite{Zhang2014RSS} matches two consecutive LiDAR scans (LiDAR Odometry) and registers the new scan to a map (LiDAR Mapping). \figsourceC{\protect\citet{Zhang2014RSS}}{2014}{RSS}.
	}
	\label{fig:visual_odometry_loam}
\end{figure}

The best performing methods on KITTI use 3D point clouds from LiDAR for ego-motion estimation (\tabref{tab:kitti_odometry_lidar}). 
\citet{Zhang2014RSS} split the SLAM problem into LiDAR-based odometry at high frequency with low accuracy and LiDAR-mapping at low frequency with high accuracy, as illustrated in \figref{fig:visual_odometry_loam}. Their LiDAR-based odometry approach matches two consecutive LiDAR scans, whereas their LiDAR-based mapping approach matches and registers the new scan to a map, resulting in low drift and low computational complexity at the same time. \citet{Zhang2015ICRA} extend this work by combining visual odometry at high frequency with LiDAR-mapping at low frequency, which allows them to further improve upon their results (\tabref{tab:kitti_odometry_lidar}).

\section{Discussion}
While localization approaches are still missing an established unified benchmark for fair comparison and evaluation of methods, a new benchmark \footnote{\url{https://www.visuallocalization.net}} based on multiple diverse datasets has recently been proposed by \citet{Sattler2018CVPR}. Based on these results, it can be concluded that current techniques still fail to perform well in challenging real-world conditions, as identified in \citep{Carlevaris-Bianco2016IJRR, Sattler2018CVPR}. One possible direction towards higher recall and more robustness is to incorporate deep CNN features encoding high-level information. For instance, \citet{Schoenberger2018CVPR, Radwan2018RAL} demonstrate that localization accuracy in challenging conditions can benefit from a semantic understanding of the environment. 

\begin{figure*}[t!]
\includegraphics[width=0.096\linewidth]{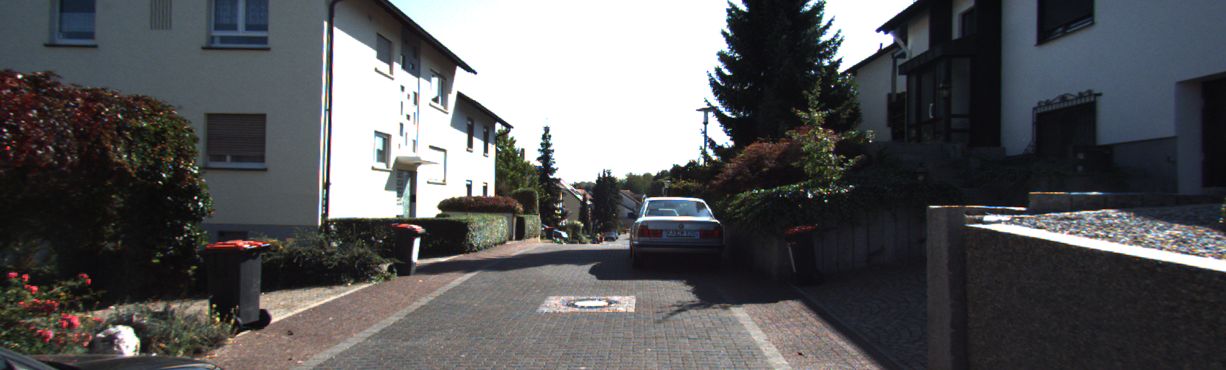}~%
\includegraphics[width=0.096\linewidth]{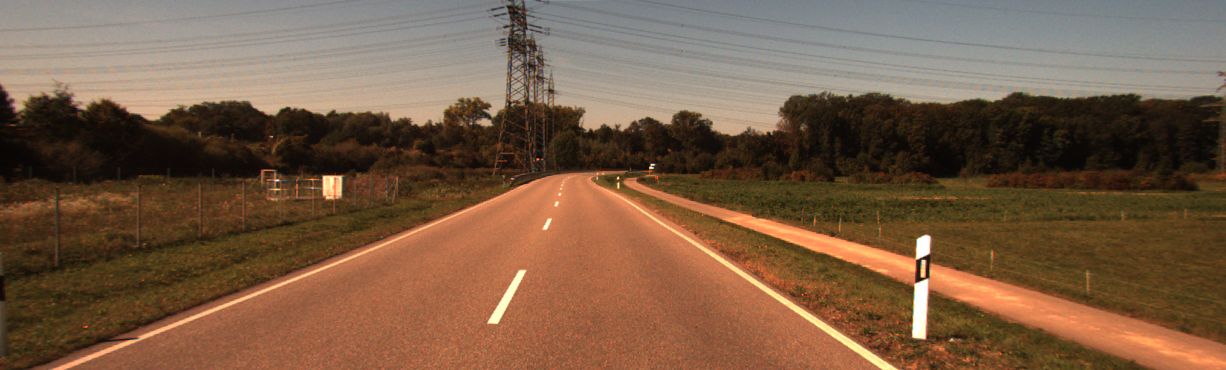}~%
\includegraphics[width=0.096\linewidth]{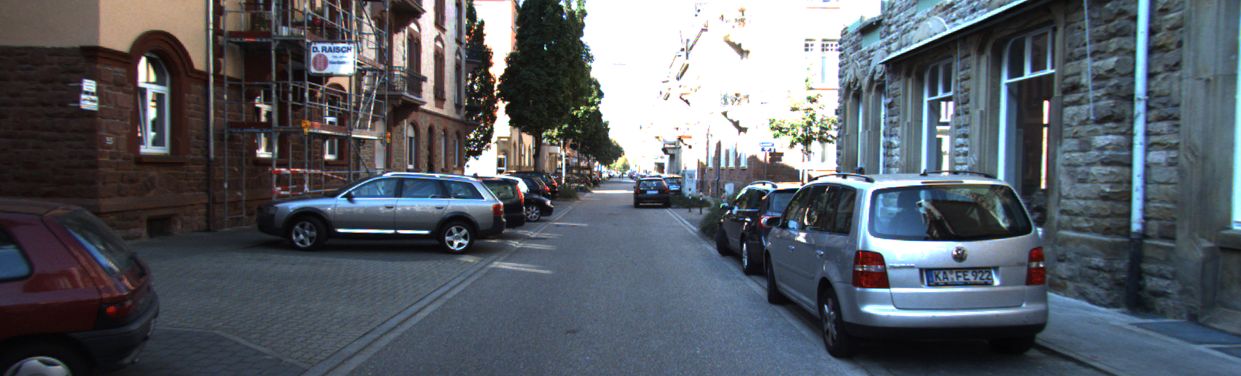}~%
\includegraphics[width=0.096\linewidth]{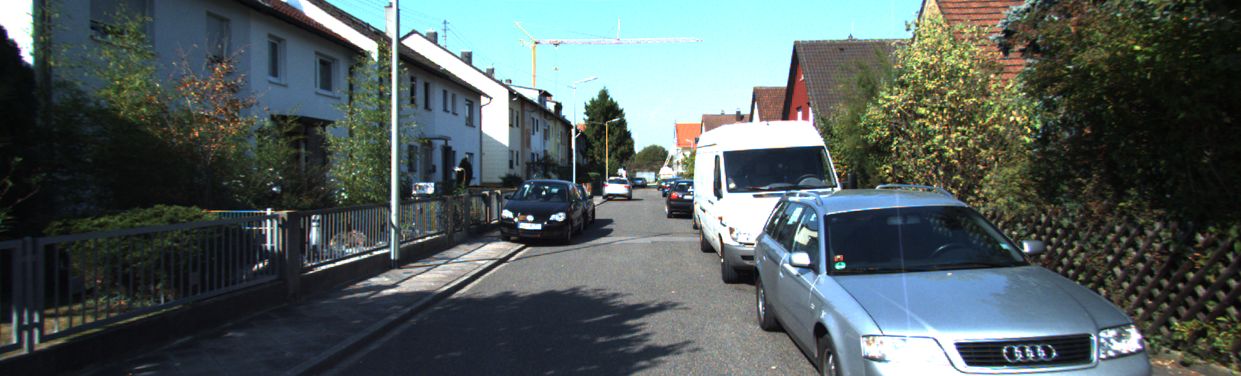}~%
\includegraphics[width=0.096\linewidth]{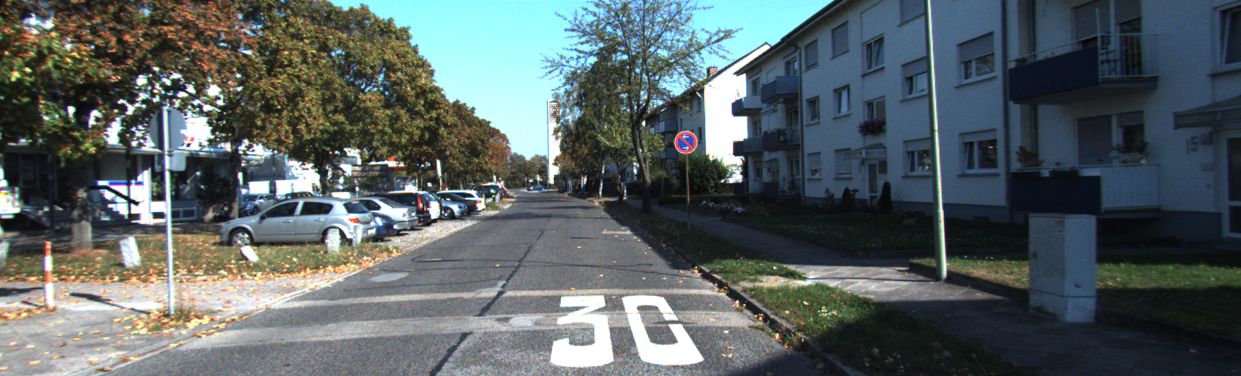}~%
\includegraphics[width=0.096\linewidth]{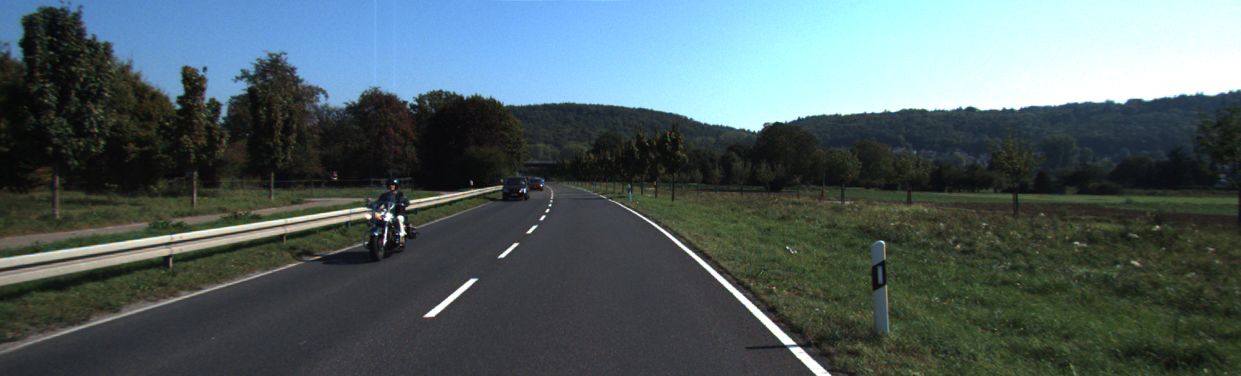}~%
\includegraphics[width=0.096\linewidth]{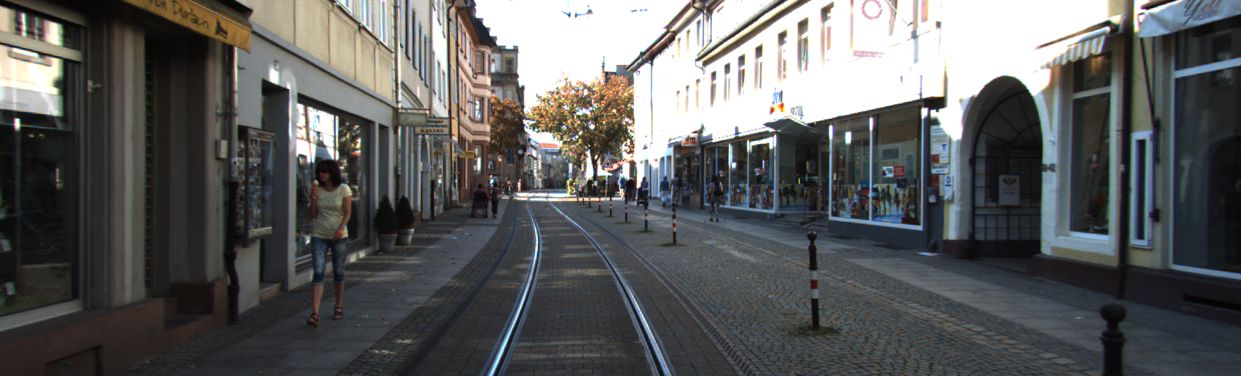}~%
\includegraphics[width=0.096\linewidth]{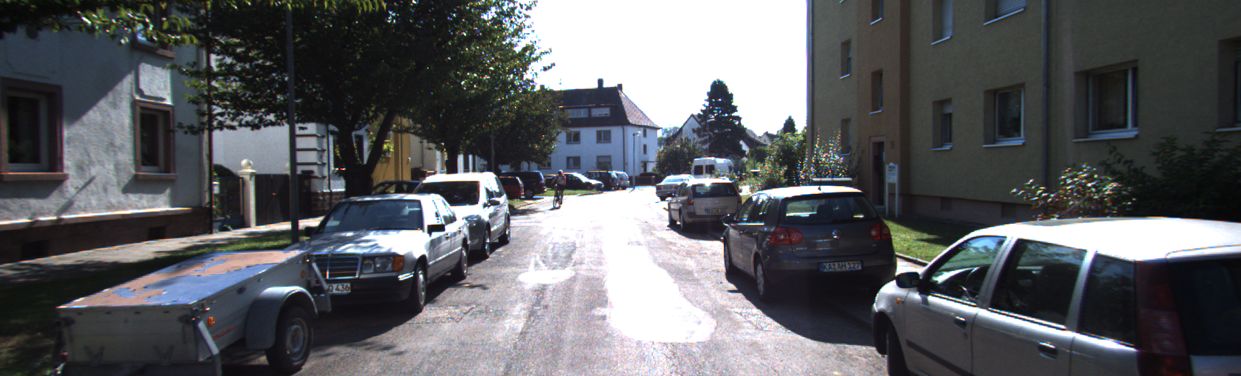}~%
\includegraphics[width=0.096\linewidth]{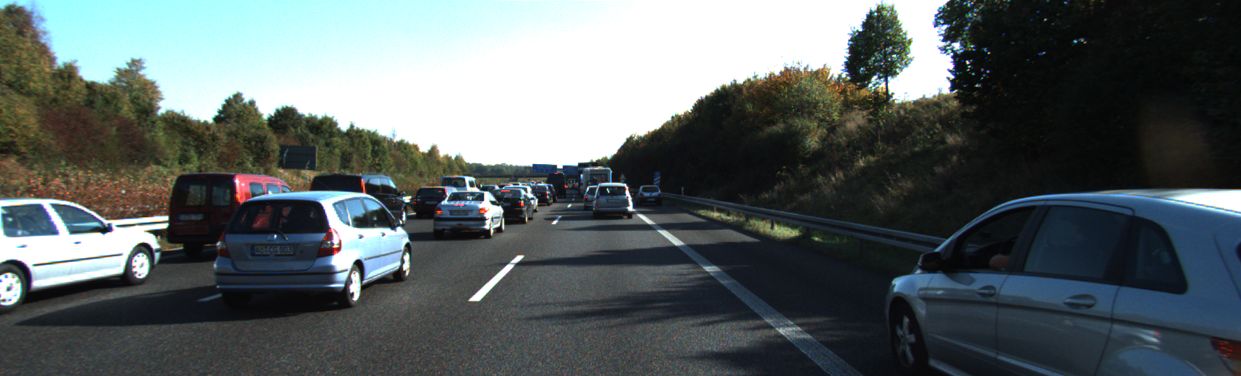}~%
\includegraphics[width=0.096\linewidth]{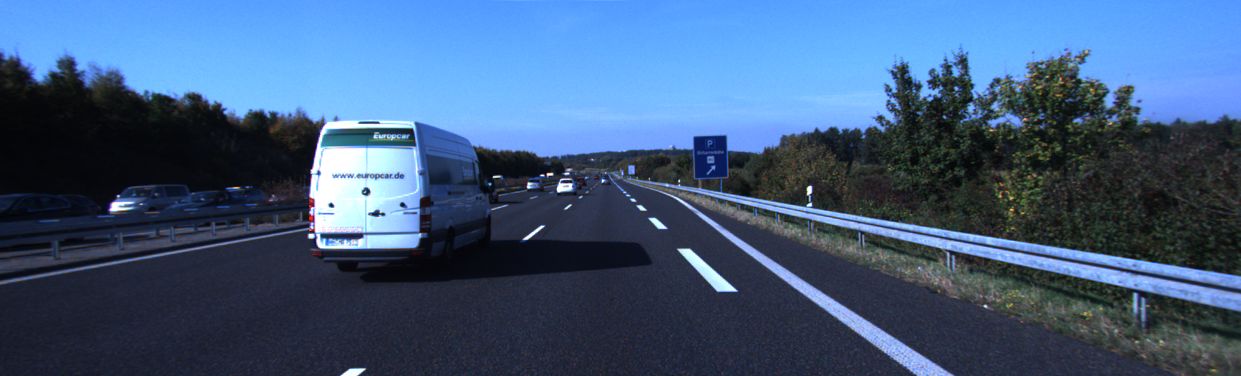}\\%
\includegraphics[width=\linewidth]{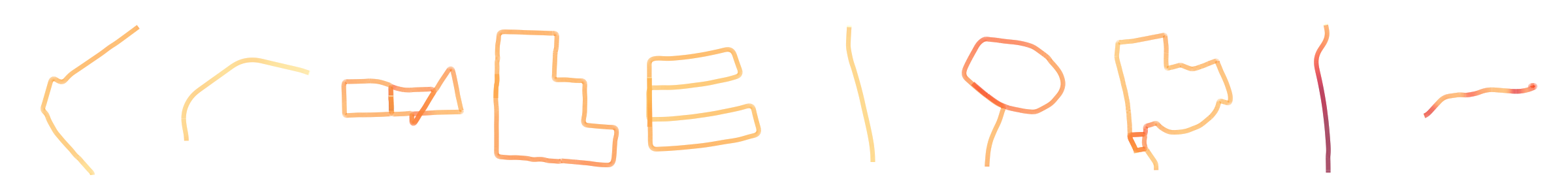}\\%
\includegraphics[width=\linewidth]{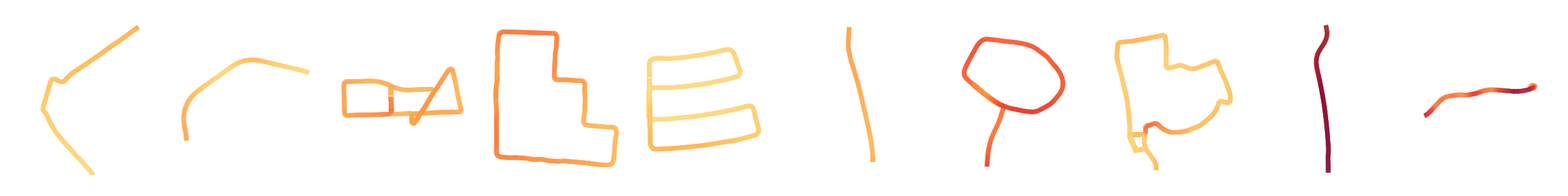}\\%
\includegraphics[width=\linewidth]{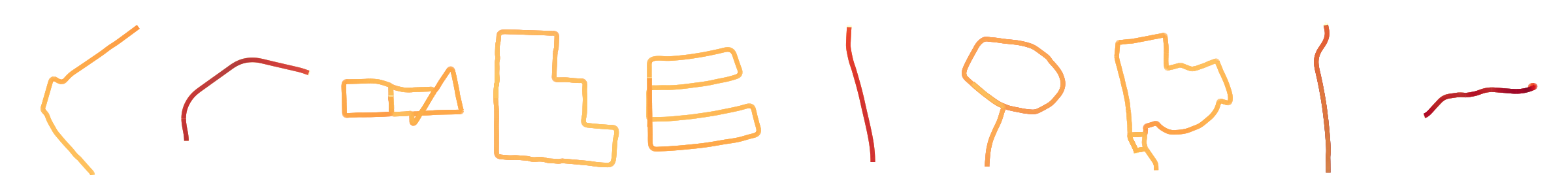}\\%
\caption{{\bf KITTI Odometry.} From top-to-bottom: example image from the sequence, average translational error, average rotational error and speed. Averages are computed over 400 meter long trajectories and for the 15 best performing methods published on the KITTI website. Darker colors (\ie red) indicate larger errors or higher speed.}
\label{fig:visual_odometry_qualitative_results}
\end{figure*}

In ego-motion estimation, monocular visual odometry methods can not yet compete with approaches using 3D information on the KITTI dataset. While LiDAR provides the richest source of information, stereo-based methods also achieve competitive results. 
In \figref{fig:visual_odometry_qualitative_results}, we visualize the average translational and rotational errors of the best performing visual odometry methods on the KITTI benchmark. The second row shows the translational error, and the third row shows the rotational error while the last row shows the speed. The highest translational and rotational errors are usually observed in case of strong turns. Furthermore, the error is correlated with speed and the amount of independently moving objects in the scene, which causes a decrease in the number of matched features in the background. While large errors can be observed for crowded highway scenes (second from right), only moderate errors occur when the highway is empty (right and second from left). Larger errors can also be observed in very narrow environments (fourth from right) where feature displacements are large.
Overall, the most accurate motion estimation is achieved using 3D information. However, it is remarkable that state-of-the-art stereo-based methods achieve competitive results using cheap passive stereo sensors in comparison to more expensive LiDAR scanners. 

	\chapter{Scene Understanding}
\label{chap:scene_understanding}
\section{Problem Definition}
One of the basic requirements of autonomous driving is to fully understand the surrounding area, such as a complex traffic scene. The complex task of outdoor scene understanding involves several sub-tasks such as depth estimation, scene categorization, object detection and tracking, event categorization, and more. Each of these tasks describes a particular aspect of a scene. It can be beneficial to model some of these aspects jointly in order to exploit the complementary nature of the different cues in the scene and to obtain a more holistic understanding. The goal of most scene understanding models is to obtain a rich but compact representation of the scene including its elements, \eg layout, traffic participants, and their relation with each other.

In contrast to modeling these problems in 2D, 3D reasoning allows geometric scene understanding and results in a more informative representation of the scene in the form of 3D object models, layout elements, and occlusion relationships. In this section, we will focus on a subset of 3D scene understanding techniques that are particularly relevant to the autonomous driving task, excluding works on scene graph estimation or image tagging.
One specific challenge in this context is the interpretation of urban and sub-urban traffic scenarios. Compared to highways and rural roads, urban scenarios comprise dynamic objects, a large degree of variability in the geometric layout of roads and crossroads, and an increased level of difficulty due to ambiguous visual features, occlusions, and challenging illumination conditions.

\begin{figure}[t]
	\centering
	\includegraphics[width=0.80\columnwidth]{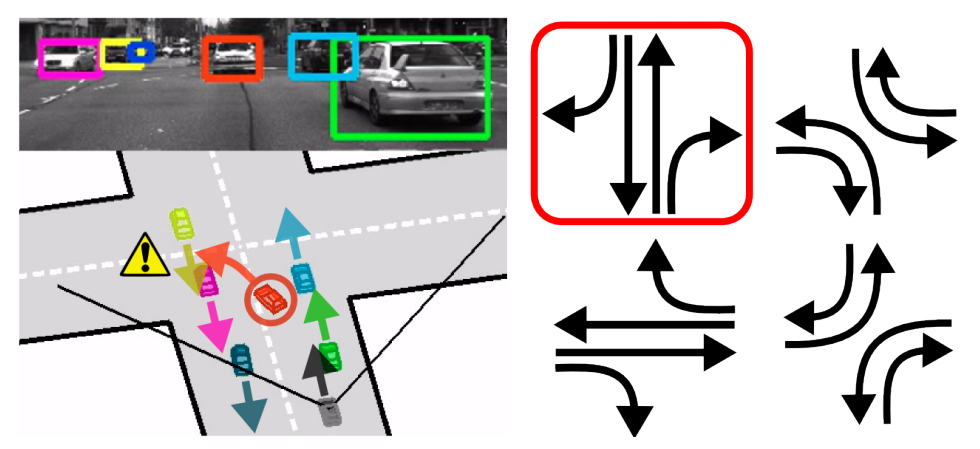}
	\caption[Scene Understanding using Traffic Patterns]{\textbf{Scene Understanding using Traffic Patterns.} \protect\citet{Zhang2013ICCV} propose to explicitly account for traffic patterns to improve scene layout and activity estimation results (right, correct situation marked in red). \figsourceC{\protect\citet{Zhang2013ICCV}}{2013}{IEEE}.
	}
	\label{fig:scene_understanding_traffic_patterns}
\end{figure}

\section{Methods}
While early work in computer vision \citep{Roberts1963, Hanson1978, Brooks1981IJCAI, Ohta1985} already tackled the scene understanding problem from various perspectives, \eg using a block world assumption \citep{Roberts1963} or via bottom-up top-down inference \citep{Ohta1985}, most approaches relied on heuristics rather than learning and were not able to generalize to complex real-world scenes. In contrast, modern approaches try to learn complex relationships directly from data. In their pioneering work, \citet{Hoiem2007IJCV} infer the overall 3D structure of an outdoor scene from a single image. The surface layout is represented as a set of coarse geometric classes with certain orientations such as support, vertical, and sky. These elements are inferred by learning an appearance-based model for each class. \citet{Oliveira2016RAS} propose a time-varying 3D representation using a set of planar polygons as primitives. Given 3D LiDAR point clouds, they find the support plane using RANSAC followed by a clustering of inliers to separate instances.

\subsection{Road Topology and Traffic Participants}
For autonomous driving, understanding the road topology and other traffic participants in the scene is of utmost importance. \citet{Ess2009BMVC} use semantic segmentation as an intermediate representation to extract the road topology and to detect crosswalks and other traffic participants. In addition, their intermediate representation simultaneously encodes the spatial layout of the scene. \citet{Wojek2008ECCV} detect vehicles and track them with a temporal filter based on a linear motion model. They also estimate the camera motion and propagate it to the next frame using a dynamic Conditional Random Field model for joint labeling of object and scene classes. However, \citep{Ess2009BMVC, Wojek2008ECCV} only infer a topological model of the scene and not a geometric model.

\citet{Wojek2010ECCV} extend \citep{Wojek2008ECCV} to a probabilistic 3D scene model that encompasses multi-class object detection, object tracking, scene labeling, and reasoning about geometric relations. \citet{Geiger2014PAMI} jointly reason about the 3D scene layout of intersections as well as the location and orientation of vehicles in the scene. They present a probabilistic generative model capturing the scene topology, geometry, and traffic activities by leveraging vehicle tracks, semantic labels, scene flow and occupancy grids.

Apart from 3D primitive-based representations, there exist other ways of representing a street scene.
A more fine-grained model of the road is proposed by \citet{Topfer2015TITS}. The complex road scene is hierarchically decomposed into roads, lanes, and finally road-edges and lane-markings. This allows them to infer a more expressive model of the road compared to \citep{Geiger2014PAMI}. 
\citet{Seff2016ARXIV} define a list of road layout attributes such as the number of lanes, drivable directions, distance to intersections, etc. They first automatically collect a large-scale dataset for these attributes by leveraging existing street view image databases and online navigation maps (\eg OpenStreetMap). Based on this dataset, they train a deep convolutional network to predict each attribute from a single street view image. 

\begin{figure}[t]
	\centering
	\includegraphics[width=1.00\columnwidth]{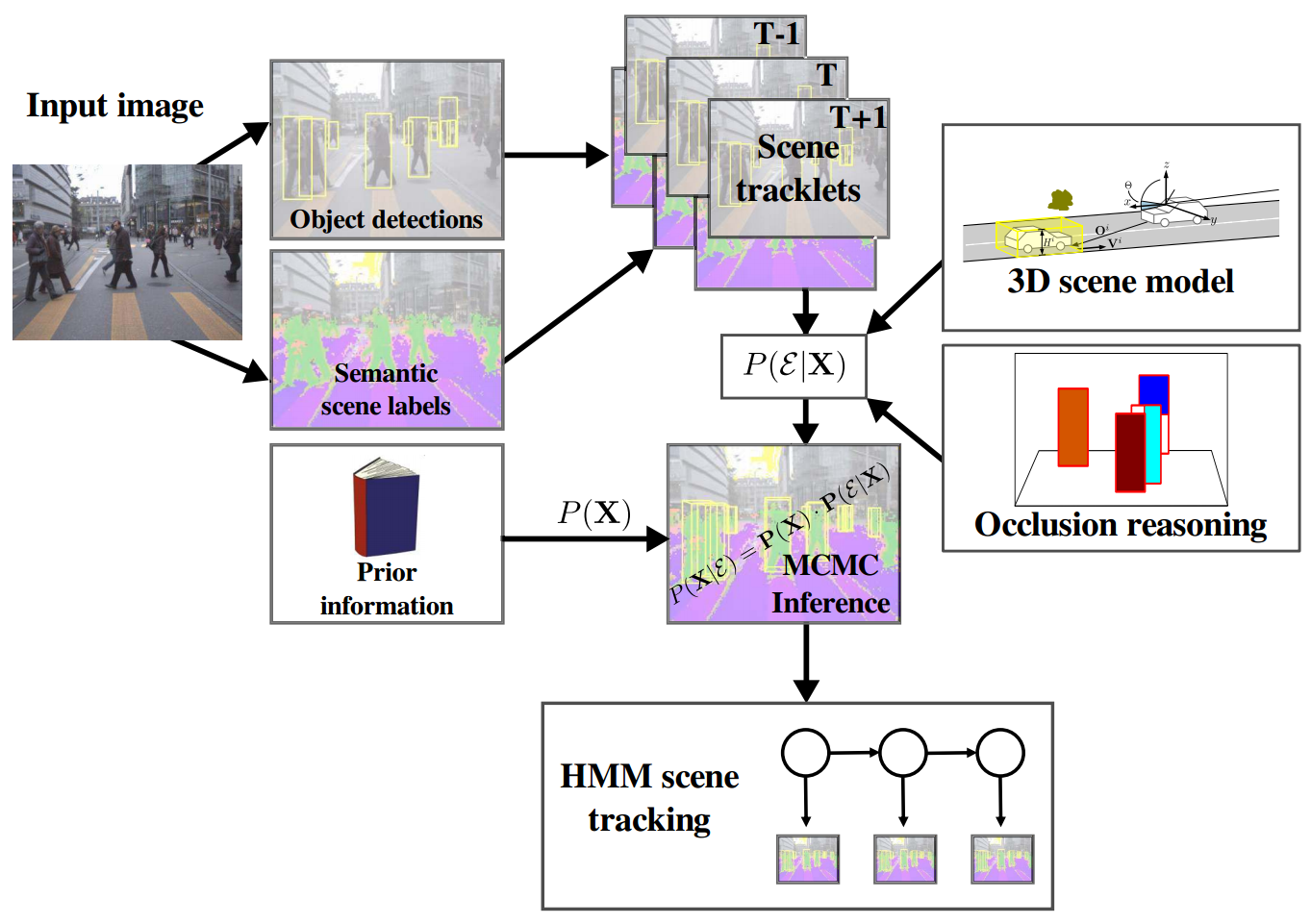}
	\caption[Physical Relationships for Scene Understanding.]{\textbf{Physical Relationships for Scene Understanding.} Overview of combined object detection and tracking system with explicit occlusion reasoning by \protect\citet{Wojek2013PAMI}. \figsourceC{\protect\citet{Wojek2013PAMI}}{2013}{IEEE}.}
	\label{fig:scene_understanding_det_track}
\end{figure}

\subsection{Physical and Temporal Relationships}

While the detection of traffic participants is addressed in our review on object detection (\chpref{chap:detection}) and object tracking (\chpref{chap:tracking}) approaches, scene understanding systems aim at integrating object detection and tracking with physical constraints and model the temporal behavior and relationship between traffic participants and the scene. \citet{Pellegrini2009ICCV} model interactions between pedestrians (social behavior) and the scene (collisions) in a multi-target tracking formulation. \citet{Kuettel2010CVPR} model spatio-temporal dependencies of moving agents in complex dynamic scenes by learning co-occurring activities and temporal rules between them. However, both approaches assume a static observer and a long observation period, \ie the scene must be observed for a significant period of time before making a decision, therefore it is not applicable to autonomous systems. 

In contrast, \citep{Wojek2011CVPR, Wojek2013PAMI, Zhang2013ICCV} consider a moving vehicle as observer and construct expressive 3D scene models by reasoning about occlusions and traffic patterns.  
\citet{Wojek2011CVPR, Wojek2013PAMI} integrate multiple object part detectors \citep{Wojek2010ECCV} into the 3D scene model for explicit object-object occlusion reasoning (\figref{fig:scene_understanding_det_track}). In addition, they enforce physically plausible trajectories by pruning geometrically infeasible detections. \citet{Zhang2013ICCV} propose a more expressive generative model of 3D urban scenes similar to \citep{Geiger2014PAMI}. While the independent tracklets in \citep{Geiger2014PAMI} can lead to implausible inference results, they reason about high-level semantics in the form of traffic patterns to avoid this problem (\figref{fig:scene_understanding_traffic_patterns}) and force the solution to conform to traffic rules. This allows them to significantly improve scene estimation and vehicle-to-lane association results.
\citet{Wang2019CVPR} propose a top-view representation for complex road scenes that can be inferred from a single camera using a deep neural network.

\section{Discussion}
While early work on scene understanding struggled to infer expressive models of the real world, learning-based approaches led to models with increasing expressivity, ranging from simple 2D models to represent road topologies and objects  \citep{Ess2009BMVC, Wojek2008ECCV}, to more complex 3D models \citep{Oliveira2016RAS, Geiger2014PAMI} which also incorporate physical \citep{Wojek2013PAMI,Wang2019CVPR} and temporal \citep{Wojek2011CVPR, Wojek2013PAMI, Zhang2013ICCV} constraints. As motivated in \citep{Seff2016ARXIV}, more expressive models can reduce the dependency on high definition maps. However, the level of expressiveness needed in autonomous driving remains an open question and the accuracy achieved by state-of-the-art scene understanding models is still limited. In addition, a unified evaluation of scene understanding approaches is difficult due to the varying complexity of models and the different challenges they tackle.

 	\chapter{End-to-End Learning for Autonomous Driving}
\label{sec:end_to_end_learning}
\section{Problem Definition}
Current state-of-the-art autonomous driving systems in industry are composed of numerous modules, \eg detection (of traffic signs, lights, cars, pedestrians), segmentation (of lanes, facades), motion estimation, tracking of traffic participants, reconstruction etc. The results from these components are then typically combined in a planning module that feeds the control.
However, this requires robust solutions to many open challenges in scene understanding in order to solve the problem of manipulating the car direction and speed. Furthermore, auxiliary loss functions are required to train each module (\eg object detection, semantic segmentation) independently, hence ignoring the actual goals of the driving task which include travel time, safety, and comfort.

As an alternative, several methods consider autonomous driving as an end-to-end learning problem. In these approaches, the tasks of perception, planning, and control are combined, and a single model is trained end-to-end using a deep neural network. Most end-to-end autonomous driving systems map from sensory inputs, such as front-facing camera images, directly to driving actions such as steering angle. 

\section{Methods}

End-to-end driving methods are typically trained from expert demonstrations to learn a driving policy that imitates the behavior of an expert or using reinforcement learning to explore the environment by trial and error (often in simulation). In the following sections, we first introduce the most relevant approaches proposed in the literature. We then discuss methods that combine ideas from behavior cloning and reinforcement learning. Finally, we discuss approaches that propose intermediate representations and demonstrate how driving models can be transferred from simulation to the real-world.

\subsection{Behavior Cloning}
\begin{figure}[t]
	\centering
	\includegraphics[width=1.00\columnwidth]{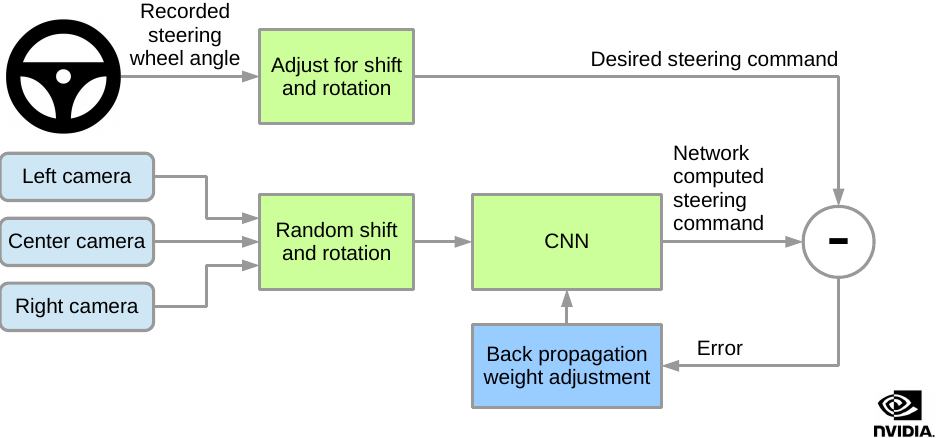}
	\caption[End-to-end Learning for Lane Following]{\textbf{End-to-end Learning for Lane Following.} A block diagram of an end-to-end model for lane following proposed by \protect\citet{Bojarski2016ARXIV}. Conditioned on the image, a CNN estimates a steering command which is compared to the expert
	command for tuning the CNN weights in order to bring the CNN output closer to
	the desired output. \figsource{\protect\citet{Bojarski2016ARXIV}}}
	\label{fig:Bojarski2016ARXIV}
\end{figure}
Behavior cloning approaches learn to map sensor observations, such as RGB images, to desired driving behavior by learning to clone the behavior of an expert. Thus, these approaches fall into the category of supervised learning techniques. Most commonly, a deep neural network is employed to represent the mapping from observations to expert actions. In the 1980s, \citet{Pomerleau1988NIPS} propose ALVINN, the first demonstration of imitation learning for self-driving vehicles using a small fully connected neural network.
30 years later, \citet{Bojarski2016ARXIV} propose a deeper end-to-end deep convolutional neural network for lane following, illustrated in \figref{fig:Bojarski2016ARXIV}, that maps images from the front-facing camera of a car to steering angles, given expert data. \citet{Xu2017CVPR} propose an alternative approach and exploit large scale online datasets from uncalibrated sources to learn a driving model. Specifically, they formulate autonomous driving as a future ego-motion prediction problem. They claim that predicting ego-motion instead of vehicle control allows their approach to generalize better to new platforms. Their deep learning architecture combines FCNs and LSTMs, and learns to predict the motion path given the current state of the agent.

Another problem with behavior cloning approach is that the training data is collected using an off-policy expert teacher, \ie the training data is collected by rolling out the expert policy, which is different from the policy being learned. As collecting expert demonstrations for all possible situations is not practical, the training trajectories do not cover all possible states. At test time, the rollout of the behavior cloning policy thus causes it to move to a different distribution of states compared to the one it was trained on. 
Due to this covariate shift between the training and test time trajectories, the behavior cloning agent's errors compound when drifting away from the expert demonstrations.
In other words, the vehicle is likely to encounter new situations it has not been trained for and therefore acts wrongly.

In contrast, in on-policy rollout, training data is collected using the current policy being learned. \citet{Ross2010AISTATS} propose DAgger to alleviate covariate shift by iteratively collecting corrective expert actions for the states visited by rolling out the currently learned driving policy. The driving policy parameters are then trained using the data collected on-policy.
However, doing on-policy rollouts with an imperfect policy has the disadvantage of drifting and potentially reaching dangerous states, thus requiring a simulator for safe training. ~\citet{Laskey2017CORL} claim to provide a safer way of generating training data using expert policy with small amounts of noise injected to approximate the errors of on-policy rollout. They achieve this by iterating between learning a noise model that minimizes the covariate shift and generating data for training the behavior cloning agent.

\begin{figure}[t]
	\centering
	\includegraphics[width=1.00\columnwidth]{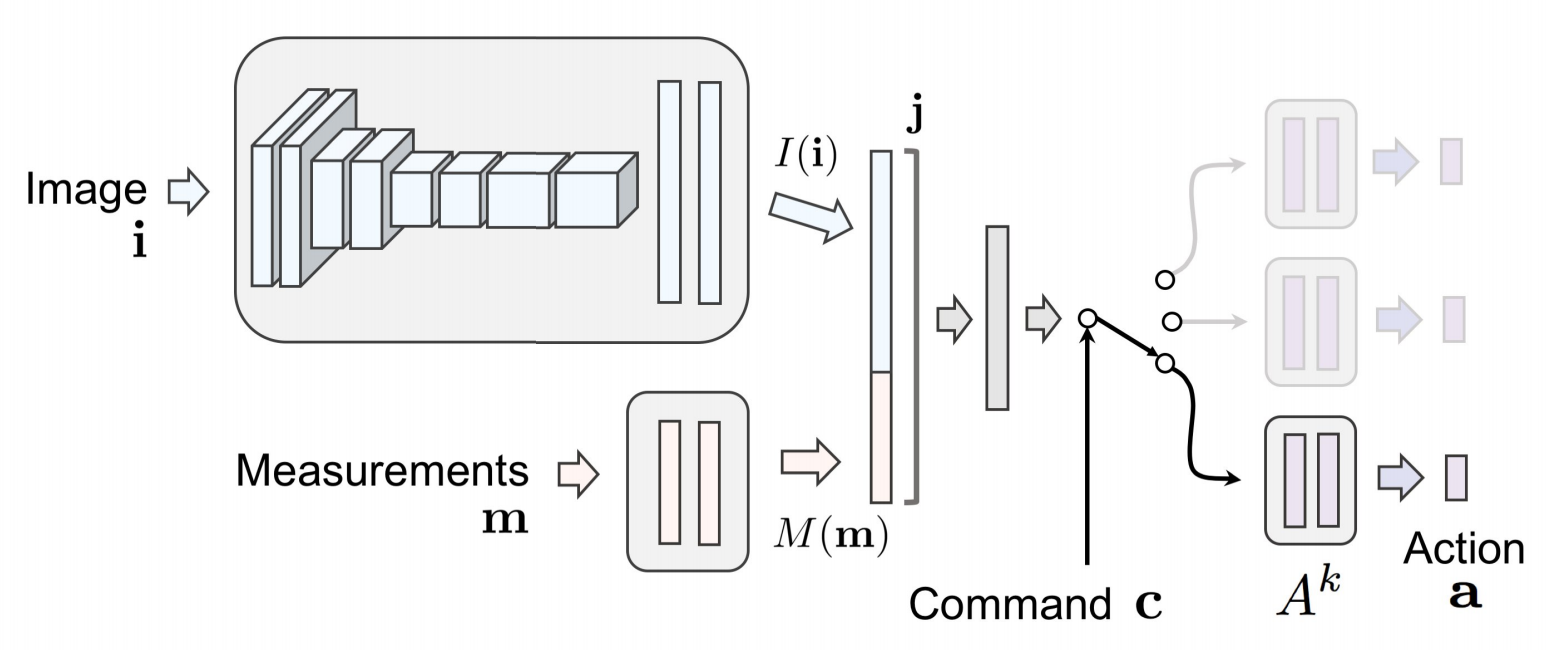}
	\caption[Goal-conditional Behavior Cloning]{\textbf{Goal-conditional Behavior Cloning.} Architecture of goal-conditional end-to-end behavior cloning for autonomous driving proposed by \protect\citet{Codevilla2018ICRA}. The goal command acts as a switch that selects between specialized sub-policies that correspond to different commands such as lane following, turning left or turning right. \figsourceC{\protect\citet{Codevilla2018ICRA}}{2018}{IEEE}.}
	\label{fig:Codevilla2018ICRA}
\end{figure}

Besides the drifting problem during test time, behavior cloning-based driving systems have other limitations. Sensor input alone is often not sufficient to uniquely infer control. Consider intersections, for example, where multiple possible actions are valid (left, right, straight). Without conditioning on the goal, all three options are acceptable. Thus, some of the behavior cloning agents, such as the one by \citet{Bojarski2016ARXIV}, require human intervention for lane changes or turns. To alleviate this limitation, \citet{Codevilla2018ICRA} propose a conditional imitation learning framework to learn a driving policy for steering and throttle control from a high-level navigational input in addition to the observations from the camera (\figref{fig:Codevilla2018ICRA}). The high-level navigational input represents the driver's intention, such as the direction to take at the next intersection, which cannot be recovered from sensory input alone.

\citet{Codevilla2019ARXIV} identify other limitations of behavior cloning approaches related to generalization performance. They observe that in contrast to typical supervised learning tasks, the generalization performance for behavior cloning does not scale with training data. Moreover, they identify significant variance in performance when varying the model initialization or the order in which training examples are sampled from the dataset. 

\begin{figure}[t]
	\centering
	\includegraphics[width=1.00\columnwidth]{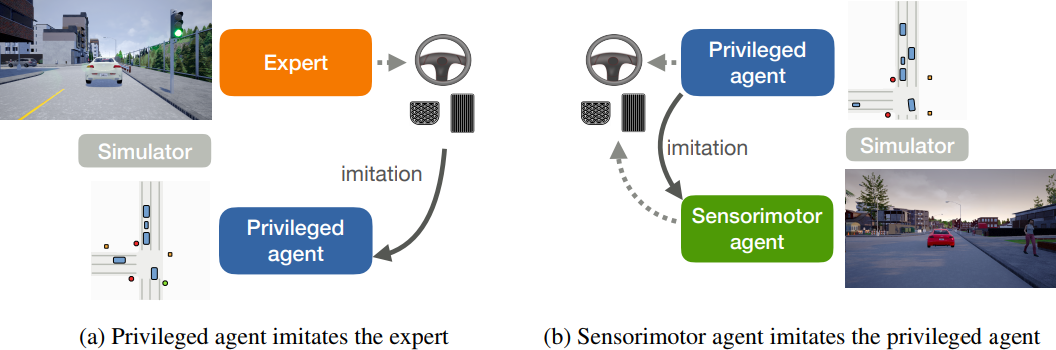}
	\caption[Learning by Cheating.]{\textbf{Learning by Cheating.} \citet{Chen2019CORL} propose to first learn an agent with privileged information (a) which afterwards teaches an agent without access to to privileged information (b) that learns to imitate the privileged agent. \figsourceC{\citet{Chen2019CORL}}{2019}{CoRL}}
	\label{fig:Chen2019CORL}
\end{figure}

\citet{Chen2019CORL} show that imitation learning can be simplified by decomposing it into two stages, as illustrated in \figref{fig:Chen2019CORL}. They first train an agent that has access to privileged information. This privileged agent cheats by observing the ground-truth layout of the environment and the positions of all traffic participants. In the second stage, the privileged agent acts as a teacher that trains a purely vision-based sensorimotor agent. The resulting sensorimotor agent does not have access to any privileged information and does not cheat. They demonstrate that this approach substantially outperforms the state of the art on the CARLA benchmark and the recent NoCrash benchmark, attaining the best performance to date.

\subsection{Reinforcement Learning}
\begin{figure}[t]
	\centering
	\includegraphics[width=1.00\columnwidth]{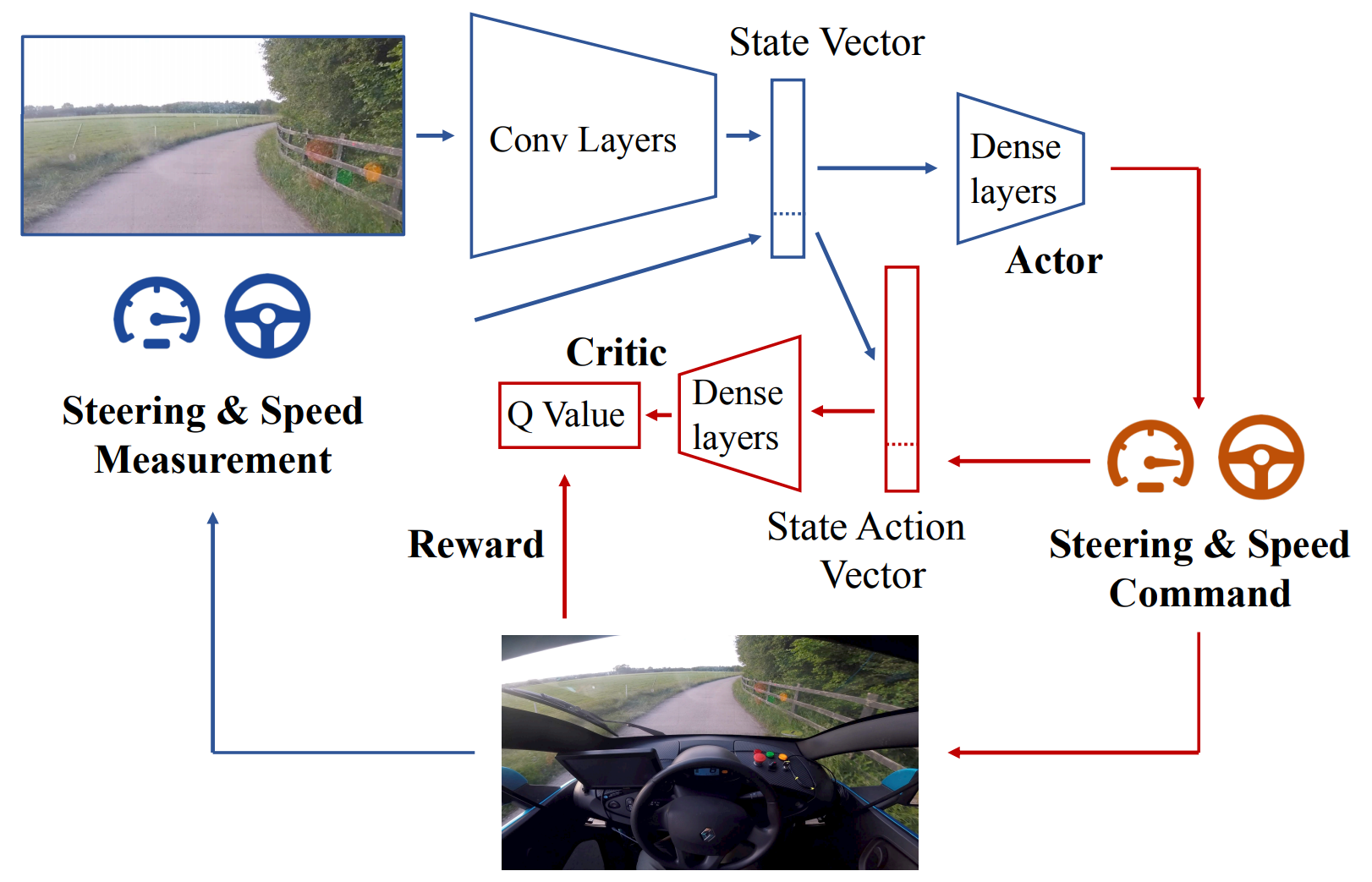}
	\caption[Reinforcement Learning.]{\textbf{Reinforcement Learning.} Block diagram of the autonomous system proposed by \protect\citet{Kendall2018ARXIV}. The system is trained end-to-end using only the reward from the environment. The value function learned by the critic network is used to update the actor's policy network parameters to increase the reward and improve the policy's performance. \figsource{\protect\citet{Kendall2018ARXIV}}.}
	\label{fig:Kendall2018ARXIV}
\end{figure}

Approaches based on reinforcement learning (RL) learn to drive by training an agent that tries to maximize a user defined reward which the agent receives while interacting with the environment. In the autonomous driving application, the reward is defined by specifying the driving agent’s preferences and goals.
\citet{Dosovitskiy2017CORL} propose a reinforcement learning method that trains a deep network based on a reward function provided by the CARLA simulator which combines speed, distance traveled towards the goal, collision damage, overlap with sidewalk and overlap with the opposite lane. For training the agent, they use the asynchronous advantage actor-critic (A3C) algorithm \citep{Konda1999NIPS} which uses the value function learned by the critic to update the actor's policy. \citet{Dosovitskiy2017CORL} observe that the RL agent performs significantly worse compared to a behavior cloning agent trained using conditional imitation learning \citep{Dosovitskiy2017CORL} despite the fact that the RL agent was trained on a significantly larger set of visual observations.
Recently, \citet{Kendall2018ARXIV} showed first promise in learning to drive in the real-world using a reinforcement learning agent (\figref{fig:Kendall2018ARXIV}). They use the deep deterministic policy gradients algorithm for training the RL agent and define the reward as the distance traveled by the vehicle without the safety driver taking control.

The aforementioned methods are trained using model-free reinforcement learning. The disadvantage of model-free methods is that they are often data inefficient and require a large number of interactions with the environment. In contrast, model-based reinforcement learning approaches learn a model of the environment dynamics from observational data and then exploit this model for training a driving policy. Model-based methods have been shown to significantly reduce the number of environment interactions required to learn an effective policy. However, model-based methods also typically require an interactive environment as a dynamics model trained on a fixed set of demonstrations may make incorrect predictions outside the training domain. The interactive training environment is however not practical in the real-world where such interactions are expensive and dangerous. To alleviate this problem, \citet{Henaff2019ARXIV} propose to train a model-based policy which is encouraged to produce actions which the forward dynamics model is confident about. They achieve this by training the policy network to minimize an uncertainty cost which represents the mismatch between the states it induces and the states in the trained data.

\subsection{Combined Methods}

Behavior cloning methods are easy to train in a supervised fashion. However, they are poor at exploring the environment and therefore require extensive on-policy data augmentation using methods like DAgger~\citep{Ross2010AISTATS}.
RL approaches, in contrast, do not require per-frame supervision and are better at exploration. However, they are inefficient to train and require a simulator or non-practical trial and error runs in a real environment as well as careful design of the reward function. Therefore, several methods have been proposed to combine the strengths of both approaches.

\citet{Liang2018ARXIV} propose an approach to alleviate the low exploration efficiency of RL for large action space. They achieve this by constraining the policy search space by initializing the weights of the policy network of an RL algorithm by a network trained to clone the expert behavior. They observe significant improvements on the CARLA benchmark over agents trained using RL from scratch.
\citet{Li2018ARXIVa} propose an approach that learns to clone only the best behaviors of several sub-optimal teachers. They estimate the best teacher by estimating the value function of each sub-optimal teacher. The sub-optimal teachers are defined using several simple controllers over the planner output. Therefore, they do not require expert teachers for labeling data and allow for better exploration compared to learning from a single expert teacher. In addition, learning from multiple sub-optimal teachers leads to faster training compared to pure RL agents as exploration only happens from feasible states.
The requirement to specify the reward function limits the practical use of Reinforcement Learning. An accurate specification of the reward  requires tedious and computationally inefficient hyper-parameter tuning. \citet{Sharifzadeh2016NIPSWORK} propose to learn the unknown reward function of the driving behavior from expert demonstrations by applying Inverse Reinforcement Learning (IRL).
In contrast to behavior cloning approaches that directly learn the observation-control mapping in a supervised fashion, Inverse Reinforcement Learning approaches claim to offer better generalization by learning a reward function that explains the expert behavior. 

\subsection{Intermediate Representations}

\begin{figure}[t]
	\centering
	\includegraphics[width=1.00\columnwidth]{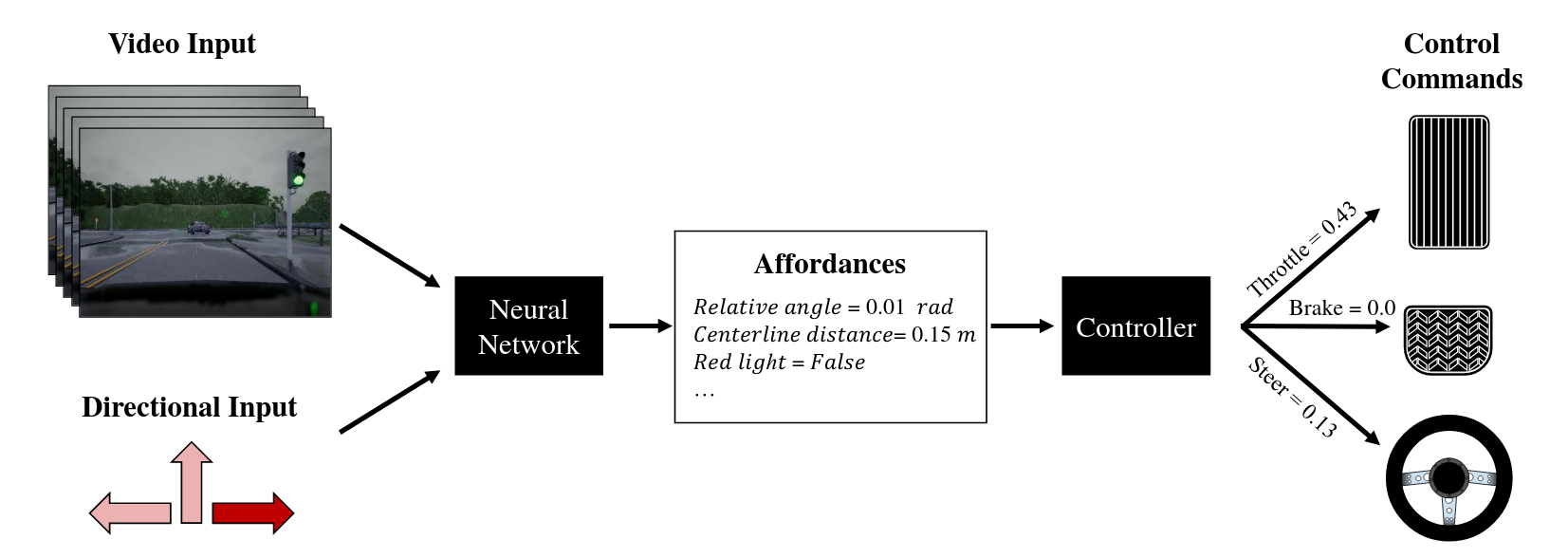}
	\caption[Conditional Affordance Learning]{\textbf{Conditional Affordance Learning.} The input video and the high-level directional commands are fed into a neural network which predicts a set of affordances such as presence of red traffic lights or distance to the lane center. These affordances are used by a controller to compute the control output. \figsourceC{\protect\citet{Sauer2018CORL}}{2018}{CoRL}.}
	\label{fig:Sauer2018CORL}
\end{figure}

Instead of directly learning a mapping from pixels to actions, \citet{Chen2015ICCVa} present an approach which first estimates a small number of human interpretable, pre-defined affordance measures such as the angle of the car relative to the road, the distance to the lane markings, and the distance to cars in the current and adjacent lane. These predicted affordances are then mapped to car actions using a rule-based controller to enable autonomous driving in the TORCS car racing simulation \cite{Wymann2015}.
The advantage of mid-level representations is that the network predicting the mid-level representations can be trained and validated before deploying them. In addition, the mid-level representations are more interpretable compared to traditional behavior cloning approaches.
Similarly, \citet{Sauer2018CORL} estimate several affordances from sensor inputs in order to drive a car, as illustrated in \figref{fig:Sauer2018CORL}. In contrast to \citet{Chen2015ICCVa}, they consider the more challenging scenario of urban driving using the CARLA simulator \citep{Dosovitskiy2017CORL}. In CARLA, the agent needs to obey traffic rules such as speed limits, red lights, avoid colliding with obstacles on the road and navigate at junctions with multiple possible driving directions. \citet{Sauer2018CORL} realize their driving agent by expanding the set of affordances to cover the most important aspects of urban environments. Similar to \citet{Chen2015ICCVa}, they use a rule-based controller to map affordances to vehicle controls.

Recently, \citet{Zhou2019ARXIV} studied the significance of using intermediate representations pursued in computer vision research such as depth, segmentation, optical flow for improving several sensorimotor tasks such as urban driving. They observed that an agent that takes as input one or more of these intermediate representations along with the image learns significantly better sensorimotor control than an agent which uses just the raw image as input. They observed significant improvements even when the intermediate representations were noisy predictions by a simple deep network.
\citet{Bansal2018ARXIV} propose a perception module that translates raw sensor observations to a mid-level representation. Their representation includes a top-down rendering of the environment where 2D boxes of vehicles are drawn along with a rendering of the road information and traffic light states. They use this mid-level representation as input to a recurrent neural network (RNN) which outputs the control command. 
Similarly, \citet{Wang2019ARXIV} infer the depth and poses of the objects present in the scene from front-facing camera images and project the objects into an overhead view. They train a behavior cloning agent over the concatenation of front-facing and overhead images and observe improved performance over an agent trained only on front-facing images.
In the same spirit, \citet{Muller2018ARXIV} train a driving policy in CARLA with mid-level representations as input. Specifically, they used binary segmentation estimated from a scene segmentation network as input to the driving policy network and observed improvements over an agent trained on raw camera images.

Similar to the aforementioned methods, \citet{Mehta2018ARXIV} also propose to use intermediate visual affordances such as ``distance to intersection'', and action primitives such as ``slow down'' as input to the driving policy network. However, in contrast to the aforementioned works, they predict visual affordances and action primitives as an auxiliary task to the driving control task.
They claim that predicting representations which are crucial for the driving decision allow the policy network to learn superior internal representations leading to more efficient training and better generalization.
\citet{Kendall2018ARXIV} studied the importance of using an intermediate representation for state representation instead of raw pixels for learning a reinforcement learning-based driving policy. They observed significant improvements in data efficiency in training the driving policy using a compressed representation of the raw image, obtained using a Variational Autoencoder (VAE)

\subsection{Transferring from Simulation to the Real World}

\begin{figure}[t]
	\centering
	\includegraphics[width=0.75\columnwidth]{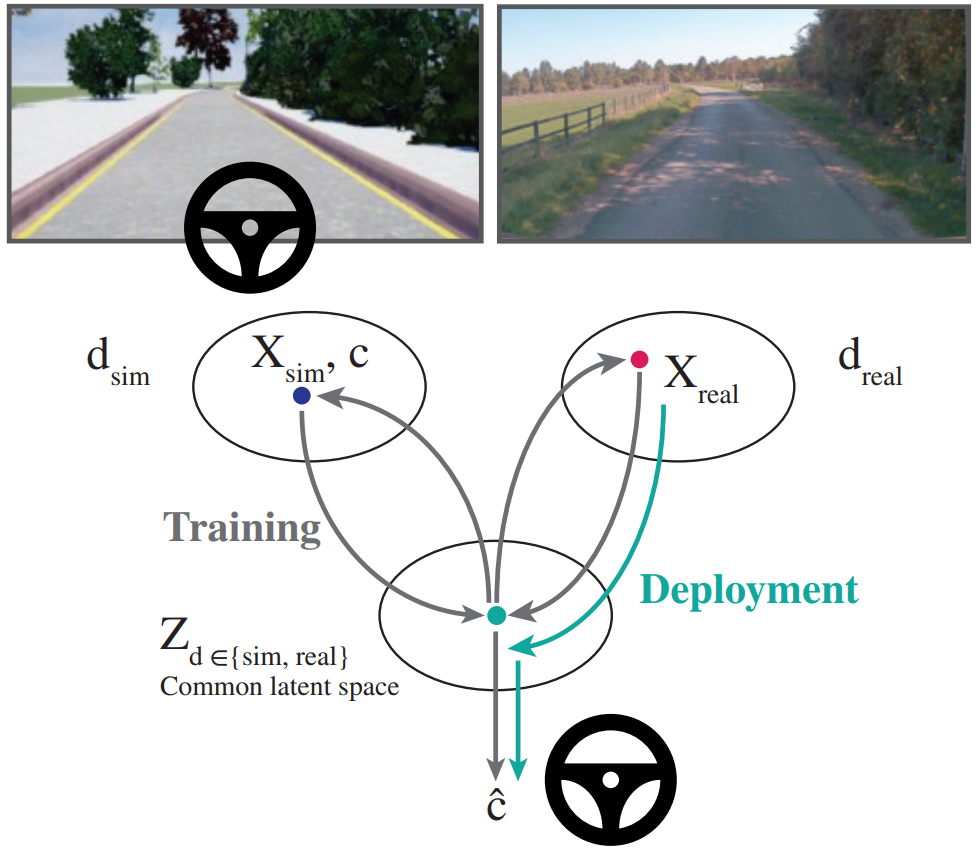}
	\caption[From Simulation to the Real World.]{\textbf{From Simulation to the Real World.} \protect\citet{Bewley2018ARXIV} proposed a model for end-to-end driving by learning to translate between simulated and real-world images, jointly learning a control policy from the common latent space $Z$ using expert labels in simulation. 
	Their method does not require real-world control labels and is able to learn a policy which can be transferred with improved generalization to real-world driving.
	\figsource{\protect\citet{Bewley2018ARXIV}}}
	\label{fig:Bewley2018ARXIV}
\end{figure}

One major limitation of reinforcement learning is the necessity of a simulation environment for trial and error. Thus, during training only synthetic data is considered and the models usually do not generalize to real data. To address this problem,
\citet{Pan2017BMVC} propose to transfer a reinforcement learning agent trained in a virtual environment to the real-world.
More specifically, they learn an image translation network to translate non-realistic simulated images to realistic images. Their translation network is composed of two conditional GANs, the first for segmenting virtual images from the simulator, and the second for translating the segmented images to their realistic counterparts.
In the same spirit, \citet{Bewley2018ARXIV} propose to train an image-to-image translation network for transferring a driving policy from simulation to real-world without any real-world control labels (\figref{fig:Bewley2018ARXIV}). In contrast to \citet{Pan2017BMVC}, which uses an explicit semantic segmentation as intermediate representation, they use an implicit latent structure as intermediate representation. They propose two autoencoder-like networks for translating between domains where a common latent space is learned through direct and cyclic losses. Their control network is trained using behavior cloning by passing the latent code as input to the control network.

\section{Datasets}

\begin{figure}[t]
	\centering
	\includegraphics[width=1.00\columnwidth]{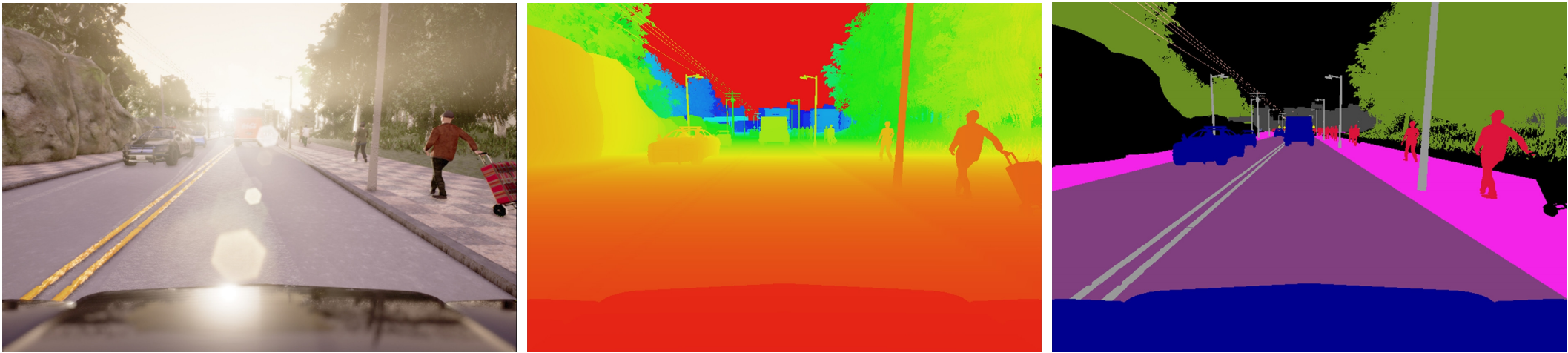}
	\caption[Sensor Modalities in CARLA]{\textbf{Sensor Modalities in CARLA.} From left to right: RGB image, ground-truth depth and ground-truth semantic segmentation. Additional sensor models can be plugged in via the provided API. \figsourceC{\protect\citet{Dosovitskiy2017CORL}}{2017}{CoRL}.}
	\label{fig:Dosovitskiy2017CORL}
\end{figure}

As behavior cloning approaches can be trained on offline expert demonstrations, several public datasets have been introduced in the last few years to train and evaluate such methods. 
The comma.ai dataset \citep{Santana2016ARXIV} provides 7.25 hours of driving data with a camera in the windshield, capturing images of the road at 20Hz. The dataset also provides observations from several other sensors such as car speed, steering angle, GPS, gyroscope, and IMU. However, the dataset only contains training examples for highway scenarios, and is therefore not suitable for learning a driving policy which operates in more challenging situations such as in cities.
The Berkeley DeepDrive Video dataset \citep{Xu2017CVPR} comprises 10,000 hours of driving in cities, on highways, in towns, and in rural areas. The dataset has been recorded using forward-facing dash cameras along with observations from sensors such as GPS, IMU, gyroscope, and magnetometer. As discussed in the previous section, online rollouts of the driving policy is an important requirement for training and evaluation of most end-to-end learning methods. However, deploying a partially trained model in a real environment to collect training data is both dangerous and impractical. Therefore, realistic driving simulators are a key requirement for training and evaluating these models. 

As one of the first open-source simulators, the TORCS racing car simulator \citep{Wymann2015} has been used for learning and evaluation of road lane following by \citet{Chen2015ICCVa}. However, the TORCS environments is simplistic, lacking complexities such as traffic participants, junctions, etc.

In contrast, CARLA \citep{Dosovitskiy2017CORL} provides a more realistic, complex and flexible open-source simulator for autonomous driving that enables training and validation in urban driving conditions. It provides high quality images along with ground-truth depth and semantic segmentation as pseudo-sensors, as illustrated in \figref{fig:Dosovitskiy2017CORL}.
In order to replicate the complex nature of urban driving, the environments in CARLA exhibit realistic urban street layouts with traffic rules, intersections, buildings, pedestrians, street signs and other traffic participants. The simulator also provides different weather and lighting conditions in order to evaluate the generalization ability of the driving agent. CARLA also provides a benchmark based on four increasingly difficult driving tasks and is actively expanded in terms of the environments, assets and agents it provides.

However, existing real-world datasets and synthetic simulators often fail to capture the long tail of the distribution which covers important but rare situations. These rare events can only be effectively captured with a large fleet of vehicles that log these situations in real-world driving. Tesla's Autopilot system \citep{TeslaSoftware9} is a dormant logging-only mode that can be queried for multiple instances of rare failure situations so that the model can be trained to avoid such failures. In addition, Shadow Mode allows Tesla to validate the Autopilot system running in the background in real situations. However, data from Tesla vehicles are proprietary and hence not released to other companies or public research institutions.

\section{Metrics}
There are no standard metrics and benchmarks for autonomous driving and thus most methods usually evaluate on their own set of metrics and datasets. The most popular benchmark CARLA \citep{Dosovitskiy2017CORL} evaluates on two metrics. First, the percentage of successfully completed episodes under the four different conditions provided by CARLA. And second, the average distance (in kilometers) driven between two infractions. Infractions include driving on the opposite lane, driving on the sidewalk, colliding with other vehicles, colliding with pedestrians, and hitting static objects.
\citet{Codevilla2018ECCV} use CARLA to analyze the correlation between offline and online metrics for evaluation of autonomous driving agents. They observe that offline metrics such as the squared or absolute error of the steering angle are poorly correlated with online metrics such as the success rate of reaching the goal. Their work highlights the tension between imitation learning and reinforcement learning. While reinforcement learning allows to train for the desired goal, training an imitation learning agent is significantly easier and does not require potentially unsafe exploration.

\section{Discussion}
A common characteristic of most end-to-end driving methods is the need to collect online training data. While behavior cloning methods have shown promising results by learning purely from expert demonstrations, minimizing the covariate shift between the expert trajectories and the agent's policy is still an open problem. 
Similarly, reinforcement learning approaches require millions of trial and error runs and thus can only be safely applied in simulation environments. 
Moreover, as shown by \citep{Codevilla2018ECCV}, offline evaluation metrics are poorly correlated with online driving performance.
Therefore, safe training and validation of end-to-end learning models require further development of realistic simulators such as CARLA~\citep{Dosovitskiy2017CORL} on which these methods can be trained before transferring the resulting policies to the real-world.
Flexibility is another desirable characteristic when developing simulators: the resulting simulations should allow for highly diverse and complex scenarios, yet also model the long tail of the data distribution to capture rare events.
While realistic simulation environments are important, there will likely remain a domain gap between simulated and real data.
Therefore, another critical direction of future research is the design of end-to-end learning methods which can be robustly transferred from simulated environments to the real-world.
Furthermore, the lack of interpretability of end-to-end driving networks prevents deeper insights into the modes of operation (in particular legal relevant failure cases) and thus requires further investigation.

	\chapter{Conclusion}
This book provides a comprehensive survey on problems, datasets, and methods in computer vision for autonomous vehicles. Towards this goal, we considered the historically most relevant literature as well as the state of the art on several relevant topics, including recognition, reconstruction, motion estimation, tracking, scene understanding, and end-to-end learning. We discussed open problems and current research challenges in each of these areas and also provided a novel in-depth analysis of the KITTI benchmark.

While self-driving vehicles have a long history, it remains difficult to make predictions when self-driving vehicles will hit the consumer market. Traditionally, the problems involved in achieving or surpassing human-level performance on this task have been underestimated. Difficulties include the high accuracy that needs to be attained, the robustness required for safe self-driving as well as adverse weather conditions (snow, rain, night). Furthermore, most self-driving systems rely on accurate HD maps for localization and detection of static infrastructure, which are hard to create and to maintain up-to-date. In addition, some of the most challenging scenarios are less structured (parking areas, complex roundabouts) and thus need to be mastered without HD maps. Pedestrians pose another challenge to self-driving vehicles as their behavior is often erratic, and communication with them can be key for making a driving decision. Other challenges include complex planning tasks such as merging into traffic and negotiating with other vehicles. Further, several ethical and legal questions need to be addressed before self-driving vehicles can be deployed in large numbers on public roads.

From a technical perspective, modular pipelines offer the advantage of parallelization, interpretability, and ease of introducing prior knowledge. However, human-engineered modules often rely on heuristics or intuitions, which may be inaccurate or wrong. Learning driving policies from data is an attractive alternative, however bridging the gap to modular and interpretable systems as well as attaining human-level performance remain unsolved problems to date. A particularly challenging problem is generalization to unseen environments and to handle rare events for which little data is available.

We are at an exciting time where self-driving technology receives considerable attention and progress is fast. At the same time, it is of prime importance that we stay objective and cautious with the claims that we make in order not to gamble people’s trust in this new technology or put people's lives at stake.
Writing a survey on this rapidly evolving field was a major tour de force.
We are well aware that some of the approaches surveyed in this work might be outdated in the near future. However, some of the works presented in this survey will stand the test of time and will be remembered as landmarks in the development of autonomous vehicles.
We hope that this survey, in combination with our online navigation tool\footnote{\url{http://www.cvlibs.net/projects/autonomous_vision_survey}}, will become useful references, encourage new research, and ease the entry for beginners starting in this exciting field.

\section{Acknowledgement}
We thank Raghudeep Gadde, Varun Jampani, Yiyi Liao, Despoina Paschalidou, Jörg Stückler, Torsten Sattler, Siyu Tang, and Osman Ulusoy for sharing their expert knowledge and giving us valuable feedback on early versions of the draft.
We also highly appreciate the help of Davide Scaramuzza, Daniel Maturana, and many others in the community for their feedback and suggestions for related work.
Finally, we would like to thank all researchers who gave us permission to use the figures from their papers and greatly appreciate the support of Benjamin Coors and Jonas Wulff who provided additional illustrations of their work.

	\backmatter  %
	
	\printbibliography
	
\end{document}